# LEAF DISEASES DETECTION USING DEEP LEARNING METHODS

**A THESIS SUBMITTED TO**
**THE GRADUATE SCHOOL OF NATURAL AND APPLIED SCIENCES**
**OF**
**ANKARA UNIVERSITY**

**by**

**El Houcine EL FATIMI**

**IN PARTIAL FULFILMENT OF THE REQUIREMENTS**
**FOR THE DEGREE OF**
**DOCTOR OF PHILOSOPHY IN**
**COMPUTER ENGINEERING**

**ANKARA**
**2023**

# ABSTRACT

Ph.D. Thesis

LEAF DISEASES DETECTION USING DEEP LEARNING METHODS

El houcine EL FATIMI

Ankara University
Graduate School of Natural and Applied Sciences
Department of Computer Engineering

Supervisor: Prof. Dr. Recep ERYİĞİT

Leaf diseases are a prevalent issue in the realm of plant health and can have a profound impact on crop yields. Therefore, detecting early signs of disease is essential for optimal plant health and productivity. Unfortunately, current methods for detecting leaf diseases are ineffective, and there is much room for improvement in this area. For example, existing techniques are typically slow and may require human intervention to identify possible disease symptoms. Deep learning methods have the potential to overcome these limitations and provide more accurate detection and identification of leaf diseases than traditional methods. However, many challenges still need to be overcome before these methods can be implemented in practice. In particular, researchers must develop resilient and precise deep learning models capable of accurately detecting and identifying different types of leaf diseases in large datasets of leaf images. The models should also be trained to perform well with images of varying quality and clarity, such as those captured under challenging environmental conditions or at different stages of a disease's development, these models should also be able to recognize and distinguish between similar diseases and other benign conditions. To address these challenges, we need to improve existing deep learning methodologies, including convolutional neural networks (CNNs), to generalize better across datasets and yield improved detection accuracy for various leaf diseases and conditions. In this study, our main topic is to devlop a new deep-learning approachs for plant leaf disease identification and detection using leaf image datasets. We also discussed the challenges facing current methods of leaf disease detection and how deep learning may be used to overcome these challenges and enhance the accuracy of disease detection. Therefore, we have proposed a novel method for the detection of various leaf diseases in crops, along with the identification and description of an efficient network architecture that encompasses hyperparameters and optimization methods. The effectiveness of different architectures was compared and evaluated to see the best architecture configuration and to create an effective model that can quickly detect leaf disease. In addition to the work done on pre-trained models, we proposed a new model based on CNN, which provides an efficient method for identifying and detecting plant leaf disease. Furthermore, we evaluated the efficacy of our model and compared the results to those of some pre-trained state-of-the-art architectures. A parameter-tuning algorithm was developed to identify the optimal performance of each model. In this work, model building and testing will be carried out using open datasets from the literature as well as collected datasets for the plant that do not exist in the literature. In this work, we also discussed the effect of datasets on the effectiveness of deep learning models. The proposed leaf disease detection approaches were successfully implemented and evaluated, and very satisfactory performance results were obtained.

## ÖZET

Doktora Tezi

DERİN ÖĞRENME YÖNTEMLERİNİ KULLANARAK YAPRAK HASTALIKLARININ TESPİTİ

El houcine EL FATIMI

Ankara Üniversitesi
Fen Bilimleri Enstitüsü
Bilgisayar Mühendisliği Anabilim Dalı

Danışman: Prof. Dr. Recep ERYİĞİT

Yaprak hastalıkları, bitki sağlığı alanında önemli bir zorluk teşkil etmekte ve mahsul verimi üzerinde derin bir etki yaratmaktadır. Bu hastalıkları erken aşamalarında tespit etmek, optimum bitki sağlığını korumak ve maksimum verimliliği sağlamak için çok önemlidir. Bununla birlikte, yaprak hastalıklarının tespiti için kullanılan mevcut yöntemler genellikle yetersiz kalmakta, etkisizlik göstermekte ve önemli ölçüde geliştirilmeye ihtiyaç duymaktadır. Mevcut teknikler tipik olarak yavaşlıklarıyla karakterize edilir ve bazen potansiyel hastalık belirtilerini tanımlamak için insan müdahalesi gerektirir.

Derin öğrenme yöntemleri, geleneksel yöntemlere kıyasla yaprak hastalıklarının daha hassas ve güvenilir bir şekilde tespit edilmesi ve tanımlanması potansiyelini sunarak bu sınırlamaların üstesinden gelme konusunda umut vaat etmektedir. Bununla birlikte, bu yöntemlerin pratik kullanıma etkili bir şekilde entegre edilebilmesi için çeşitli engellerin aşılması gerekmektedir. Araştırmacılar, yaprak görüntülerinden oluşan kapsamlı veri setleri içinde çeşitli yaprak hastalıklarını hassas bir şekilde tespit edip tanımlayabilen sağlam ve doğru derin öğrenme modelleri oluşturma zorluğuyla karşı karşıyadır. Bu modeller, olumsuz çevresel koşullar altında veya hastalık gelişiminin çeşitli aşamalarında çekilenler de dahil olmak üzere, değişen kalite ve netlikteki görüntülere uyum sağlayabilmelidir. Ayrıca, bu modeller benzer hastalıkları ve diğer zararsız koşulları tanıyabilme ve ayırabilme yeteneğine sahip olmalıdır.

Bu çok yönlü zorlukları etkili bir şekilde ele almak için, özellikle evrişimli sinir ağlarının (CNN'ler) optimizasyonu ve geliştirilmesine odaklanarak, mevcut derin öğrenme metodolojilerinin kapsamlı bir şekilde iyileştirilmesini üstlenmemiz gerekmektedir. Bu dikkatli iyileştirme çabası, farklı karmaşıklık, ölçek ve görüntü kalitesi derecelerini kapsayan çeşitli veri setleri genelinde bilgi ve içgörüleri genelleştirme yetenekleri açısından bu sinir ağlarının kapasitesini artırma zorunluluğundan kaynaklanmaktadır. Aynı zamanda çabalarımız, hastalık tespitinin genel hassasiyetini ve etkinliğini artırmaya, her biri kendine özgü görsel imzalarla karakterize edilen zengin bir yaprak hastalıkları dokusu ve tespit görevine nüanslar ve karmaşıklıklar getirebilecek çeşitli çevresel koşullar boyunca uygulanabilirliğini genişletmeye yöneliktir. Temel olarak, iyileştirme süreci, sürekli gelişen bitki sağlığı ve yaprak hastalığı tespiti ortamının ortaya çıkardığı karmaşık zorlukların üstesinden gelebilecek derin öğrenme metodolojileri için daha çok yönlü ve dirençli bir temel oluşturmayı amaçlamaktadır. Temelde, özellikle CNN'ler çerçevesinde derin öğrenme metodolojilerinin iyileştirilmesi, bu hesaplama sistemlerini geleneksel sınırlamaların sınırlarını aşacak şekilde güçlendirmeyi amaçlayan çok yönlü bir çabayı temsil etmektedir. Bu sistemlerin, karmaşık ve dinamik bir alan olan yaprak hastalıklarının tespitinde üstünlük sağlamak için gerekli esneklik ve zeka ile donatılması ve böylece bu kritik tarımsal soruna daha doğru, verimli ve uyarlanabilir çözümlerin önünün açılması amaçlanmaktadır.

Bu çalışmanın ana amacı, yaprak görüntüsü veri setlerini kullanarak bitki yaprak hastalıklarının tanımlanması ve tespit edilmesi için yeni derin öğrenme yaklaşımlarının geliştirilmesi etrafında dönmektedir. Mevcut yaprak hastalığı tespit yöntemlerinin karşılaştığı mevcut zorlukları incelemekte ve derin öğrenmenin hastalık tespit doğruluğunu artırmak için nasıl bir çözüm olarak hizmet edebileceğini açıklamaktadır. Bu çabanın bir parçası olarak, kritik hiperparametreleri ve optimizasyon tekniklerini kapsayan verimli bir ağ mimarisinin tanımıyla birlikte, mahsullerdeki çeşitli yaprak hastalıklarının tespiti için yenilikçi yöntemler önerilmektedir. Bu çalışma, hızlı ve doğru yaprak hastalığı tespiti yapabilen bir model oluşturmayı amaçlayarak, en etkili yapılandırmayı belirlemek için farklı ağ mimarilerini sistematik olarak karşılaştırmaktadır.

Ayrıca araştırma, bitki yaprak hastalıklarının tanımlanması ve tespit edilmesi için etkili bir araç sunan CNN'e dayanan yeni bir modeli tanıtmaktadır. Modelin etkinliği kapsamlı bir şekilde değerlendirilmiş ve önceden eğitilmiş son teknoloji mimarilerden elde edilen sonuçlarla karşılaştırılmıştır. Her modelin performansını optimize etmek için bir parametre ayarlama algoritması geliştirilmiştir.

Ayrıca, bitki hastalıklarının tanımlanması ve sınıflandırılmasına ilişkin mevcut araştırmaların öncelikle tek kanallı ve aynı çözünürlüklü görüntülere odaklandığı belirtilmektedir. Bu sınırlama, kesin hastalık tespiti için gerekli olan kapsamlı bilgiyi yakalamada yetersiz kalabilir. Sonuç olarak, bu tez, hastalık sınıflandırma hassasiyetini ve verimliliğini artırmak için birden fazla bilgi kanalından yararlanarak yaprak hastalıklarını otomatik olarak sınıflandıran derin çok ölçekli evrişimli sinir ağı (DMCNN) çerçevesini bir çözüm olarak ortaya koymaktadır.

Bu Araştırmada, hem genel kullanıma açık veri setlerini hem de yeni derlenen veri setlerini kullanarak model geliştirme ve test etme çalışmaları yürütmekte ve literatürdeki bir boşluğu doldurmaktadır. Ayrıca bu çalışma, çeşitli veri setlerinin derin öğrenme modellerinin etkinliği üzerindeki etkisini araştırmakta, bu modellerin farklı veri setleriyle karşılaştıklarında nasıl performans gösterdiğini ve daha fazla veri çeşitliliği ile performanslarının nasıl artırılabileceğini incelemektedir. Bu çaba özellikle yaprak hastalıklarının belirlenmesi bağlamında önemlidir ve yaprak hastalıkları için teşhis ve tedavi sistemlerini geliştirirken yaprak hastalıklarını tanımak için derin öğrenmenin potansiyelinin altını çizmektedir.

Önerilen yaprak hastalığı tespit yöntemleri başarıyla uygulanmış ve kapsamlı bir şekilde değerlendirilerek oldukça tatmin edici performans sonuçları elde edilmiştir. Bu araştırma, yaprak hastalıklarıyla mücadele etme ve mahsul sağlığı ile verimliliğini artırma arayışında önemli bir adım teşkil etmektedir.

## FOREWORD AND ACKNOWLEDGEMENTS

I would like to express my sincere gratitude and appreciation to my supervisor, Prof.Dr. Recep ERYIĞIT, for their unwavering support, guidance, and invaluable feedback throughout the entire duration of my Ph.D. journey. Their expertise, mentorship, and dedication have been instrumental in shaping the direction and quality of this research.

I would like to extend my thanks to my colleagues and fellow researchers for their stimulating discussions, valuable feedback, and camaraderie. I am also grateful to the staff and faculty of Ankara University, Computer Engineering Department, for providing a conducive research environment.

I would like to express my heartfelt gratitude to my family for their unwavering support, understanding, and encouragement throughout this long and challenging journey. Their love, patience, and belief in me have been a constant source of motivation.

Lastly, I would like to thank all the individuals who have supported me in any way during my Ph.D. program. Your contributions have been invaluable and greatly appreciated.

El houcine EL FATIMI
Ankara, September 2023

## TABLE OF CONTENTS

## SYMBOLS AND ABBREVIATIONS

| | |
|---|---|
| AI | Artificial Intelligence |
| CNNs | Convolutional Neural Networks |
| DL | Deep learning |
| DNN | Deep Neural Network |
| DBM | Deep Boltzmann Machine |
| DNN | Deep Neural Network |
| FCNNs | Fully Convolutional Neural Networks |
| GPU | Graphics Processing Unit |
| MLPs | Multi Layer Perceptrons |
| ReLU | Rectified Linear Unit |
| RBM | Restricted Boltzmann Machine |
| RNN | Recurrent Neural Network |
| SVM | Support Vector Machine |
| DLQP | Directional Local Quinary Patterns |
| FBFN | Fuzzy Based Function Network |
| ANNs | Artificial neural networks |
| PCNN | Plain convolutional neural network |
| DRNN | Deep residual network |
| SCNN | Shallow CNN |
| LSTM | Long Short-Term Memory Neural Network |
| MLPs | Multi-layer perceptrons |
| SLP | Single-layer artificial neural network |
| MCMC | Markov Chain Monte Carlo |
| RBF | Radial basis function |
| NaCRRI | National Crops Resources Institute |
| HSV | Hue, Saturation, Value |
| SGD | Stochastic Gradient Descent |
| DMCNN | Deep Multi-Scale Convolutional Neural Networks |
| $i$ | Input gate |
| $q$ | Input gate |
| $o$ | Output gate |
| $f$ | Forget gate |
| $c$ | Memory cell |
| $w$ | Weighted |
| $f_t$ | Activation vector for the forget gate |
| $o_t$ | Activation vector for the output gate |
| $i_t$ | Activation vector for the input/update gate |
| $\widetilde{c_t}$ | Cell input activation vector |
| $c_t$ | Cell state vector |
| $\sigma_g$ | Sigmoid function. |
| $\sigma_c$ | Hyperbolic tangent function |
| K | Kernel or filter |
| S | Feature map |
| G | Desired output |

## LIST OF FIGURES

## LIST OF TABLES

# 1. INTRODUCTION

Agriculture, with its indispensable and central significance, plays a vital part in the economic progress of the nation. But Plant leaf diseases are a major concern in agriculture, as they can cause significant yield losses and decrease produce quality, making them a primary research focus. Despite significant advances in agricultural research in recent years, there is still much that we still need to learn about plant diseases and their causes, and plant disease detection which remains a major challenge. Furthermore, inspecting all the plants on a farm manually for signs of disease is both time-consuming and expensive. However, detecting plant leaf diseases is challenging due to the wide range of symptoms that different pathogens can have and the subtle differences in how these and other factors can impact the shape and size of leaves. Therefore, it is very important to accurately detect plant diseases for crop quality and quantity.

Utilizing automatic techniques for detecting crop diseases is beneficial as it reduces the need for manual supervision, particularly in large production fields. One such technique is the employment of DL models for the detection of plant diseases. The automatic detection of leaf diseases is a significant research issue being pursued to benefit farmers because it is essential for early and rapid control of large fields of crops. Therefore, to solve this problem, in this study, we develop and implement an automated approach for the early identification of plant leaf diseases utilizing DL methods, which can identify different patterns in the images of plants leaf disease, and which shows great promise for increasing the accuracy of diagnostic procedures for plant diseases; using deep learning methods can easily automate the detection of leaf diseases and save both time and money. In addition, it would help to improve plant health management by detecting diseases as early as possible. Therefore, it is highly beneficial in the end to have various challenging projects because it contributes to the development of a system that can be easily maintained, improved, and expanded, especially as new models and algorithms are developed. As a result, this system will continue to be used as a tool for more accurate plant leaf disease detection, eventually addressing the complete set of image identification challenges.

Deep learning techniques can be developed to recognize disease symptoms automatically in images of plant leaves. These techniques have the capability to extract complex features from images and make accurate predictions, even in the data's presence of noise and variability. Employing DL for the detection of plant leaf diseases can greatly enhance the speed and accuracy of leaf disease diagnosis, minimize the reliance on manual labor, and enable real-time monitoring of crops. However, there are also challenges associated with using deep learning for leaf disease identification. These challenges include the need for high-quality data for testing and training the models, the selection and optimization of deep learning model architectures and hyperparameters, and the interpretability and explainability of the models. Despite these challenges, deep learning methods have demonstrated encouraging outcomes in leaf disease detection, and and the advancement of these techniques holds the potential to revolutionize how plant diseases are detected and managed.

Furthermore, several factors make detecting and identifying plant leaf disease difficult, including: (i) dataset images may be very closely related and thus have very similar diseases; (ii) environmental factors can influence plant leaf appearance; (iii) insufficient plant leaf datasets; (iv) image background; (v) symptom variations; and (vi) immature plant leaves do not always display apparent characteristics; they may differ in texture and color from mature plant leaves. Therefore, developing new models and algorithms will help us better understand fundamental leaf disease traits and the interactions that control these traits. It will also improve our ability to produce more efficient and accurate detection in other areas of image identification and the challenges they face. Developing new DL models to detect and identify leaf diseases will undoubtedly be challenging. However, it is an endeavor that could have substantial positive implications for both plant health and crop production, as it could provide us with the knowledge and insight necessary to make more informed decisions in the field of agriculture and allow us to develop more effective strategies for plant leaf disease prevention, identification, and management, ultimately leading to a more sustainable and productive agricultural industry. To enhance our capacity for efficient, correct, and reliable detection of leaf diseases, more research is necessary to develop DL models for their identification in the future.

In this study, we discussed the challenges facing current methods of leaf disease detection and how deep learning may be employed to overcome these challenges and enhance detection accuracy; we also studied existing methods extensively, and new alternative approaches have been proposed for plant disease identification using leaf image datasets, the proposed system provides an efficient method that includes a model that utilize advanced deep-learning methods to classify and detect leaf disease, we also presented a detailed study to determine the most efficient network architecture (hyperparameters and optimization methods) for model-setting architectures to get the ideal solution; we compared and evaluated the efficacy of various architectures until we found the best architecture configuration that allowed us to use a practical model that could detect leaf disease easily.

In reinforcement of the work done on the pre-trained models, we proposed a new CNN-based model that provides an efficient method for identifying and detecting plant leaf disease. The proposed approach is expected to help farmers solve their problems. We also presented all the processes for developing and implementing this new convolutional neural network (CNN) model for leaf disease identification in higher dimensional spaces. Then we fine-tuned our model by adjusting parameters like learning rate to ensure that our model performed similarly well on the dataset and was in the optimal solution. Furthermore, we assessed the efficacy of our deep learning model and compared the results to those of some state-of-the-art architectur; Thisis is an important insight for soft applications, such as when the models deployed to be used in an agriculture environment. Finally, we extracted the features for the best detection by varying the number and properties of layers in the image. The objective was to evaluate the feasibility of determining the architecture of the best-performing model and the best training settings for these problems. The developed model was tested on a collected dataset of Brassica seeds using various evaluation measures. This method assisted us in identifying and implementing the best-performing architectures and training settings for predicting class labels. It enables us to tune the training model and find the best combination of architecture parameters, network topology, and weight settings for predicting class labels. Furthermore, the developed model can be applied to address other problems with similar characteristics.

In this work, we also created new datasets, collaborating closely with a specialist in agriculture for expert labeling and data curation. We recognize the pivotal role of high-quality datasets in the development of robust deep learning models. Thus, we explore the impact of datasets on the performance of these models by evaluating a single model trained on different datasets of leaf images. First, we used MobileNet, a deep learning architecture model focused on image classification and a mobile platform. Then we presented a comparative study of MobileNet architecture applied to these datasets to detect leaf disease. In this context, we also discussed some issues related to automating leaf disease detection employing a single MobileNet model architecture on various datasets and how the performance of one model can change depending on the dataset changes. We used three datasets to evaluate a single model, including unhealthy and healthy classes; for each dataset, all parameters had to be constant and identical; The model's implementation was assessed using a variety of criteria, including training accuracy, validation accuracy, and test accuracy. The effect of datasets on the effectiveness of deep learning models is essential in various applications. For instance, one of the benefits of effectively comparing different experiments is the ability to achieve high performance, more accessible retraining, and longer life. In all of these cases, there are two critical questions: first, is there a difference between the datasets, or are they similar, and second, if so, what is the performance that corresponds to these differences or the similar dataset, and which variables are responsible for each case? As a result, an accurate system is designed in this thesis to give a clear idea about these mentioned problems of leaf disease detection using a single model and different datasets; in addition to conducting detailed experiments that exhibit the impact of each dataset on the model's performance is important.

Moreover, the existing studies on plant disease identification and classification predominantly concentrate on single-channel and same-resolution images, which may not be sufficient to capture the comprehensive information required for accurate disease detection. This thesis proposes a deep multi-scale convolutional neural network (DMCNN) framework for automatically classifying leaf diseases to address this limitation. The proposed approach utilizes multiple channels of information to enhance the precision and efficiency of disease classification. The DMCNN architecture comprises parallel streams of convolutional neural networks (CNNs) at different scales, that are merged at the end to form a single output.

The suggested approach is evaluated on a tomato plant images dataset that contains 10 distinct classes of diseases and compared to various existing models. According to the research findings, the proposed DMCNN model demonstrates that the proposed DMCNN model outperforms other models regarding accuracy, F1 score, precision, and recall. This study further demonstrates the potential of DL methods for automated classification of disease in agriculture, which can support early disease detection and prevent crop loss.

# 2. LITERATURE REVIEW

In this chapter, the studies in the literature on the detection of leaf diseases are comprehensively reviewed.

## 2.1 Studies on Classification of Leaf Diseases Using Deep Learning Methods

In recent years, there has been a significant increase in interest and dedicated efforts from scientists to automate the classification of plant leaf diseases utilizing image-based methods. However, despite these efforts, these diseases significantly threaten sustainable agriculture, leading to economic losses for farmers and the global economy, Moreover, despite the recent advancements in disease classification methodologies, there remains a significant need for a rigorous process involving a sizable team of experts to continuously monitor these diseases in their initial phases. This is because most current disease classification methods rely solely on visual observation by plant disease experts. However, recent studies have demonstrated that DL models using various approaches can effectively classify leaf diseases. Therefore, developing a robust disease classification system is crucial for efficiently and timely identifying leaf diseases.

Several studies have been carried out focusing on the use of deep learning-based models to classify plant leaf diseases across various crop species. As an example, Brahimi et al. (2017) introduced a study that utilizes AlexNet and GoogLeNet as deep learning models to classify and visualize symptoms of tomato disease, achieving an impressive accuracy of 97.3% to 99.2%. Ahmad et al. (2020) presented a framework to identify plant diseases in another study. The method employed Directional Local Quinary Patterns (DLQP) to compute key points from the input image, followed by training a Support Vector Machine (SVM) classifier to classify plant diseases based on the computed key points. Although this method improves disease recognition accuracy, further performance improvement can be achieved by incorporating shape and color-based information from the input sample. Furthermore, Picon et al. (2018) provided a method for agricultural disease identification in the wild, utilizing deep CNN for mobile capture devices. The system was highly effective in classifying trained datasets and repported an accuracy of 96%.

Aravind et al. (Aravind et al.,2018), proposed a research project that utilizes the AlexNet deep learning model to classify Grape diseases. The results indicated a 1.61% improvement in accuracy compared to AlexNet model, resulting in a classification accuracy rate of 97.62% for three specific diseases. In a similar study, Pantazi et al. (Pantazi et al., 2018) devised a methodology for categorizing plant diseases that incorporates the Local Binary Pattern (LBP) algorithm with the Support Vector Machine (SVM) classifier. This approach offers a significant benefit in that it possesses exceptional generalization capabilities. However, its ability to classify samples accurately can be hampered by noisy data. Nonetheless, the system attained a commendable accuracy rate of 95%.

Liang et al. (2019) conducted a study in which they compared the performance of the original Convolutional Neural Network (CNN) model with CNN combined with Support Vector Machine (SVM) for the identification of rice diseases. The results showed that both methods performed similarly well, with CNN achieving an accuracy of 95.83% and CNN with SVM achieving 95.82%. In their research paper (Agarwal et al., 2020), they proposed a CNN-based architecture specifically designed for the classification of tomato diseases, the study is noteworthy for its computational efficiency and impressive accuracy rate of 91.2%. However, the model faces the challenge of overfitting when dealing with a limited number of classes. To address this issue, some researchers have opted to utilize CNN-based models, which have shown promise in overcoming this limitation. A study by (Richey et al., 2020) presented a novel method for classifying different types of maize crop diseases using a mobile application-based approach. The researchers employed a deep learning model called ResNet50, which exhibited remarkable generalization capabilities and achieved an impressive accuracy rate of 99%. However, this approach's performance is constrained by mobile phone processing power and battery usage needs, rendering it feasible only for specific devices.

In a research publication by (Chouhan et al., 2021), the focus was on exploring innovative techniques for plant disease detection and classification. Their study specifically addressed the challenges related to the segmentation and classification of leaf diseases in Jatropha Curcas L. and Pongamia pinnata L. They proposed a cutting-edge approach that combined machine learning algorithms with advanced image processing techniques, achieving remarkable accuracy in disease segmentation. Additionally, (Chouhan et al., 2021) extended their research to another domain by introducing an automated method for the detection and classification of foliar galls in plant leaves. Their approach utilized an Internet of Things (IoT) based system integrated with fuzzy logic and neural networks to accurately identify and classify foliar galls. Furthermore, in a prior investigation (Chouhan et al., 2018), the authors devised an intelligent system for the identification and classification of bacterial diseases in plant leaves. Their approach employed a hybrid optimization algorithm coupled with a neural network, demonstrating effective computation and yielding valuable insights for disease diagnosis.

In (Sembiring et al., 2020), the authors provided a technique for classifying tomato plant diseases based on images of the leaves. They used a lightweight CNN for this purpose, the study was fast and accurate, with an accuarcy of 97.15%, but it was only tested for the identification of tomato leaf diseases and was not robust to other scenarios. In addition to the aforementioned studies, Barbedo presented a robotic system for plant diseases in (Barbedo, 2018), utilizing the GoogleNet model; this study tackled several factors and parameters affecting the network's performance with multiple crops, achieving an accuracy of 80.75%. In their notable publication (Shijie et al., 2017), Shijie et al. presented a pioneering study that aimed to classify ten tomato diseases using a combination of MSVM with VGG16. The study yielded impressive results, reporting an accuracy of 89% in disease classification. Additionally, the researchers achieved remarkable efficiency by training fine-tuned models within a reduced timeframe. Expanding on their objective to encompass various plant species, further investigations were carried out. Chen et al. introduced their project in (Chen et al., 2019), using a CNN model to recognize tea disease with an accuracy of 90.16%. Furthermore, Liu et al. presented their study in (Liu et al., 2017) by employing a CNN to determine apple diseases with an accuracy of 97.62%.

This section successfully reviewed, and reported the performance of various deep-learning approaches for plant disease classification. Table 2.1 presents an analysis of existing technique used for classifying plant leaf disease.

Table 2.1 A summary of existing plant leaf disease classification techniques

| Reference | Field | Plant | Deep learning method | Accuaracy | Approach |
|---|---|---|---|---|---|
| (Brahimi et al., 2017) | Diseases Classification | Tomato | AlexNet, GoogLeNet | 97.3–99% | Transfer learning |
| (Ahmad et al., 2020) | | Multiple | The DLQP approach with the SVM classifier | 96.53% | Transfer learning |
| (Picon et al., 2019) | | Wheat | ResNet 50 | 96% | Training from scratch |
| (Aravind et al.,2018) | | Grape | AlexNet | 97.62% | Transfer learning |
| (Pantazi et al., 2019) | | Multiple | LBP algorithm with the SVM | 95% | Training from scratch |
| (Liang et al., 2019) | | Rice | CNN | 95.83% | Training from scratch |
| (Agarwal et al., 2020) | | Tomato | CNN | 97.2% | Transfer learning |
| (Richey et al., 2020) | | Maize | ResNet50 | 99% | Transfer learning |
| (Chouhan et al., 2020) | | Jatropha Curcas L. and Pongamia Pinnata L. | Hybrid neural network and seven different deep Learning methods | 99.6% | Training from scratch |
| (Sembiring et al., 2020) | | Tomato | Lightweight CNN | 97.15% | Transfer learning |
| (Barbedo, 2018) | | Multiple | GoogLeNet | 87% | Transfer learning |
| (Shijie et al., 2017) | | Tomato | VGG16 | 89% | Transfer learning |
| (Chen et al., 2019) | | Tea | CNN | 90.16% | Training from scratch |
| (Liu et al., 2017) | | Apple | CNN | 97.62% | Training from scratch |

Upon reviewing the literature analysis presented in Table 1, it becomes apparent that DL models encompassing diverse concepts have demonstrated remarkable effectiveness across various facets of plant leaf disease classification. These models range from traditional methods of leaf disease recognition to more recent techniques, including convolutional neural networks and transfer learning, and have demonstrated promising results for plant leaf disease classification. Nevertheless, it is imperative to conduct additional research to comprehensively assess the performance of deep learning models in various domains and gain a deeper understanding of their limitations. This will facilitate the improvement of accuracy in leaf disease recognition, particularly in the fields of plant leaf disease identification.

The implementation of deep learning in the detection of plant leaf diseases has the potential to revolutionize the current diagnosis of plant leaf diseases by providing more accurate and faster results, which would allow for better preventive measures to be taken, thus reducing the losses caused by leaf diseases in agriculture and the environment. Therefore, it will be interesting to investigate deep learning methods for detecting plant leaf disease. In the upcoming section, we will delve into the studies documented in the literature that focus on plant leaf disease detection utilizing DL techniques.

## 2.2 Studies on Detection of Leaf Diseases Using Deep Learning Methods

The continuous advancements in deep learning methods have brought about a revolution in the field of image recognition. This progress allows for highly accurate detection and identification of objects in images. As a result, DL approaches have now expanded their use to various agricultural and farming applications following their successful application in other domains. The integration of deep learning has ushered in a new era for agricultural applications, empowering them to proficiently discern and categorize various plant species, accurately identify and combat weed and disease infestations, optimize crop yields through data-driven insights, and open up avenues for countless other possibilities. In addition, the high level of accuracy provided by deep learning has allowed farmers to make more informed decisions with data, leading to better crop yields and higher profit margins while reducing the need for manual labor. Furthermore, deep learning technology has given farmers more control and accuracy over crop production.

Many efforts and scientific studies have recently been introduced and implemented to solve crop disease identification problems to find an optimum solution that can help identify the infected leaf automatically. As a result, numerous approaches based on deep learning techniques have been suggested. For instance, A study for automatic and accurate leaf detection of diseases using deep learning approaches wree presented by (Muhammad et al., 2021). This study suggested using a deep learning model using EfficientNet and 18161 tomato leaf images to classify tomato diseases. They especially compared the performance of two models, U-net and Modified U-net, with 99.95% and 99.12%, respectively. The results showed that the Modified U-net model had better performance for leaf detection than the U-net model and achieved more accurate results for tomato leaf disease detection. For the same plant, In their publication (Liu et al., 2020), Liu et al. presented a study focused on the early detection of tomato gray spot disease. The study employed the MobileNetV2 model as its foundation. A high accuracy rate and quick detection speed characterize the proposed method, making it a viable solution for large-scale tomato field monitoring and demonstrating the potential of using deep learning techniques to detect tomato diseasesThe proposed method demonstrated a commendable detection accuracy of 93.24%.

In a study by Sahu et al. (2021), an approach for detecting, classifying, and visualizing bean leaf diseases was presented. The method discussed in this study was based on two deep learning models, GoogleNet and VGG16, to extract features automatically from images, and experimental results show that GoogleNet outperforms VGG16 with 95.31% accuracy. The authors in this work conclude that using GoogleNet is an efficient method for plant disease detection and visualization because of its high accuracy and quick training time. A study on the impacts of dataset size and interactions on the prediction accuracy of deep learning models was proposed by (Alexandre et al., 2021) research analyzes the ways in which training dataset measures and interactions impact the performance of those prediction models. The models produced successful results because they were trained on simulated datasets without interactions. However, the results become much more promising when the models are trained on datasets with interactions.

In a pioneering study by (Karthik et al., 2019), a novel approach was introduced to detect diseases on tomato leaves using deep residual networks. The researchers utilized the PlantVillage dataset, which consisted of 24,001 validation images and 95,999 training images representing three specific diseases. The experimental results demonstrated the efficacy of the proposed network in leveraging CNN learning features at multiple processing stages, resulting in an impressive overall accuracy of 98%. Similarly, (Fuentes et al., 2022) proposed a deep-learning method for disease identification in tomato plant images captured by cameras with varying resolutions. The approach employed three CNN object detectors and a deep feature extractor, showcasing the effectiveness of the proposed methodology. In addition, data expansion techniques were employed to improve training accuracy and reduce false positives. Extensive testing on a diverse dataset of tomato diseases demonstrated the method's capability to successfully detect and classify nine diseases while effectively handling complex plant surroundings.

In a groundbreaking research conducted by (Ramachandran et al., 2019), a transfer-learning approach was applied to identify diseases types in cassava leaves. The authors went a step further by developing a smartphone-based CNN model for cassava disease identification, which achieved an impressive accuracy of 80.6%. Similarly, (Oyewola et al., 2021) employed a PCNN and DRNN to detect and classify five different cassava plant diseases. The study revealed that DRNN outperformed PCNN with a significant accuracy improvement of 9.25%. In another noteworthy study by (Rangarajan et al., 2020), a pre-trained VGG16 feature extractor and a multiclass SVM were utilized to classify various diseases in eggplant. The study focused on leveraging the power of deep learning techniques to accurately identify and differentiate between different types of diseases affecting eggplant plants. In addition, to estimate the performance of this study, diverse color spaces were used, including RGB, HSV, and YCbCr. This work showed that the combination of VGG16 and multiclass SVM can successfully detect eggplant diseases with an accuracy of up to 94.14%. Although this accuracy is promising, combining the VGG16 feature extractor and multiclass SVM could help detect other plant diseases with similar success.

In an original study conducted by (Sladojevic et al., 2016), a deep-learning architecture utilizing the Caffe DL framework was developed to detect and classify 13 different plant diseases. The CNN model showcased remarkable performance with an impressive accuracy rate of 96.3% in accurately identifying the diseases. In a separate research endeavor, (Geetharamani et al., 2019) presented a nine-layer CNN model specifically designed for the detection of plant diseases. The authors utilized the PlantVillage dataset and incorporated data augmentation techniques to enhance the size of the dataset for experimentation purposes.

Through comprehensive evaluation, they demonstrated that their proposed method outperformed traditional machine-learning approaches in terms of accuracy and computational efficiency. The effectiveness of the nine-layer CNN model in accurately detecting plant diseases was established through their experimental results. Furthermore, (Yang Li et al., 2020) employed a shallow CNN (SCNN) to identify diseases in leaves. The study involved extracting and classifying CNN features using SVM and RF classifiers. The SCNN model exhibited accurate detection results for all three types of diseases, highlighting its efficacy in disease identification across multiple crop species.

(Mohanty et al., 2016) identified 26 plant diseases using AlexNet and GoogleNet CNN architectures. The researchers showed that both architectures achieved good accuracy, with AlexNet surpassing GoogleNet regarding the number of diseases accurately identified. With the help of several CNN architectures, (Ferentinos et al., 2019) could accurately identify 58 distinct plant diseases. They used real-time images to test the CNN architecture as part of their approach. The authors found that the CNN architectures could accurately identify and detect different plant diseases, with an average accuracy rate of 99.49% in AleXNetOWTBn and 99.53% in VGG. In (Adedoja et al., 2019), an accuracy rate of 93.82% was attained while detecting plant leaf diseases utilizing deep CNN architecture based on NASNet. Another deep-learning-based platform is suggested in (Ai et al., 2020) for detecting crop diseases and insect pests. The authors employed CNN as the underlying deep-learning engine to find 27 crop illnesses in China's difficult mountainous areas. Chinese farmers can efficiently utilize the system because it was designed with a Java applet as the user interface. A series of studies carried out by the authors revealed recognition accuracy of 86.1%. (Zeng et al., 2020) created a DL-based system for assessing the severity of the citrus disease. For training six DL models, including SqueezeNet-1.1, AlexNet, Inception-v3, DenseNet-169, VGG13, and ResNet-34, a dataset of 5406 pictures of infected citrus leaves was employed. In addition, the data augmentation techniques utilized to improve the amount of the training dataset in addition to the initial training dataset, which helped the models learn more effectively. In order to find which models are better suited to identifying the severity of citrus disease, the scientists analyzed the performance of these six models. Using the Inception-v3 model, the best detection was 92.60% based on the Mask Region model (He et al., 2017).

(Jiang et al.,2019) suggested a technique for detecting apple leaf disease. The R-CNN DL model for object instance segmentation can identify items of interest in an image while producing a segmentation mask for each instance. A CNN model is trained to recognize common apple diseases utilizing a dataset including 2029 images of sick apple leaves. The identification accuracy was estimated to be 78.8%, taking into account the CNN model's very small training dataset. As seen in Table 2.2, which summarizes affiliated work on plant leaf disease label, several studies have been conducted to develop different approaches for the automated detection of leaf diseases employing images of leaves.

Table 2.2 A summary of existing plant leaf disease detection techniques

| **Reference** | **Field** | **Plant** | **Deep learning method** | **Accuaracy** | **Approach** |
|---|---|---|---|---|---|
| (Muhammad et al.,2021) | **Diseases Detection** | Tomato | U-net and Modified U-net | U-net= 99.12% U-net = 99.95% | Transfer Learning |
| (Liu et al.,2020) | | Tomato | MobileNetv2-YOLOv3 | 93.24% | Transfer Learning |
| (Sahu et al., 2021) | | Beans | GoogleNet, VGG16 | 95.31% | Transfer Learning |
| (Karthik et al., 2021) | | Tomato | CNN | 98% | Transfer Learning |

Table 2.2 A summary of existing plant leaf disease detection techniques (continue)

| Reference | Field | Plant | Deep learning method | Accuaracy | Approach |
|---|---|---|---|---|---|
| (Ramachandran et al., 2019) | Diseases Detection | Cassava | Single-shot multi-box (SSD) model And MobileNet detector and classifier | 80.6% accuracy on İmages 70.4% accuracy on video | Transfer Learning |
| (Oyewola et al., 2021) | | Cassava | PCNN, DRNN | DRNN outperformed PCNN by 9.25%. | Transfer Learning |
| (Rangarajan et al., 2021) | | Eggplant | VGG16 | 94.14%. | Transfer Learning |
| (Sladojevic et al., 2016) | | Multiple | Finetuned CNN architecture | 96.3% | Training from scratch |
| (Geetharamani et al., 2019) | | Multiple | Nine-layer deep CNN | 96.46% | Training from scratch |
| (Li et al., 2020) | | Maize, apple, and grape | CNN with SVM and RF | 94% | Transfer Learning |
| (Mohanty et al., 2016) | | Multiple | AlexNet and GoogleNet | AlexNeT= 99.27% GoogleNet=99.34% | Transfer Learning |
| (Ferentinos et al., 2019) | | Multiple | AlexNetOWTBn and VGG | AleXNetOWTBn=99.49% VGG = 99.53% | Transfer Learning |
| (Adedoja et al., 2019) | | Multiple | NASNet-based deep CNN | 93.82% | Transfer Learning |
| (Ai et al., 2020) | | Multiple | CNN | 86.1%. | Transfer Learning |
| (Zeng et al., 2020) | | Multiple | DenseNet-169, AlexNet, ResNet-34, and VGG13 | The best detection was 92.60%. | Transfer Learning |
| (Jiaang et al.,2019) | | Apple | CNN | 78.8% | Transfer Learning |

According to the literature, deep learning models using various approaches efficiently detected diseases on plant leaf images. Nevertheless, studies of a few essential techniques, particularly those developed from scratch, were not so common; a large number of studies used only a pre-trained model-based transfer learning approach; in addition, they face the same problem with almost the same techniques, which necessitates working on multiple projects with multiple methods. Therefore, To fully realize the potential of deep learning models for detecting different diseases on plant leaves, Moreover, the researchers need to work on multiple projects and experiments, employing various techniques, including those based on transfer learning and those that start from scratch. Because this suggests that deep learning models have much potential for detecting diseases on plant leaves, but more research and experimentation are needed to explore their full capabilities. We will work on this in this study to further benefit from deep learning. Furthermore, by studying the underlying deep learning techniques and considering how they can be implemented to help improve the accuracy of plant disease recognition, this work seeks to provide insights that can guide future research.

# 3. THESIS PURPOSE AND CONTRIBUTION

In the realm of agriculture, timely detection of leaf diseases holds paramount importance in preventing crop losses and ensuring food security. However, traditional methods of leaf disease detection, such as visual inspection and laboratory analysis, can be laborious, subjective, and expensive. Moreover, these methods could not suitable for detecting diseases in large-scale farming operations.

To overcome these challenges, deep learning approaches have been used to create computer vision algorithms that can automatically detect leaf diseases from images of leaves. In addition, these algorithms can process large extensive data swiftly and accurately and learn to recognize patterns in the image’s indicative of specific diseases. Overall, the integration of deep learning in leaf disease detection holds great potential for improving crop management and reducing crop losses. However, the development of more accurate and robust algorithms and the creation of more comprehensive datasets will be critical for further advancing this field.

This thesis aims to leverage deep learning techniques to create novel approaches and algorithms that can accurately and swiftly detect plant leaf diseases in real-time. We have presented our own work titled "A Novel Convolution Neural Network-Based Approach for Plant Disease detection" as an example of this approach. Our primary goal is to enable the detection of leaf diseases using real-time images, thereby improving the speed and ease of detection. Additionally, the research proposed in this thesis intends to assess and contrast the outcomes of this method with other prevailing techniques employed for the detection of plant diseases. By doing this, the accuracy of disease detection will be improved, making it easier for farmers and other agricultural stakeholders to take appropriate action when a disease is detected. As a result, the proposed research in this thesis will make an invaluable contribution to the realm of plant disease detection, which can lead to improved crop yield and greater agricultural productivity.

In this thesis, we thoroughly analyze the challenges faced by current methods of leaf disease detection and explore how DL techniques can overcome these challenges and improve disease detection accuracy. Therefore, we extensively explored and analyzed these aspects in our comprehensive study entitled "Convolutional Neural Networks in Detection of Plant Leaf Diseases: A Review." We also conduct an extensive investigation of existing methods and propose novel alternative approaches using deep learning methods and leaf image datasets.

Furthermore, we introduce a new approach called "Plant Leaf Diseases Detection Using MobileNet Model". The objective of this work is to develop an automated model capable of accurately classifying and identifying different types of diseases. To achieve this, the model utilizes MobileNet, leaf images, and an optimized network architecture. The anticipated outcome of this proposed method is to provide valuable assistance to farmers in effectively addressing their challenges. We also discussed the complete creation and implementation process for this new convolutional neural network (CNN) model. Then, by adjusting parameters, we fine-tuned our model to ensure it performed similarly well on the dataset and was in the optimal solution. In addition, we evaluated the efficacy of our model and compared the outcomes with those of some state-of-the-art architectures to evaluate the feasibility of determining the architecture of the most satisfactory performing model and the most suitable training settings for these problems.

The current literature on plant disease detection and classification has primarily focused on one single-channel and same-resolution images, which may not fully capture the comprehensive information needed for precise disease detection. To solve this limitation, this thesis introduces a new approach called "Deep Multi-Scale Convolutional Neural Networks for Automated Classification of Multi-class Leaf Diseases (DMCNN)". The proposed approach leverages multiple channels of information to enhance the efficiency of disease detection. The DMCNN architecture consists of parallel streams of convolutional neural networks (CNNs) operating at different scales, which are then merged at the final stage to produce a unified output.

While there are numerous publicly available datasets on plant leaf diseases in the literature, there is a scarcity of datasets based on natural images. To address this gap, we have collected and prepared a novel dataset comprising common plant disease types that were previously absent in the literature. This dataset is expected to facilitate the research of scholars working in this field, enabling them to develop novel approaches for the diagnosis and treatment of plant diseases, validate existing approaches, and compare their accuracy and performance.

Furthermore, the dataset was meticulously curated by capturing pictures of affected plants in diverse locations, under varying light and environmental conditions. This approach ensures a broader representation of numerous types of plant diseases, making the dataset suitable for developing algorithms that accurately detect and identify plant diseases. The availability of these datasets will not only benefit researchers and farmers but also contribute to global efforts toward sustainable agriculture. It holds the potential to revolutionize the diagnosis and treatment of plant diseases, leading to improved crop yields, reduced pesticide usage, and ultimately, a more sustainable food production system. Moreover, the accessibility of this dataset will foster collaboration among researchers, promoting more comprehensive studies on plant diseases and their impact on agriculture. This collaborative environment will facilitate the development of pioneering solutions to overcome the challenges encountered by farmers, particularly in developing countries where food security is a significant concern.

In addition, this thesis attempted to explore the influence of of datasets on the implementation of deep learning models. To achieve this, we conducted an analysis using a single model trained on several datasets of leaf images. Through a comparative study titled "Impact of Datasets on The Effectiveness of Mobilenet for Leaf Disease Detection," we sought to understand how different datasets influenced the performance of the model. This study took into consideration Multiple factors, including differences in accuracy and computation time across datasets, aiming to identify the factors contributing to these variations. Additionally, the comparative study evaluated the model's performance on datasets of varying sizes, quality, and complexity. Factors such as category distribution in the dataset, labeling quality, and image diversity were also considered.

The scope of this work will be expanded to encompass diverse deep-learning models, and their efficacy will be assessed for the accurate detection of other critical pathologies. The practical senses of this research can be readily extended to address detection challenges pertaining to other plant leaf diseases, as the insights gained from this work can serve as inspiration for similar visual recognition tasks. The outcomes obtained through this research have the potential to contribute to advancements in the field of plant health management and offer valuable insights for tackling similar detection problems in agriculture and beyond.

# 4. PLANT LEAF DISEASES ANALYSIS AND ITS SYMPTOMS AND SIGNS

Environmental change has made it more challenging to detect new plant leaf diseases. Therefore, identification of the variables affecting the emergence and increased incidence of these diseases is necessary. Furthermore, we list emerging symptoms and signs of plant leaf diseases and explain how they differ according to the environment. We also presnetd some traditional methods used to detect disease in plant. Here are some common bacterial, viral, and fungal plant leaf disease symptoms and signs.

## 4.1 Bacterial Disease Symptoms and Signs

The symptoms of bacterial plant infection are similar to those of fungal plant disease. The disease is distinguished by minor, light green spots that appear water-splattered. The injuries increase and appear as dry, dead spots (Mattihalli et al., 2018). For example, bacterial leaf spots have darker or darker water-soaked areas on the foliage, occasionally with a light radiance, and are typically indistinguishable in the estimate. When the spots are dry, they have a spotted appearance. Eventually, the affected spots become large, grayish-brown areas on the leaves, with a center that may appear yellow or brown and a darker border around it. These spots can rapidly spread across the entire leaf surface, resulting in complete defoliation and potential damage to different parts of the leaf, such as young stems, fruits, and flowers.

Bacterial disease signs can be challenging to detect, including bacterial ooze, water-soaked lesions, and bacteria streaming in water from a cut stem. Furthermore, the affected leaf areas may initially appear as small, water-soaked spots, eventually expanding and taking on a more circular shape with a yellow-brown center and darker borders. The presence of these signs can be alarming. However, it is important to mention that not all of these symptoms are necessarily present in every bacterial disease. Thus, correctly diagnosing the symptoms is essential to choose an appropriate treatment. In Table 4.1, we present some typical diseases of several familiar plants. We chose rice, cucumber, tomato, and maize to describe the significant bacterial disease in these critical plants to provide an example of plants' important role in sustaining life.

Table 4.1 A glimpse of common bacterial diseases of plants

| Plant | Major Types of Disease<br>Bacterial | Reference |
|---|---|---|
| Rice | Bacterial leaf streak, Bacterial leaf blight | (Chen et al., 2021; Shrivastava et al., 2019) |
| Cucumber | Target spots, Angular spot, brown spot | (Kianat et al., 2021; Zhang et al., 2019) |
| Tomato | Bacterial wilt, canker, soft rot | (Abbas et al., 2021; Ferentinos et al., 2018) |
| Maize | Bacterial leaf streak, Bacterial stalk rot | (Sun et al.,2021; Yu et al., 2014) |

The diseases highlighted in Table 4.1 represent the wide range of bacterial diseases that affect plants and the potentially devastating consequences they can have on food production and crop yields. Consequently, understanding the nature and dynamics of bacterial diseases in plants is essential for successfully managing and cultivating crops, especially the essential staples such as rice, cucumber, tomato, and maize, which are critical to human nutrition and food security. Furthermore, understanding these diseases can help inform agricultural research and development initiatives, which have been identified as critical strategies to enhance food production and improve crop yields sustainably. As a result, we must also comprehend the symptoms of other diseases, such as fungal and viral infections, to effectively prevent the spread of crop-related diseases and optimize agricultural productivity.

### 4.2 Fungal Disease Symptoms and Signs

Fungi are among the most common plant pathogens. Pathogenic fungi employ a variety of strategies to colonize crops and cause illness. Some fungi kill their hosts and start eating dead tissue, while others colonize living tissue (Mattihalli et al., 2018). Some fungi can also suppress the immune system of their hosts, making them vulnerable to attack by other organisms. Additionally, some fungi form symbiotic relationships with plants to feed off their resources, allowing the fungus to grow and spread more quickly.

There are several common fungal plant disease symptoms, such as birds-eye spots on berries, damping off of seedlings, phytophthora, leaf spots on leaves. Leaf spot, also known as septoria brown spot, and chlorosis which causes yellowing of the leaves. These symptoms can vary in severity depending on the type of fungus, the plant species infected, and environmental factors such as weather. However, it is essential to note that the existence of any of these symptoms does not guarantee the presence of fungal plant disease, and a plant can exhibit symptoms similar to those caused by fungal diseases even when the cause is something else, such as nutrient deficiency or insect infestation. As a result, it is critical to ensure that any fungal plant disease diagnosis is correct to employ the most appropriate management strategy. When diagnosing a fungal plant disease, it is essential also to consider all possible signs present, such as stem rust (wheat stem rust), leaf rust, sclerotinia (white mold), and powdery mildew. Therefore, it is essential to look for multiple indicators of fungal disease to make a correct and accurate diagnosis. Table 4.2 displays the prevalent fungal diseases affecting four commonly cultivated plants.

Table 4.2 A glimpse of common fungal diseases of plants

| | **Major Types of Disease** | |
|---|---|---|
| **Plant** | **Fungal** | **Reference** |
| Rice | False smut, Rice stripe blight, rice blast | (Shrivastava et al., 2019; Chen et al., 2021) |
| Cucumber | Powdery mildew, downy mildew gray mold, anthracnose, black spot | (Kianat et al., 2021; Zhang et al., 2019) |
| Tomato | Late blight, early blight, leaf mold | (Abbas et al., 2021; Ferentinos et al., 2018) |
| Maize | Rust disease, Leaf spot disease, gray spot | (Sun et al., 2021; Yu et al. 2014) |

## 4.3 Viral Disease Symptoms and Signs

The most challenging plant leaf infections to assess are those caused by infections. Infections have no visible symptoms that can be observed quickly and are frequently confused with pesticide damage and nutritional deficiencies. With yellow or green stripes or spots on the foliage, aphids, leafhoppers, whiteflies, and creepy cucumber crawlies are common carriers of this disease. Leaves can be twisted or wrinkled, which could hinder development. In order to properly diagnose and treat these infections, it is essential to inspect the plant for any visible symptoms of the disease, such as crinkled leaves, mosaic leaf patterns, stunting, and yellowed leaves, and then to consider the environmental conditions in which the plant is being grown, such as temperature, humidity, and sunlight, as well as the type of plant itself (Mattihalli et al., 2018). Also, the most critical factor in treating viral diseases is prevention, as it is much easier to prevent the spread of a virus than cure an infected plant. Preventive measures, such as rotating crops and disposing of infected plant materials, are critical in reducing the spread of viruses. These factors can help identify viral diseases and determine an appropriate treatment. Table 4.3 shows several popular plants' common viral diseases.

Table 4.3 A glimpse of common Viral diseases of plants

| Plant | Major Types of Disease<br>Viral | Reference |
|---|---|---|
| Rice | Rice black-streaked dwarf virus, Rice leaf smut | (Shrivastava et al., 2019; Chen et al., 2021) |
| Cucumber | Spot virus, mosaic virus | (Kianat et al., 2021; Zhang et al., 2019) |
| Tomato | Tomato yellow leaf curl virus | (Abbas et al., 2021; Ferentinos et al., 2018) |
| Maize | Crimson leaf disease, rough dwarf disease | (Sun et al., 2021; Yu et al. 2014) |

From the table presented above, it is apparent that there is a significant overlap between fungal, bacterial, and viral plant disease symptoms. This overlap makes it challenging to determine the type of disease a plant may have based on its symptoms alone, requiring careful examination and diagnostic tests to diagnose a plant's disease accurately; this is a challenging task when dealing with a plant with multiple types of disease, as it can be challenging to determine which of the symptoms is the primary cause, as well as any secondary causes. Therefore, a thorough examination and diagnostic tests must be carried out to determine the type of disease, its primary and secondary causes, and how best to treat the plant.

## 4.4 Relationship Between Crop Types and Pathogen Groups

The relationship between crop types and pathogen groups can be complex and dependent on various factors such as the region, climate, and agricultural practices. However, there are some general associations between certain crops and pathogen groups. For example, rice is commonly associated with bacterial blight caused by Xanthomonas oryzae, while wheat is often susceptible to fungal diseases such as Fusarium head blight caused by Fusarium graminearum. Similarly, soybean is prone to diseases caused by both fungal pathogens such as Phomopsis and bacterial pathogens such as Pseudomonas syringae, while corn is commonly affected by fungal pathogens such as Aspergillus flavus that produce mycotoxins. According to a study by (Del Ponte et al., 2009), crop type can significantly impact pathogen group prevalence and diversity. For example, their research found that soybean crops had a higher prevalence of fungal pathogens than other crops. Additionally, they observed that maize crops were more likely to be infected by viral pathogens. Similarly, a study by (Osdaghi et al., 2017) investigated the relationship between crop type and pathogen groups in Iranian almond orchards. They found that fungal pathogens were the most prevalent group, and their occurrence was significantly influenced by orchard location and the type of almond cultivar. These studies suggest that the relationship between crop types and pathogen groups is complex and context-dependent, influenced by factors such as geographical location, crop management practices, and environmental conditions. Understanding these relationships is essential for developing effective strategies to manage plant diseases and maintain crop productivity.

To provide a clear picture of the relationship between crop types and pathogen groups and how fungal, viral, and bacterial diseases spread and dominate in plants and crops, In Figure 4.1, we show the type of association of crops with groups of infections, as well as the number of diseases studied by their causal agents, suggesting that fungal, viral, and bacterial diseases are more frequent and geographically widespread than other diseases, such as nematodes, algae, abiotic organisms, and Prague, which are less common in crops, these data are extracted from 100 studies related to plant diseases (Additional data will be discussed in the data extraction section).

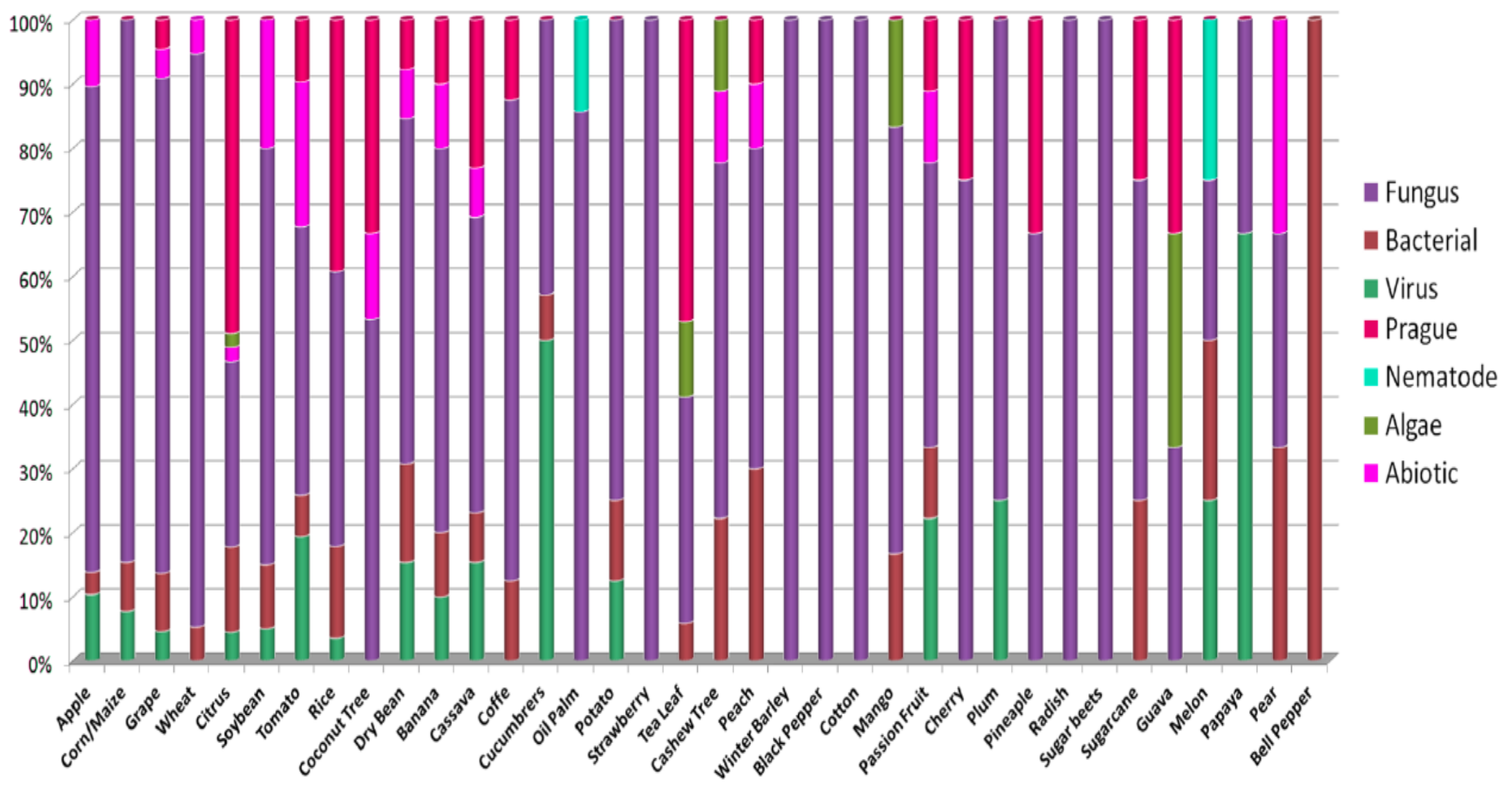

Figure 4.1 Relationship between crop types and pathogen groups

This suggests that for crops in particular, fungal, viral, and bacterial diseases are far more prevalent than those caused by other agents, and this is reflected in the number of studies conducted on each type of disease, highlighting the importance of being vigilant in monitoring crop health, and implementing disease-prevention strategies to protect against fungal, viral and bacterial diseases, in order to reduce the risk of crop loss and associated economic damage. Therefore, it is important to understand the specific behavior of each type of disease to minimize the impact on crop yields and to develop effective and sustainable disease management strategies that can be implemented promptly. In conclusion, the data presented in Fig 11 highlights the importance of understanding the behavior of different types of diseases in relation to crops, as well as their geographic prevalence, in order to protect crop health and reduce economic losses, and ensure the successful and sustainable management of agricultural crops in the long term.

In conclusion, any gardener or agriculturist must be familiar with fungal, bacterial, and viral plant disease symptoms. They should also be familiar with new technologies, such as deep learning, that can help accurately diagnose plant diseases, so they can take the necessary steps to prevent and treat them.

### 4.5 Traditional Detection Methods for Plant Leaf Disease

Plant leaf diseases significantly threaten agricultural productivity and food security worldwide. Therefore, detecting plant leaf diseases early is essential for effective disease management, as it allows for timely intervention to effectively curb the disease's further propagation. Traditional detection methods have been used for many years and rely on a combination of visual inspection, sample collection, and laboratory analysis.

Visual inspection involves visually examining plants for symptoms and signs of disease (Wolff et al., 2014), such as leaf spots, discoloration, and wilting. This method is relatively straightforward and does not require any specialized equipment. However, it relies heavily on the observer's expertise, and some diseases may not have visible symptoms in their early stages.

Sample collection involves collecting plant tissue or soil samples from the affected area and sending them to a laboratory for analysis (Karami et al., 2014). The laboratory can use various techniques to detect plant pathogens, including microscopy, serological assays, culture-based assays, and nucleic acid-based assays. Microscopy involves examining the pathogen's structure under a microscope, while serological assays detect specific proteins produced by the pathogen. Culture-based assays involve growing the pathogen on a suitable medium, while nucleic acid-based assays detect the pathogen's genetic material.

Serological assays involve testing plant tissue for the presence of specific antibodies or antigens associated with plant pathogens (Agrios., et al 2005). These assays can be done using ELISA (enzyme-linked immunosorbent assay) or other techniques. Serological assays are highly specific and sensitive and can be used to detect low levels of pathogens in plants.

Culture-based assays involve growing plant pathogens in a laboratory using specialized media (Mngxing et al., 2018). This method can be used to identify the specific pathogen causing the disease. Culture-based assays are highly specific and can provide accurate results, but they are time-consuming and require specialized equipment and expertise.

Traditional detection methods for plant leaf diseases have been widely used for many years and have proven to be effective. However, they do have some limitations, which include :

1. **Time-consuming:** Traditional detection methods often require sample collection and laboratory analysis, which can be time-consuming and delay disease identification.
2. **Subjective**: Visual inspection relies heavily on the observer's expertise and can be subjective. Different observers may interpret symptoms and signs differently, leading to inconsistencies in disease identification.
3. **Limited detection capabilities:** Some pathogens may be difficult to detect using traditional methods, and some diseases may not have visible symptoms in their early stages. This can result in delayed or inaccurate diagnoses.
4. **Limited specificity**: Traditional detection methods may be unable to differentiate between various strains or species of pathogens, leading to misdiagnosis or underestimation of disease severity.
5. **Cost**: Sample collection and laboratory analysis can be costly, particularly when multiple samples need to be analyzed.

6. **Expertise required:** Traditional detection methods often require specialized knowledge and expertise to interpret the results accurately. This can limit their usefulness for non-experts or those with limited training.
7. **Environmental impact**: Some traditional detection methods, such as chemical analysis or pesticide use, can negatively impact the environment and human health.
8. **Inability to provide real-time monitoring:** Traditional detection methods usually involve manual inspection of plant samples, which may not provide real-time monitoring. This can limit the ability to detect diseases early and prevent their spread.
9. **Invasive**: Some traditional detection methods, such as tissue culture and DNA sequencing, involve invasive techniques that can damage or destroy the plant. This can be problematic for plants that are rare, endangered, or have significant economic value.
10. **Limited accessibility**: Traditional detection methods may not be accessible to all farmers, especially those in remote or underprivileged areas. This can limit their ability to diagnose and treat plant diseases effectively.

Despite these limitations, traditional detection methods remain an essential tool for identifying plant leaf diseases. They provide a cost-effective and reliable way to detect and diagnose plant diseases and allow for timely intervention, enabling effective control measures to limit the continued spread of leaf diseases. With the emergence of new technologies, traditional detection methods are being complemented by newer methods that are faster, more accurate, and less subjective, providing a more comprehensive approach to disease detection and management.

In the following section, we will provide detailed information about deep learning and its essential role in plant leaf disease detection so that one can better understand how this technology is revolutionizing the way we detect, diagnose, and treat plant diseases.

# 5. DEEP LEARNING

Deep learning is a prominent discipline within the field of machine learning that revolves around creating and optimizing artificial neural networks. These networks are designed to emulate the structure and functionality of the human brain. The term of Deep learning (DL) was introduced as a new field of research within machine learning in 2006. It was first referred to as "hierarchical learning" (Mosavi et., 2017), and then Hinton et al. (Hinton et al., 2006) introduced "deep learning," that that focuses on the concept of "artificial neural networks" (ANN). After that, deep learning became popular, resulting in a renaissance in neural network research.

DL utilizes artificial neural networks to create a computer model which mimics the function of biological neural networks; therefore, deep learning emerged historically from artificial neural network research. As a result, it is also referred to as "new generation neural networks" at times. Furthermore, deep learning structures are based on human-like thinking and decision-making processes, allowing them to understand the world around us and similarly make decisions for humans, such as in voice recognition, natural language processing, computer vision, and robotics (Bengio et al., 2009).

DL technology has become an incredibly popular and significant subject within the fields of artificial intelligence, data science, and analytics because of its capacity for learning from the provided data. Furthermore, as it can produce significant results in various classification and regression issues and datasets, several corporations, like Google, Microsoft, Nokia, etc., actively analyze it (Karhunen et al., 2015). In addition, DL technology is being used in a diverse set of applications ranging from natural language processing to image recognition to autonomous driving and medical diagnosis, providing powerful insights from massive datasets, optimizing complex decision-making, and even generating predictive analytics. With its remarkable capacity to process large quantities of data efficiently, DL technology is becoming increasingly popular and gaining a tremendous amount of attention from researchers, technology providers, and even investors.

Deep learning can be defined as an artificial neural network, with the term "deep" usually referring to the number of layers within the network. Consequently, a neural network with multiple layers is commonly referred to as a "deep" neural network. When artificial neural networks first appeared in the 1970s, they had only a few layers. However, the number of layers is becoming increasingly common today (Tesauro., 1992), resulting in a deeper, more intricate, and more advanced artificial neural network with a greater capacity for precise data identification and classification. Besides the number of layers, deep learning can also be improved by training more sophisticated algorithms, introducing data from various sources, and taking advantage of more powerful computers, making it an invaluable tool for various tasks. This is due to the tremendous advances in artificial intelligence over the last few decades, which have enabled us to build and use more advanced neural networks than ever before, with the ability to quickly identify and classify data that was previously unidentifiable. These advances in artificial intelligence have allowed us to push the boundaries of deep learning, providing a powerful tool for businesses, researchers, and everyday users alike; they have also allowed us to develop more efficient, reliable, and accurate models that may be employed to solve a wide variety of issues.

Three types of DL techniques exist: unsupervised, supervised, and semi-supervised, each of which can be used to solve various problems. Unsupervised learning is used to detect patterns in data and make predictions without the need for a label. In contrast, supervised learning uses labeled data to learn from. Semi-supervised integrates labeled and unlabeled data, allowing machines to learn more complex tasks than just labeled or unlabeled data alone. For example, unsupervised learning is used for clustering, such as identifying groups in data, and density estimation, which can be used to identify anomalies or outliers in the data. Semi-supervised learning is helpful for tasks such as classification and regression, as it allows machines to learn more complex patterns by using both labeled and unlabeled data. Supervised learning is the widely used method, as it can solve simple and complex tasks such as object recognition, image classification, and natural language processing. These learning techniques can be utilized in various applications, depending on the task, such as autonomous driving, medical imaging, natural language processing, and more.

In the literature, deep learning and image processing-based studies for detecting plant diseases are mainly based on classifying attributes extracted from features, including texture, color, and shape of plant leaves. However, both the determination of the features and the methods of extracting the elements from the images affect the success of the results. The fact that classification accuracy depends on feature extraction methods is one of the important drawbacks of these approaches. Recently, the determination and extraction of features have been largely solved with deep CNNs and used vastly in classification-based identification studies in different fields. The popularisation of CNNs in image-based recognition started with the success of the results and computational performance of the AlexNet CNN model, which operates on graphics cards. Convolutional neural networks models such as Resnet, VGG, Googlenet, and Mobilnet, developed for image-based classification, are widely used in many fields, from health to agriculture.

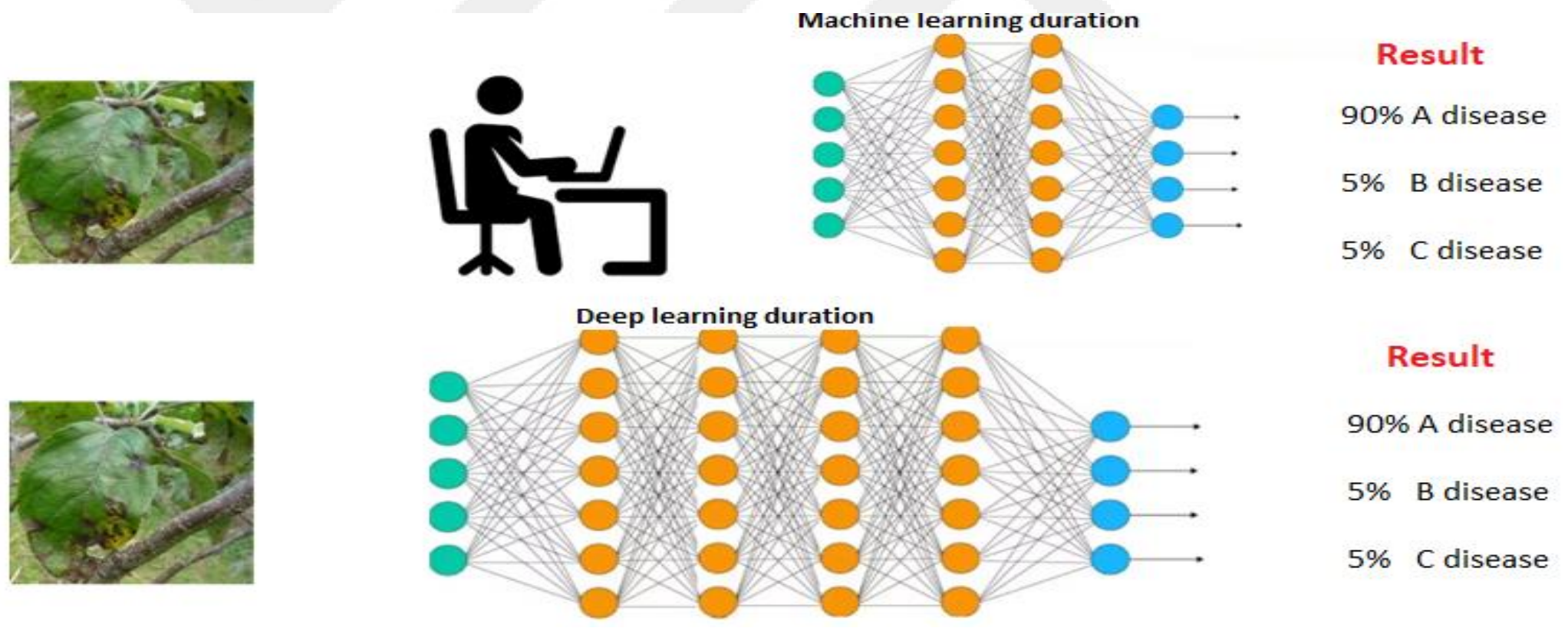

Figure 5.1 One of the main advantages of deep learning methods over machine learning in image-based learning is that the determination of image features is done by the learning network

Extensive evidence has demonstrated that deep learning networks exhibit superior generalization capabilities when it comes to images, especially on unseen data. This advantage arises from their ability to extract intricate low-level features from images, a task that can be challenging for traditional machine learning algorithms. For instance, a deep learning network may be able to identify objects in an image, even if they are inundated with other elements. This ability to extract features makes deep learning networks a powerful tool for object recognition and detection. Moreover, deep learning networks can learn abstract representations of images, allowing them to perform tasks such as classification with much greater accuracy than traditional machine learning algorithms, even if the images given are incomplete or distorted. Additionally, the ability to utilize convolutional layers in a deep learning network gives it an extra edge over traditional machine learning algorithms, as these layers enable the network to better process information from various levels of abstraction, allowing it to recognize objects or features in an image regardless of the size or position. Furthermore, deep learning networks can operate on large datasets with higher efficiency, enabling them to quickly identify patterns or features that may otherwise be difficult to detect, allowing them to rapidly adapt to changes in the environment and quickly identify new objects or features, making deep learning networks a powerful tool for both object recognition and detection.

In conclusion, DL is a form of artificial intelligence based on the neural network structure and processing of complex data, making it an effective tool for analyzing and analyzing data similarly to humans. Therefore, deep learning is a rapidly growing field, and it has become increasingly accessible to researchers and practitioners due to its scalability and efficiency compared to traditional machine learning methods.

## 5.1 Artificial Neural Networks

An artificial neural network is an essential computational model composed of various processing elements that receive inputs and generate outputs per predetermined activation functions. ANN is a subset of machine learning that has grown in importance in recent research and development. McCulloch and Pitts first presented the concept of "artificial neural networks" (ANNs) in 1943 (McCulloch et al., 1943). However, it was not until 1986, when Rumelhart et al. created the backpropagation algorithm, that ANNs became popular (Moller et al., 1993; Rumelhart et al., 1986). Since then, ANNs have been used for a wide range of tasks. From solving complex problems in robotics and computer vision to more general classification tasks such as handwriting recognition and language processing, ANNs have proven to be practical tools for researchers, providing a way to tackle challenging problems with relatively simple techniques.

Artificial neural networks (ANNs) constitute one of the most extensively used artificial intelligence (AI) techniques (Medina et al., 2004). They have applications in various fields, including pattern recognition, natural language processing, control systems, and data mining. For example, artificial neural networks are used in classification, clustering, pattern matching, function approximation, prediction, control, optimization, and search operations. ANNs are composed of nodes, or neurons, connected by weighted edges that can be adjusted according to the output produced by the network for a given input. The idea behind ANNs is to mimic the human brain by modeling its structure and behavior and making it possible to solve problems that would be too difficult for a single algorithm, such as those that involve a large amount of data or complex relationships between different variables.

Various methods can train artificial neural networks, such as backpropagation, learning vector quantization, genetic algorithms, or fuzzy logic. Through these methods, ANNs can learn by recognizing patterns in data and adjusting their internal parameters to optimize their performance, allowing them to complete a wide variety of tasks, ranging from prediction to classification and generalization. ANNs can also be used to model complex non-linear relationships, making them especially suitable for applications where data is ever-changing and predictions must be made quickly and accurately. In addition to learning and adapting quickly, ANNs are much more reliable than traditional methods when making predictions. Furthermore, some steps must be taken in order to use ANN, as shown in Figure 5.2 It is important to carefully follow these steps to ensure accurate and reliable results from ANN. It is also crucial to continuously monitor and adjust the network as needed to maintain its accuracy.

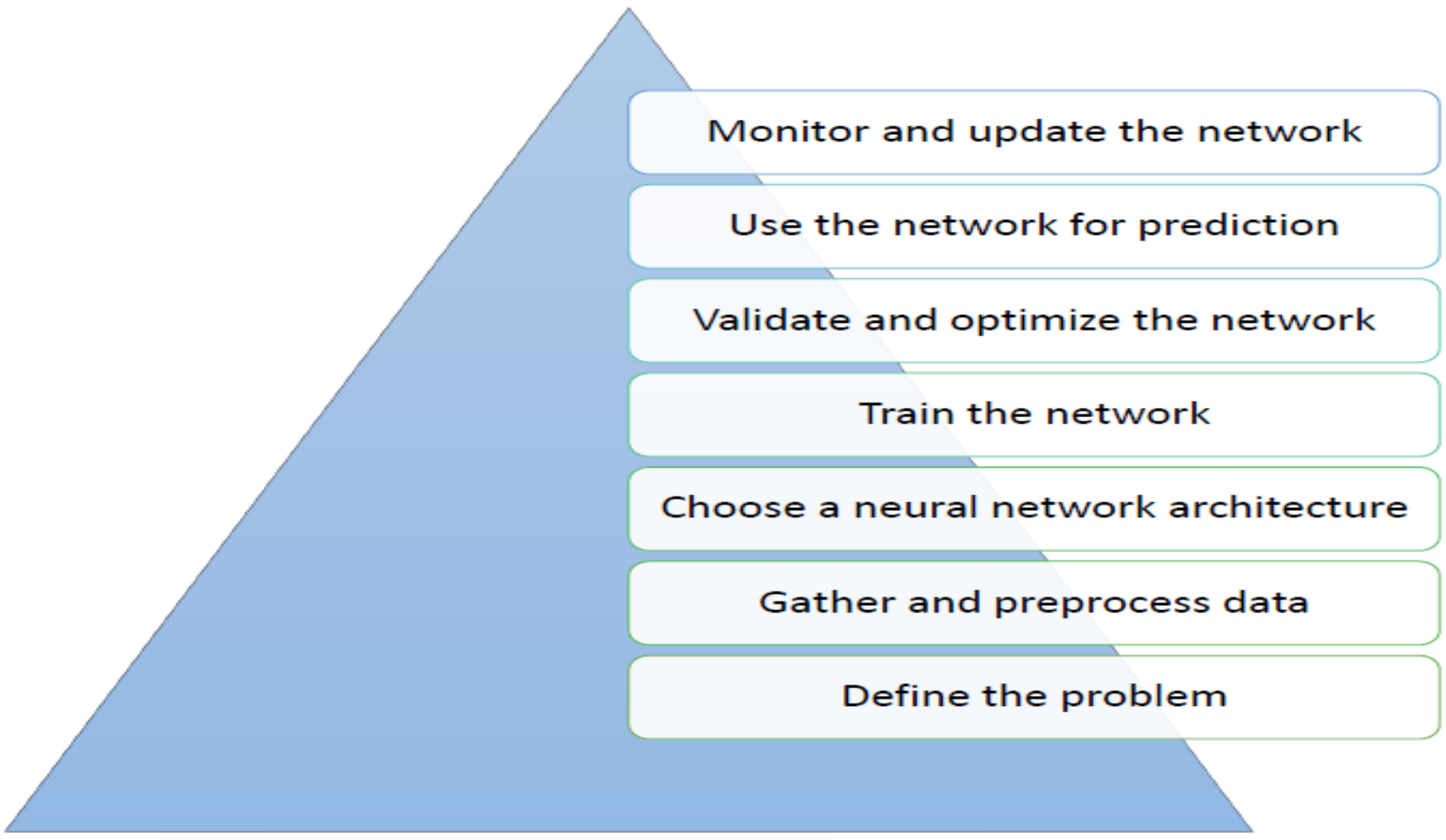

Figure 5.2 Key steps to use an artificial neural network

An ANN is built from a network of connected units known as "artificial neurons," loosely modeled after the neurons or units in the biological brain. These neurons are organized into multiple layers and are connected by weighted edges that store the data used in the calculations of the ANN. Each layer of neurons has its own set of weights and biases, and the ANN learns by changing these weights and biases as it is exposed to different inputs. The ANN employs an optimization algorithm like gradient descent to learn and discover the most suitable weights and biases. This iterative process aims to minimize the disparity between predicted and actual outputs, allowing the ANN to effectively predict outputs based on the given inputs.

The artificial neural network is made up of five layers: the input layer, the output layer, the hidden layers, weights, and biases, and the activation function, which are all essential components of an ANN (Medina et al., 2004; Stanford., 2007). The input layer accepts inputs from the outside world, which are then passed on to the hidden layers of neurons, where the weights and biases are used to determine the output of the ANN. The output layer then passes the result to the activation function, which decides whether a neuron should be activated or not and what type of output it should produce. Finally, the activation function produces the output from the neural network, which can be used to make decisions or predictions, Figure 5.3 present a basic architecture of ANN.

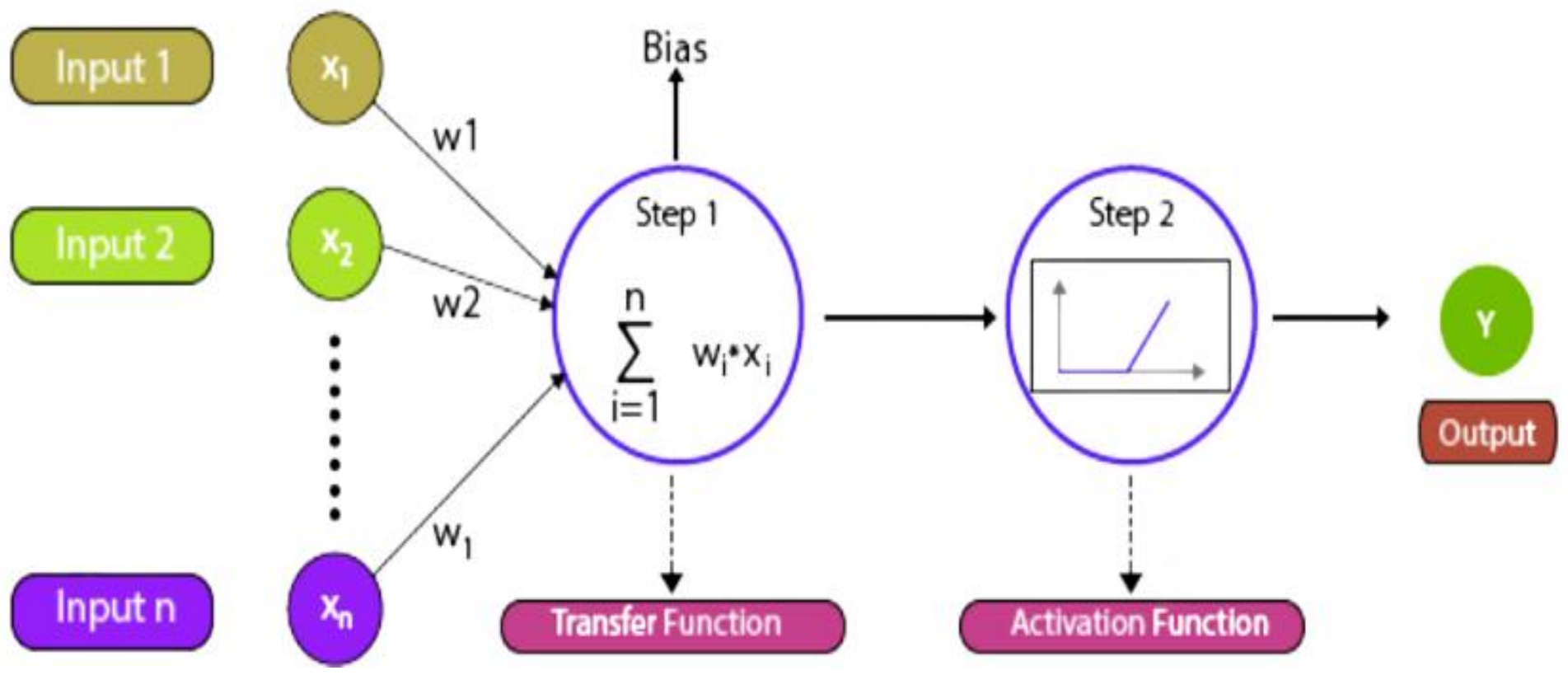

Figure 5.3 Artificial Neural Networks Architecture

### 5.1.1 History of artificial neural networks

McCulloch and Pitts first presented the concept of "artificial neural networks" (ANNs) in 1943 (Meculloch et al., 1943). However, Donald Hebb presented "The Organization of Behavior" in 1949, highlighting that neural pathways are reinforced each time they are used, a concept fundamental to how humans learn. The Hebb rule, developed by Donald Hebb, demonstrated that the number of connections in neural networks is specifically related to learning (Hebb, 1949). Hebb's rule suggested that the strength of synaptic connections could increase or decrease based on learning, meaning that the neurons used together became more connected. In contrast, those used less frequently became less connected.

The first neurocomputers were created in the 1950s, and as computers advanced in the 1950s, it became possible to simulate a hypothetical neural network. Nathanial Rochester from IBM's research laboratories took the first step in this direction.

The Cornell Laboratory of Aeronautics completed the modeling of perceptron work on the IBM 704 computer in 1957. First, Frank Rosenblatt introduced the Perceptron, a single-layered neural network model, in 1958. A perceptron is comprised of input and output layers. The Perceptron's functionalities, however, were limited. Then Rosenblatt wrote an article titled "Perceptron: A Probable Model of Storage and Organization of Information in the Brain" (Rusenlatt, 1958) in 1958 to explain his theories and assumptions regarding perception procedures.

Bernard Widrow and Marcian Hoff of Stanford University created the "ADALINE" and "MADALINE" models in 1959. The names are derived from their use of Multiple ADAptive LINear Elements, which is typical of Stanford's fondness for acronyms.

ADALINE was created to recognize binary patterns to predict the next bit while reading streaming bits from a phone line. MADALINE became the first neural network to be used on a real-world problem, employing an adaptive filter to eliminate echoes on phone lines. Even though the system is as old as air traffic control systems, it continues to be in commercial use (Averkin et al., 2018).

Widrow and Hoff created a learning procedure in 1962 that examines the value before the weight adjusts it (i.e., 0 or 1) using the rule: Weight Change = (Pre-Weight line value) * (Error / (Number of Inputs)). It is based on the idea that, even if one active perceptron has a significant error, the weight values can be adjusted to distribute it across the network or at least to adjacent perceptrons.

In 1969, Minsky and Papert demonstrated that single-layer artificial neural networks were insufficient for the special-or (XOR) operation. This problem causes the ANN studies to be postponed for some time.

The Kohonen algorithm was created in 1974. Unsupervised learning is the basis of Kohonen's algorithm. After a year, In 1975, the first multilayered network, an unsupervised network, was created.

Interest in the field was reignited in 1982. Caltech's John Hopfield published an article for the National Academy of Sciences. His approach was to use bidirectional lines to develop more valuable machines. Previously, neuron connections were just one way. In the same year, Reilly and Cooper used a "Hybrid network" with multiple layers, every layer employing various problem-solving strategies. In 1982, there was also a joint US-Japan conference on Cooperative/Competitive Neural Networks.

The problem in 1986, with multiple layered neural networks in the news, was how to extend the Widrow-Hoff rule to multiple layers. Three different groups of researchers, one of whom included David Rumelhart, a member of Stanford's psychology department, developed similar ideas, which are now known as backpropagation networks because they distribute pattern recognition errors all through the network.

### 5.1.2 General structure of artificial neural networks

As discussed in the previous section, ANNs are computational models derived from the human brain. In other words, it is the mathematical modeling of human brain logic. After going through some processes, the main goal is to provide an output that aligns with our objective. ANNs have hundreds or thousands of artificial neurons, just like the human brain has billions. ANNs are used for a variety of tasks and have different basic architectures:

#### 5.1.2.1 Single-Layer Artificial Neural Networks

A single-layer artificial neural network (SLP) is a category of neural network consisting of a single layer of artificial neurons, each with weights adjusted during training. SLPs, also known as perceptrons, were among the first neural network architectures developed. The Perceptron is the fundamental component of ANNs. Furthermore, The Perceptron consists of five components:

- **Inputs**: Our independent variables (x) are as follows.
- **Weight**: Weight parameters (w) govern the strength of the link between inputs and neurons. It also represents the effect of an independent variable on the outcome.
- **Bias value(b):** This constant value allows the output value to be controlled. Furthermore, when all inputs are zero, the process can still proceed.

- **Activation Functions**: The activation function (f) describes the neuron's output under specific conditions.
- **Output**: The dependent variable (y) is the desired outcome. Perceptrons divide the result into two classes, 1 and 0.

We can show the process by formulating it as *y=f(x×w+b)*. Figure 5.4 depicts an example of ANN structure.

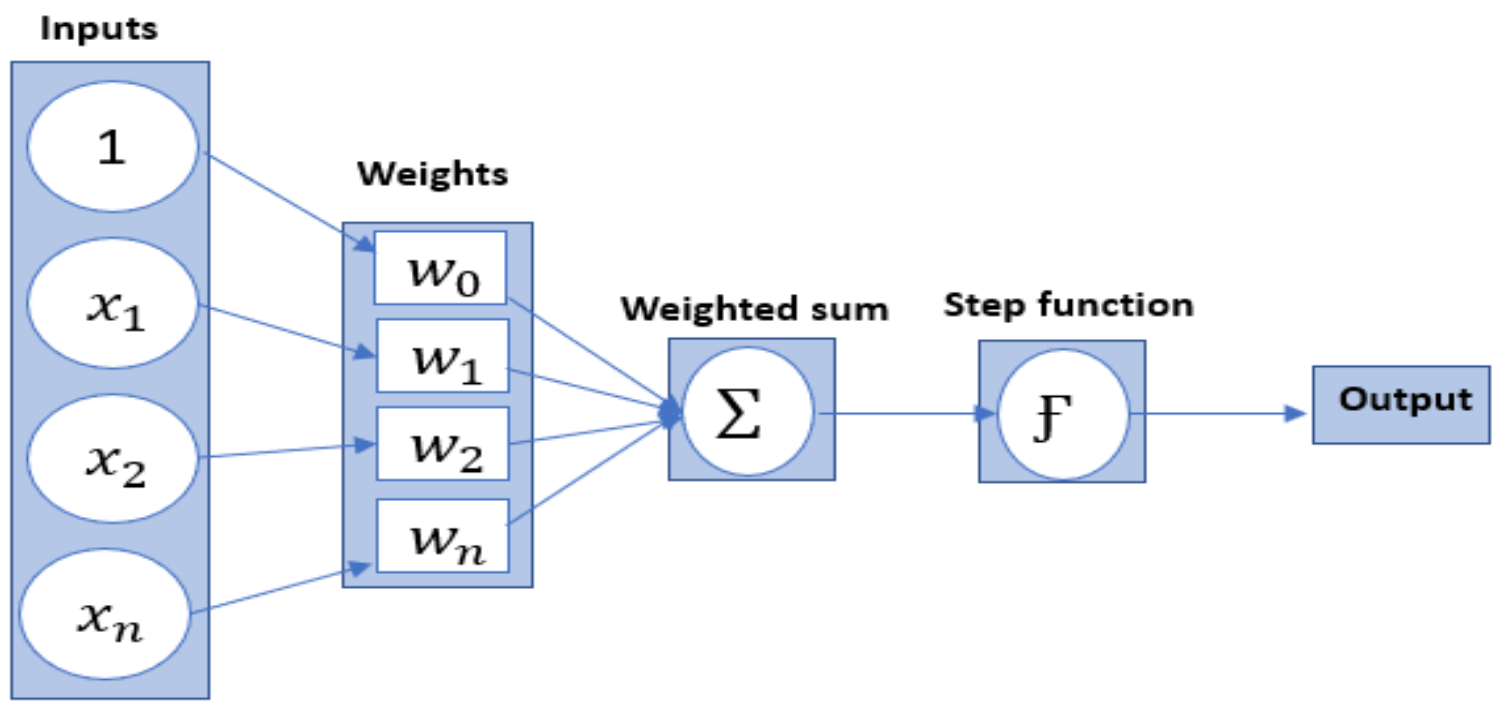

Figure 5.4 Example of single-Layer Artificial Neural Networks architectures

The weighted sum is calculated by multiplying the weights and inputs and then adding them, as shown in equation 5.1. This value includes the bias value. *(Bias=b= x0× w0)*

$$\boldsymbol{z} = \sum_{i=0}^{n} x_i w_i = x_0 w_0 + x_1 w_1 + \cdots + x_n w_n \tag{5.1}$$

The result is obtained by applying the activation function to the weighted sum(z). As an activation function, perceptrons use the step function. If the weighted sum matches the step function;

*If z>0, then result is 1,*

*If z≤0, then result is 0.*

The output formula with activation function presenting using the following equation:

$$Output(o) = \begin{cases} 1, & if\ w * x + \mathrm{b} > 0, \\ 0, & otherwise \end{cases} \quad (5.2)$$

During training, the weights are adjusted to reduce the error between both the network's actual output and the desired output. This is typically done using a supervised learning algorithm, such as gradient descent, where the weights are adjusted in the direction that minimizes the error.

Single-layer neural networks are useful for addressing simple classification problems using data that can be separated linearly. However, they have limited expressive power and need help learning more complex patterns. As a result, multi-layer neural networks, which can learn more complex representations, are often used for more challenging tasks.

#### 5.1.2.2 Multi-Layer neural networks

Multi-layer neural networks, also known as multi-layer perceptrons (MLPs), have more than one layer. Therefore, they can be utilized for non-linearly separable problems and perceptrons. They accomplish this through the activation functions used in their layers. The activation functions cause neurons' output to be nonlinear. As a result, it is possible to solve more complicated problems. (ANNs devolve into a linear regression model without the activation function.) Activation functions introduce complexity and nonlinearity into ANNs, allowing them to solve more complex problems than those solvable by linear regression models. They also provide the ability to represent complex relationships between inputs and outputs, such as nonlinear decision boundaries. Figure 5.5 depicts an example of MLPs structure.

MLPs' fundamental layers are as follows:

1. **Input layer**: consists of 1 neuron per input x.
2. **Hidden layers (one or more)**: The number of neurons in each layer varies depending on the problem.
3. **Output layer**: The total number of neurons is defined by the problem.

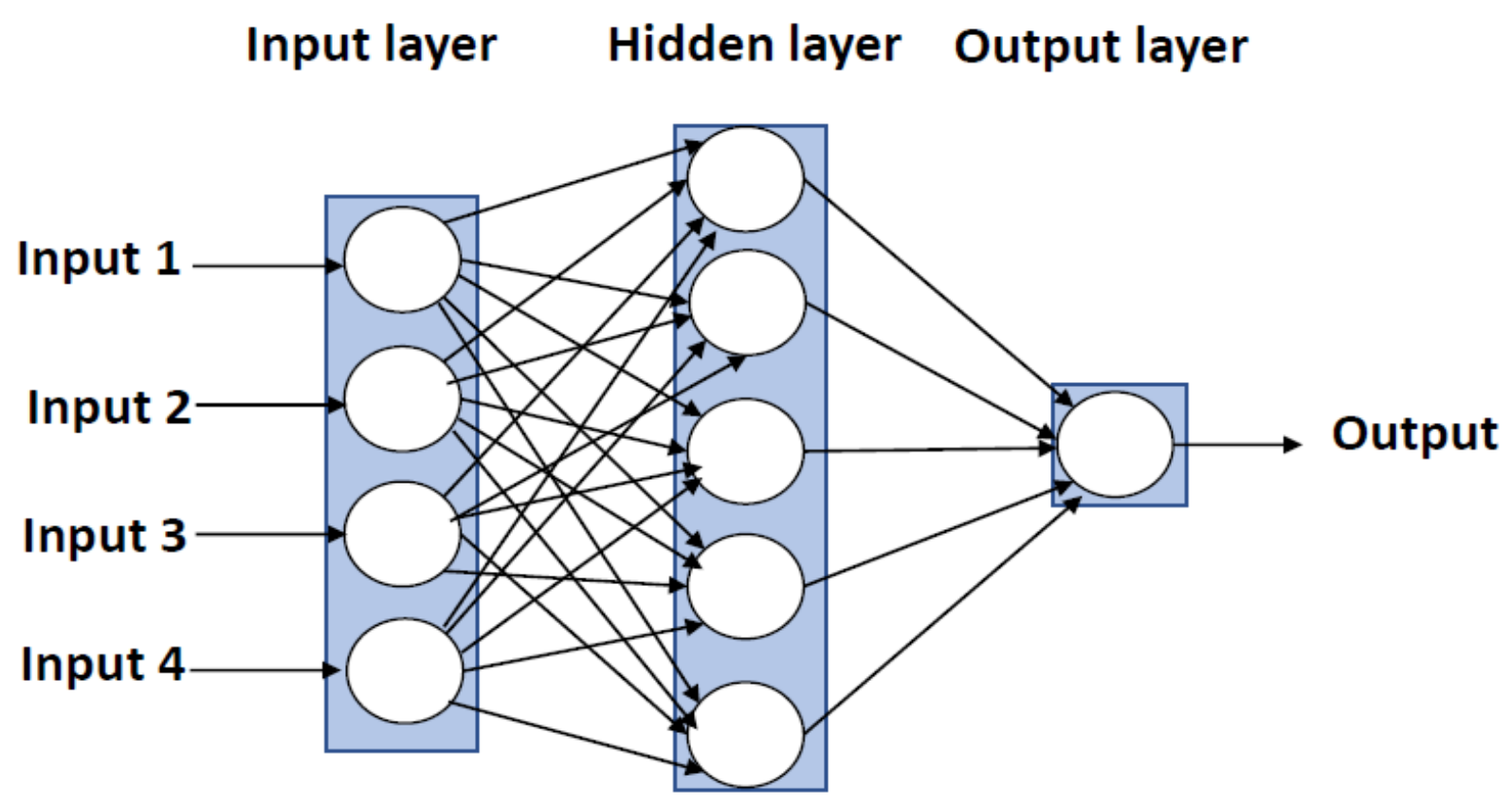

Figure 5.5 Example of Multi-Layer Perceptron architectures

In multi-layer perceptron architectures, we apply the same process used in the perceptron to each neuron of the hidden layers by using the following:

1. A weighted sum(z) is calculated.
2. It is transferred to a related hidden neuron, and then the neuron's activation function.

In the subsequent step, the outputs of the hidden layers are transmitted to the output layer. As mentioned earlier, the number of neurons in the output layer is determined by the specific problem being addressed.

- **Regression**: is composed of one neuron.
- **Binary Classification**: is composed of one neuron.
- **Multi-label Classification:** 1 neuron for each label
- **Multi-class Classification**: 1 neuron in the output layer for each class.

Multi-layer neural networks have more than one layer of neurons in the network, which allows them to learn complex patterns and solve difficult problems. Furthermore, they are more efficient than single-layer neural networks in dealing with large number of data since they can process the data more quickly while making fewer mistakes in complex tasks; the critical advantage of multi-layer neural networks is their ability to "generalize" the input data, which means that the data can be used to identify patterns in more tasks.

This ability to generalize data means a multi-layer neural network can be trained with various inputs. It can recognize patterns in tasks where the input data has never been seen before and make accurate predictions. In addition, multi-layer neural networks are more robust and can better handle noisy data than single-layer neural networks. Thus, multi-layer neural networks have become an invaluable tool for artificial intelligence.

#### 5.1.2.3 Long short-term memory neural network (LSTM)

The concept of an "LSTM network" encompasses a distinctive type of neural network architecture where the neurons in the hidden layer possess memory cells. This unique design allows the long-short-term memory (LSTM) network to effectively store and analyze information over extended periods. By leveraging an array of memory cells, LSTM networks gain the ability to learn from experiences and accumulate knowledge over time, making them invaluable for diverse applications, including text analysis, natural language processing, and image recognition. LSTM networks are particularly useful for learning long-term dependencies due to the ability of their hidden layers to store information in memory cells. Furthermore, the information stored in memory cells can be used to provide context to data and guide the decision-making process, allowing LSTM networks to make better predictions than other types of neural networks.

An LSTM network is made up of multiple layers. Each layer within the network is carefully crafted to fulfill a particular function in the network's overall operation. Layers in an LSTM network are often organized sequentially, with each subsequent layer built on the result of the prior layer For example, the input layer of an LSTM network takes in raw environmental input in the form of images, sounds, or text. The first hidden layer of the network processes the inputs and identifies important features within the data. Next, the hidden layer uses this information to produce a set of outputs which are referred to as candidate values for the inputs in the next layer. The next network layer receives these candidate values and compares them to the expected values for each input to determine which ones are correct. This entire process is repeated multiple times over multiple training cycles until all of the hidden layers have been fully trained.

As the name implies, LSTM networks combine features of both traditional memory and neural networks. Like other neural network models, LSTM networks use a series of interconnected nodes to process and store information. However, unlike typical neural networks, LSTM networks can also store information for long periods.

The equations for an LSTM cell's forward pass with a forget gate's compact forms are (Ruihui et al., 2019):

$$f_t = \sigma_g(W_f x_t + U_f h_{t-1} + b_f) \tag{5.3}$$

$$i_t = \sigma_g(W_i x_t + U_i h_{t-1} + b_i) \tag{5.4}$$

$$o_t = \sigma_g(W_o x_t + U_o h_{t-1} + b_o) \tag{5.5}$$

$$\widetilde{c_t} = \sigma_c(W_c x_t + U_c h_{t-1} + b_c) \tag{5.6}$$

$$c_t = f_t \odot c_{t-1} + i_t \odot \widetilde{c_t} \tag{5.7}$$

$$h_t = o_t \odot \sigma_h(c_t) \tag{5.8}$$

Where, The initial values are $c_0 = 0$ and $h_0 = 0$ , the operator $\odot$ indicates the Hadamard product . The subscript $t$ indicates the time step. $W_q$ and $U_q$ include the input and recurrent connection weights, respectively. where $q$ can be the input gate $i$, and output gate $o$, the forget gate $f$ or the memory cell $c$.

The rest of the variables are defined as follow:

- $f_t \in (0,1)^h$ : activation vector for the forget gate
- $o_t \in (0,1)^h$: activation vector for the output gate
- $i_t \in (0,1)^h$: activation vector for the input/update gate
- $\widetilde{c_t} \in (-1,1)^h$: cell input activation vector
- $o_t \in \mathbb{R}^d$ input vector to the LSTM unit
- $c_t \in \mathbb{R}^d$: cell state vector
- $\sigma_g$: sigmoid function.
- $\sigma_c$: hyperbolic tangent function.

#### 5.1.2.4 Hopfield network

The Hopfield Network refers to a network of neurons that is completely interconnected and has connections between every single neuron. By changing the value of neurons to the desired pattern, the network is trained using input patterns. Then, its weights are calculated. The weights are kept the same. After a network has been trained for one or more patterns, it will eventually converge on those patterns. It is unique compared to other neural networks.

Hopfield Network is also called as a feed-forward neural network. It stands as one of the most accessible and widely employed neural networks. The Hopfield network has a fixed number of inputs, hidden, and output nodes. All of them have the same dimension. Each of these nodes can be active or inactive. The network output is the sum of the products of all the nodes. This equation is as follows. $G=W*H-Wout$ where G is the desired output, W refers to the weight matrix, H is the hidden layer output matrix, and Wout is the output matrix. The Hopfield network does the job of finding the weights that best reproduce the desired output. Here, it is the answer to a linear equation system. This equation can solve various problems like clustering, classification, feature selection, etc.

Moreover, a Hopfield network can be a solution for any one of these problems. It can learn using the self-organizing property. So, it can solve complex problems without human intervention. Also, it has only local connection weights. Therefore, it needs less computational power and is applied to a wide range of problems, like finding patterns in data, image classification, machine learning, pattern recognition, etc.

In other terms, a Hopfield network function's input and output patterns are in a discrete line form. These discrete vector patterns can either be binary (0,1) or bipolar (-1,1) in nature. The network's weights are symmetrical, and there are no self-connections, i.e. $w_{ij} = w_{ji}$ and $w_{ij} = 0$. Weights will be adjusted while the discrete Hopfield network is being trained. We already know that binary and bipolar input vectors are both possible (Kitchenham et al., 2010). Thus, the following relation may be used to update weights in both cases:

- **Case 1**: Binary input patterns

  For a set of binary patterns $sp,\ p = 1\ to\ p$

  Here, $sp = s_1p, s_2p, \ldots\ldots, s_ip, \ldots\ldots, s_np,$ And the Weight Matrix is provided by

$$w_{ij} = \sum_{p=1}^{p} [2s_i\,(p) - 1][2s_j(p) - 1]\ for\ i\#j$$

- **Case 2:** Bipolar input patterns

  For a set of binary patterns $sp,\ p = 1\ to\ p$

  Here, $sp = s_1p, s_2p, \ldots\ldots, s_ip, \ldots\ldots, s_np,$ And the Weight Matrix is provided by

$$w_{ij} = \sum_{p=1}^{p} [s_i\,(p)][s_j(p)]\ for\ i\#j$$

In summary, Hopfield networks can be used then for pattern recognition, associative memory, and optimization problems. They are particularly useful for problems where the input data is noisy or incomplete, as well as for problems where certain rules or constraints constrain the output. Hopfield networks are often used for associative memory, where the network is trained to associate a set of input patterns with a set of output patterns. Once the network is trained, it can retrieve the output pattern given an input pattern, even if the input pattern is noisy or incomplete.

Hopfield networks may also be utilized to solve optimization issues that aim to find a function's least or maximum value. The network is trained to represent the function as the energy of a physical system, and the function's minimum or maximum value corresponds to the system's stable state. The network can be iteratively updated to converge to a stable state, representing the optimal solution to the optimization problem.

Overall, Hopfield networks are useful for problems where some patterns or rules need to be learned and recognized and where there is some degree of noise or uncertainty in the input data. They have demonstrated successful applications in diverse fields, such as image and speech recognition, constraint satisfaction, and optimization tasks.

#### 5.1.2.5 Recurrent neural network (RNN)

Recurrent refers to neural network connecting neurons in the hidden layers. Therefore, memory is present in it. Furthermore, the hidden layer neuron constantly gets activation from the bottom layer and its prior activation value. So, in recurrent neural networks, all the neurons are connected and remember their past activations. This makes the network more powerful at recognizing patterns, especially when the input is noisy.

Recurrent neural networks are widely utilized in various applications including: image recognition, machine translation, and text recognition. This is because they can capture the temporal dynamics of natural phenomena such as speech or handwriting. However, their performance can be enhanced by incorporating recurrent reinforcement learning (RL) techniques, which can teach them how to perform tasks more efficiently in the future. By applying RL techniques, the recurrent neural network can learn to recognize patterns more quickly and accurately in various scenarios, including recognizing objects in images, accurately translating language, and identifying text.

In addition, incorporating RL techniques can help recurrent neural networks learn to make better decisions in dynamic and uncertain environments, allowing them to be more adaptive and effective in various applications. Applying RL techniques to recurrent neural networks can improve their performance in several ways. For example, RL algorithms can optimize the learning parameters of a recurrent neural network by continually tweaking the values of these parameters until they reach optimal performance while allowing the network to modify its behavior in response to environmental changes or data. This process of continually adapting and adjusting the parameters of a recurrent neural network through reinforcement learning algorithms can significantly enhance the network's capabilities, allowing it to recognize patterns more effectively, make better decisions in dynamic or uncertain environments, and overall perform better.

There are four important steps to using RNN:

1. We start by modeling the neural network as a feed-forward net with input and output layers.
2. Then, we add a recurrent layer between both the input and the output layers, so that the neural  network is now modeling a recurrent process.
3. We train neural network utilizing standard feed-forward machine learning algorithms.
4. The resulting recurrent neural network will be more efficient and accurate than a feed-forward net because it can more accurately model the underlying recurrent process.

How to change a feed-forward network into a recurrent neural network is described below:

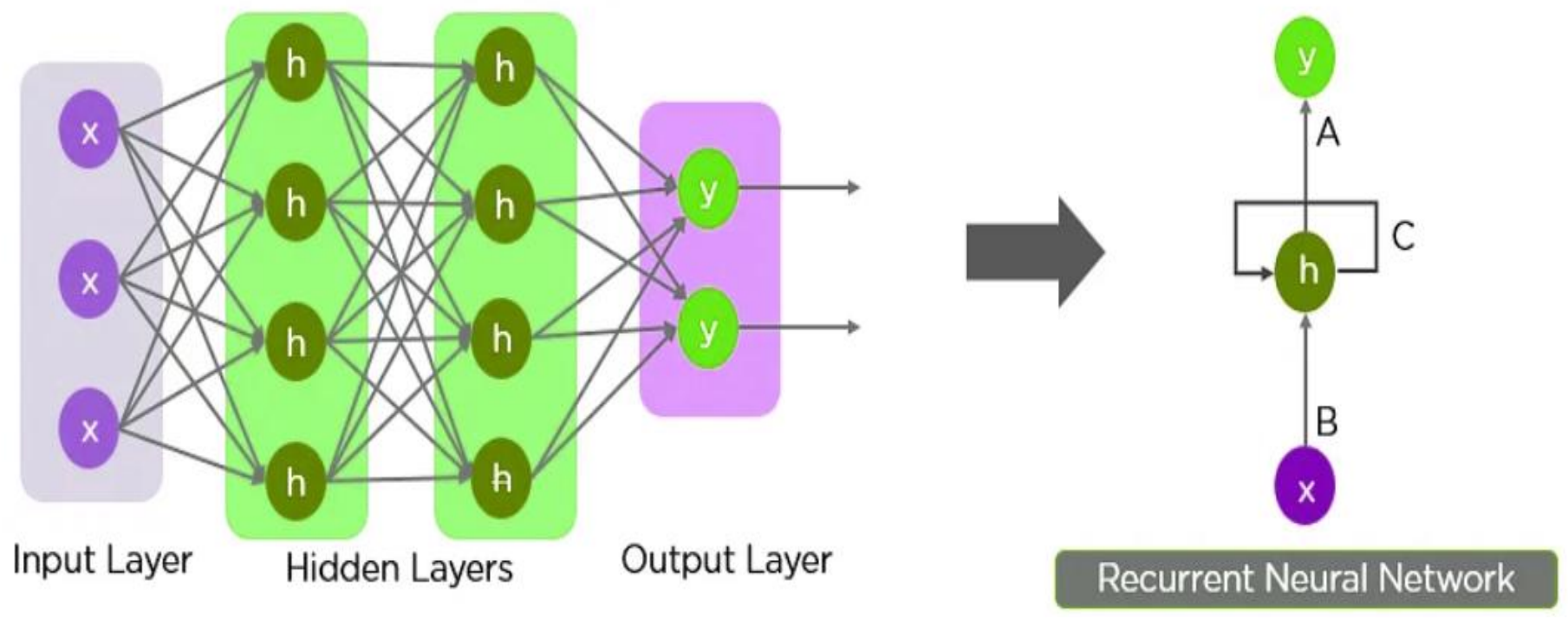

Figure 5.6 Basic RNN structure

A single layer of recurrent neural networks is created by compressing the nodes from several neural network layers. The parameters of the network are A, B, and C.

#### 5.1.2.6 Boltzmann machine neural network (BMNN)

The difference between these networks and the Hopfield network is that certain neurons are input while others are concealed by nature. Furthermore, the weights are learned using the backpropagation process after being randomly initialized. Boltzmann Machine Neural Network is based on the same ideas but comprises a finite number of layers and can solve problems more accurately than the Hopfield network. The Boltzmann machine is a type of neural network that Ludwig Boltzmann first proposed in 1884 (Tuto, 2022).

The basic idea is that a neural network can be considered a collection of tiny machines with its own set of weightss, that make up the entire network together. Each of these individual machines is responsible for learning and changing its set of weights. By adjusting the weights of these individual machines, the Boltzmann machine can learn patterns and find optimal solutions for problems, making it a powerful tool for solving complex problems.

Here are the key components and steps involved in a Boltzmann Machine Neural Network:

1. **Architecture**: A BM composed of a set of input and also hidden units connected through symmetric weights. The input units represent the input data, and the hidden units learn to extract features from the input data.
2. **Stochastic Binary Units:** The units in a BM are stochastic binary units, which means that they can take on two states (0 or 1) with a certain probability. The activation energy of the unit determines the probability of a unit being in state 1.
3. **Energy Function:** The energy function of a BM is defined in terms of the states of the individual units within the network and the weights that connect these units. The negative sum of the products of the states and the weights between the units gives the energy of a particular configuration of the units.
4. **Learning:** The learning algorithm of a BM is called Contrastive Divergence (CD), which is a form of unsupervised learning. During training, the BM learns to reconstruct the input data by adjusting the weights between the units. The objective is to reduce the variation between the reconstructed data and the original data.

5. **Sampling**: After the BM has been trained, it can create new input data samples by sampling from the probability distribution defined by the energy function. This is done by running a Markov Chain Monte Carlo (MCMC) algorithm, which repeatedly updates the states of the units based on the probability distribution.
6. **Applications**: BMs have been widely utilized in various applications, including image recognition, speech processing, and natural language processing. BMs have also been used in deep learning as building blocks for more complex models such as Deep Boltzmann Machines (DBMs) and Restricted Boltzmann Machines (RBMs).

Boltzmann Machine Neural Networks are a generative model that can learn patterns and relationships in data. They are particularly useful for unsupervised learning tasks where the input data does not have explicit labels or categories. However, training BMs can be computationally expensive, and their effectiveness for certain tasks is still an active area of research.

#### 5.1.2.7 Radial basis function

These networks resemble feed-forward neural networks, except that the activation function for these neurons is the radial basis function. Radial basis function (RBF) neurons have the ability to encode complex non-linear relationships within input and output values, which is why they are especially well-suited to pattern recognition and function approximation tasks, including speech recognition, image recognition, and object detection, making them an important tool in the creation of artificial intelligence. In addition, RBF neurons have several advantages over traditional feed-forward networks. They are computationally more straightforward and require less training data than traditional neural networks, making them an attractive choice for many applications.

Here are the key components and steps involved in a RBF Neural Network:

1. **Architecture**: The RBF comprises three layers: an input layer that receives input data, a hidden layer that computes the activations of hidden units, and an output layer that generates the network's output.
2. R**adial Basis Function:** The radial basis function is a mathematical function commonly utilized to determine the activation of hidden units in a neural network. Typically, the RBF takes the form of a Gaussian function, which measures the distance between the input data and a set of predefined centers.
3. **Centers:** The centers of the radial basis function are the points in the input space around which the Gaussian function is centered. The centers can be randomly initialized or learned during training using clustering algorithms like k-means.
4. **Weights**: The weights between the hidden and output layers are computed using linear regression or another supervised learning algorithm. During training, the weights are adjusted to minimize the distinction among predicted and actual output.
5. **Training:** During training, the input data is entered into the network, and the weights are adjusted to reduce the difference among predicted and actual output. The training algorithm can be supervised or unsupervised, depending on the task.
6. **Prediction:** Once the RBF neural network has been trained, it can predict new input data by passing it through the network and obtaining the output.

The RBF neural network is a powerful machine-learning valuable model for function approximation and pattern recognition. It is simple to implement and can be trained efficiently using gradient descent or other optimization algorithms. However, the RBF neural network's performance can be sensitive to the choice of the radial basis function as well as the quantity of hidden units, and careful tuning of these parameters is often necessary to achieve good performance.

Figure 5.7 summarizes a general example of these types of neural network architecture that are commonly used for machine learning tasks. However, it is essential to point out that the actual architecture and design may differ depending on the specific task, dataset, and application requirements.

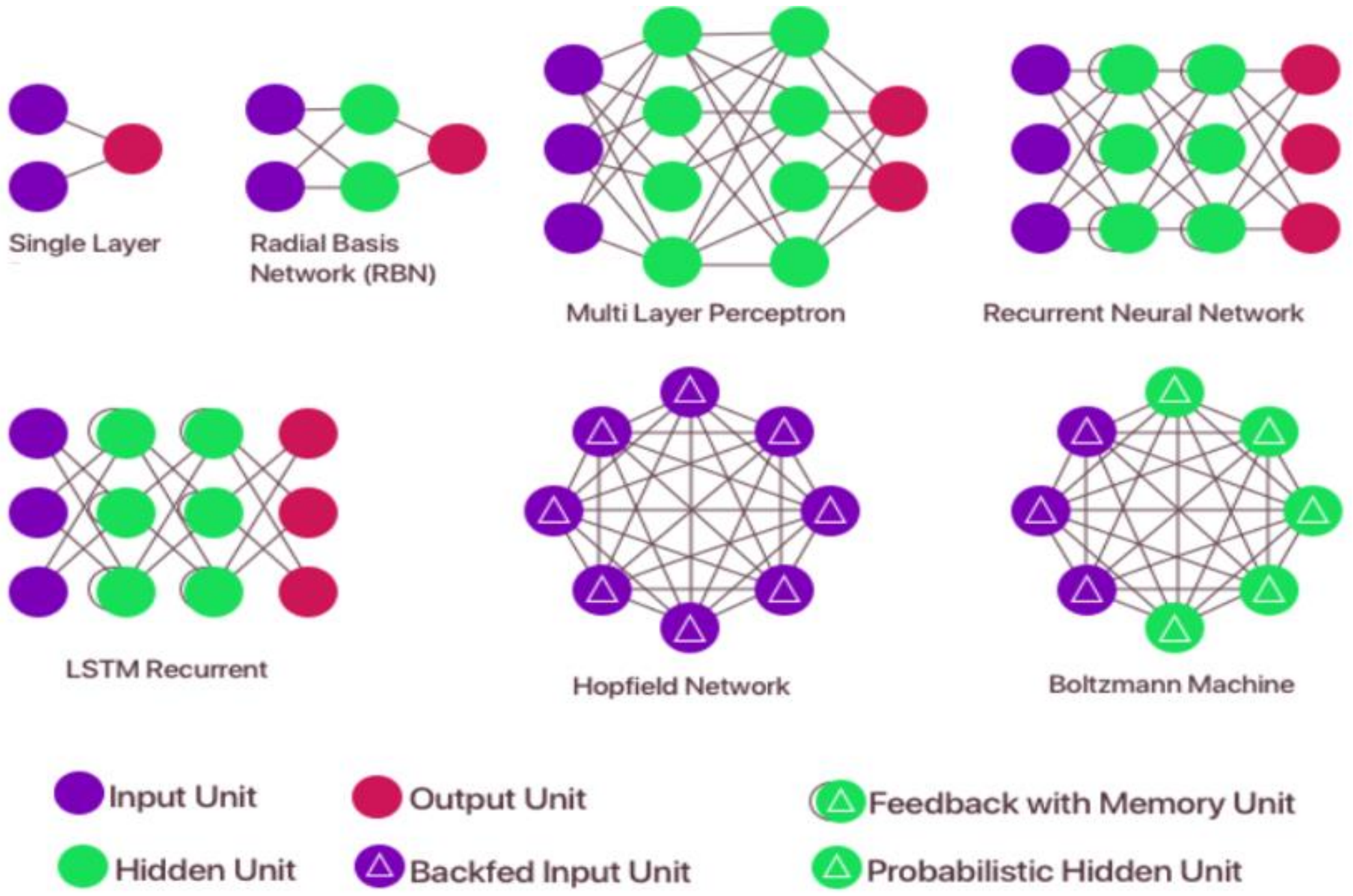

Figure 5.7 An illustration of many neural network architectures

## 5.2 Convolutional Neural Networks (CNN)

In the previous section, we explored the architecture of artificial neural networks (ANNs), which typically consist of an input, hidden, and output layer. Each neuron has a different weight and bias value, which undergoes a transformation using an activation function. If the resulting value exceeds a predefined threshold, the neuron passes the information to the subsequent layer; otherwise, it retains it. ANNs propagate data forward from one layer to the next. During the training phase, the objective is to optimize the weights and biases within each neuron to minimize a specific cost function. In addition, the aim is to generate output values that closely approximate the actual labels. This process involves iteratively adjusting the parameters based on the training data. Convolutional neural networks (CNNs), a type of multi-layer neural network architecture, are specifically designed to extract relevant features from various input types, including images and text. The core component of CNNs is the convolution operation, which involves applying a set of grid-structured weights to the input data. This operation helps capture complex relationships within the data, enabling subsequent layers to classify or regress the input effectively.

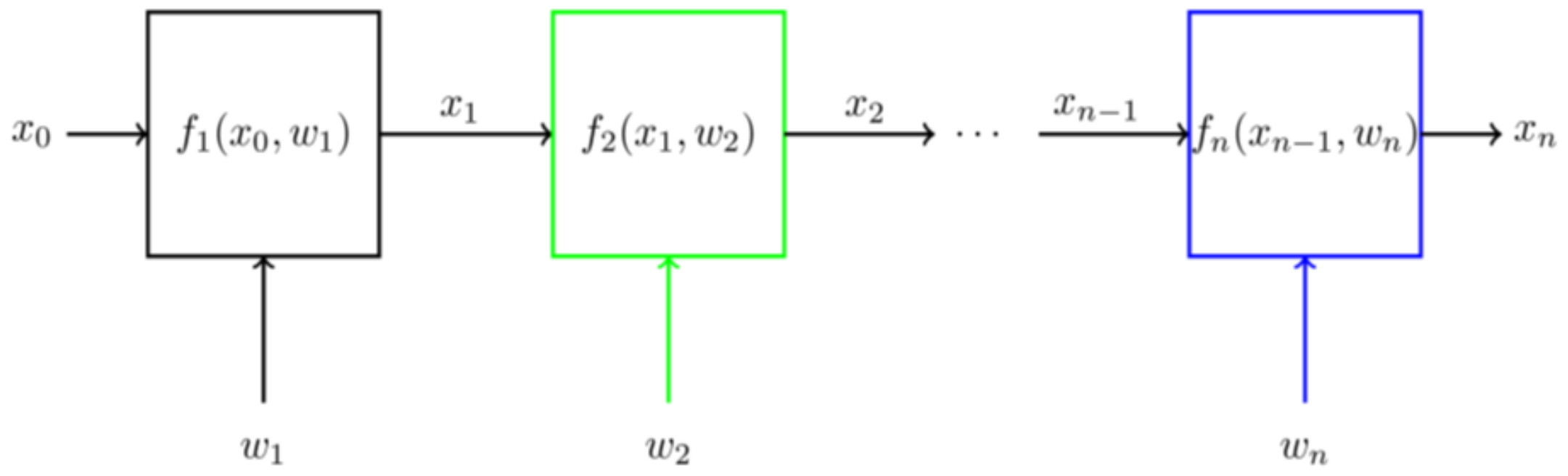

Figure 5.8 ANN Computational Architecture

CNNs, also known as convolutional neural networks, have emerged as a prevalent and extensively utilized type of artificial neural network (ANN) in tasks involving image recognition and processing. The inception of CNNs can be traced back to the groundbreaking work of LeCun et al. with the introduction of LeNet-5 in 1998 (LeNet-5. 1998). However, it is worth noting that the foundation for CNNs was laid earlier by Hubel and Wiesel in 1962, as evidenced by their pioneering research (Hubel et al., 1962). An extensive literature review confirms that CNNs are the prevailing deep learning models for image classification, as highlighted by the insightful findings of Traore et al. (Traore et al., 2018). These advancements in CNN architectures have propelled their widespread adoption and utilization across various domains. CNNs have revolutionized the field of image data classification and set the standard for deep learning models, providing a powerful and effective tool for working with high-dimensional datasets. The development of CNNs by Hubel and Wiesel has profoundly affected the field of deep learning, revolutionizing how data is analyzed and classified and allowing us to achieve levels of accuracy and complexity previously thought impossible. Since their inception in 1962, CNNs have come a long way in capability and complexity and can now achieve highly accurate classifications across a wide range of image datasets. This development has opened up new possibilities for working with image datasets and enabled researchers to tackle some of the most complex problems in the field, such as object recognition and scene understanding.

The optimization process in CNN architectures involves leveraging the back-propagation algorithm to iteratively adjust the weights and biases, thereby minimizing the associated cost function. CNN models have demonstrated their efficacy in diverse applications, such as recognizing and classifying zip code digits from the United States Postal Service, as documented by LeCun et al. in their notable work (LeCun et al., 1998). Similarly, in the realm of plant disease detection, CNNs analyze datasets consisting of leaf images, extracting critical features that aid in accurate diagnosis. Toda et al. illustrate how CNNs mimic human decision-making by capturing textures and colors of plant leaf lesions for disease detection (Toda et al., 2019). The distinguishing feature of CNNs lies in their ability to autonomously learn essential features from leaf images using a multitude of non-linear filters, thereby outperforming models that rely on manually engineered features, as demonstrated by (Singh et al., 2018).

CNNs are becoming more prevalent in image segmentation, image classification, and object recognition. Various organizations have successfully employed them in domains such as healthcare, web services, and mail services. As demonstrated in (Abdel-Hamid et al., 2014; Kmilaris et al., 2018). CNN can process a variety of data types, including images, videos, audio, and natural language. These data can be very complex, and extracting useful information from them can be difficult. However, when combined with the powerful features of convolutional neural networks, it becomes possible to remove meaningful features from this data and use them to make intelligent decisions. In addition to the various data that can be utilized as input for convolutional neural networks, the architecture of these networks also makes them well-suited to address complex problems.

CNNs, being a type of feedforward neural network, exhibit a distinctive structure characterized by a hierarchical arrangement of layers, including convolution, pooling, and fully connected layers. This architectural design draws inspiration from the receptive field found in the visual cortex of humans, with layers stacked to form a deep network. The CNN's journey commences with a convolution layer and progresses through subsequent layers encompassing pooling, ReLU correction, and culminating in a fully connected layer. Canziani et al. provide an extensive and comprehensive exploration of the different layer types in CNNs and their respective functions (Canziani et al., 2016). To visualize the structure, Figure 5.9 depicts the architecture of a CNN model.

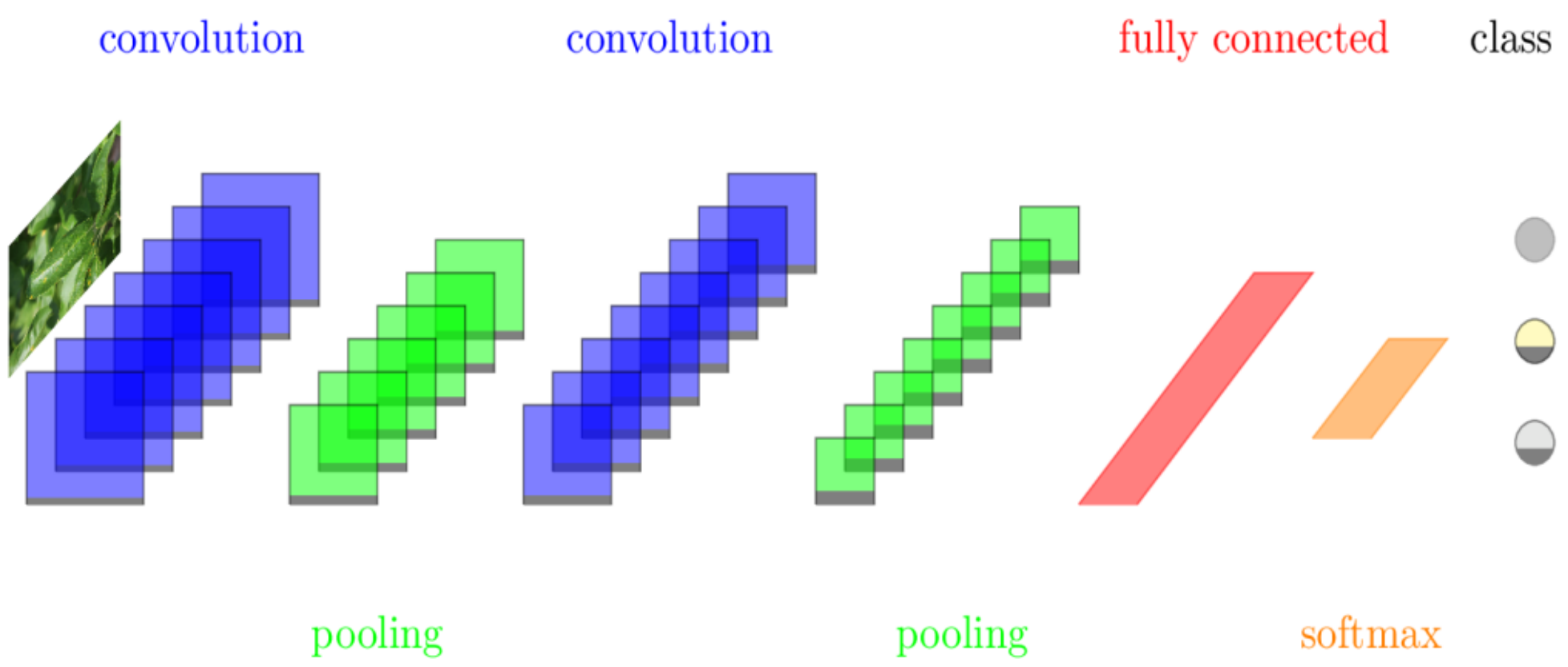

Figure 5.9 A common CNN architecture

CNNs have proven to be highly effective in extracting powerful features from diverse input data. The unique architecture of CNNs enables them to efficiently and accurately capture relevant features from various types of input data. Specifically, the convolutional layers employ filters to extract distinctive features from input images, which are subsequently downsampled by the pooling layers to reduce dimensionality. The fully connected layers leverage these high-level features to classify input images into predefined classes, enabling CNNs to make informed decisions based on the acquired knowledge. This makes CNNs well-suited for various applications, including image classification, object recognition, and segmentation (Schmidhuber et al., 2015). Furthermore, the architectural design of CNNs plays a pivotal role in feature extraction and utilization for robust classification and detection, as highlighted by the research conducted by (Alom et al., 2019).

Building a CNNs from scratch can be a time-consuming process. Therefore, researchers often rely on successful and popular architectures to build their models. Among these, GoogleNet, AlexNet, VGG, and Inception-ResNet have been widely used (Palo, 2016). Each architecture has its own unique benefits and is more suitable for certain scenarios (Canziani et al., 2016). It should be mentioed that most of these architectures come with pre-trained weights, which means that they have been previously trained on specific datasets and have learned to provide efficient analysis for certain domains of problems (Pan et al., 2010). For example, ImageNet is a well-known dataset used for pre-training DL architectures (Dong et al., 2009). Another popular dataset for pre-training is PASCAL VOC (Bahrampour et al., 2015), which contains annotated images with bounding boxes that can be used to train object detection networks.

In the realm of deep learning, researchers have access to a plethora of platforms and tools to experiment with. TensorFlow, Theano, Keras (an API based on TensorFlow and Theano), PyTorch, Caffe, Pylearn2, TFLearn, and the Deep Learning Matlab Toolbox are among the most commonly used platforms and tools available (Bahrampour et al., 2015). Moreover, several of these tools (such as Caffe and Theano) integrate popular deep learning architectures (including AlexNet, VGG, and GoogleNet) either as libraries or classes, making them easier to use for researchers. These platforms and tools provide researchers with the necessary support to design and test their deep learning models efficiently and effectively. Furthermore, there are several distinct libraries that are tailored to deep learning, such as Torch7 and its community-driven LuaDeep library, as well as NVIDIA's CUDA libraries, which are specifically designed to make use of the parallel processing capabilities of modern GPUs, allowing researchers to build and train complex deep learning models quickly and efficiently. Although these tools provide useful libraries, frameworks, and architectures to expedite research in deep learning, they also present a barrier for new researchers who are unfamiliar with them, resulting in a need for additional resources that can help them become acquainted with them and better understand their inner workings.

### 5.2.1 Convolution layer

The convolutional layer forms the backbone of CNNs and serves as a computational powerhouse within the network. It comprises essential components, including input data, filters, and feature maps. By incorporating the convolutional layer, CNNs can effectively capture an extensive range of features from input images. This expanded receptive field enhances the model's ability to detect and extract meaningful information. In traditional image processing approaches, examining every pixel in an image is a straightforward yet computationally intensive method. However, such an exhaustive process can significantly slow down the training process. Moreover, considering the sparsity of image pixels, many of them being zeros, they may not convey the essential features required for object recognition. Hence, the convolutional layer optimizes the computation by selectively focusing on relevant regions and extracting valuable features for subsequent analysis.

The convolution operation involves the application of a filter to the image and sliding it over it to extract more specific and distinctive features that are unique to the object. The filter criteria learned from data make it simpler for the detector to fine-tune the essential features (e.g., edges, shapes) necessary for determining the output and allow for building a deeper and more complex neural network (Zhang et al., 2020). The output results of a convolution operation are known as a feature map or activation map. The cross-correlation operation inspired the convolution operation, and an instance of a 2D convolutional operation can be found in the equation below (Goodfellow et al., 2020):

$$S(i,j) = I(i,j) * K(i,j) = \sum_m \sum_n I(i+m, j+n) K(m,n) \tag{5.3}$$

Where I is the input, K is the kernel or filter, S is the feature map, (i, j) is the entry position in a 2-D matrix, and m and n are the length and width of the kernel. For example, if we apply convolution on a 2 by 2 kernel, The calculation of the first entry of the feature map can be determined, as shown in Figure 5.10.

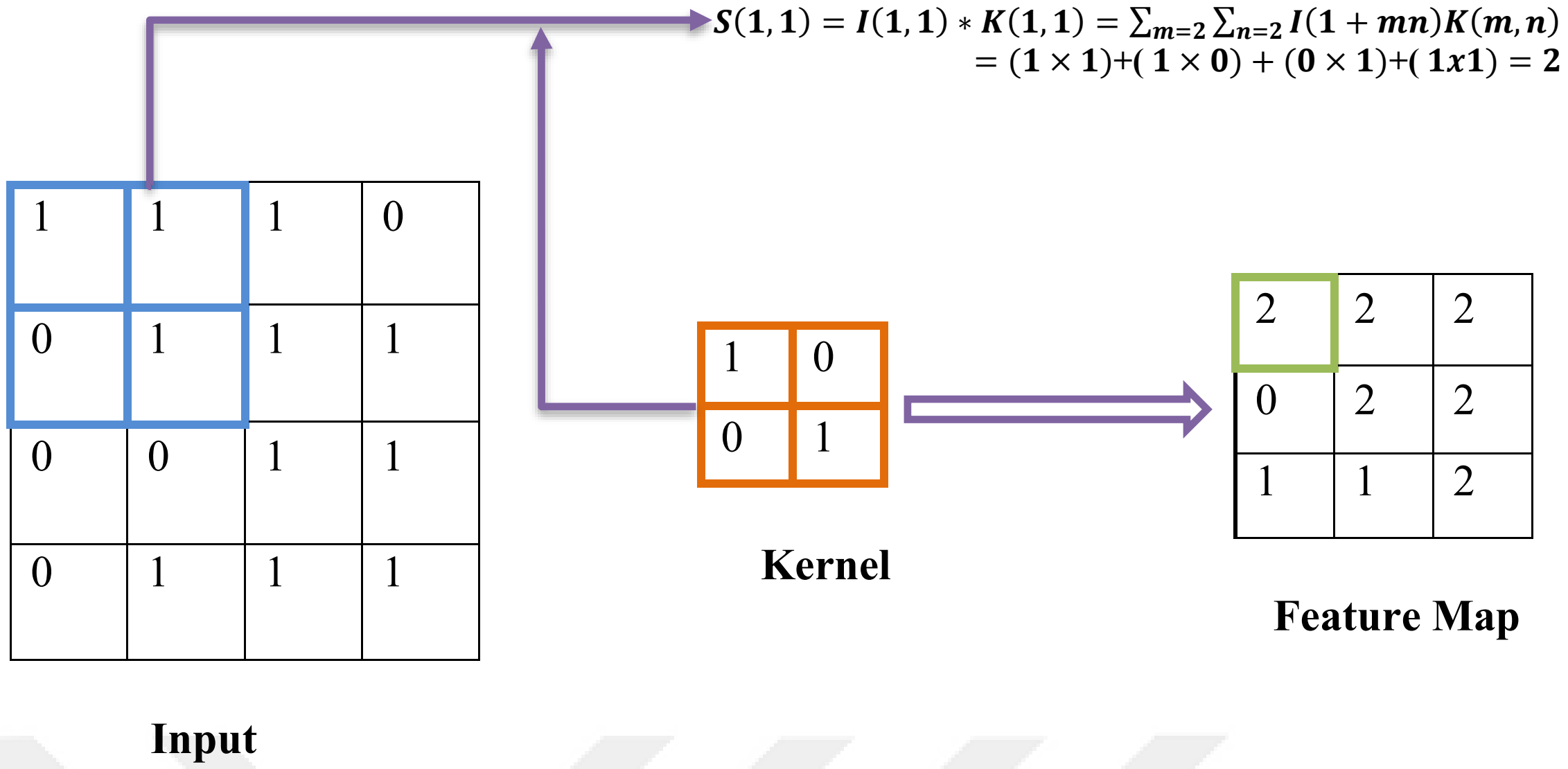

Figure 5.10 2x2 kernel convolutional operation

To conclude, the convolutional layer refers the basis of a CNN and is where most computation occurs. The convolutional layer performs convolutions using the filter to generate a feature map which is from the input data, which is then fed into the next network layer. The convolutional layer is a vital network component, allowing it to recognize patterns in the input data and extract features from it. Through this process, the convolutional layer can detect a wide range of features from the input data, from high-level features such as shapes and objects to low-level features including edges and corners, providing the network with a rich set of features that can use to make predictions. This process of detecting features from the input data makes the convolutional layer a potent tool for deep learning applications, and it is a fundamental component of various exesting deep learning models.

### 5.2.2 Pooling Layer

The pooling layer plays a crucial role in image compression by substituting the output of the feature map with a summary statistic derived from nearby outputs. Max-pooling, for example, utilizes a sliding window to compute the maximum value, while average pooling computes the average value within the window. This downsampling process significantly reduces redundant pixels, resulting in accelerated training and decreased memory consumption. By incorporating the pooling layer in the design of deep neural networks, the complexity of image information can be effectively streamlined, as highlighted in the research by (Goodfellow et al.,2020).

The pooling layer, also known as subsampling, minimizes the spatial size of the convolution layer's output to decrease the number of parameters and computations in the network. It also regulates overfitting. The pooling layer is typically utilized between two convolution layers or between fully connected layers. This layer primarily employs two methods: max pooling and average pooling. In a given input window, max pooling utilizes the highest values, whereas average pooling uses the average values.

In Max Pooling, a kernel is passed over the input volume, and the maximum value within each region covered by the kernel is taken as the output value. This assists in extracting the most essential features from the input volume. And in Average Pooling, a kernel is passed over the input volume, and the average value within each region covered by the kernel is taken as the output value. This helps in reducing the effect of noise in the input volume. Both Average Pooling and Max Pooling can be utilized in different parts of a CNN architecture to extracting the most essential features from the input volume and reduce the spatial dimensions of the output volume. Pooling has two essential parameters: pooling size and stride. A larger pooling window and stride will result in more aggressive downsampling and a smaller output feature map. Conversely, a smaller window and stride will result in less downsampling and a larger output feature map. Figure 5.11 shows an example of pooling.

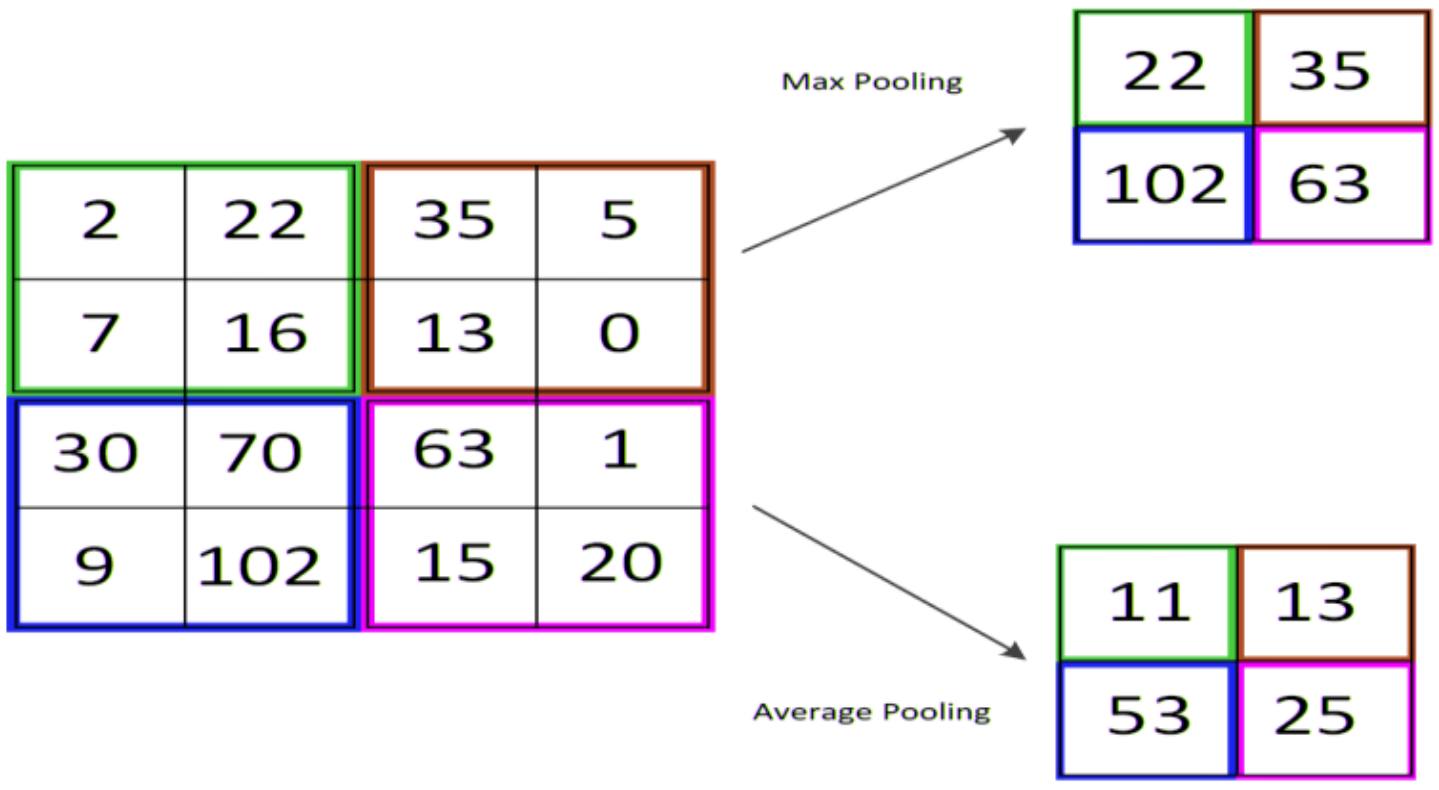

Figure 5.11 Example of pooling

Overall, pooling layers are useful for minimizing the spatial dimensions of the feature maps, which helps to dercrease overfitting and speed up computation. However, too much pooling can result in information loss, so it is important to balance the amount of downsampling with the network's ability to retain essential features.

### 5.2.3 Fully connected layer

The fully connected layer, found in convolutional neural networks (CNNs), plays a crucial role in improving class scores and enabling accurate predictions. Acting as the "final layer" of the network, it consists of a 1xN-sized matrix that connects to all neurons in the preceding layer. By incorporating the fully connected layer, which is an integral part of feed-forward neural networks, CNN architectures can enhance their ability to analyze input data. In the CNN, the output of the final pooling or convolutional layer undergoes flattening and is then passed as input to the fully connected layer. Within this layer, the flattened input undergoes a matrix multiplication operation, followed by the application of an activation function. As a result, final output scores or probabilities for different classes are generated, allowing the network to make predictions based on the learned features extracted from the input data. This utilization of the fully connected layer empowers CNNs to refine class scores, leverage learned features, and make accurate predictions, contributing to their effectiveness in various applications (Kizrak et al., 2018; Ülker et al., 2017).

The fully connected layer plays a pivotal role in many convolutional neural network architectures, as it is responsible for extracting non-linear and hierarchical features from the input data. This layer comprises a set of neurons that are initialized with random weights and bias terms and are then trained through a learning process. The output of the layer is determined by a non-linear activation function, which enables the model to capture complex patterns in the data. To further improve the model's performance, a fully connected layer can also incorporate a dropout mechanism, which randomly drops neurons to prevent over-fitting and promote generalization. The inclusion of this supplementary regularization layer aids in promoting the model's ability to generalize well to a novel, unseen data instances.

This ensures that the CNN learns the most important features of the data rather than fitting to a large amount of noise and keeps the model from becoming overly dependent on specific neurons. With all of these features, the fully connected layer is an essential part of any CNN architecture and can be an effective tool for building accurate and reliable models; when used correctly, a fully connected layer can increase the accuracy and precision of a CNN model by providing an effective way to extract essential key features from the data while maintaining the ability of model to generalize to new data, ultimately leading to more accurate predictions.

In summary, the fully connected layer role in a CNN serves as a classifier, taking the features that the convolutional and pooling layers have learned and producing a final prediction for the input image. It is typically composed of one or more dense layers and can be computationally expensive but can be optimized through techniques such as global average pooling or dimensionality reduction.

Here an example of a simple CNN architecture and a fully connected layer. We suppose we have an input image of size 32x32x3 (width, height, and channels). The architecture of CNN can be:

- Convolutional layer 32 filters of size 3x3 and stride of 1
- After the convolutional layer, the ReLU activation function is typically applied.

- A max pooling layer is applied with a pool size of 2x2 and a stride of 2.
- The second convolutional layer has 64 filters with a size of 3x3 and a stride of 1, Like the previous layer, it uses the ReLU activation function.
- The next layer is another max pooling layer with the same parameters as the previous one.
- After the pooling layer, the output is flattened to create a one-dimensional vector.
- Then, a fully connected layer with 128 neurons is added to the architecture to make high-level decisions based on the learned features.
- Using ReLU activation function
- To mitigate overfitting, a dropout layer is implemented.
- Additionally, the architecture includes an output layer with 10 neurons, corresponding to the 10 classes in the multi-class classification problem. The softmax is used in the output layer to convert the final layer's outputs into class probabilities.

In this CNN architecture, the fully connected layer plays a vital role in the classification process. It takes the output of the flatten layer, which is a one-dimensional vector of size 64x64x64 (depending on the specific parameters of the previous layers). The fully connected layer consists of 128 neurons, each connected to every neuron in the preceding layer. The output of the fully connected is then passed through a ReLU activation function, which presents non-linearity to the model and enhances its ability to learn complex patterns from the data. Additionally, a dropout layer is incorporated after the ReLU activation function to prevent overfitting, by randomly dropping out some neurons during training, adding an extra layer of regularization.

The last layer of the architecture comprises 10 neurons, corresponding to the number of classes in the classification task. Using the softmax activation function, the outputs of this layer are transformed into a probability distribution, indicating the likelihood of the input image belonging to each class. This enables the model to provide class predictions based on the highest probability value among the output neurons.

Throughout the training process, the backpropagation algorithm is employed to adjust the weights and biases of the fully connected layer, as well as the other layers in the CNN. Acting as a classifier, the fully connected layer utilizes the features acquired from the convolutional and pooling layers to classify the input image into one of the predefined classes. By iteratively updating the parameters based on the error between predicted and actual labels, the model gradually improves its ability to accurately classify new images.

### 5.2.4 Activation functions

The activation function is a fundamental component of neural networks, with a crucial role in determining the activation status of individual neurons. By evaluating the weighted sum of inputs and incorporating a bias, the activation function introduces non-linearity to the neuron's output. This non-linearity is essential for enabling intricate and adaptable decision-making within the network. The activation function contributes to the network's ability to capture complex patterns and relationships in the data, facilitating the network's learning and prediction capabilities.

In multilayer networks, linear activation functions convert expressions into non-linear expressions. For example, after calculation, $y = f(x, w)$ matrix multiplication of our linear function in the form of each neuron's weight transforms it into a non-linear value in layers. The functions normalize the output of the hidden layers because the learning process is back-derivative. Softmax, sigmoid, ReLU, TanH, ELU, PReLU, SoftPlus, and Swish are examples of standard activation functions. ReLU, Softmax, and Sigmoid are the most commonly used.

Activation functions play an essential role in deep learning models by introducing non-linearity, which is needed to model complex problems. In general, these activation functions provide the neural network with more options to better map the input to its output and give the model more flexibility to learn from the data. Table 5.1 shows an example of the most commonly used activation functions. And Figure 5.12 depicts the most used activation functions (Firat unv, 2018).

Table 5.1 An example of common activation functions

<table>
<tr><th colspan="2">Function</th><th>Function Graph</th></tr>
<tr><td>Sigmoid</td><td>When the activation function for a neuron is a sigmoid function it is a guarantee that the output of this unit will always be between 0 and 1. Also, as the sigmoid is a non-linear function, the output of this unit would be a non-linear function of the weighted sum of inputs. It is used for binary classification, and is usually located in the last layer.<br><br>$$f(x) = \frac{1}{1 + e^{x}}$$</td><td>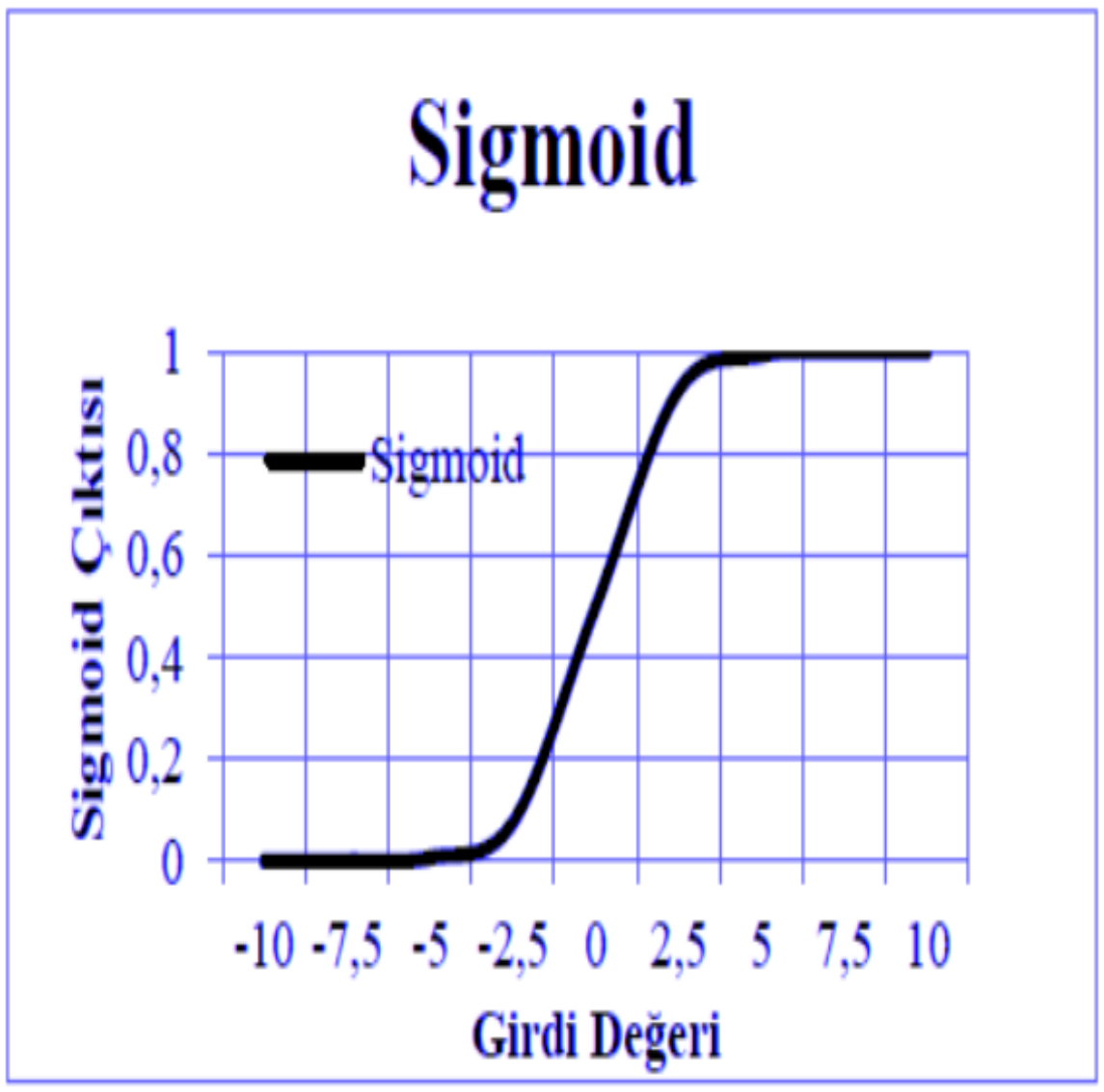
</td></tr>
<tr><td>Softmax</td><td>The softmax activation function converts the neural network's unprocessed outputs into a vector of probabilities, which is effectively a probability distribution over the input classes. Consider an N-class multiclass classification issue. Therefore, in a multiclass problem, Softmax assigns a decimal probability to each class. These decimal probabilities must sum up to 1.0. Training converges faster than it would without this additional constraint.<br><br>$$f(x) = \frac{e_i^{x}}{\sum_i^k e^{x}}\ ,\ \ i = 0,1,2, \dots, k$$</td><td>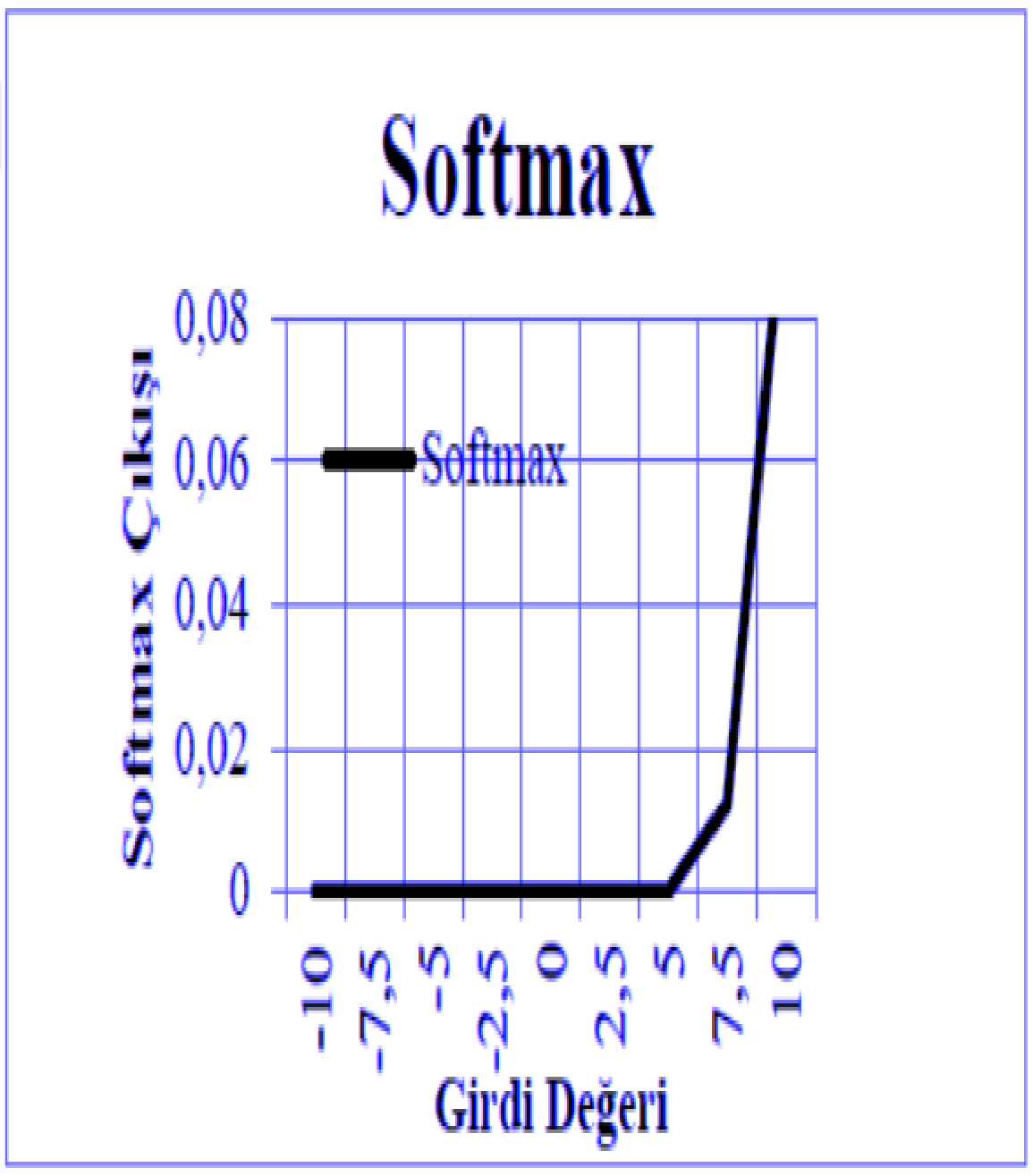
</td></tr>
</table>

Table 5.1 An example of common activation functions (continue)

| | Function | Function Graph |
|---|---|---|
| **ReLU** | ReLU is a non-linear function. It only involves a comparison of its input and the value 0. It also has a derivative of 0 or 1, depending on whether the input is negative or not. <br> $f(x) = \begin{cases} x, & if\ x \geq 0, \\ 0, & if\ x < 0 \end{cases}$ | 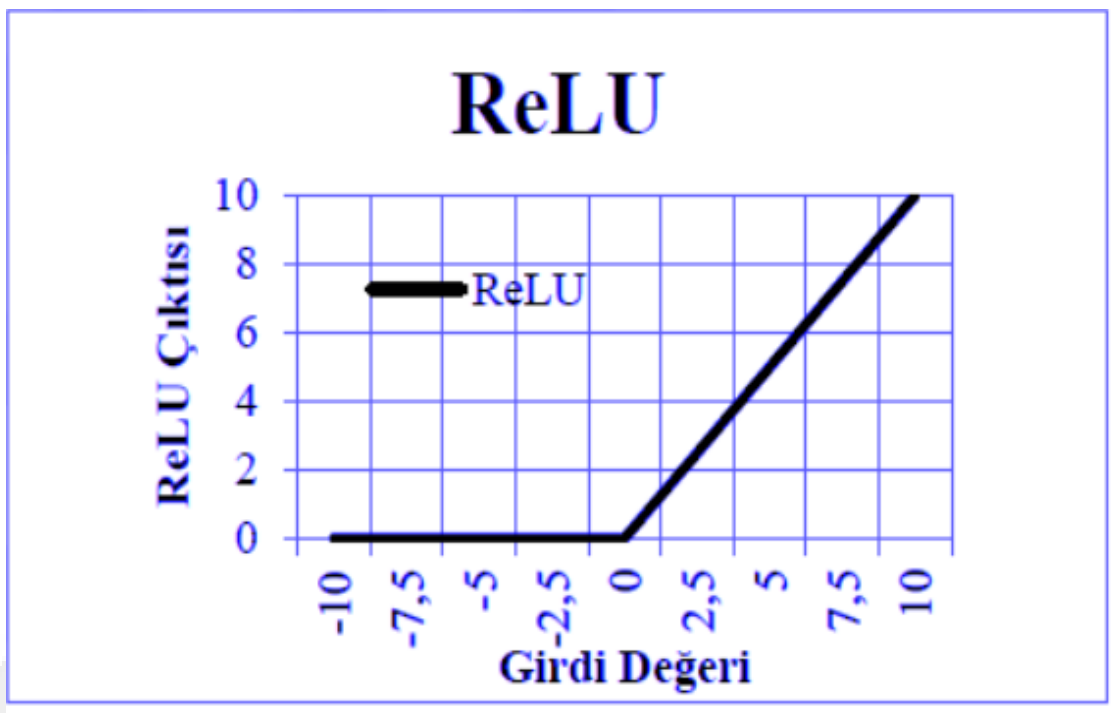  |
| **Step** | In neural networks, the unit step activation function is a common feature. The output assumes a value of 0 for a negative argument and 1 for a positive argument. For example, the function is as follows: The output is binary, and the range is between (0,1). <br> $\hat{y} = \begin{cases} 1, & if\ \mathrm{wx} + \mathrm{b} \geq 0, \\ 0, & if\ \mathrm{wx} + \mathrm{b} < 0 \end{cases}$ | 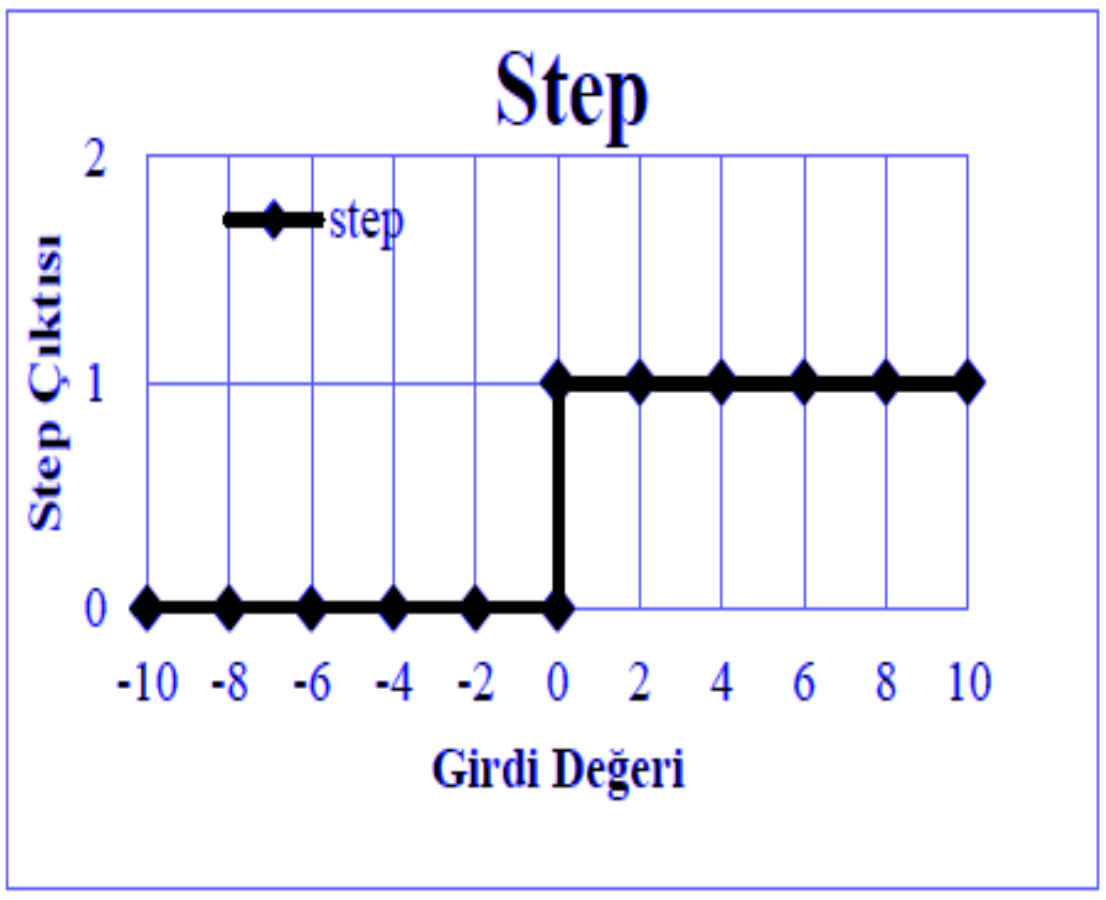  |
| **Tanh** | The tanh function is primarily used for classification between two classes. This activation function, which was popular in the early days of artificial neural networks, is a non-linear function that produces output in the interval [-1,1]. <br> $\mathrm{f(x)} = \tanh(\mathrm{x}) = \frac{2}{1 + e^{-2x}} - 1$ | 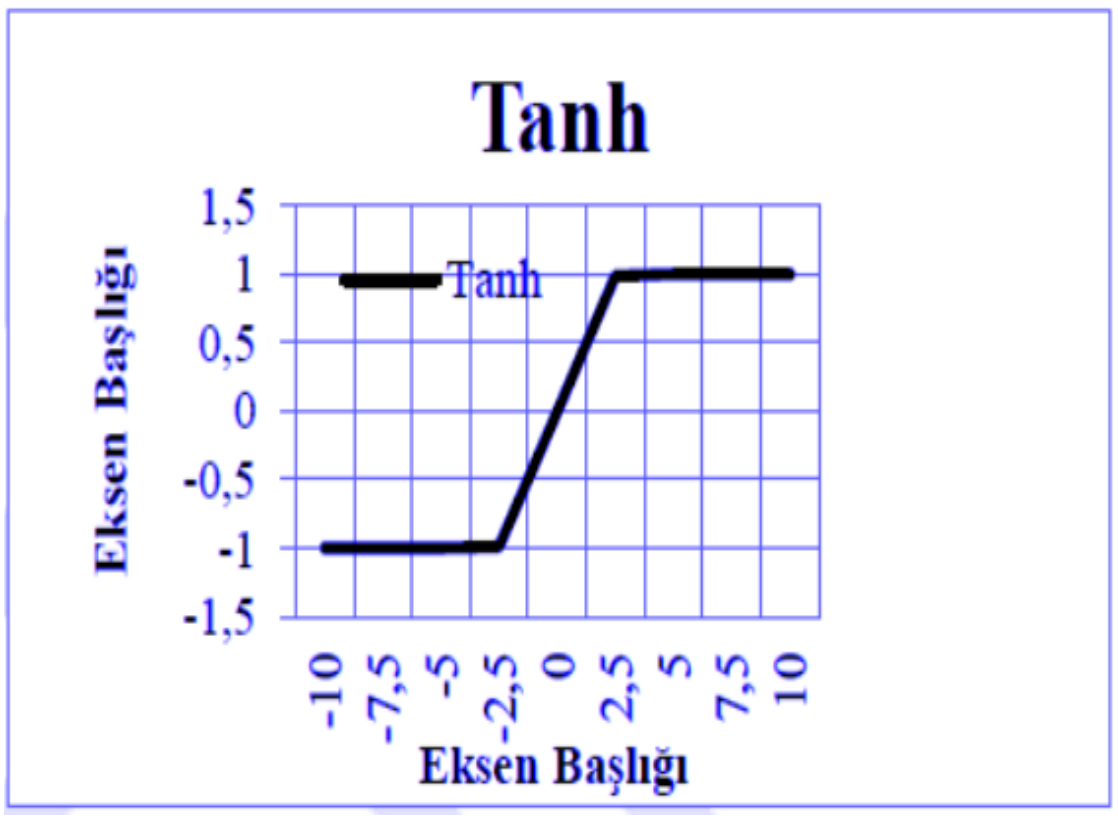  |

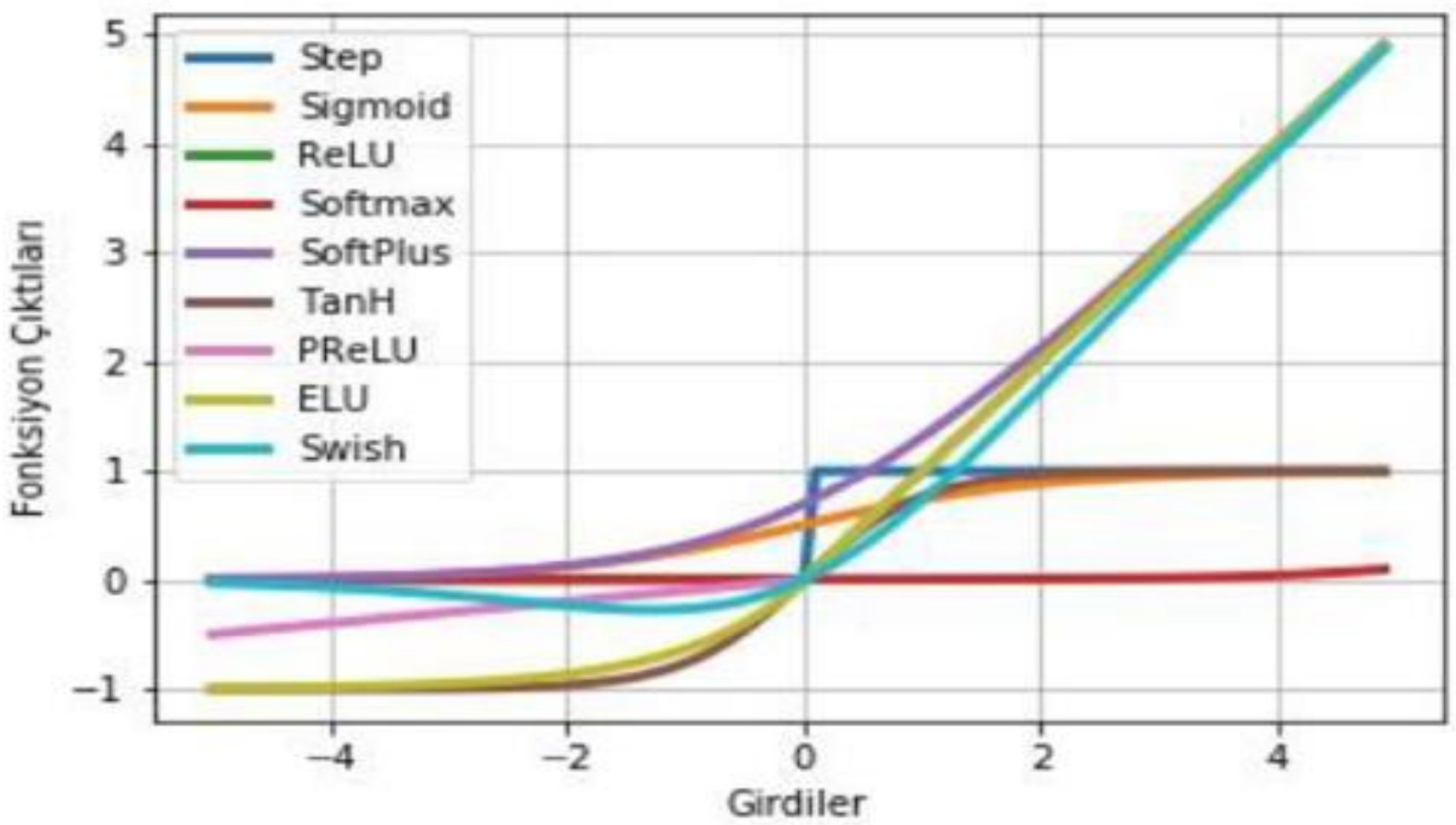

Figure 5.12 Commonly used activation functions

### 5.2.5 Evolution of CNN architecture

CNNs have gained significant popularity in the field of biologically motivated Artificial Intelligence (AI) methodologies. The origins of CNN can be traced back to Hubel and Wiesel's (1959, 1962) neurobiological experiments (Firat unv, 2018; Hubel et al., 1959). Their work served as a framework for many cognitive models, with CNN replacing nearly all of them. Consequently, numerous efforts have been dedicated over the years to enhance CNN's performance.

The field of convolutional neural networks (CNNs) has witnessed significant advancements since the inception of LeNet-5 in 1998. These innovations can be broadly classified into structural reformulation, parameter optimization, and regularization techniques (Lacun et al., 1998). However, current works have shown that the most substantial improvements in CNN performance come from fundamental restructuring and the creation of novel building blocks (Khan et al., 2019).

This systematic literature review (SLR) focuses on new architecture proposals based on the groundbreaking work of (Khan et al., 2019) and (Goodfellow et al., 2016). The cutting-edge CNN architectures are designed to tackle various tasks of computer vision including classification, and object detection. Through the utilization of convolutional and subsampling layers, input images are efficiently encoded, followed by a classifier approach that accurately estimates class probabilities in classification scenarios. Furthermore, these classification layer models can also serve as effective feature extractors for tasks involving segmentation and recognition (Alom et al., 2019).

In the past years, the field of semantic segmentation has witnessed the introduction of several models comprising encoding and decoding stages, as mentioned in the work of Alom et al. (Alom et al., 2019). The Fully Convolutional Network (FCN), initially proposed by (Long et al., 2015), stands as one of the pioneering segmentation models. Building upon this foundation, researchers have developed various novel architectures, including UNet by (Ronneberger et al., 2015), DeepLab by (Chen et al., 2018), SegNet by (Badrinarayanan et al., 2017), and R2U-Net by (Alom et al., 2019). Distinct from classification and segmentation, object detection involves the dual challenge of identifying the object category (classification) and localizing its position in the image (regression). Noteworthy models have been introduced to tackle this problem, such as YOLO by (Redmon et al., 2016), Region-based CNN (R-CNN) by (Girshick et al., 2014), SSD: Single Shot MultiBox Detector by (Liu et al.,2020), Focal for Dense Object Detection by (Lin et al.,2017), and Fast R-CNN by (Wang et al., 2017). These models have demonstrated their efficacy in various object detection scenarios.

It is important to highlight that CNN architectures have demonstrated remarkable robustness and effectiveness in tasks involving object segmentation, detection, and classification. However, these architectures often demand a substantial amount of training data to effectively learn the multitude of parameters and deep layers necessary for achieving generalization. To address the challenge of limited data availability, several approaches have been introduced in the literature. These include Transfer Learning (Shao et al., 2015), Customizing Layers (Sledevic et al., 2019), Optimizing Hyperparameters (Diaz et al., 2017), and Data Augmentation (Shorten et al., 2019). These techniques aim to enhance the performance of CNNs by leveraging existing knowledge, tailoring the network architecture, fine-tuning hyperparameters, and artificially expanding the training dataset. The evolution of deep CNN architectures can be visually depicted, as illustrated in Figure 5.13, showcasing the advancements and refinements made in the field.

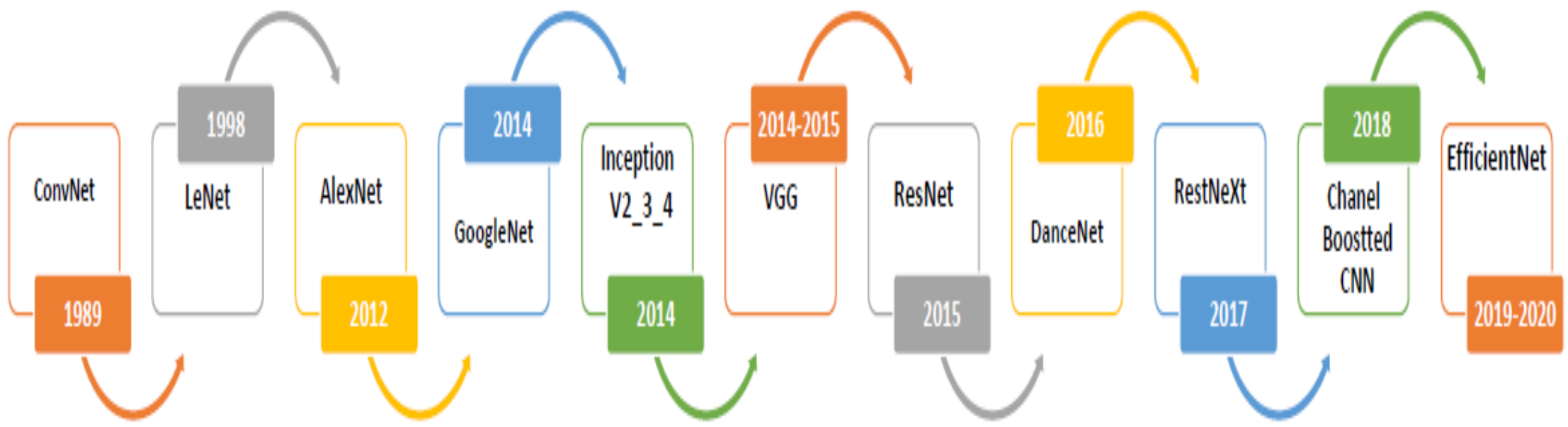

Figure 5.13 CNN evolution from ConvNet to current architectures

### 5.2.6 Comparison of popular CNN frameworks

CNN frameworks are crucial components in projects that employ the CNN architecture, contributing significantly to the development of advanced tools that streamline complex programming challenges and offer enhanced expertise. The availability of a diverse range of frameworks has provided researchers with a multitude of options and platforms to conduct experimental studies in the field of deep learning (Shorten et al., 2019). These frameworks are meticulously designed, each with its unique characteristics and tailored for specific purposes, empowering researchers to explore and exploit the potential of CNNs in their respective domains. By leveraging these frameworks, researchers can unlock new possibilities and make substantial advancements in the field of deep learning.

From faster model training and deployment to debugging and profiling capabilities, the use of CNN frameworks provides a distinct advantage over the traditional methods used for building deep learning models, enabling researchers to speed up and refine the development process, resulting in models that are faster and more accurate than ever before. This, in turn, has revolutionized the research and development of deep learning applications by allowing for greater agility and speed in prototyping new models, as well as providing new insights that weren't possible with previous methodologies.

Therefore, using CNN frameworks is transforming the research and development process, enabling researchers to create DL models more quickly and accurately than before and leading to deeper understanding and innovative solutions to the problems that deep learning seeks to address. In this context, it is essential to compare the different deep learning frameworks available for implementing CNNs to choose the one that best meets the project's requirements. The most well-known CNN frameworks are summarized in Table 5.2.

Table 5.2 CNN framework comparison

| Framework | Programming Language | Open Source | Operating System Compatibility | Interface |
|---|---|---|---|---|
| Torch (Ferentinos et al., 2018) | C, Lua | Yes | Linux, Windows, macOS | C, C++ |
| Caffe (Jia et al., 2014) | C++ | Yes | Linux, Windows, macOS | Python C++, MATLAB |
| Keras (Chollet et al., 2017) | Python | Yes | Linux, Windows, macOS | Python |
| TensorFlow (Abadi et al., 2016) | Python, C++, CUDA | Yes | Linux, Windows, MacOS | Java, Python, JavaScript |
| Matlab Toolbox (Kim et al., 2017) | MATLAB, C++, C, Java | No | Linux, Windows, MacOS | MATLAB |
| Deeplearning4j (Amara et al., 2017) | Python, Java | Yes | Linux, Windows, MacOS | Python Java, Clojure |

When choosing a deep learning framework for implementing CNNs, it is important to consider the project's specific needs, such as the dataset's size, the CNN model's complexity, and the required computational resources. It is also important to consider personal preference and familiarity with the framework, as this can impact development speed and ease of use.

For example, if the project requires high computational performance and the use of GPUs, PyTorch may be a good choice as it offers built-in support for CUDA. On the other hand, if the project involves building and training complex CNN models, TensorFlow's low-level APIs may provide more flexibility and control. On the other hand, if the project requires quick prototyping and easy-to-use high-level APIs, Keras may be a good choice.

Ultimately, the choice of deep learning framework should be made based on the project's specific needs and the developers' skills and experience. It is also worth considering the availability of resources and community support for the chosen framework, as this can impact the speed of development and the quality of the final model.

DL advancements supply various tools and platforms for implementing CNN in various applications, such as agriculture. As a result, the framework described above has been used in various agricultural applications and studies, each tailored to the study conditions of the author, project size, dataset, and complexity. These applications range from the use of CNN for crop classification and disease identification to the monitoring and optimization of water resources in agricultural production, providing much-needed decision support for farmers and agricultural production; in each of these applications, CNN has been used to optimize the decision-making process and provide actionable results. In conclusion, using convolutional neural network frameworks has opened a new realm of possibilities in agricultural applications, allowing for large datasets and fast and accurate processing.

### 5.2.7 Comparison of popular CNN models

Convolutional neural networks (CNNs) have revolutionized computer vision, becoming essential for various image-processing tasks. There are several popular CNN models available, each with its collection of advantages and drawbacks. These models have been designed over the years to address diverse computer vision tasks such as image segmentation, object detection, and classification. In this context, it is essential to compare the different deep learning frameworks available for implementing CNNs to choose the one that best meets the project's requirements. Here is a brief comparison of some of the most commonly used CNN models:

1. **LeNet-5:** LeNet-5, pioneered by Yann LeCun in the 1990s, stands as one of the earliest CNN models to emerge. It consisted of seven layers and was originally designed for handwriting recognition. While LeNet-5 is less accurate than newer models, it is still used as a baseline for comparison (Lecun et al., 1998).
2. **AlexNet:** AlexNet introduced by Alex Krizhevsky in 2012, and was the first CNN to win the ILSVRC. It consists of eight layers and utilize several new methods, such as ReLU and dropout, to improve accuracy (Krizhevsky et al., 2012).
3. **VGGNet:** VGGNet, a noteworthy contribution from the University of Oxford in 2014. It comprises up to 19 layers and uses small 3x3 filters to improve accuracy. VGGNet is known for its simplicity and ease of implementation (Simonyan et al., 2014).
4. **GoogLeNet/Inception:** GoogLeNet, also known as Inception v1, was developed in 2014 by Google. comprising 22 layers, and uses several novel techniques, such as inception modules and auxiliary classifiers, to improve accuracy while reducing the number of parameters (Szegedy et al., 2015).
5. **ResNet:** ResNet, or residual network, was developed by Microsoft in 2015. It consists of up to 152 layers and uses residual connections to address the vanishing gradient problem. ResNet is currently one of the accurate CNN models availabl (Alto et al., 2015).
6. **MobileNet:** a notable CNN architecture, was presented by Howard et al. in their paper published in 2017 (Howard et al., 2017). It is specifically optimized for mobile devices that have limited computing power, and it utilizes depthwise separable convolutions, which exist more efficiently compared to traditional convolutions. MobileNet has achieved remarkable performance on the ImageNet dataset, with a top-5 error rate of just 5.6%, making it a state-of-the-art solution (Howard et al., 2017).
7. **EfficientNet**: EfficientNet was introduced in 2019 by (Tan et al., 2019). It is known for its efficiency and effectiveness and uses a novel scaling method that balances width, depth, and resolution, attains superior accuracy while utilizing fewer parameters. EfficientNet has demonstrated cutting-edge performance, achieving an exceptional top-5 error rate of merely 1.7%.

Choosing the appropriate CNN model relies on the specific demands and criteria of the project at hand, including factors such as accuracy, speed, and available resources. However, AlexNet, VGGNet, EfficientNet, and ResNet are currently the most popular and widely used CNN models. Other popular CNN models include InceptionNet and DenseNet, each with their unique architecture and strengths in different areas of computer vision project. Thus, it is essential to consider the trade-offs between accuracy and speed when selecting a CNN model for a particular project. (Canziani et al., 2016) give a great comparison of various CNN architectures, as seen in Figure 5.14. The vertical axis displays ImageNet classification accuracy in the top 1. The number of procedures to classify an image is displayed on the horizontal axis. The size of the network's parameter count is directly correlated with the size of the circle.

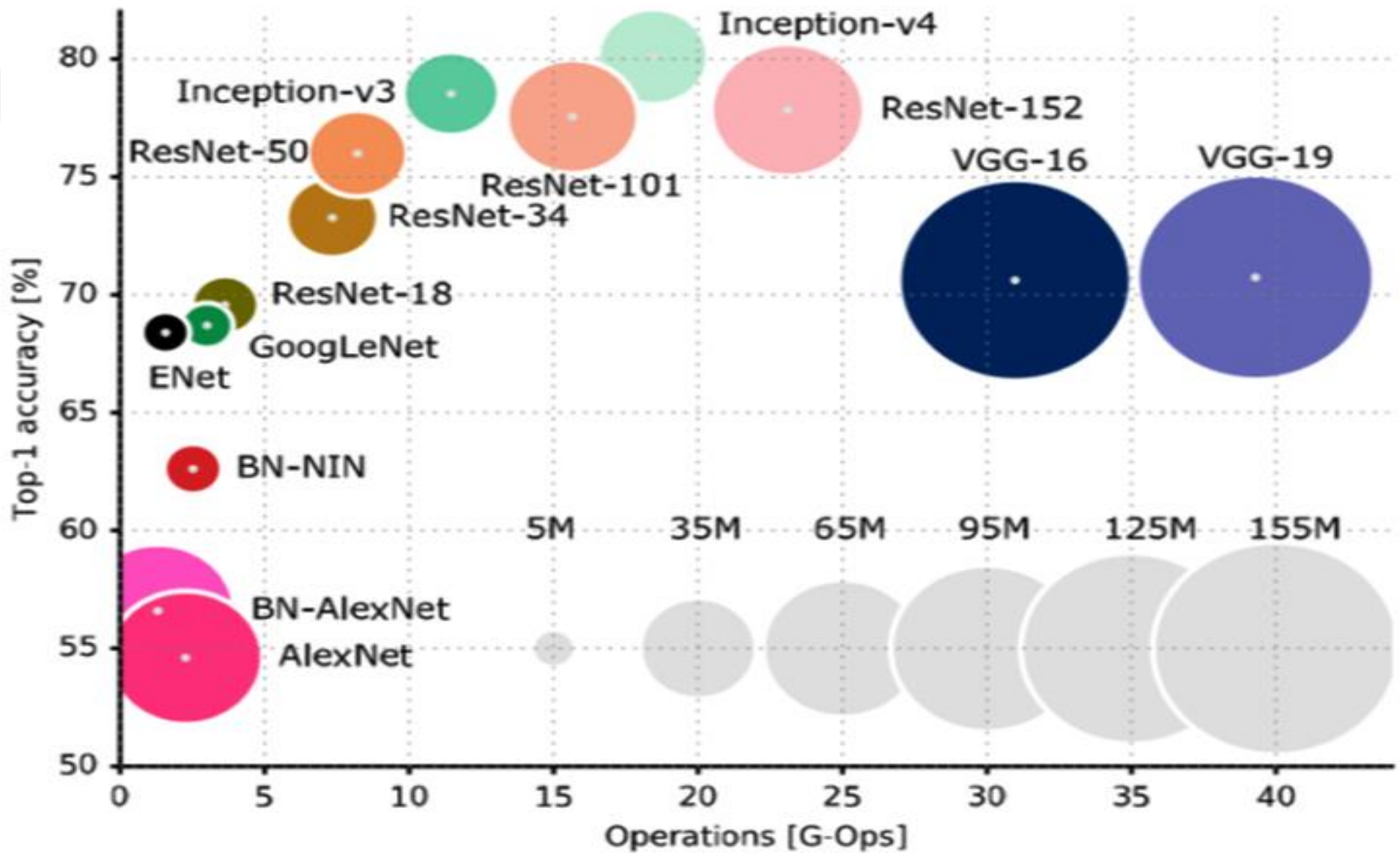

Figure 5.14 Comparison of the most common CNN architectures

This comparison can be useful for choosing an appropriate CNN architecture for a particular application, depending on the available computational resources and the desired level of accuracy. The study also highlighted the importance of efficient architectures that can achieve high accuracy with fewer parameters and operations, which is essential for real-time applications, including autonomous driving and robotics.

### 5.2.8 Interpretability and explainability of deep learning models

Interpretability and explainability of DL models are critical in plant leaf disease detection because understanding how a model makes its predictions can provide valuable insights into the underlying biological processes and inform decision-making in agriculture (Srinivasan et al., 2013). In the context of leaf disease detection, transparency is essential, as the decisions made by these models may have a significant effect on agricultural practices and food security. In order to ensure that these models are used responsibly and ethically, it is essential to provide transparency in the decision-making process, including access to the data used, the algorithms employed, and the factors that influence the model's output.

One approach to increasing the interpretability and explainability of DL models in plant leaf disease detection is using visualization techniques. For example, saliency maps can visualize which parts of an image the model is paying attention to when making its predictions. Furthermoe, this can help identify which features are most important for disease detection and can provide insights into the underlying biological mechanisms. Another approach is to use explainable AI (XAI) techniques to make deep learning models more transparent and interpretable. For example, LIME (Local Interpretable Model-Agnostic Explanations) is a popular XAI technique that can explain individual predictions by generating a local, interpretable model around each prediction.

Interpretability and explainability are also crucial for ensuring that DL models in detecting leaf diseases are fair and unbiased. For example, if a model is biased towards certain types of plant diseases, it can lead to incorrect diagnoses and ineffective treatment. By understanding how a model arrives at its predictions, researchers can identify and correct biases in the model.

Overall, increasing the interpretability and explainability of deep learning models in leaf disease detection can lead to more effective and transparent models, ultimately improving agriculture decision-making and leading to better plant health.

Interpreting deep learning models can be challenging due to their intricate nature of the models and large amount of data they are trained on. Some of the primary obstacles in comprehending DL models include the following:

1. **Black-box nature**: Deep learning models are frequently referred to as "black boxes", meaning their internal workings are not readily understandable or explainable. Furthermore, this can make understanding why the model made a particular decision or prediction difficult.
2. **Non-linear and high-dimensional models**: Deep learning models are often non-linear and high-dimensional, making it challenging to identify the most important features or patterns the model uses to make decisions.
3. **Lack of interpretability tools**: There is a lack of standard tools and techniques for interpreting deep learning models. While various approaches have been proposed, it can be difficult to know which approach is best suited for a particular task or model.
4. **Balance between performance and interpretability**: There is frequently a trade-off among a deep learning model's performance and interpretability. More interpretable models may sacrifice some accuracy, while highly accurate models may be less interpretable.
5. **Overfitting**: DL models could be susceptible to overfitting, wherein the model becomes excessively complex and starts to incorporate noise from the training data. This can make understanding the model's true underlying patterns and features difficult.

Researchers are developing new approaches for interpreting deep learning models to overcome these challenges, including visualization techniques, sensitivity analysis, activation maximization, gradient-based methods, and model-specific interpretation methods. Additionally, a growing interest is in developing more interpretable deep learning models that can balance accuracy with interpretability. Ultimately, improving the interpretability of deep learning models will be important for building trust and understanding in using these models in various applications, including plant leaf disease detection. By using these techniques, researchers can gain insights into the behavior of deep learning models and increase their interpretability. Furthermore, this can help build trust and understanding in using these models in various applications, including plant leaf disease detection.

### 5.2.9 Pre-Trained network

A pre-trained model is one that was created previously by someone else and trained on an extensive dataset to address a minor issue. Instead of starting from scratch, AI users can use and implement a pre-trained model as a preliminary step that they do not have to invest time, energy, and resources in creating their model from scratch and can instead focus on fine-tuning the pre-trained model for their use case. Pre-trained models can therefore be seen as an efficient shortcut to achieving better results with less effort. The idea behind pre-training is to involve utilizing a pre-existing network architecture that has been trained on a vast dataset, then fine-tuning the model utilizing a smaller dataset relevant to the current task. However, it is still important to remember that while they provide a starting point, they require tweaking and fine-tuning to successfully apply in any given use case. Nevertheless, pre-trained models are a great asset in artificial intelligence, and they have the potential to significantly accelerate the development of AI solutions, allowing users to focus on the critical task of customizing their AI system for their own needs.

Pre-trained models are potent tools for AI users, allowing them to utilize the knowledge and experience of experienced AI developers to quickly and efficiently build upon existing models, rather than starting from scratch. The key steps to utilize a pre-trained network are as follows:

1. **Identify the task:** The first step is identifying the task we want to solve including, object detection, image classification or sentiment analysis.

2. **Choose a pre-trained model:** Once we have identified the task, we must choose a pre-trained model trained on a common basis or dataset. There are numerous pre-trained models available, and the model chosen will be determined by the task at hand as well as the size of the dataset.
3. **Fine-tune the model:** After choosing a pre-trained model, we must fine-tune it on our specific dataset. This entails retraining the model's last few layers on our data while preserving the earlier frozen layers. This enables the model to adapt to the unique characteristics of our data and improve its accuracy.
4. **Evaluate the model**: Once it has been fine-tuned, we must evaluate its performance on a validation dataset. This will give us an insight into how well the model performs and whether further adjustments are needed.
5. **Use the model:** We can predict new data once we are satisfied with its performance. The pre-trained model will have learned the relevant features and patterns from the original dataset and can be used to predict new data with similar features.

A pre-trained network model can save time and effort in developing machine learning applications. However, choosing the right model and fine-tuning it appropriately is important to achieve the best possible performance on our specific task.

Various researchers used pre-trained model architectures, including convolutional neural networks (CNNs), to develop innovative artificial intelligence (AI) solutions across various domains, including agriculture. As an illustration, (Naik et al., 2022) deployed a dozen distinct deep-learning models to categorize five different types of chili diseases. The VGG19 model attained the highest precision score of 83.54%, sans data augmentation. Conversely, DarkNet53 demonstrated the most remarkable outcome with data augmentation (accuracy = 98.63%). Furthermore, in their study, Partel et al. (2020) investigated the utilization of various models, including YOLO-v3 (Redmon et al., 2018), Faster R-CNN (Ren et al., 2015), ResNet-101, ResNet-50, and Darknet-53 (Redmon et al., 2018), to develop an intelligent sprayer capable of performing real-time control on plants.

A groundbreaking study by (Sahu et al., 2021) introduced a unique methodology for effectively classifying and identifying diseases in bean crops. Their research highlighted the advantages of fine-tuning pre-trained networks over training models from scratch. By fine-tuning hyperparameters, the accuracy of GoogleNet improved significantly from 90.1% to an impressive 95.31%, while VGG16 exhibited notable progress from 89.6% to 93.75%. These findings underscored the substantial impact of transfer learning, demonstrating the network's ability to leverage learned features across different problem domains. In a parallel study, Mukti et al. leveraged transfer learning techniques using ResNet50 to identify plant leaf diseases (Mukti et al., 2013). Their extensive dataset comprised a staggering 87,867 images, which were carefully split into 80% for training and 20% for validation. Remarkably, their approach achieved an outstanding accuracy of 99.80%, highlighting the efficacy of transfer learning in this domain. Moreover, (Arya et al., 2019) conducted an intriguing investigation into the potential of convolutional neural networks (CNNs) to detect plant diseases. They compared the performance of various CNN architectures, including AlexNet and a shallow CNN, in detecting diseases in potato and mango leaves using pre-trained models. The results demonstrated the superior accuracy of the AlexNet approach, achieving an impressive 98.33%, compared to the shallow CNN's accuracy of 90.85%.

### 5.2.10 Training from scratch

Training from scratch in deep learning has gained popularity as a powerful tool for solving complex problems in recent years. Training from scratch necessitates a developer collecting a large labeled data set and configuring a network architecture capable of learning the features and model. This technique is particularly useful for new applications with many output categories. However, it is a less common approach because t necessitates a large amount of data, causing training to take days or weeks. Despite this, training from scratch can provide the highest accuracy and precision in specific applications and is especially useful in applications where large data sets are available, such as medical imaging and natural language processing. Despite its complexity and challenges, training from scratch is often the best option for applications that require precision and accuracy, as it allows for greater accuracy in predicting the output categories. Training from scratch in deep learning can also have advantages over transfer learning regarding flexibility. It applies to many domains and can produce better results with limited data.

Furthermore, "training from scratch" is an approach to deep learning that involves training a model from raw data with no prior knowledge of the data required, allowing the model to learn and identify patterns and insights within the data.

This approach has many advantages, as it allows for a more generalizable model to be created that can better adapt to data from various domains and more tailored models to be created that can recognize intricate patterns in the data. These advantages, combined with the fact that it does not necessitate previous knowledge of the data and applies to many different domains, makes training from scratch a highly desirable approach for deep learning applications. The most important steps to use training from scratch in a deep learning model are:

1. **Data Collection and Preprocessing:** Collecting relevant and sufficient data for the problem and preprocessing it to make it suitable for training. This may involve data normalization, cleaning, augmentation, and dividing into training and validation sets.
2. **Model Architecture:** Choosing the right architecture for the model, which involves deciding the number and type of layers, as well as the activation functions employed. The choice of architecture is heavily influenced by the nature of the problem at hand and the intricacy of the data involved.
3. **Hyperparameter Tuning:** Setting the model's hyperparameters, which include learning rate, number of epochs, batch size, regularization techniques, and optimization algorithms. These hyperparameters can significantly affect the training process and also the model's performance.
4. **Training the Model:** Training the model requires providing the training data to the model and adjusting the model's weights iteratively until the model could accurately predict the target which is variable. Also, this may require several iterations and adjustments of the hyperparameters to achieve the desired results.

5. **Evaluating the Model:** After training the model, it must be evaluated on a validation dataset to ensure it can generalize well to new data. This may involve comparing the predicted and actual values utilizing metrics inlcuding accuracy, F1 scores, recall, and precision.
6. **Deployment:** After successfully training and validating the model, it is ready to be deployed for making predictions on new, unseen data. This may involve integrating the model into an application or website or creating an API for others.

Training a deep learning from scratch can be an arduous and intricate process. It demands a thorough understanding of the problem, the available data, and the model architecture, as well as meticulous tuning of the hyperparameters to attain optimal performance.

Various researchers have utilized model architectures trained from scratch, such as (Lu et al., 2017) who developed a novel DCNN-based method for detecting rice diseases. Their proposed model was trained from scratch on a dataset consisting of 500 images of both healthy and unhealthy rice leaves collected from rice fields. Through a 10-fold cross-validation approach, their CNN model achieved an impressive accuracy of 95.48% in identifying ten common rice diseases. Notably, the CNN model outperformed traditional machine-learning models by a significant margin. Similarly, (Zhang et al., 2018) suggested a CNN model with six layers and three fully connected layers to detect broad-leaf weeds. By conducting a comparative analysis between CNN and SVM, it was revealed that their proposed CNN model achieved an impressive accuracy of 96.88% in identifying weeds, surpassing the accuracy of 89.4% achieved by the SVM model. These findings conclusively demonstrate the superior performance of the CNN model over the SVM model in detecting leaf weeds in pastures.

In their research, (Milioto et al., 2017) developed a CNN model for wise discrimination that integrated vegetation identification and accurate plant identification for essential weeds. They trained the model using multi-spectral data and evaluated its performance on images from diverse beet fields. Through the analysis of various combinations of convolutional and fully connected layers, the team successfully created an efficient and reliable model. Remarkably, their model achieved superior performance by utilizing three convolutional and two fully connected layers, surpassing the performance of other models. It is noteworthy that the team did not rely on geometric priors, such as crop row patterns, to achieve these exceptional results. Similarly, (Dyrmann et al., 2016) employed a CNN model to detect plant species in color images encompassing 10,413 early-stage crop species. They devised a novel system that incorporated max-pooling layers, convolutional layers, fully connected layers, batch normalization, activation functions, and residual layers. The network attained an impressive classification accuracy of 86.2%, highlighting the effectiveness of a meticulously designed model in accurately identifying plant species.

In the realm of plant disease detection, a team of researchers (Pearlstein et al., 2016) embarked on a study that explored the utilization of CNNs. Their approach involved training the model on synthetic image data and subsequently evaluating its performance on real-world data. The CNN architecture they employed consisted of five convolutional layers and two fully connected layers, which demonstrated the model's ability to accurately identify crop plants, even in the presence of occlusions. In a separate study by (Nkemelu et al., 2018), the focus shifted to plant seedling classification using CNNs. The researchers conducted a comparative analysis between CNNs and traditional algorithms such as K-Nearest Neighbor (KNN) and Support Vector Machine (SVM). Remarkably, the CNN outperformed both methods, achieving an impressive accuracy of 92.6%, while KNN and SVM scored 56.84% and 61.47%, respectively. The authors employed a CNN architecture consisting of 6 convolutional and 3 fully connected layers, and they assessed the model's performance on both original and pre-processed images.

To summarize, training from scratch in deep learning was widely used in various applications. It has become a powerful tool for solving complex problems, leading to improved performance, especially in applications based on CNN. In addition, training from scratch in deep learning has become an essential and popular research topic due to its ability to improve performance in diverse tasks including image classification, object detection, language translation, and various other applications.

# 6. FUTURE DIRECTIONS OF DEEP LEARNING IN AGRICULTURE

As per the latest research, CNN has been utilized to discuss various agricultural-related problems. However, there is still an unexplored scope for the implementation of CNN in other agricultural-related issues such as crop phenology, water erosion assessment, seed identification, leaf and soil nitrogen content, herbicide use, food diseases or defects, and detection of plant water stress. Additionally, there are many other potential areas where CNN can be used effectively, such as utilizing aerial imagery through drones to monitor the efficiency of seed production, improving the quality of wine production by harvesting crops at the optimum maturity level, monitoring animals and their movements for overall welfare and detecting diseases. Moreover, CNN can also be applied to various computer vision scenarios in agriculture.

There is a growing need for advanced models in many research fields, including environmental informatics. In particular, previous studies in this area could have benefited from the use of more advanced models like Recurrent Neural Networks (RNN) or Long Short-Term Memory (LSTM) architectures. These models possess the capability to exhibit dynamic temporal behavior, allowing them to remember and forget information over time or when needed. In the context of environmental informatics, such models could be used to better understand climate change, forecast weather conditions and phenomena, estimate the environmental impact of physical or artificial processes, and more. By leveraging the capabilities of these advanced models, researchers can gain deeper insights into complex environmental phenomena and make more informed decisions about environmental management and policy. In addition, such models can learn from the past and use this information to make predictions or decisions, allowing environmental informatics to approach complex problems more accurately while also considering the temporal dependencies of the processes involved. By employing these more advanced models, environmental informatics can achieve unprecedented levels of accuracy in predicting and responding to climatic changes. Thus, environmental informatics can benefit significantly from the use of more advanced models, icluding Recurrent Neural Networks (RNN) or LSTM architectures. These models excel in capturing the temporal dynamics of climatic data and integrating the behaviors of multiple environmental processes into a unified model, allowing the user to make better-informed decisions.

Deep learning algorithms can be employed in smart agriculture to monitor the crops' water and temperature levels. Additionally, farmers can view their fields from any location in the world. This intelligent agriculture powered by DL is quite effective. With deep learning, farmers can receive detailed information about their crops, which allows them to be better prepared for droughts, floods, and other potential disasters. It also provides an opportunity to maximize their yield and gain higher profits. Furthermore, DL can also be used to detect pests and diseases in the crops, thus helping to reduce the use of potentially harmful chemical pesticides, thereby making agricultural production more sustainable and environmentally friendly. In addition to all these applications, deep learning also can transform the way we produce food and manage our natural resources. Here are some additional deep learning applications in agriculture in the future:

1. **Precision Agriculture:** Deep learning algorithms can analyze data from sensors, satellites, and drones to monitor crop growth, soil health, and water use. This can help farmers optimize their inputs and improve yields while reducing waste.
2. **Crop Disease Detection:** Deep learning models have proven to be highly effective in detecting plant diseases with remarkable accuracy and speed. This capability empowers farmers to promptly respond and mitigate the further spread of diseases, leading to reduced crop losses and better crop management.
3. **Crop Yield Prediction**: Deep learning algorithms can effectively estimate agricultural yields by analyzing data from numerous sources, such as weather patterns, soil conditions, and plant genetics. This can help farmers plan their harvests and optimize their production.
4. **Climate Change Adaptation:** Deep learning can analyze large amounts of data on weather patterns and climate change to predict how crops will respond to changing conditions. This can help farmers adapt their practices and plant suitable crops for a changing climate.

5. **Food Security:** Deep learnng can help optimize food production, reduce waste, and improve the quality of crops, which can help increase food security for a growing global population.

Deep learning has the potential to transform agriculture by providing farmers with more precise, data-driven insights into their operations. By harnessing the power of deep learning, we can produce more food with less waste and environmental impact while ensuring food security for generations to come. In addition to DL applications in agriculture, AI technology has revolutionized the agricultural sector, offering diverse applications and transformative opportunities. For example, it may significantly reduce resource and labor problems and serve as a potential tool for organizations to manage the complexity of current agriculture. Therefore, big businesses need to start investing in this area. Figure 6.1 provides a review of the existing AI applications in agriculture

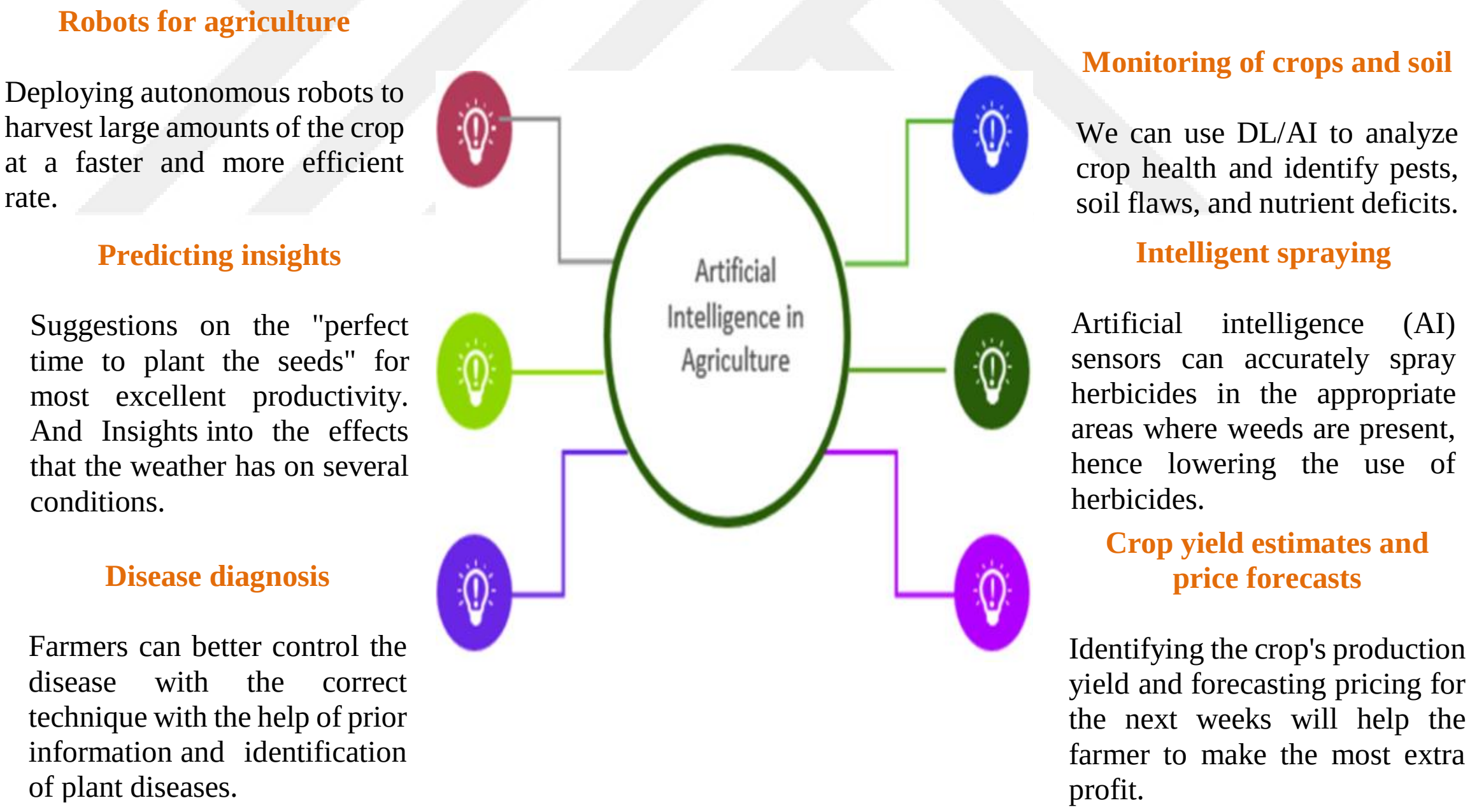

Figure 6.1 Artificial intelligence applications in Agriculture

It is clear from the figure that AI applications have become an integral part of the agricultural industry, enabling precision farming, yield optimization, and resource utilization. In addition, these applications can track soil health, identify weeds and other pests, measure crop growth, and more. However, there are many challenges to overcome before we can fully embrace the power of AI technology in our field. We need more data to train deep learning algorithms and identify patterns and trends from available data. We also need greater access to high-performance computing resources to run these algorithms and interpret the results meaningfully. And finally, we need to focus on developing better interfaces and workflows that allow the data sets produced by farmers to be more accessible for AI applications in the future. While these challenges seem daunting, we are optimistic that we will be able to overcome them in the coming works. Therefore, while AI has great potential to transform agriculture, our primary goal is to help farmers incorporate this technology into their daily lives in a way that is both beneficial and meaningful for them. By working together to leverage the full potential of AI, we are confident that we can help make agriculture more sustainable for future generations.

## 7. APPLICATIONS OF DEEP LEARNING IN AGRICULTURE: DETECTION OF PLANT LEAF DISEASES

Deep learning (DL) techniques have significantly advanced image analysis, particularly CNNs (Tugrul et al., 2022). Numerous studies focusing on the automatic detection of leaf diseases have been conducted; in these studies, deep learning techniques have proven to be highly successful for plant disease detection and classification, providing higher accuracy levels than traditional methods such as manual observation, morphological analysis, and digital image processing; this is mainly because deep learning technique such as CNNs can extract complex features from images, reducing the need for manual feature extraction and allowing the system to learn from data without the need for manual feature engineering. However, it is important to note that applying DL for the automatic identification of leaf diseases can be challenging due to the wide variety of plant species and the various conditions in which these plants may grow.

Deep learning has numerous applications in agriculture, and one of the most promising is detecting plant leaf diseases. Here are some specific applications of deep learning in this field:

1. **Early Detection:** Deep learning models could detect leaf diseases early, even before visible symptoms appear. This can help farmers take timely action to help limit the disease's propagation and decrease crop loss.
2. **Increased Accuracy:** Traditional methods of plant disease detection rely on expert visual inspection, which can be a laborious and subjective process. However, the use of deep learning models can achieve high levels of accuracy in detecting diseases and classifying leaves into multiple disease categories.
3. **Rapid Diagnosis:** Deep learning models can diagnose a plant leaf disease within a few seconds or minutes, compared to days for traditional methods. This can save farmers time and resources and increase crop management efficiency.
4. **Decision support system**: Deep learning algorithms may be employed to help farmers make decisions. The algorithms can analyze data from sensors and cameras in the field to recommend when and how to treat crops to prevent disease outbreaks.

5. **Reduced Costs**: Deep learning models can reduce the cost of disease detection by eliminating the need for expensive equipment and expert services. This can make disease detection more accessible to small farmers and increase the adoption of precision agriculture.
6. **Real-Time Monitoring:** Deep learning models can be integrated with drones or other sensors to monitor crops in real time. This can provide farmers with up-to-date information on the health of their crops and help them make informed decisions on crop management.

Numerous research studies have demonstrated the effectiveness of using deep learning models in detecting plant diseases, and the results have been promising. By combining various approaches and techniques, there is a tremendous opportunity for the development of automated imaging methods in the field of plant disease identification. These technologies have the potential to enhance farming methods, advance agricultural practices, and promote food security while minimizing manual labor, expenses, and time spent on disease diagnosis.

Over the past five years, the researcher has presented numerous approaches that utilize deep learning techniques, particularly CNNs, for identifying plant leaf diseases. Remarkably, most of these publications were released after 2016, underscoring the innovative and state-of-the-art nature of this approach in agriculture. Figure 7.1 illustrates the quantity of research papers published from 2013 to 2022, focusing on automatically detecting plant leaf diseases through deep learning models. The graph depicts a peak in in research focused on automatic disease detection in the year 2016, confirming the nascent and pioneering nature of the field.

The data was collected by performing a comprehensive keyword search across multiple databases, encompassing journal articles published from 2013 to 2022, including MDPI, Springer, Google Scholar, and ScienceDirect.

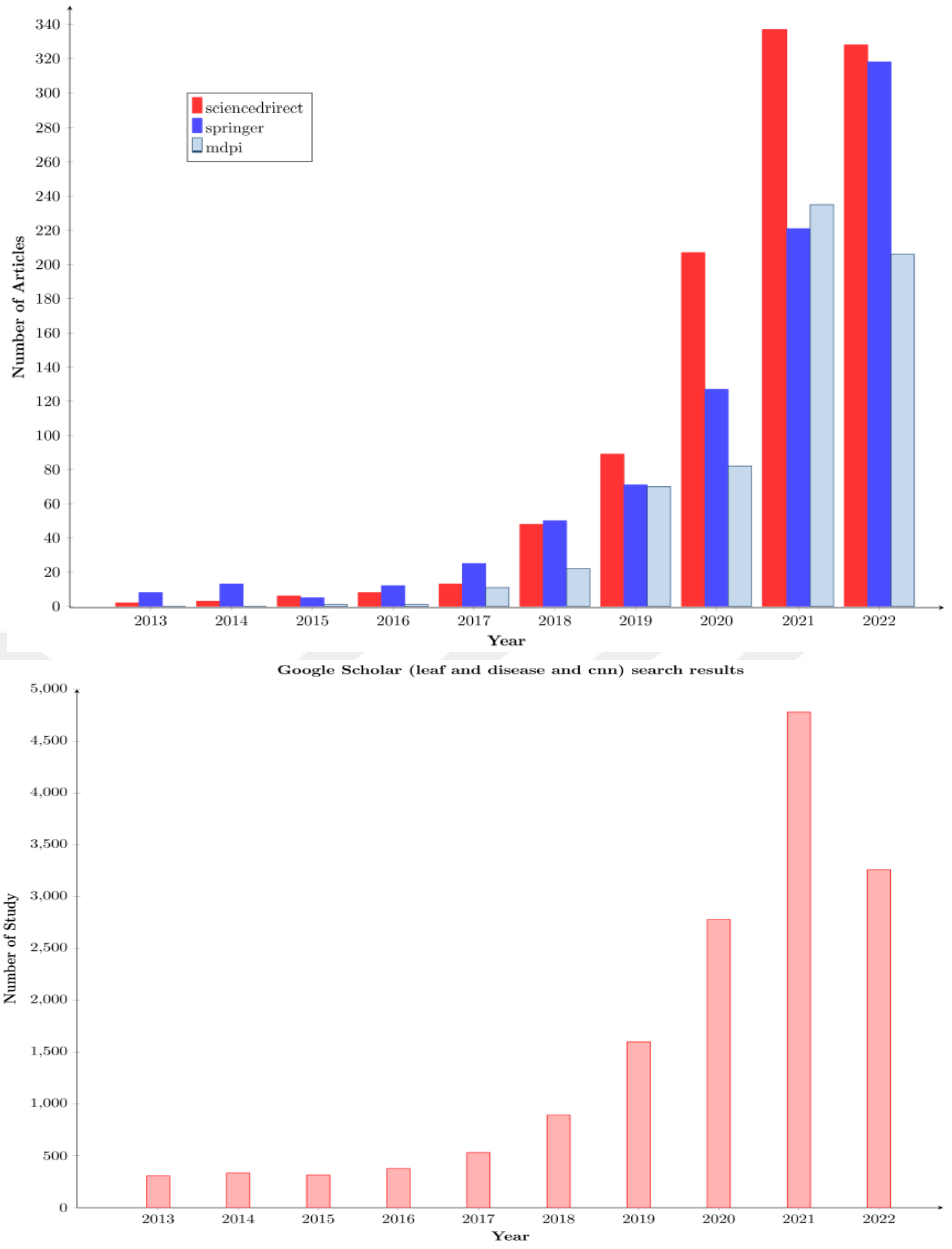

Figure 7.1 Number of articles published between 2013 and 2022 using DL-based models to identify plant leaf diseases

The thesis primarily focused on detecting and classifying plant diseases using image-based techniques. The author's pre-trained and custom deep learning models were thoroughly examined in this study. Furthermore, the models' efficacy was evaluated using a variety of datasets containing both healthy and diseased plant images, and their performance was measured using various accuracy metrics.

The study found that both pre-trained and custom DL models were able to identify plant diseases with a high level of precision, indicating that DL models are a promising tool for of plant diseases detection and providing researchers with an effective and method for identifying, classifying, and understanding the complexity of various plant diseases. Moreover, the results of this study show the potential for deep learning applications in agriculture, particularly the potential of CNN-based models in diagnosing and classifying plant diseases.

To facilitate the readers in their selection process and enable them to compare different DL models, Table 7.1 summarizes and clarifies some research examples and necessary details about deep learning applications in agriculture.

Table 7.1 DL methods comparison in the detection of plant leaf diseases

| Refference | Plant | Dataset | Model | Accuracy | Year |
|---|---|---|---|---|---|
| (Kawasaki el al., 2015) | cucume | self | CNN | 94.90 | 2015 |
| (Ioffe et al.,2015) | rice | self | CNN | 95.48 | 2015 |
| (Mohanty et al., 2016) | multiple | Plant Village | GoogLeNet | 99.35 | 2016 |
| ((Nachtigall et al., 2016) | apple | Plant Village | AlexNet | 97.30 | 2016 |
| (Lu et al., 2017) | wheat | self | VGG-FCN-VD16 | 97,95 | 2017 |
| (Lu et al., 2017) | rice | self | DCNN | 95.48 | 2017 |
| (Brahimi et al., 2017) | tomato | self | GoogLeNet | 99.18 | 2017 |
| (Liu et al., 2017) | apple | Plant Vilage | AlexNet | 97.62 | 2017 |
| (Amara et al., 2017) | banana | Plant Village | LeNet | 99.00 | 2017 |
| (Ramcharan et al. 2017) | cassava | self | Inception-v3 | 93.00 | 2017 |
| (Wang et al. 2017) | apple | Plant Village | VGG16 | 90.40 | 2017 |
| (Cruz et al., 2017) | olive | Plant Village | LeNet | 99.00 | 2017 |
| (Cruz et al., 2017) | potato | PlantVillage | VGG | 96.00 | 2017 |
| (Ha et al., 2017) | radish | self | VGG-A | 93.30 | 2017 |
| (Dang at al., 2017) | radish | self | GoogLeNet | 90.00 | 2017 |
| (Durmus et al., 2017) | tomato | Plant village | AlexNet | 95.60 | 2017 |
| (Ferentinos et al., 2018) | multiple | Plant Village | VGG | 99.53 | 2018 |

Table 7.1 DL methods comparison in the detection of plant leaf diseases (continue)

| Refference | Plant | Dataset | Model | Accuracy | Year |
|---|---|---|---|---|---|
| (Arivazhagan et al.,2018) | mango | self | CNN | 96.67 | 2018 |
| (Rangarajan et al., 2018) | tomato | Plant Village | AlexNet | 97.49 | 2018 |
| (Nandhini et al., 2018) | banana | self | CNN | 93.60 | 2018 |
| (Mukti et al., 2019) | wheat | self | ResNet-50 | 96.00 | 2019 |
| (Chen et al., 2019) | multiple | Plant Village | ResNet50 | 99.80 | 2019 |
| (Liang et al., 2019) | tea | self | LeafNet | 90.16 | 2019 |
| (Sibiya et al., 2019) | rice | self | Lenet5 | 95.83 | 2019 |
| (Howlader et al., 2019) | maize | Plant Village | CNN | 92.85 | 2019 |
| (Singh et al., 2019) | guava | self | DCNN | 98.74 | 2019 |
| (Mishra et al., 2019) | mango | self | MCNN | 97.13 | 2019 |
| (Fang et al.,2019) | multiple | Plant Village | DCNN | 88.46 | 2020 |
| (Darwish et al., 2020) | multiple | Plant Village | ResNet-50 | 95.61 | 2020 |
| (Mkonyi et al., 2020) | maize | Kaggle | VGG16 | 98.20 | 2020 |
| (Karlekar et al., 2020) | tomato | self | VGG16 | 91.90 | 2020 |
| (Yin et al., 2020) | soybean | self | CNN | 98.14 | 2020 |
| (Rangarajan et al., 2020) | pepper | self | ResNet50 | 88.38 | 2020 |
| (Ahmad et al., 2020) | eggplant | self | VGG16 | 99.40 | 2020 |
| (Liu et al., 2020) | plum | self | Inception-v3 | 92.00 | 2020 |
| (Vallabhajosyula et al., 2020) | grape | self | DICNN | 97.22 | 2020 |
| (Hassan et al., 2021) | multiple | Kaggle | CNN | 100.00 | 2021 |
| (Yadav et al., 2021) | multiple | Plant Village | EfficientNetB0 | 99.56 | 2021 |
| (Atila et al., 2021) | peach | self | CNN | 98.75 | 2021 |
| (Wang et al., 2021) | multiple | Plant Village | EfficientNet | 98.42 | 2021 |
| (Sahu et al., 2021) | cucumbe | self | Efficient-B5-SwinT | 99.25 | 2021 |
| (Subetha et al., 2021) | bean | Kaggle | GoogleNet | 93.75 | 2021 |
| (Indu et al., 2021) | apple | kaggle | VGG19 | 87.70 | 2021 |

Table 7.1 DL methods comparison in the detection of plant leaf diseases (continue)

| Refference | Plant | Dataset | Model | Accuracy | Year |
|---|---|---|---|---|---|
| (Ahmad et al., 2021) | tomato | Plant Village | AlexNet | 99.86 | 2021 |
| (Naik et al., 2021) | tomato | self | Inception v3 | 99.60 | 2021 |
| (Pandey et al., 2022) | chili | self | SECNN | 99.12 | 2022 |
| (Jin et al., 2022) | multipl | self | DADCNN-5 | 99.93 | 2022 |
| (Javidan et al., 2022) | grape | self | InceptionV1 | 96.13 | 2022 |
| (Zeng et al., 2022) | grape | Plant Village | GoogleNet | 94.05 | 2022 |
| (Yu et al., 2022) | maize | Plant Village | GhostNet | 92.90 | 2022 |
| (Zhang et al., 2022) | maize | Plant Village | LDSNet | 95.40 | 2022 |
| (Wei et al., 2022) | apple | kaggle | Resnet | 95.80 | 2022 |
| (Pandey et al., 2022) | chili | Plant village | SECNN | 99.28 | 2022 |
| (Pandey et al., 2022) | apple | Plant village | SECNN | 99.78 | 2022 |
| (Pandey et al., 2022) | maize | Plant village | SECNN | 97.94 | 2022 |
| (Pandey et al., 2022) | pepper | Plant village | SECNN | 99.19 | 2022 |
| (Pandey et al., 2022) | potato | Plant village | SECNN | 100.00 | 2022 |
| (Pandey et al., 2022) | tomato | Plant village | SECNN | 97.90 | 2022 |
| (Hanh et al., 2022) | soybean | self | R-CNN | 83.84 | 2022 |
| (Ravi et al., 2022) | multiple | kaggle | Resnet | 99.89 | 2022 |
| (Li et al., 2022) | multiple | Plant Village | EfcientNet-B3 | 98.91 | 2022 |
| (Sun et al., 2022) | cassava | kaggle | CNN | 87.00 | 2022 |
| (Jiang et al., 2022) | apple | self | ConvVIT | 96.85 | 2022 |
| (Memon et al., 2022) | multiple | kaggle | EfficientNet | 99.70 | 2022 |
| (Chen et al., 2022) | wheat | Plant Village | Inception-v3 | 92.53 | 2022 |
| (Russel et al., 2022) | cotton | self | CNN | 98.53 | 2022 |
| (Russel et al., 2022) | cassava | self | ResNet-50 | 89.70 | 2022 |
| (Russel et al., 2022) | multiple | Plant village | CNN | 98.61 | 2022 |
| (Gaikwad et al., 2022) | multiple | Mepco Tropic Leaf | CNN | 90.02 | 2022 |
| (Prabu et al., 2022) | multiple | self | AlexNet | 86.85 | 2022 |
| (Kurmi et al., 2022) | mango | self | MobilenetV2 | 99.43 | 2022 |
| (Nagi et al., 2022) | pepper | Plant Village | CNN | 95.80 | 2022 |

Table 7.1 DL methods comparison in the detection of plant leaf diseases (continue)

| Refference | Plant | Dataset | Model | Accuracy | Year |
|---|---|---|---|---|---|
| (Nagi et al., 2022) | potato | Plant Village | CNN | 94.10 | 2022 |
| (Nagi et al., 2022) | tomato | Plant Village | CNN | 92.60 | 2022 |
| (Subramanian et al., 2022) | grape | Plant Village | CNN | 98.40 | 2022 |
| (Gajjar et al., 2022) | maize | Kaggle | InceptionV3 | 99.66 | 2022 |
| (Xu et al., 2022) | multiple | self | CNN | 96.88 | 2022 |
| (Singh et al., 2022) | multiple | Plant Village | CNN | 99.86 | 2022 |
| (Ruth et al., 2022) | maize | Plant Village | AlexNet | 99.16 | 2022 |
| (Pandian et al., 2022) | multiple | Kaggle | CNN | 99.00 | 2022 |
| (Pandian et al., 2022) | multiple | Plant Village | CNN | 98.41 | 2022 |
| (Borhani et al., 2022) | multiple | Plant Village | DCNN | 99.79 | 2022 |
| (Yakkundimath et al., 2022) | wheat | Plant Village | ResNet152 | 95.00 | 2022 |
| (Wu et al., 2022) | rice | self | VGG16 | 92.24 | 2022 |

As shown in Table 7.1, Much research has been conducted on the application of deep learning methods in agriculture. However, most researchers use similar model architectures and reach similar experimental outcomes. As a result, further research is necessary to address new requirements and conduct experiments with additional datasets and novel architectures. Without such efforts, there is a risk of duplicating existing work, making it imperative to explore new avenues and push the boundaries of the field. The comparison of model architectures demonstrates that experimental settings, datasets, and data size influence the choice of suitable DL techinque and model. In addition to the application of DL in agriculture, Figure 7.2 gives an insight into the use of different technological applications in detecting plant leaf diseases.

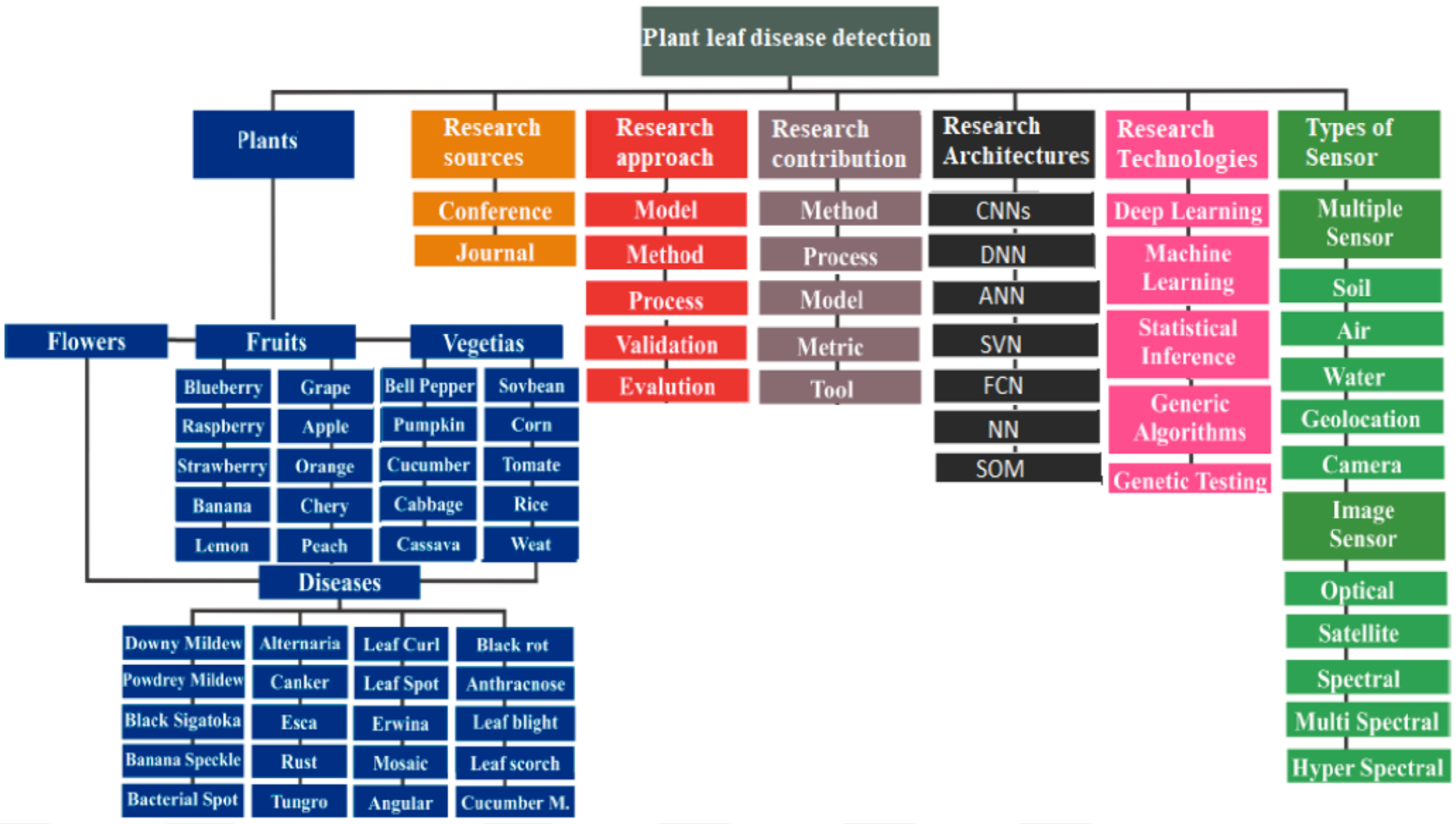

Figure 7.2 Examples of many technology uses for identifying plant leaf diseases

Deep learning applications in plant disease area are generally based on three critical points: classification, detection, and segmentation. Classification involves sorting different types of plant diseases into distinct categories based on their unique characteristics. Detection requires identifying the presence of a disease using various techniques such as image analysis, spectral analysis, or machine learning algorithms. Segmentation involves separating diseased regions of a plant from healthy regions for a more detailed analysis of the disease's spread and severity. The classification of plant diseases can be improved by leveraging deep learning algorithms including convolutional neural networks that could recognize complex patterns and features. Detection can be enhanced by combining various sensors and technologies, such as hyperspectral imaging, thermal sensing, and drones. Finally, segmentation can benefit from using advanced image processing techniques such as watershed segmentation or deep learning-based algorithms. Figure 7.3 shows an example of each of these three important applications.

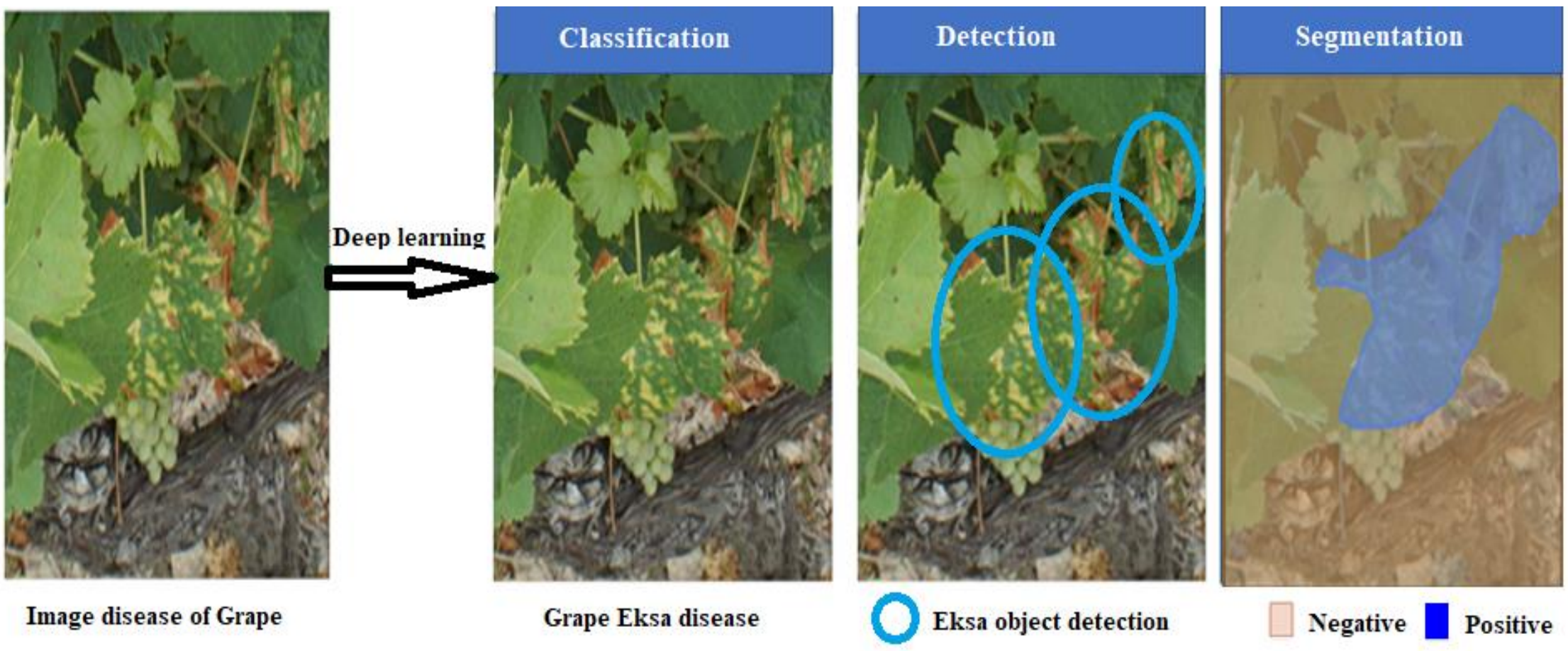

Figure 7.3 Example of deep learning application in plant diseases areas

Plant leaf disease detection and classification involves several stages, including data collection, preprocessing, training with deep learning algorithms, and ultimately displaying the results using visual aids or diplomas. The accuracy and reliability of the classification model are determined by input data number and the chosen features. In addition to data preparation and training, the effectiveness of leaf disease detection methods can also be improved by incorporating advanced technologies such as deep learning. These techniques enable more accurate and automated analysis of large datasets, which will help in simplifying the task of detecting and managing different plant diseases for researchers.

During the data preparation stage, the relevant features of the plant leaves, such as shape and color, must be extracted using computer vision techniques. Once the data is ready, an appropriate deep-learning algorithm can be trained and tested to classify plant leaf diseases accurately. When the trained model is deployed, it can be integrated into a mobile app or web tool that allows farmers and researchers to quickly identify and respond to outbreaks of plant diseases utilizing simple images of the affected leaves. However, it is important to continue updating the model with new data to improve accuracy and account for emerging diseases, Figure 7.4 depicts several stages used to apply deep learning in the plant disease area. The first stage involves collecting a large dataset images including healthy and diseased plants. The next stage entails training a DL model on these datasets and, finally, the model deployment to identify plant diseases.

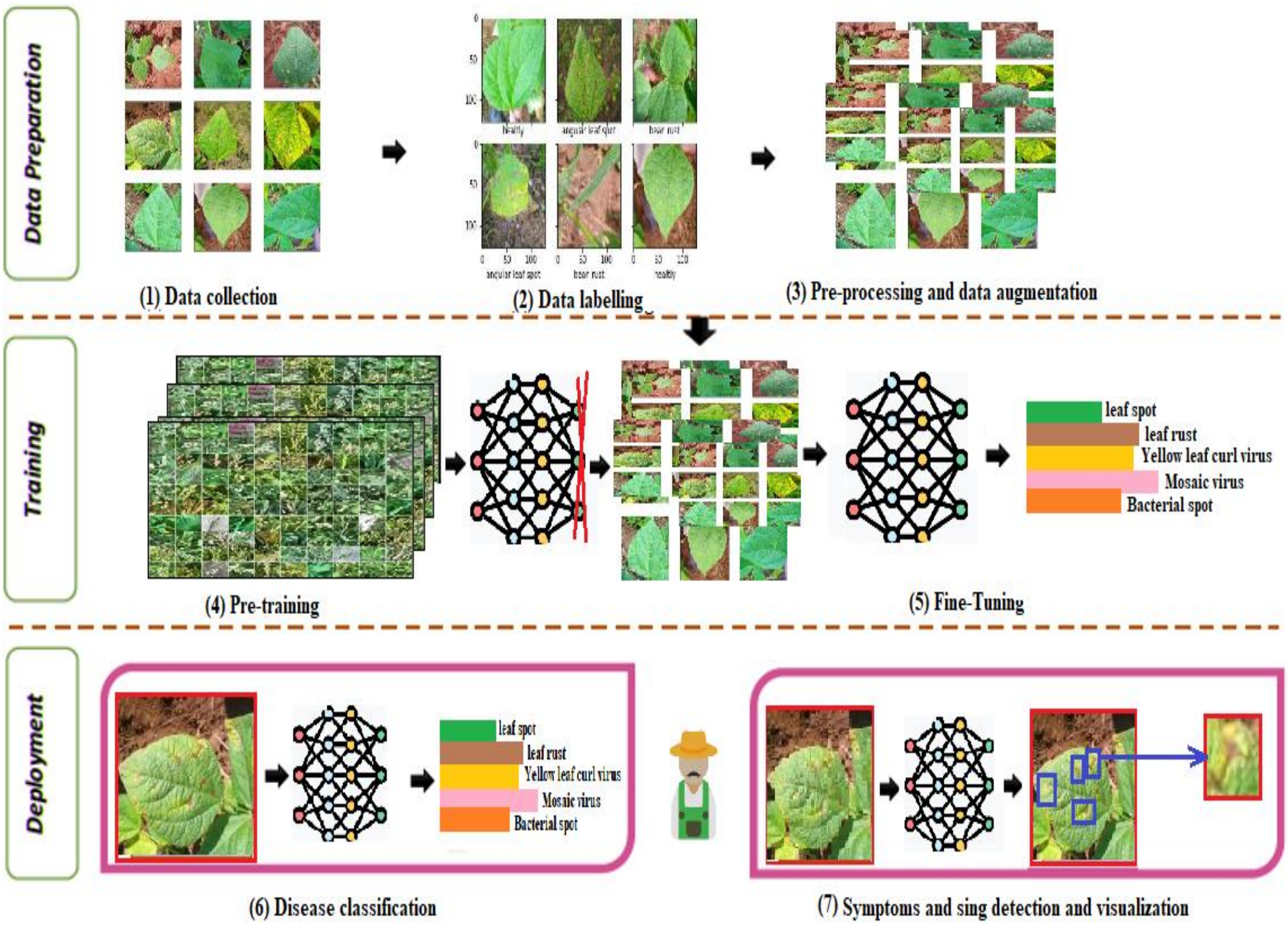

Figure 7.4. Several stages used to apply deep learning in the plant disease area

In the literature (see Table 11), most studies aimed to achieve the highest performance based on accuracy and loss. In addition, after a detailed study of the architecture they used, it was observed that most of the models adopted DL techniques, including CNNs; these techniques allowed for more reasonable feature extraction and modeling of complex relationships among the data, and almost all the studies were based on the idea of achieving the best performance. Overall, the studies prioritize accuracy and loss while utilizing deep learning techniques like CNNs and RNNs for improved feature extraction and complex relationship modeling. Despite proposed architecture changes, many still adopt the same system idea for achieving optimal performance. Figure 7.5 depicts the dataflow diagram that was utilized by the majority of research to carry out their work.

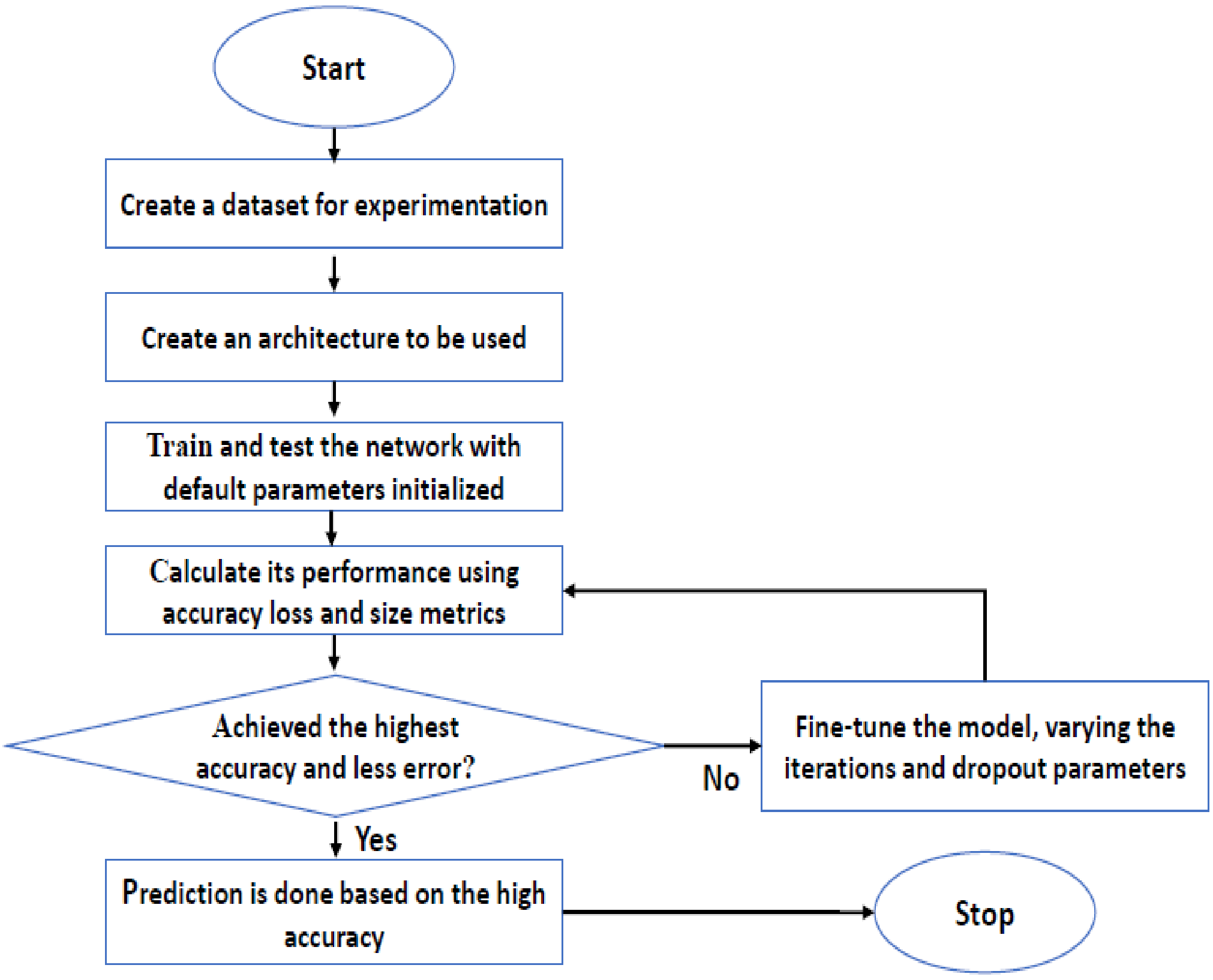

Figure 7.5 The dataflow diagram that most studies used to perform their system

It can be concluded, that deep learning methods are now become the go-to approach for achieving high accuracy in various fields. However, there is a need to explore alternative techniques that can improve model interpretability and reduce reliance on large datasets. Furthermore, some studies explored transfer learning in combination with deep learning techniques to enhance their models further. However, while some proposed changes to the architecture, the general focus remained on achieving the best performance through these techniques. Ultimately, these findings suggest the significance of deep learning in improving accuracy and loss in various fields.

Deep learning techniques are crucial for achieving optimal performance in many studies. While proposed architecture changes have yet to alter this approach drastically, it will be interesting to see how future research continues to evolve. However, it is important to note that deep learning also brings its own set of challenges and limitations, such as data unavailability, model interpretability, and potential biases. Addressing these issues will be vital in ensuring the practical and ethical use of deep learning in various applications.

Overall, continued research and development in deep learning hold great promise for advancing numerous fields and improving our understanding of complex phenomena.

Deep learning approaches have shown great promise in the accurate detection of plant leaf diseases, and they can be applied in conjunction with different rearch methods such as validation and evaluation. By using deep learning algorithms, researchers can train models to accurately identify and classify various types of plant diseases by using images of leaves, which can help improve crop yields and reduce the use of pesticides. Furthermore, there are various research methods to apply in plant leaf disease detection (Canziani et al., 2016), such as validation and evaluation research. Validation research is focused on determining whether a particular method or tool can accurately detect the existence of a disease in plant leaf. On the other hand, evaluation research assesses the effectiveness of different detection methods in real-world scenarios and compares their performance to identify the most reliable approach; therefore, various deep learning approaches can be used with these research methods. These research methods are crucial for developing accurate and efficient plant disease detection systems, which may assist farmers detect and prevent the spread of diseases in their crops. Additionally, using deep learning approaches can be apply in one of these reseacrh area, leading to more effective disease management strategies. Table 7.2 shows different research methods used in the literature with descriptions.

Table 7.2 Research method

| **Research techniques** | **Description** |
|---|---|
| Evaluation Research | Methods are implemented and thus fulfill the evaluation of the technique. This implies that the method's use in practice (the solution implementation) is demonstrated, along with the advantages and disadvantages of doing so (implementation evaluation). This includes identifying issues in the area as well. |
| Validation Research | Researched methods are novel and have yet to be used in actual practice. Methods might be experimental, such as work done in a lab. |

Table 7.2 Research method (continue)

| Research techniques | Description |
|---|---|
| Experience Papers | Experience papers typically document how a particular task or project was accomplished in practice, with an emphasis on the author's own experience and insights gained. |
| Philosophical Papers | These articles organize the area using a taxonomy or conceptual framework to demonstrate a fresh way of looking at things that already exist. |
| Opinion Papers | These papers represent an individual's subjective assessment on whether a particular practice is negative or positive. They do not use comparable research and study approaches. |

In the next section, we will present the outcomes of our review of 100 studies focused on detecting different types of plant diseases utilizing specialized data extraction and analysis techniques. Additionally, we outline a range of problems and potential solutions related to the use of DL in plant disease detection. Our report also examines the pros and cons of using DL in agriculture, along with an overview of publicly available plant leaf datasets and significant issues and solutions related to plant leaf disease detection.

### 7.1 Data Extraction and Analysis Procedures

Within this segment, we present specific data collected from 100 studies aimed at supporting researchers in their quest for knowledge and understanding regarding the application of DL techniques in agriculture. This data encompasses a range of factors, including the plant species, dataset sizes, and model architectures employed by the various studies examined, as well as the results that were achieved, and ultimately, whether or not the application of DL in agriculture proved to be successful. Furthermore, researchers can use the data provided in this section to gain insight into how deep learning is applied in agriculture, the challenges and opportunities that exist, and to draw parallels between the performance of DL technique in different agricultural scenarios.

The data can inform researchers' decisions on which dataset, plants, and model architectures to use in their agricultural applications. It can also provide insight into how DL methods can be used in agricultural contexts and how their results compare with other approaches. In addition, the data collected from these 100 studies can also provide a baseline to compare and contrast the work and better understand the pros and cons of a different approachs. Finally, through the data collected from these 100 studies, researchers can gain invaluable insight into using deep learning in agricultural contexts.

### 7.1.1 Search methodologies

In this thesis section, we analyzed the most current studies regarding the application of DL in agriculture.Therefore, this study was carried out in two key stages: the first included collecting 128 prior studies and papers that address DL in relation to the area of agriculture, and the second included a detailed review and analysis of the research that had been collected.

During the first stage, we conducted a comprehensive and systematic literature search in several databases such as Springer, MDPI, ScienceDirect, and Google Scholar to identify relevant studies on DL in agriculture published during the last five years (see algorithm 1). After collecting the studies, we proceeded to the second stage, which entailed a detailed review and analysis of the collected research with these research questions:

- The method and approach presented.
- The isssue was highlighted.
- Data sets employed
- Results achieved.
- Potential limitations of the study.
- Problems and solutions.

In addition to these research questions, we submitted these search strings to the relevant search databases and filtered the search results based on the search parameter criteria given in Table 7.3. The search results were reviewed and assessed for relevance to the research question. Articles that met the criteria were included in the final review set, while those that did not meet the criteria were excluded from further analysis. After that, the included articles were critically appraised to evaluate their findings' methodological quality and relevance. These steps of the process were repeated until all relevant articles were identified. The search results were then synthesized and analyzed to identify patterns or trends in the literature that addressed our research question. The synthesized results were evaluated to assess their applicability to the research question, and conclusions were derived from the patterns identified from the literature review. Finally, the outcome of this analysis was used to generate and extract data and develop conclusions that could be used to guide future research in this area.

Table 7.3 Criterion for selection

| | **Type of Creterion** | | |
|---|---|---|---|
| **No** | **Inclusion** | **Exclusion** | **Description** |
| CT1 | x | - | Studies presenting novel deep learning models or architectures. |
| CT2 | x | - | Studies that use deep learning to identify plant diseases |
| CT3 | x | - | Studies presenting strategies or techniques for applying deep learning to detect plant diseases. |
| CT4 | - | x | Studies in which DL models are not the primary approach |
| CT5 | - | x | Studies that are not written in English |
| CT6 | - | x | Studies with no acceptable data for extraction |
| CT7 | - | x | Studies lacking a full text |
| CT8 | - | x | Duplicate publishing from different sources |

The Springer, MDPI, ScienceDirect, and Google Scholar databases are the main targets of the studies search strategy. These databases were chosen because of their DL-related solid effect factors. Therefore, a focused search string must be built to get the most out of these electronic search databases. To that end, keywords were selected based on the research objectives, and tailored search terms were used to maximize the accuracy of the results, while a mix of Boolean operators was used to connect the selected keywords and apply restrictions, ensuring that the search results only included the most relevant and appropriate paper. As a result, the search strings were constructed to deliver many high-quality results. To build a search string, we define a general (pseudo) search string that will subsequently be adjusted based on the search database. Once the keywords were chosen, the search string was adjusted to meet the specific requirements of each search database, such as the syntax of Boolean operators (AND and OR), phrase searches and truncation, since each electronic search database has its own specific syntax.

Additionally, each database offers a limited number of searchable fields such as the titles, abstracts, and keywords, making it necessary to include in the search string different words, as well as synonyms and related terms, that may help capture the full range of studies related to our topic of interest. After all the adjustments were made to the search string, it was ready to be used in each database. Finally, the adjusted search string was entered into the relevant search database. The resulting hits were used as input for further screening, which aimed to determine whether the retrieved documents were relevant to our research topic. This process was repeated for all databases being searched. Algorithm 1 depicts the whole procedure for identifying relevant works.

**Algorithm 1** Pseudocode for creating search strings

Databases ← *[Springer, Google_Scholar, MDPI, Science_Direct]*

{**Initialize keywords**}

Plant_keywords ← *[Cassava, Tea, Radish, Plum, Pepper, Rice, Wheat, Bean, Apple, Maize, Soybean, Tomato, Cucumer, Potato, Grape, Mango, Cotton, Guava, Peach, Olive, Chilli, Eggplant, Banana] /* using* ***OR*** *boolean operator*/*

```
Disease_keywords ← [Disease, infection, Disorder, Viral, Fungal, Bacterial]
System _keywords ← [Machine_Learning, Deep_Learning, CNN, VGG, MobileNet,
                    SVM]
Target_keywords  ← [Classification, Detection, Identification, Diagnosis]
Search_String  ← ""{Search string}
for plant in Plant_keywords  do
   for  disease in Disease_keywords do
      for  target in Target_keywords do
         for system in  System_keywords do
            Search_String = plant AND disease AND target AND system
            for  database in Databases do
                 papers  ← databases. search(Search_String)
           end for
          end for
      end for
    end for
end for
```

We conducted these search strings into the relevant search databases and filtered the search results using the criteria listed in Table 7.3. The filtered results were then evaluated to identify potential sources that met the exclusion and inclusion criteria outlined in the study protocol.

This method enabled us to identify potential research studies relevant to our research topic, as well as articles and reviews that may have included additional information, allowing us to build a comprehensive and detailed understanding of the research area.

After collecting the studies to conduct the data extraction method for the application of deep learning in agriculture, we selected 125 studies about deep learning and excluded 25 similar publications. Next, 100 research studies that matched the requirements for our analysis goals were summarized after reviewing the other publications; after summarizing the 100 studies that met our criteria, we identified and analyzed key themes, such as the types of plants used as other relevant information. Finally, we concluded and made recommendations based on our analysis. A flow diagram of the process for choosing literature studies was shown in Figure 7.6.

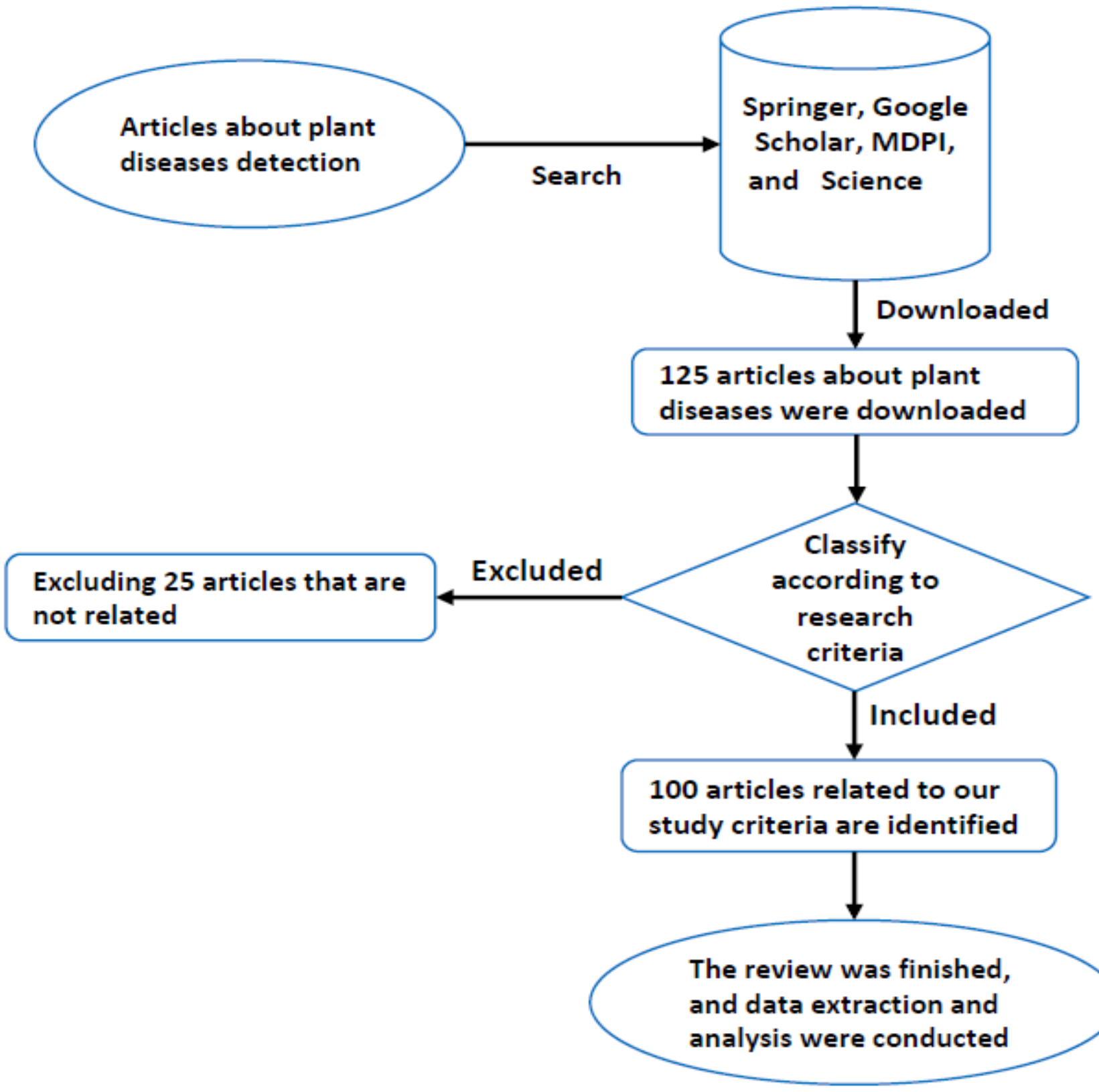

Figure 7.6 The selection procedure for the literature review

Analyzing DL's performance is a crucial component of this research. As a consequence, we reviewed and evaluated several essential papers. By analyzing and extracting a huge quantity of data, we could also analyze various DL technologies and compile a list of the most significant advantages and disadvantages that impact DL performance. The main problems and limitations identified in the prior study were also studied and addressed in the following section.

### 7.1.2 Data extraction

In order to extract data and prepare a detailed analysis of deep learning applications in agriculture by using the methodology existing in the search methlogies section, we reviewed 100 of the most appropriate DL methods articles from the previous five years on identifying various plant leaf diseases, focusing on identifying diseases in crop plants such as wheat, rice, maize, olive, etc. We analyzed the accuracy and efficacy of each of these approaches and the distribution of the study depending on the plant used and the viability of the suggested techniques for deployment in real-world scenarios. In our analysis, we found that most deep learning applications in agriculture focused on the classification and recognition of plant diseases rather than their treatment and that the accuracy of these DL methods varied depending on plant and disease studied. Figure 7.7 compares different architectural designs based on plant type and accuracy obtained from 100 reviewed studies.

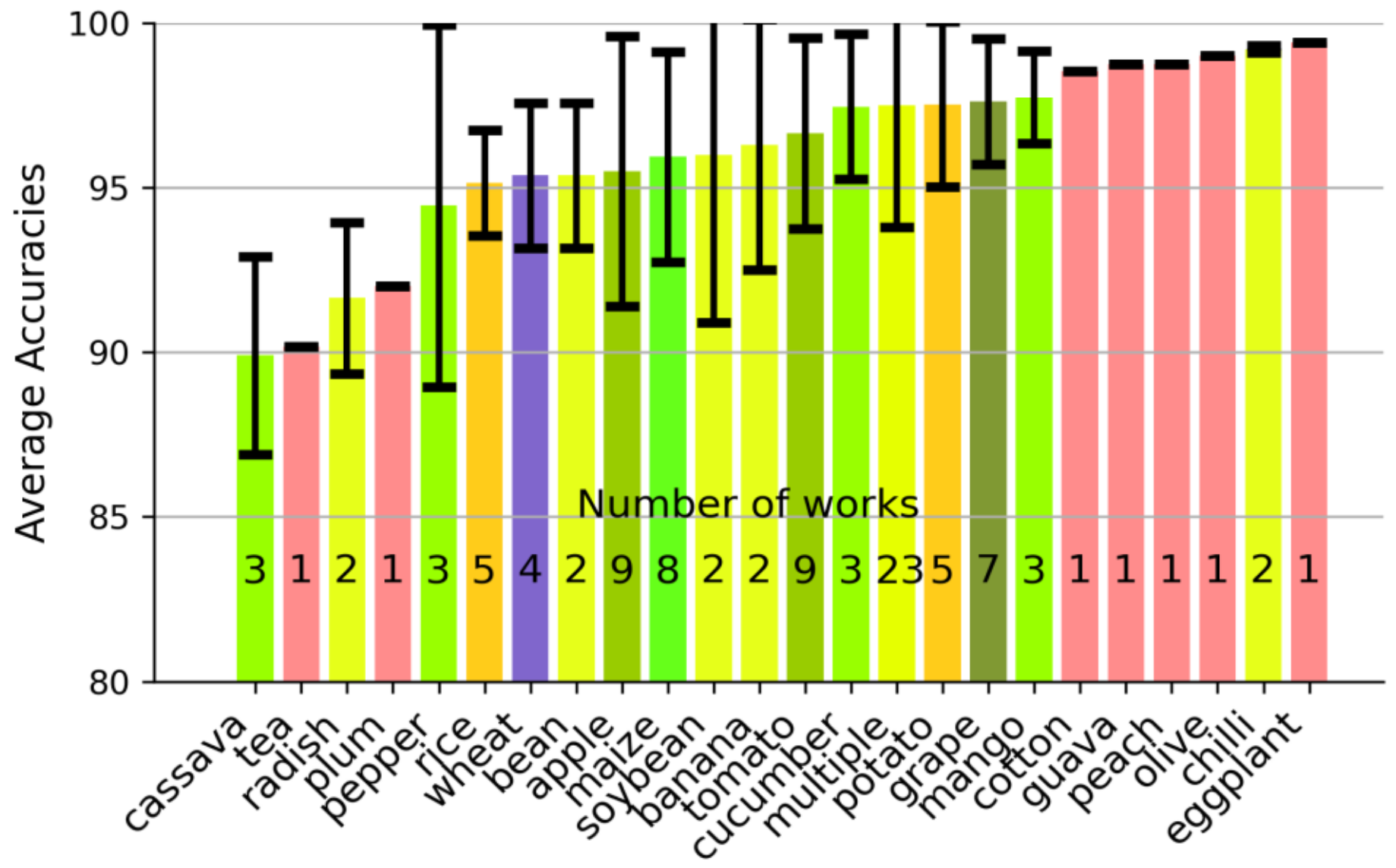

Figure 7.7 DL model architectures comparison in terms of plant type and accuracy

From the results shown in Figure 7.7, it is clear that there have been many successful implementations of DL in the area of agricultural image analysis, with a wide variety of architectures being applied to different types of plants and crops, demonstrating the potential for deep learning to be used as a powerful tool in agriculture. Furthermore, the accuracy of the trained models is quite promising and indicates that deep learning can be an effective method for automatically identifying and classifying diseases in plants, potentially leading to improved efficiency in identifying and managing plant diseases. However, the results from Figure 7.7 also indicate that there are still many open questions and challenges when it comes to deep learning in agriculture, such as the development of more efficient architectures, improving accuracy on different types of plants, and better integration into existing agricultural workflows.

In addition to comparing DL model architectures by accuracy and plant type, we presented in Figure 7.8 the distribution of percentages and the number of crops commonly utilized in 100 analyzed papers that utilized deep learning models to identify plant leaf diseases.

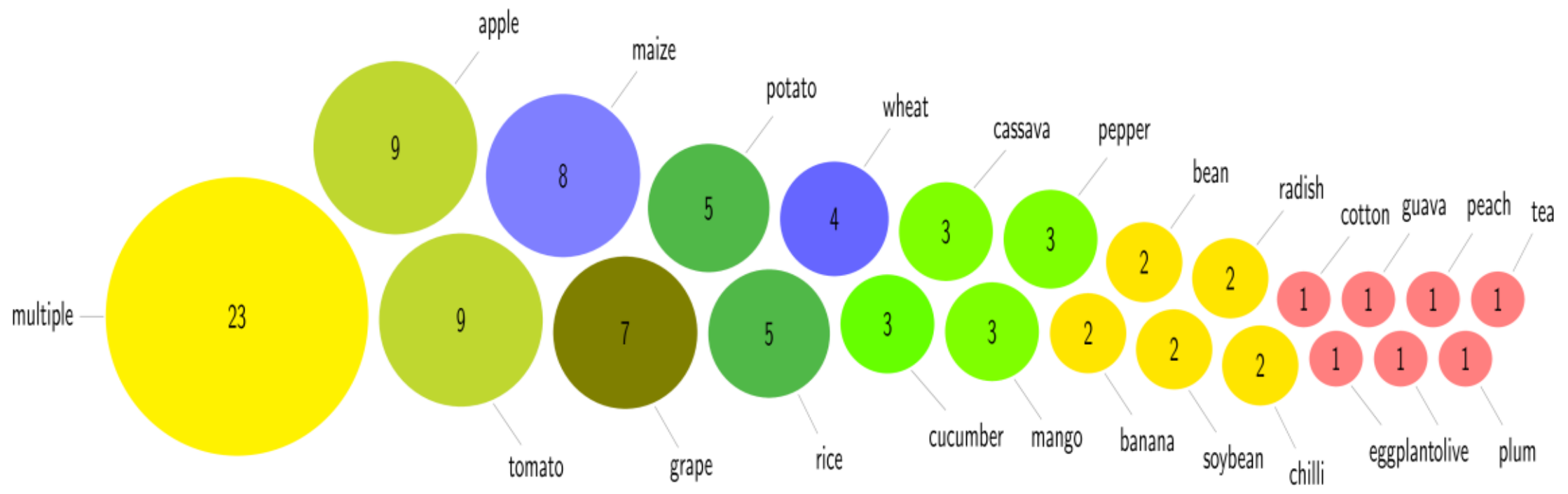

Figure 7.8 Analysis of plant diversity in 100 DL-based studies for ıdentifying plant leaf diseases

With a percentage of 19.2%, it is clear that most authors employed variety of crops (more than one species of the crop) throughout their studies and used datasets with various plant phenotypes. In the 100 summarized research studies, tomato and apple are the second most frequently utilized crop (11.5%). In contrast, the least used crops include tea, plum, guava, peach, and cutton (1.9%), demonstrating that while some essential crops, such as maize, maybe the most popular, it is by no means the only crop studied. Furthermore, the fact that these essential crops are not being studied more often highlights the importance of diversifying research to encompass a wider range of crop species, allowing for a more comprehensive understanding of the complex interactions between crops and their environment. Furthermore, the research studies on multiple crops show various plant phenotypes, from morphological and developmental traits to environmental adaptability and disease resistance; this indicates that scientists are beginning to understand the importance of understanding different plants' characteristics in order to have a comprehensive understanding of how they interact with their environment and what can be done to improve crop yields, leading to more efficient and sustainable agricultural practices in the future.

In conclusion, the summarization of 100 research studies has shown that the majority of researchers prefer to work on multiple crops that include more than one species. Moreover, while tomato and apple is the second most frequently studied crop, other crops such as tea, plum, guava, and peach are being studied less often, and all of these crops play a critical role in the global agricultural system, providing essential nutrients for people around the world. Therefore, researchers must continue to study all types of crops to fully understand the intricacies of their genetic makeup and how they interact with the environment, enabling them to develop new agricultural practices that are more efficient, sustainable, and environmentally friendly.

Alongside outlining the most frequently employed plant species across the 100 CNN-based studies reviewed, we also evaluated the distribution of deep learning (DL) algorithms utilized in the research. Figure 7.9 presents a breakdown of the number and prevalence of studies classified by the DL algorithm implemented during the phase of approach development. Results revealed that the most commonly adopted architecture for plant disease detection using CNN was the recently developed algorithm, utilized in 33 of the 100 studies (representing 33% of the summarized research). Additionally, our review identified VGG as the second most commonly used CNN algorithm, with MobileNet, LeNet, and ConVIT being the least utilized algorithms for plant leaf disease detection.

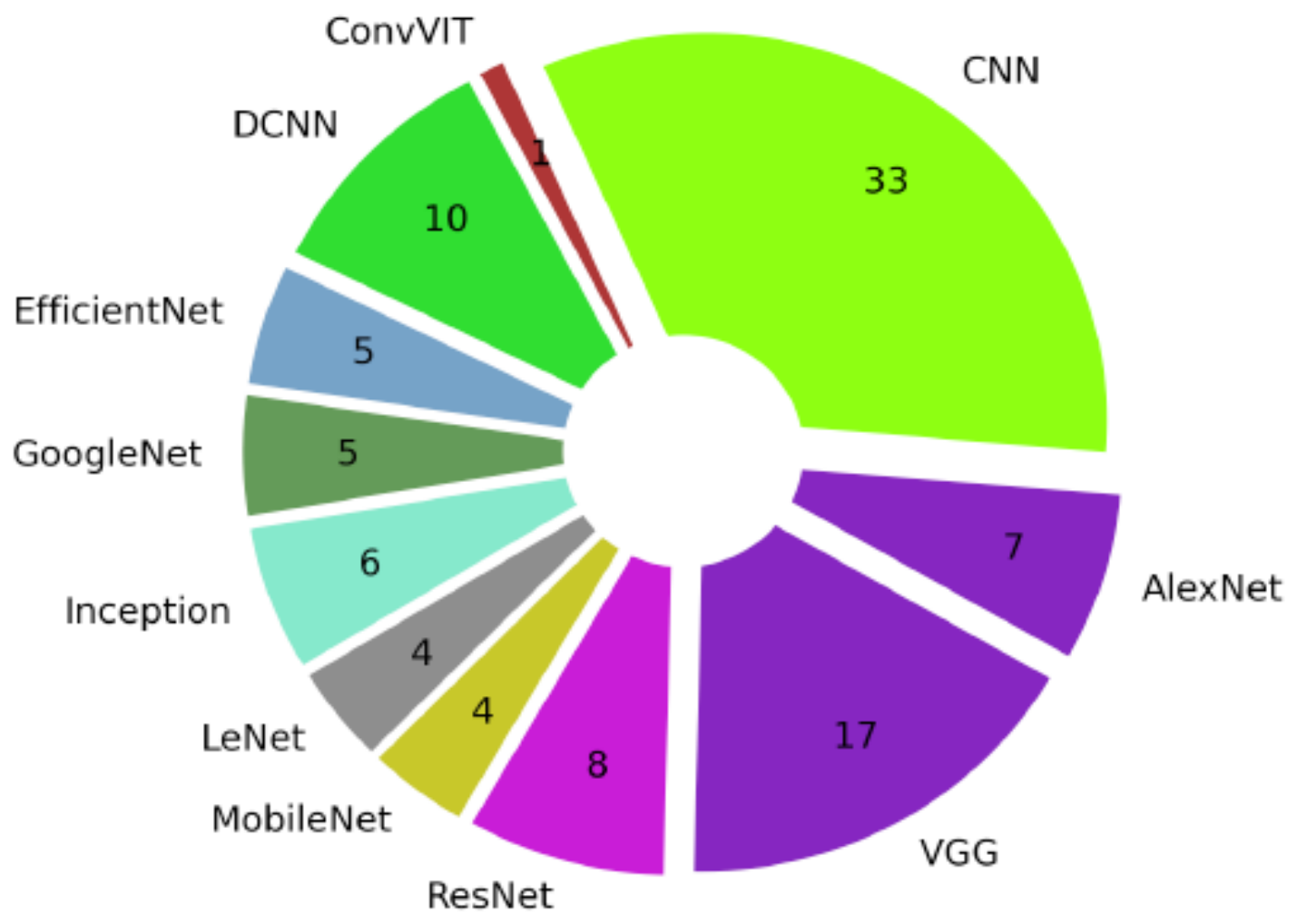

Figure 7.9 Distribution of the most commonly used CNN algorithm in the 100 studies reviewed

It is evident that CNN-based architectures dominate the field of leaf disease detection, even though the algorithms utilised for plant disease detection vary greatly, with more than 11 different algorithms being used across the 100 summarized studies. Overall, this demonstrates a steady growth over time in the utilization of CNN-based architectures in the field of leaf disease detection, and is likely to continue to do so as more researchers look for ways to improve accuracy and speed; this suggests that CNN architectures are the preferred model to use when attempting to the detection and classification of leaf diseases. However, other models, such as decision trees and SVM, could also be utilized to identify plant diseases in the future, and further research into the efficacy of these models should be conducted. In conclusion, many model architectures have been utilized effectively within the realm of detecting and classifying plant diseases. However, CNNs remain the most popular choice due to their higher accuracy and speed.

Furthermore, we presented in Figure 7.10 an overview of the accuracy metrics associated with seven distinct DL architectures employed across the 100 reviewed publications focused on identifying plant leaf diseases. It was clear that by the 100th epoch of training, nearly all models had converged and reported accuracy levels greater than 95%. Furthermore, compared to other models like ConvVIT, EffecientNet, and MobileNet, models like AlexNet, CNN, and VGG produced the highest accuracy; this was likely due to the more complex architecture of the AlexNet, CNN, and VGG models, as their deeper structures enabled them to effectively extract features from the dataset and provide more accurate predictions, compared to simpler models such as ConvVIT, EffecientNet, and MobileNet.

Furthermore, these deeper architectures exhibited a greater parameter complexity, as they contained many more layers than the simpler models, allowing them to fine-tune their parameters more accurately, which in turn resulted in better accuracy when making predictions, resulting in the highest accuracy levels at the 100th epoch of training. Figure 7.10 depicts the accuracy distribution of DL architectures utilized in the 100 reviewed studies.

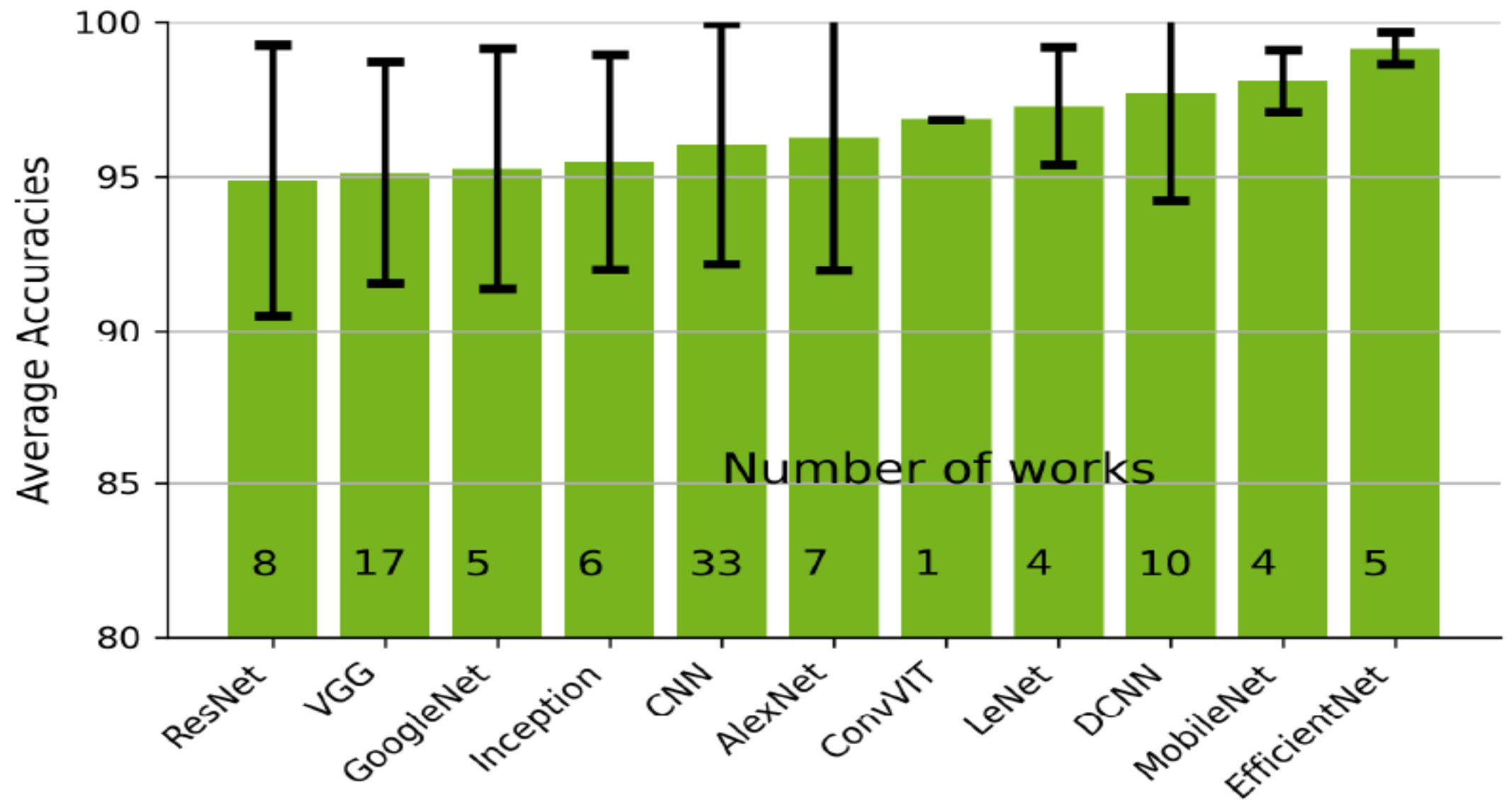

Figure 7.10 DL architectures accuracy distribution in 100 reviewed studies

These findings demonstrate that the AlexNet, CNN, and VGG models' more complex architecture is advantageous when compared to simpler architectures in terms of accuracy since they can better extract features from the dataset and accurately fine-tune their parameters for better predictions. However, simpler architectures should be noticed, while more complex architectures can produce higher accuracy. Despite having fewer parameters and computations, they can still achieve accuracy levels comparable to cutting-edge approaches, making them a viable option for applications with limited computing resources and data availability. In addition to the choice of deep learning architecture, presented in Figure 7.10, the experimental parameters configuration options in these 100 reviewed studies were presented in the table 7.4.

Table 7.4 Setup choices for the experimental parameters in 100 reviewed studies

| Training paramters | Options |
|---|---|
| Selection of training method | Training from scratch or Transfer learning |
| Selection of dataset type | Grayscale, Colour, segmeted Leaf |
| Selection of training and testing set distribution | Test:20%, Train:80% or Test:50%, Train:50% or Train:60% Test:40% or Train:20%, Test:80% |

These parameter configurations in the table included a selection of training mechanism, dataset type, and dataset distribution. As shown, the parameters explored in the studied deep learning architectures are manifold and range from training mechanism to dataset type and distribution, making it difficult to choose the optimal parameter configuration, especially when considering the fact that different parameter configurations can significantly influence the effectiveness of deep learning architectures, This is further compounded by the fact that different parameter configurations can produce drastically different results when evaluating the performance of deep learning architectures. Moreover, while some parameter configurations may perform better than others on certain datasets, they may be less effective on others, requiring parameter configurations to be continually modified and adjusted to produce the best performance results. As such, it is important to consider a range of parameter configurations when selecting a deep learning architecture, as the best configuration for one dataset may not be suitable for another, necessitating the need to carefully analyze and compare different parameter configurations to ensure the best performance of a DL architecture, and this is what we will discuss in the following sections.

The choice of an appropriate dataset is a significant factor in determining the effectiveness of DL methodologies in identifying plant diseases. To this end, we analyzed the distribution of studies according to the datasets employed, taking into account the key characteristics of the most commonly utilized datasets featured in the reviewed studies. By correlating data for each dataset, we were able to generate an accurate distribution of usage across the 100 studies considered. Figure 7.11 visually presents the distribution the year-wise distribution and the prevalence of the most widely used datasets identified within our review.

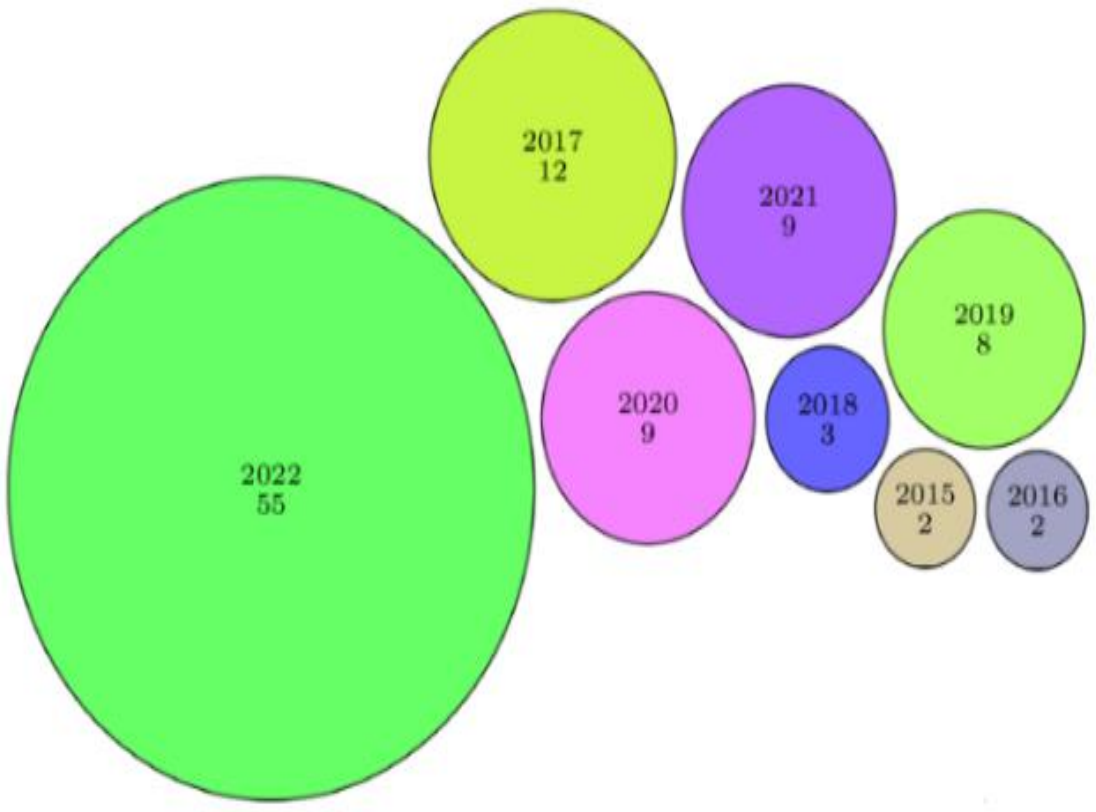

**(a)**

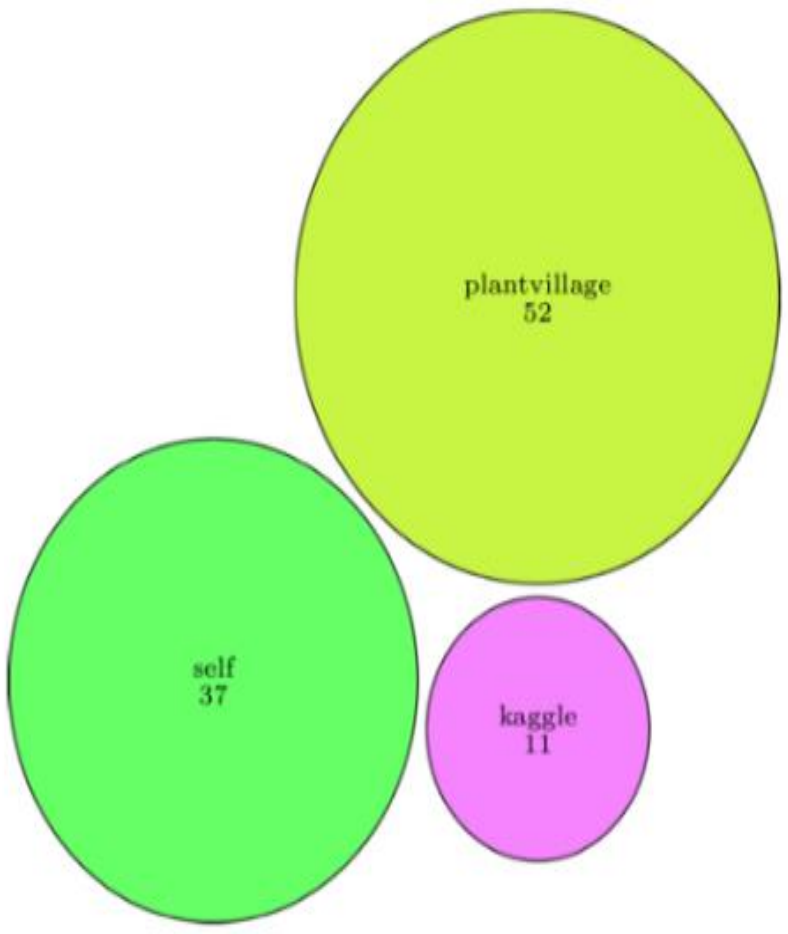

**(b)**

Figure 7.11 Distribution of year and dataset across 100 reviewed studies

According to our findings, most studies used PlantVilage, Kaggle, and self-collected datasets. Overall, this data suggests that the range of datasets used in studies related to disease detection via CNNs has increased dramatically over the years spanning from 2016 to 2020, with a more significant number of studies utilizing self-collected datasets, indicating a trend toward researchers collecting more data for their studies rather than relying on existing datasets. This trend can be attributed to the growing recognition of the need for large, high-quality datasets to train models accurately and efficiently and to the advancement of technologies that have made it easier for researchers to collect data, such as through sensors, mobile applications, and drones. The following section will go over plant datasets that are publicly available.

### 7.2 Publicly available plant leaf datasets

In this section, we provide a number of popular and publicly accessible datasets. Researchers and data scientists frequently use these datasets to explore, build models, and create machine-learning solutions. They can be used for various purposes, from identifying trends and correlations to developing predictive models, and can be easily accessed from various sources.

- **Flavia dataset**: Lee et al. introduced the Flavia dataset, which contains images of individual leaves from 32 diverse plant species, and is intended for use in plant leaf recognition. (Lee et al., 2015).
- **MalayaKew dataset**: As reported by LifeCLEF et al. (2019), is a collection of images depicting single leaves from 44 plant species, and it serves the purpose of facilitating plant recognition tasks.
- **LifeCLEF**: Contains information about the plant's geographic distribution, identity, and uses. Since 2014, the dataset has been growing with data from volunteers all over the world (University P. S, 2019).
- **PlantVillage:** Krohling et al. in (Krohling et al., 2020) established Plant Village as a research and development initiative at Penn State University, with the aim of providing open access to all agricultural knowledge that can aid people in growing crops. The Plant Village dataset contains 61,486 unaltered images of various crops, each labelled with its corresponding disease and categorized into 39 classes.
- **Plant Pathology:** The Plant Pathology dataset (Lim et al., 2019) comprises 3651 RGB images of various apple foliar disease symptoms taken throughout the 2019 growing season from commercially produced cultivars in an unsprayed apple orchard at Cornell AgriTech. There are 187 images with complex disease symptoms among the 3651 RGB images, 1200 images of apple scab, 1399 images of cedar apple rust, and 865 of healthy leaves.
- **BRACOL Dataset:** BRACOL is a Brazilian arabica coffee plant species dataset used to identify and quantify coffee diseases and pests (Hendrycks et al., 2019). It includes 1747 images of arabica coffee leaves infected with leaf miner, brown spot, leaf rust, and Cercospora spot.

With the increasing availability of publicly available datasets, it has become easier for researchers and data scientists to access high-quality data and apply it to their projects. In addition to being freely available, these datasets often contain large amounts of data, which can provide great insights and help develop better models, thus making them invaluable resources in developing data-driven projects. The use of these datasets is quickly becoming a crucial part of data science, as the insights gained from analyzing them are often invaluable for gaining an understanding of the problem domain and developing better models, as well as for testing the accuracy of existing models and ensuring that any conclusions drawn from the data are valid as such, these datasets are not only useful for data science projects but can also provide a valuable resource for researchers and businesses alike. For the best use of these datasets, it is essential for users to consider the data quality and any potential biases or inaccuracies present in the dataset and to ensure that any inferences derived from the data are reliable and accurate. In addition, it is essential to consider the privacy and ethical implications of using these datasets, as there is often sensitive personal information present in them, and any misuse of this data could have serious repercussions.

## 7.3 Major Challenges and Solutions in Plant Leaf Disease Detection

There had been no notable advances in plant disease classification prior to 2015. However, deep learning (DL) has emerged as a prevalent approach for plant disease identification, becoming a forefront technology in this field since 2016. Convolutional neural networks (CNN) are the most commonly utilized DL-based methods for plant leaf disease classification. Despite their success, challenges still exist in developing effective DL techniques for the classification and detection of plant diseases. This section provides an overview of these challenges, along with potential solutions, and emphasizes the need for continued research in this evolving field.

### 7.3.1 Insufficient plant leaf datasets

A major obstacle in using DL for plant disease identification is the requirement for datasets that are both sufficiently diverse and large in size. All of the other problems mentioned are caused in part by this condition. With a large and diverse dataset, the model can identify plant diseases accurately. However, it may have yet to see enough examples of certain diseased plants to recognize them, resulting in many misidentifications. Developing more diverse datasets is essential for deep learning applications in plant disease identification to become more accurate and reliable. To ensure the accuracy and reliability of DL applications in plant disease identification, researchers need to work to create datasets that are larger and more diverse. These datasets must contain images of plants from different regions and climates, with a wide range of variations in terms of soil type, environmental factors, growth conditions, disease characteristics, and different species and plant varieties.

The primary challenge in utilizing DL methods for leaf disease identification and classification lies in the requirement for extensive datasets. Insufficient dataset size can significantly impede practical implementation, resulting in inaccurate results despite the model's effectiveness. Unfortunately, agricultural researchers face limited access to publicly available databases and often need to create their image dataset from scratch, exacerbating the issue. Moreover, due to external factors conditions including the weather, data collection may be time-consuming and require several work days. To effectively overcome these issues, agricultural researchers should consider different factors, such as the diversity of the data, the quality of the images, and other resources.

The following solutions are available to address the issue of a limited and inadequate dataset:

1. **Data augmentation techniques:** To enhance the variety of data during the training process, data augmentation techniques create artificial samples from the initial dataset. Image augmentation, on the other hand, generates new data from existing data, thereby aiding in the training of deep neural network models. This technique helps improve the model's generalizability and allows it to learn more robust features. Image augmentation techniques can include transformations such as rotating, shifting, cropping, and flipping the images in order to create new training data from the existing dataset, thus allowing the model to learn from the same image multiple times in various positions and orientations. These techniques not only enhance the size of the training dataset but also allow for a more in-depth understanding of the data by enabling the model to learn from different views of an object, scene, or image.

By using these augmentation techniques, a DL model can learn more meaningful features from the training data, resulting in improved accuracy when making predictions on unseen data. In addition, these techniques also allow for better generalization of the model by improving its ability to recognize patterns in unseen data that it may not have seen in the training data; this ultimately leads to improved accuracy and more robust models that may be utilized for image classification, and object detection.

Augmentation techniques such as Fast Auto Augment (Cubuk et al., 2020), AugMix (Ho et al., 2019), Rand Augment (Liu et al., 2017), and population-based augmentation (Sladojevic et al., 2016) are the latest methods used to increase dataset diversity. For instance, in a study by (Liu et al., 2017), By applying data augmentation techniques, the dataset size was expanded from 1053 to 13,689 images. Similarly, (Chen et al., 2019) employed perspective transformation and rotation techniques, increasing the image count from 4483 to 33,469 and improving accuracy as the dataset grew. (Barbedo, 2016) utilized the image resizing technique, which helped expand from 1567 to 46,409 images, resulting in a 10.83% accuracy improvement over the non-expanded data.

2. **Transfer learning:** Transfer learning is an approach in machine learning technique that involves leveraging a pre-trained model as a foundation for a new model in a different project. By reusing previously trained networks, only a few layers need to be retrained on new datasets, reducing the amount of data required for training (Chen et al., 2019). This significantly minimize the time and resources needed to build a new model. Transfer learning has become a crucial tool in machine learning, enabling faster and more efficient model building with smaller datasets that may not be sufficient for traditional approaches. For instance, (Chen et al., 2019) demonstrated the effectiveness of transfer learning by developing INC-VGGN DL architectural features for plant disease identification, by modifying a pre-trained VGGNet. The suggested model attained an impressive accuracy of 91.83% on Plant Village and demonstrated a 92.00% accuracy on their proprietary dataset.

3. **Citizen science**: Citizen science as an idea was first proposed in 1995. This strategy is used in scientific studies to collect data from non-professional people. In order to classify plant diseases and pests, farmers send the images they have collected to a server. An expert then labels and analyzes the images to determine what they represent (Chen et al., 2019).
4. **Data sharing:** Another method of increasing datasets is through data sharing. Globally, numerous studies are currently being undertaken on accurate disease detection. If the various datasets are shared, the dataset will become more accurate. This situation will promote more critical and satisfying findings from the study. Data sharing allows datasets to be larger and more accurate, which can lead to more significant discoveries. Therefore, researchers must share their datasets to promote the growth of knowledge and understanding while also enabling the replication of research, enhancing the credibility of the field, and allowing for more efficient use of resources. By making data sharing a priority in the research field, scientists will be able to better analyze and interpret results, allowing for more meaningful conclusions to be drawn. Furthermore, researchers will have access to a wider range of data points, allowing them to gain a much deeper understanding of the problem at hand.

### 7.3.2 Image background

The impact of the background of an image on detection is one of the problems researchers encounter. However, this impact is frequently unclear due to the overlap of numerous factors. The organizing process and how plants interact with one another are what stand out the most. In addition, the type of background and its degree of complexity can also influence detection, thus creating a challenge for researchers when attempting to use image analysis techniques. Backgrounds are thus an essential factor in image analysis and can greatly influence the accuracy of the results, it is essential to consider the interactions between plants and how they are affected by one another to ensure precise and reliable outcomes in the analysis. To understand the effect of the background on detection, researchers have to consider several factors, such as the variability of colors, shapes, and textures present in the background, as well as the complexity of the background and its degree of homogeneity.

To address the issue of image backgrounds, segmentation techniques are proposed as a necessary solution in situations where the packed background in real-time image collection may contain features that resemble the area of interest. Failing to address this issue could result in the model learning background features during training, leading to inaccurate identification results. By employing techniques including edge detection and blob detection, segmentation aims to isolate areas of interest in an image and reduce the impact of the background on the foreground. This creates a more uniform background, which is easier for the model to classify. Moreover, segmentation can help to reduce the amount of information processed by the model and enhance the accuracy of classification.

In contrast, organizing the image collection is something that some researchers are interested in. Because it produces comparatively homogenous backgrounds, the background is typically preserved in this situation. Therefore, it has no impact on detection and might even increase the precision of detection; this is due to the fact that object detection algorithms often rely on background subtraction or object segmentation in order to differentiate the foreground from the background. As such, organizing an image collection in a way that preserves the background is advantageous for object detection algorithms. It reduces the need for background subtraction or object segmentation, simplifying the task significantly, and allowing for a more efficient and accurate detection.

### 7.3.3 Computational resources

Deep learning algorithms require significant computational resources, which may limit their practicality for use in resource-limited settings. These algorithms typically require large amounts of memory, processing power, and storage to train and deploy models, and they can be computationally expensive and time-consuming to run.

However, there are several approaches to mitigate these challenges and make deep learning more accessible in resource-limited settings. Some of these approaches include:

1. **Model Optimization:** There are several techniques for optimizing deep learning models to reduce their computational requirements, such as reducing the network size, using sparsity-inducing methods, and using quantization techniques.
2. **Cloud computing:** Cloud computing platforms can provide on-demand access to scalable computational resources, which helps reduce the costs and infrastructure requirements of running deep learning models.
3. **Distributed computing:** Distributed computing frameworks such as Apache Spark and Apache Hadoop can be utilised to distribute the training and inference of DL models across multiple machines, which can help to speed up computation and reduce resource requirements.
4. **Edge computing:** Edge computing involves running computations on local devices, such as smartphones, tablets, or IoT devices, rather than in a centralized location. This can reduce the need for large bandwidth and processing power and enable real-time data processing.
5. **Transfer learning:** Transfer learning involves utilizing pre-trained models and fine-tuning them for specific tasks rather than training models from scratch. This can lead to a reduction in the amount of data and computational resources required for training models.

Overall, while deep learning algorithms require significant computational resources, several strategies can be used to make them more practical and accessible for use in resource-limited settings.

### 7.3.4 Symptom variations

Symptoms are the plant's effects and changes in appearance. It may result in a significant appearance, color, or functionality difference. The manifestation of plant disease symptoms is influenced by the intricate interplay between diseases, plants, and the environment, as highlighted by (Zhang et al., 2019). Typically, disease symptoms may exhibit similarities, but various natural factors such as sunlight, temperature, wind, humidity, and others can also impact the symptoms. The dynamic interplay among diseases, plants, and environmental conditions can result in changes in symptoms, posing challenges in data collection and accurate reporting.Therefore, it is difficult to determine why certain symptoms manifest or appear more in some climates and less in others; this challenges scientists to research and understand these diseases and makes it more difficult to develop treatments or preventive measures; this underscores the importance of understanding not just the disease itself but also its relationship to the environment. Therefore, it is necessary to find ways to detect and treat diseases that are not only effective but also consider the various environmental factors at play. In other words, the complexity of disease dynamics across environments presents scientists with unique challenges and opportunities that require a deep and holistic understanding of the interconnections between the environment, diseases, and their effects on people's health.

Plant disease identification presents a significant challenge due to the presence of multiple diseases that can manifest and merge on the same plant leaves. This creates a scenario where symptoms can change rapidly, leading to difficulty in identifying disease types (Zhang et al., 2019). Additionally, similar symptoms may occur in a broad range of diseases, further complicating the identification process (Barbedo, 2018). Addressing this issue requires diversifying the database in real-world applications (Makerere AI, 2020). Researchers has been increasingly utilizing this approach to enhance data diversity effectively. This method is more practical as it enables scientists to collect all variations and disease data quickly and efficiently.

In conclusion, leaf disease detection is a essential aspect of agriculture, also DL algorithms have shown significant potential in accurately identifying and classifying diseases. However, several challenges still need to be addressed to improve the accuracy and practicality of disease detection models. By addressing these challenges through the use of standardization, transfer learning, ensemble learning, and edge computing, we can overcome the limitations of current disease detection methods and improve crop health and productivity.

In the following sections, we will discuss several experimental studies conducted, as well as the results obtained for detecting and identifying plant leaf diseases using diffirent DL algorithms.

# 8. EXPERIMENTAL STUDIES AND RESEARCH RESULTS

Several experimental studies and the outcomes achieved through the utilization of different deep learning algorithms for detecting and identifying plant leaf diseases were discussed in this section. We investigated the feasibility and effectiveness of using deep learning algorithms to accurately detect and identify different leaf diseases from images collected from real-world scenarios. In addition, We have shown the possibility and the potential of deep learning algorithms as a robust tool for detecting and identifying plant leaf diseases, showcasing promising results in regard to both speed and accuracy. Furthermore, we have conducted a detailed study on leaf disease detection utilizing a Mobilenet Model. We also presented a study about datasets' impact on mobile networks' effectiveness for bean leaf disease detection. Furthermore, we introduced a new CNN system for seed image classification and identification. We also analyzed the advantages of utilizing deep learning algorithms for the detection and identification of leaf diseases over traditional methods.

Additionally, we discussed the importance of data collection and feature engineering in developing effective deep-learning models, which is an essential factor in achieving better results for detecting and recognizing plant leaf diseases. Finally, we highlighted the potential of deep learning algorithms in agricultural applications. We demonstrated how they could help farmers detect and identify diseases in their crops more accurately, quickly, and accurately than traditional methods. Overall, our study has shown the potential of deep learning algorithms in agricultural applications and illustrated how they could be utilized to accurately and efficiently identify leaf diseases, providing farmers with an innovative and efficient way to monitor their crops for potential diseases. The results of the experiments have shown that deep learning algorithms have great potential for accurately detecting and identifying plant leaf diseases from digital images, providing reliable and precise results quickly and efficiently.

### 8.1 Plant Leaf Diseases Detction Using Mobilenet Model

Plant leaf diseases can cause significant damage to crops, leading to reduced yields and economic losses. In addition, traditional detection methods for plant leaf diseases can be time-consuming, costly, and may require specialized expertise. Furthermore, to address these challenges, advanced methods such as CNNs and other deep-learning approaches were developed to automate identifying plant leaf diseases.

Using a Mobilenet model for plant leaf disease detection is an example of a deep learning technology that enables the automatic detection and classification of plant diseases based on images of leaves. This approach involves training a Mobilenet model using a plant leaf dataset of labeled healthy and unhealthy images, which can then be used to detect and classify new images of leaves as either healthy or diseased.

The use of a Mobilenet model for plant disease detection has several advantages over traditional detection methods:

1. It is faster and less expensive than traditional methods, as it does not require physical sampling or laboratory analysis.
2. It can be more accurate and consistent than human visual inspection, as the model can be trained to detect subtle visual differences that may be difficult for the human eye to discern.
3. It can be easily deployed on mobile devices, making it available to farmers in isolated or underdeveloped locations without traditional detection techniques.

The adoption of MobileNet in plant leaf disease detection is acknowledged for enhancing accuracy, speed, and accessibility in disease identification within the field. Moreover, MobileNet is specifically engineered to be efficient and optimized for mobile and embedded applications. Furthermore, this makes it a promising plant leaf disease detection approach in field applications.

This section presents a new approach to identifying and classifying leaf diseases utilizing MobileNet, a CNN renowned for its effectiveness in mobile applications. This study used bean leaf images as an example plant leaf dataset. The MobileNet network was trained on this dataset by applying transfer learning techniques and fine-tuning, resulting in the accurate identification and classification of bean leaf diseases.

The suggested method leverages MobileNet and the open-source TensorFlow library, Incorporated within this study is a comprehensive comparison and analysis of various MobileNet architectures, encompassing the exploration of hyperparameters and optimization methods to develop smaller and more effective models. Various architectures were assessed to create an efficient model able to quickly identifying and classifying diseases into distinct categories. The datasets employed in this study consist of leaf images obtained from diverse locations in collaboration with the National Crops Resources Institute (NaCRRI) (Howard et al., 2017), the dataset comprises two classes representing different types of bean leaf diseases and one class representing healthy bean leaves. Our research objective was to devise a novel deep learning-based solution utilizing MobileNet and TensorFlow to accurately identify different bean leaf diseases, serving as an effective tool for early disease detection and diagnosis.

With the goal of creating an automated model, this work aims to harness the power of MobileNet, along with a dataset of bean leaf images and an efficient network architecture to accurately classify and identify various types of diseases. Because beans are widely cultivated, they are susceptible to diseases that affect their production; this is the motivation behind the idea that classifying and detecting leaf disease is the solution to saving bean crops and their productivity. To achieve this goal, the automated model must exhibit the capability to accurately detect various types of diseases, differentiate them from healthy leaves, and also determine the severity of each disease and its potential impact on bean production. After creating the model and training it, the next step will be to evaluate its performance using large datasets from actual leaf samples. The evaluation process will assess the model's accuracy, recall, precision, and F1 score to determine how well the model can classify and identify disease types in leaves. These metrics provide important insights that will help us improve our model and reduce the time it takes to identify diseases in leaves. In the end, the model will help us make better farming decisions by allowing us to quickly detect diseased leaves and make timely management decisions to help improve crop yields and prevent the spread of diseases to other plants.

There are several applications where it is crucial to compare and evaluate different MobileNet architectures to identify plant disease by uitilizing a single dataset. First, comparing different MobileNet architectures can lead to a more efficient and accurate classification of leaf diseases and has some unclear benefits, including high performance, a longer life, and simpler retraining. And also comparing the different MobileNet architectures can lead to a better understanding of how specific MobileNet architectures perform on certain tasks, allowing us to identify which architecture is best suited for a given task and which architecture is more cost-effective and efficient.

The materials and methods section of the study provided details on the system configuration, dataset, and training procedure. The results and discussion section, on the other hand, presented the experimental setup and discussed the obtained results.

#### 8.1.1 Research materials and methods

This section provides a description of the techniques, datasets, and performance evaluation criteria employed in the proposed approach.

##### 8.1.1.1 Dataset and system confıguration

The dataset used in the study comprises 1296 bean leaf images collected from real fields. These images are classified into three distinct categories: bean rust, healthy bean, and angular leaf spot. This public dataset was introduced by Tensorow and selected from GitHub (Makerere AI, 2020). It was annotated by specialists from NaCRRI in Uganda who identified the disease in each image. The dataset was released on January 20, 2020, and was split into 80% training, 10% testing, and 10% validation sets. The annotated images can be utilized for different purposes, including crop disease detection and classification, and the development of AI-based predictive models for real-time disease detection and classification.

The Makerere AI Bean Leaf Disease dataset is a valuable asset in the agriculture industry, with the potential to revolutionize crop monitoring and disease detection. The images, collected from the Makerere University farm in Kampala, Uganda, have been manually annotated by experts and have a resolution of 512 x 512 pixels, making them ideal for AI-driven tasks such as crop disease detection and classification. Therefore, in our study scenario, we converted each image to 128 by 128 pixels per MobileNet's input requirement to give an appropriate trainer model to enhance disease prediction. Figure 8.1 displays examples of the leaf images categorized by the classes employed in this study.

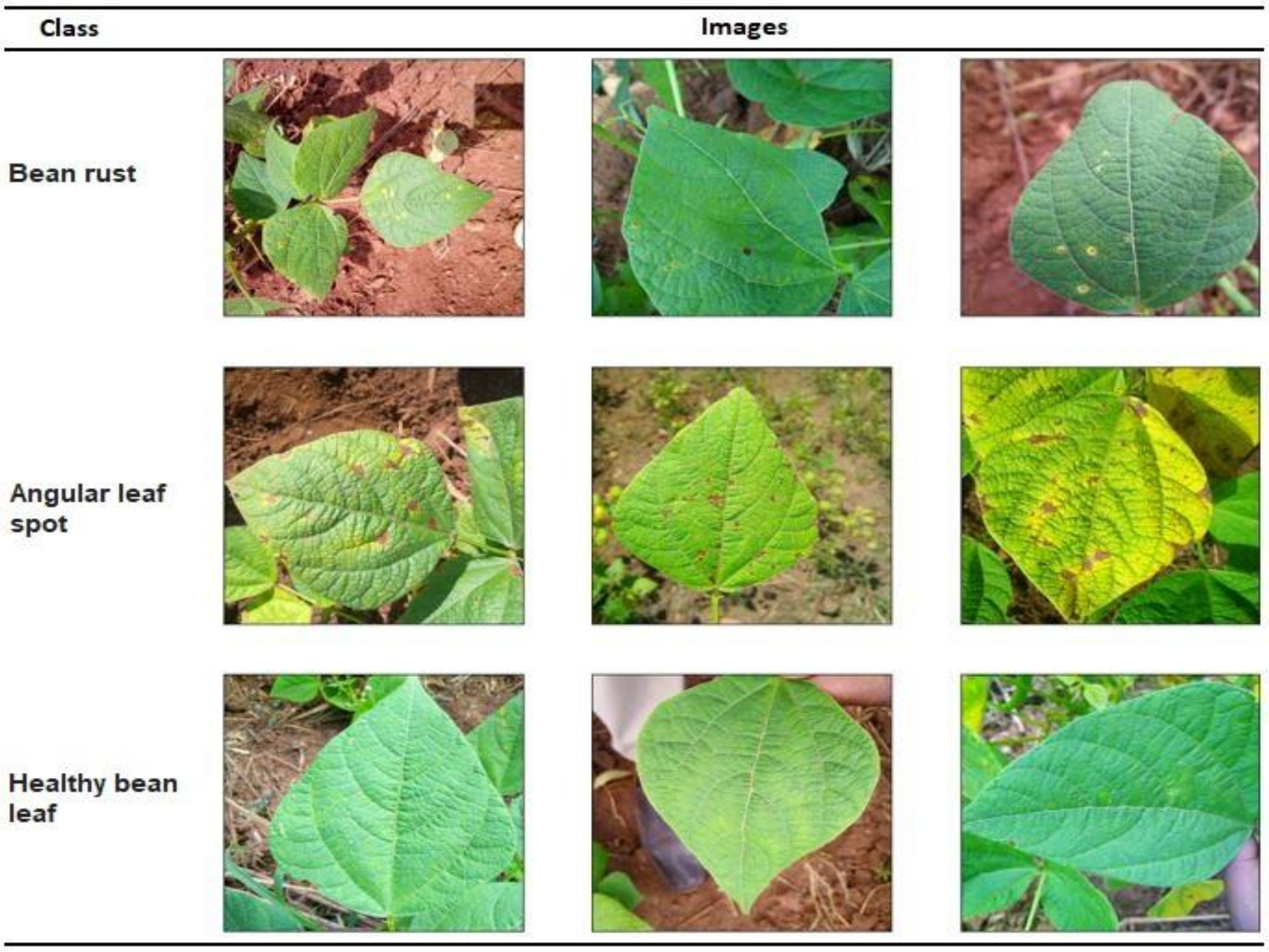

Figure 8.1 Some examples from bean leaf diseases classes

This publicly available dataset comprises leaf images that have been evaluated by experts from NaCRRI to determine the presence of either healthy leaves, bean rust disease, or angular leaf spot disease. As illustrated in Figure 8.1, some images have a background that is primarily composed of overlapping leaves from the same plant, and the image boundaries vary across groups within the same class. To train this dataset, we will utilize a CNN model that employs the MobileNet network architecture, and we will evaluate the results based on four performance assessment criteria: recall, F1-score, precision, and accuracy. We expect that after the training, the model will deliver superior measures of recall, F1-score, precision, and accuracy. For a comprehensive description of the dataset used, refer to Table 8.1.

Table 8.1 Description of bean leaf diseases dataset

| Class | Symptoms | Dataset |
|---|---|---|
| Bean rust | Rust may appear on any part of the plant that is above ground, although it is most common on the undersides of the leaves. | 436 |
| Angular leaf spot | Typically, rust first appears as lesions that are surrounded by the veins of the leaf and appear water-soaked. These lesions have an angular shape. As they progress, the lesions may turn yellow, then brown, and the affected leaf tissue may eventually become entirely brown. | 432 |
| Healthy | - | 428 |

#### 8.1.1.2 Model architecture implementation

In this section, we outline the experimental configuration of our model, which utilizes the MobileNet architecture in conjunction with the TensorFlow framework. Several steps must be taken to implement DL architectures, starting with dataset collection and ending with performance analysis, evaluation, and classification. In our case, we divided the model into several stages, including data analysis and creating an input channel, to develop a classifier capable to accurately classify the health status of bean leaves, distinguishing between diseased and healthy conditions. As MobileNet's learning strategy aligns with supervised learning in deep learning, we labeled the data for training purposes, as depicted in Figure 8.2. Subsequently, we used the TensorFlow framework to define our MobileNet architecture, allowing us to train and optimize it to accurately classify the bean leaves diseases.

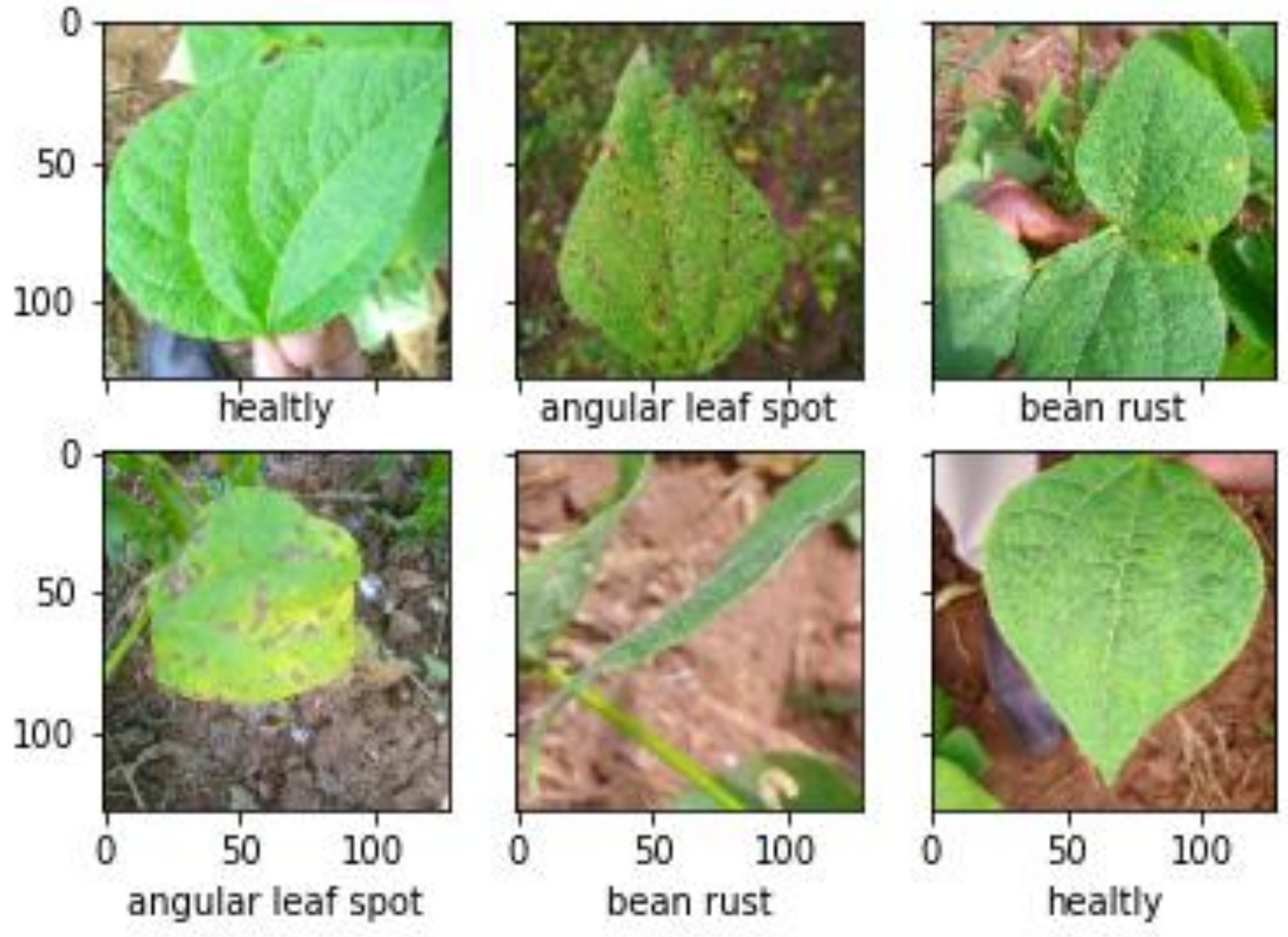

Figure 8.2 An example of a labeled dataset

Similar transformations are used to develop the validation and test pipelines. For example, examining the imbalance between disease classes is an excellent idea to see if one has noticeably fewer samples than the other. However, for this study, we used a public dataset that had already been divided into three balanced classes. We decided to use the public dataset as it offered three distinct, balanced classes of data, allowing us to have equal samples from each class, ensuring that the results were not skewed or biased in any way; this also confirmed that the results obtained from our models were reflective of the data's actual characteristics rather than being biased or distorted. We also considered the balance of features between the three classes to ensure that all relevant information was accounted for in each of them and that our models captured the most important characteristics of each class. Moreover, this provided us with the opportunity to use different methods to evaluate, analyze and draw insights from the dataset, making sure that our results were not only accurate but also comprehensive.

In this research, MobileNet has eight convolutional layers designed to classify images. Each image is utilized several times during the training phase. The learning algorithm processes each training batch only once within a single epoch before evaluating its performance on the validation set. Once the the learning algorithm has rated its performance on the validation set, it can then adjust its parameters for the next batch of images during a new epoch, repeating the process of learning from and rating the validation set multiple times until the model has achieved a satisfactory level of accuracy. This iterative training and validation process allows MobileNet to learn from the images without being subject to overfitting, thus ensuring that its performance on the testing set is as accurate as possible, giving us a model that can accurately detect and classify objects in real-world scenarios.

The research work utilizes a training set consisting of 1034 images, with each batch containing 32 instances, resulting in a total of 33 batches per step. The initial number of epochs is set to 100, but it is important to stop the model when the accuracy and loss have reached a stable state. Monitoring the training and validation loss curves, as well as the accuracy at each epoch, is crucial to ensure effective model training. Although the current number of epochs is set to 100, it is advisable to halt the training process once the accuracy and loss have stabilized. The expected behavior is that the training and validation losses decrease with each epoch, which will be closely observed throughout the training process. We will closely monitor the training and validation loss curves and accuracy at each epoch to ensure the model is trained well; we will also monitor the model to check if it has achieved a good level of generalization by looking for signs of overfitting, such as high training accuracy but low validation accuracy. After assessing the accuracy of the model on the test dataset, we can make necessary adjustments to our training parameters to optimize the model for better performance, such as decreasing the number of epochs, adding regularization techniques, changing the learning rate, and so on. Thus, it is important to consider not just the number of epochs but also other factors when training a model. In conclusion, when training a model, it is crucial to consider a range of factors.

In terms of the present experiment designs' hyperparameter settings, we utilized a filter size of 10 with correlations of 0.8 and 0.001 as the learning rate. This filter size of 10 was chosen as it allowed us to capture the most important features in our data, while the correlations of 0.8 and 0.001 were selected for their ability to optimize the accuracy of our model given the size of our data set and compute resources available, while also allowing us to train our model quickly and efficiently.

To ensure a robust comparison of results, the classification outcomes in this study were obtained by analyzing and comparing various architecture configurations (including hyperparameters and optimization methods) while maintaining similar experimental conditions. In addition, performance measures were implemented for the classification of bean crop disease, and diverse classification methods were also implemented on the testing data in the prediction. The result section will describe the obtained results.

#### 8.1.1.3 Training process

Multiple architectures and optimizers were utilized in the TensorFlow training process of the MobileNet models, including SGD, adagrad, RMSprop, nadam, and adam optimizers, utilizing asynchronous gradient descent. Nevertheless, in comparison with different models, such as Inception, MobileNets employ different regularization techniques and rely on depthwise separable convolution instead of regular convolution used in Inception V3. This leads in fewer parameters in MobileNet, but may result in a slight performance decrease. Hence, it is crucial to carefully manage weight decay on the depthwise filters due to their limited parameters. To strike an optimal trade-off between model performance and parameter efficiency, we applied optimization techniques such as varying learning rates, mini-batch sizes, and training epochs to fine-tune the model. Additionally, regularization techniques like dropout and batch normalization were incorporated to combat overfitting and enhance generalization. By reducing the model's complexity, we improved its generalization ability while maintaining performance and efficiency balance. This resulted in a significant increase in model accuracy without compromising its effectiveness.

Python and the Tensorflow CPU library are used to train and test the MobileNets architecture. On the other hand, we employed another technique called MobileNetV2 (Xiang., 2019), an improved version of MobileNet with a separable convolution as its core. It is a network architecture built on an inverted residual based on convolutional layers. A remarkable outcome was achieved by MobileNetV2, which had been pre-trained on ImageNet datasets, achieved a notable result in feature extraction from fruit images (Shehu et al., 2021). By combining the power of Python with the TensorFlow CPU library and the new MobileNetV2 architecture, we achieved a remarkable outcome in extracting features from bean leaf images. We used this new architecture to classify bean leaves into two main categories: healthy and infected, using various metrics to compare the accuracy and efficiency of the classifier. Figure 8.3 illustrates the system components used for training and testing the suggested disease screening algorithm based on MobileNet V2.

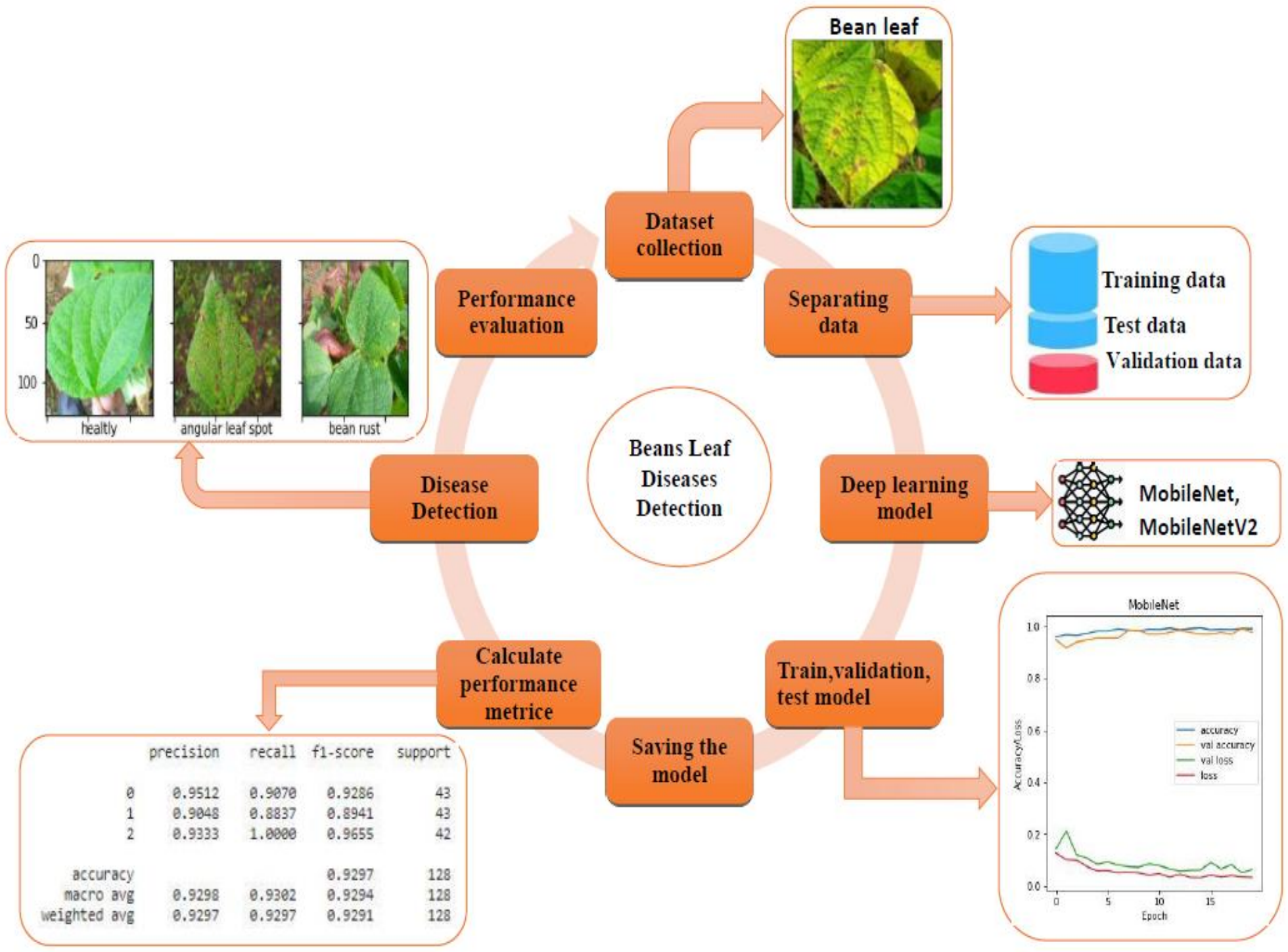

Figure 8.3 The main components of beans leaf disease classification using MobileNet / V2

The experimental analysis was conducted on a Dell N-series laptop with specifications including a 2.5 GHz Intel i6 CPU and 6 GB RAM. The implementation of the system utilized the MobileNet model in conjunction with the TensorFlow open-source library in Python. To expedite the learning process, Google Colab was employed on a personal computer equipped with a GPU, enabling accelerated computations. The inclusion of a GPU significantly reduced learning time and facilitated the processing of a substantial number of examples in each iteration of the learning process. In addition to the specifications of the laptop, the choice of software and open-source library for deep learning was fundamental in helping to process data quickly, accurately, and efficiently. Google Colab allowed for access to a wider array of hardware, enabling the utilization of a more powerful GPU, which was essential in optimizing the speed and accuracy of deep learning, resulting in more efficient processing of the data and faster training time.

### 8.1.2 Result and discussion

The aim of this experiment is to assess and examine the influence of various hyperparameters such as optimizer selection, batch size, and learning rate on the MobileNet and MobileNetV2 architectures. This testing method was carried out and compared consecutively, and each hyperparameter was tested independently to identify the best architectural setting.

All the criteria (architecture, techniques, hyperparameters, filters, etc.) were controlled under identical conditions to compare various architectures' performance. Additionally, random images were chosen for each set using the same dataset. As a result, performance varies depending on the architecture and image utilized. This experiment aimed to compare and identify the optimal hyperparameter configuration for each architecture and determine which architecture produced the highest performance, so developers could determine which architecture best suited their applications; to analyze the efficacy of the architectures and techniques, multiple tests were conducted utilizing various metrics, while training each model's architecture with different hyperparameter combinations. The accuracy of the models was evaluated on a standardized dataset, making it possible to identify the optimal parameters for any given architecture using this method.

In this experiment, the dataset was divided into three classes: healthy, angular leaf spot, and bean rust, with each class consisting of 428, 432, and 436 examples respectively. The dataset was further divided into test, training, and validation sets, with 128, 1035, and 133 examples respectively. The main objective of the experiment was to develop a robust DL model capable of accurately distinguishing between different types of bean leaf diseases. A summary of data distribution is depicted in Figure 8.4.

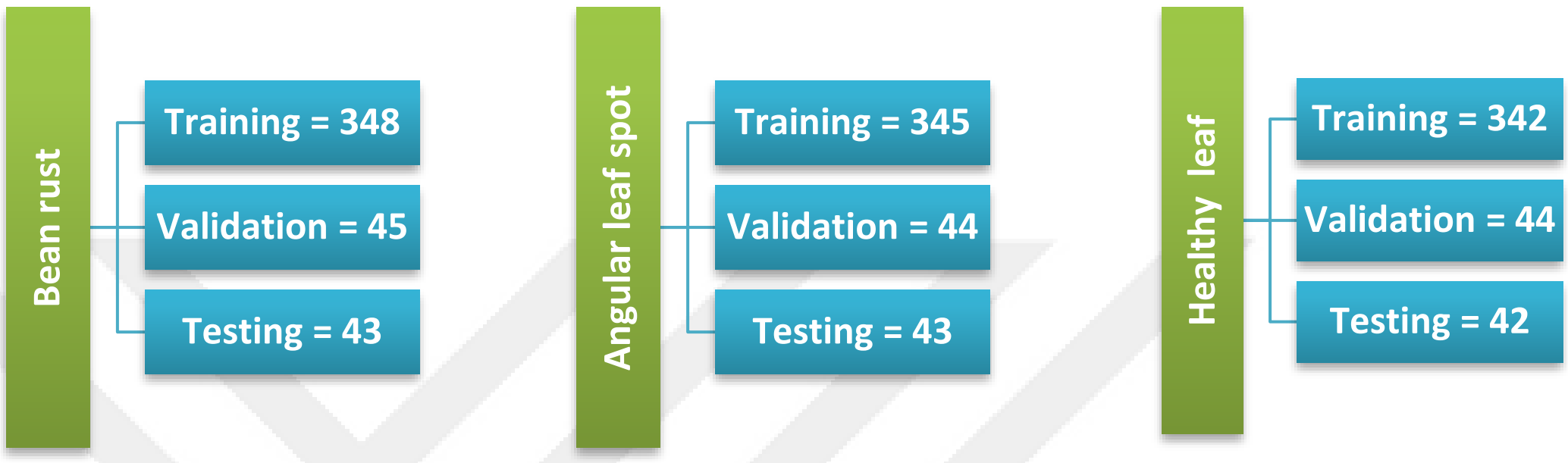

(a) (b) (c)

Figure 8.4 Dataset distribution: (a) Data distribution for Bean rust disease. (b) Data distribution for Angular leaf spot. (c) Data distribution for Healthy leaf

The following sections cover the evaluation criteria utilized to assess the implementation of the sugested approach and present the results obtained for the classification of bean leaf diseases.

#### 8.1.2.1 Optimization method

Using identical experimental conditions, which encompassed methods, architecture, hyperparameters, and dataset, this section carried out tests and evaluations on five different optimizers: nadam, adagrad, SGD, RMSprop, and adam optimizer. The performance of these optimizers was compared by assessing their accuracy on both the training and validation sets. The outcomes of this comparison are shown in Table 8.2.

In the classification results for the three classes of bean leaf diseases, among the five optimizers evaluated, SGD and Adam optimizer demonstrated high performance and accuracy compared to the others. Additionally, both optimizers were found to be more reliable and stable, as they did not experience accuracy decreases. Therefore, for this experiment based on MobileNet, both SGD and Adam optimizers were selected, as depicted in Figure 8.4. The SGD optimizer achieved the second-highest accuracy of 99.94% in identifying bean leaf diseases, while the Adam optimizer performed the best with a 100% accuracy.

Comparing the accuracies of SGD and Adam to other optimizers, the results demonstrate that the Adam and SGD optimizers were more accurate and reliable than others in identifying bean leaf diseases, indicating that the two optimizers were more suitable for use in this application. This comparison also shows that the Adam and SGD optimizers are far more effective than the other available optimizers in terms of accuracy and reliability, making them an ideal choice for the given application. The classification outcomes for the various optimizers we evaluated are shown in Table 8.2. Figure 8.5 depicts a curve that compares the accuracy, loss, validation accuracy, and validation loss performances of the Adam and SGD optimizers.

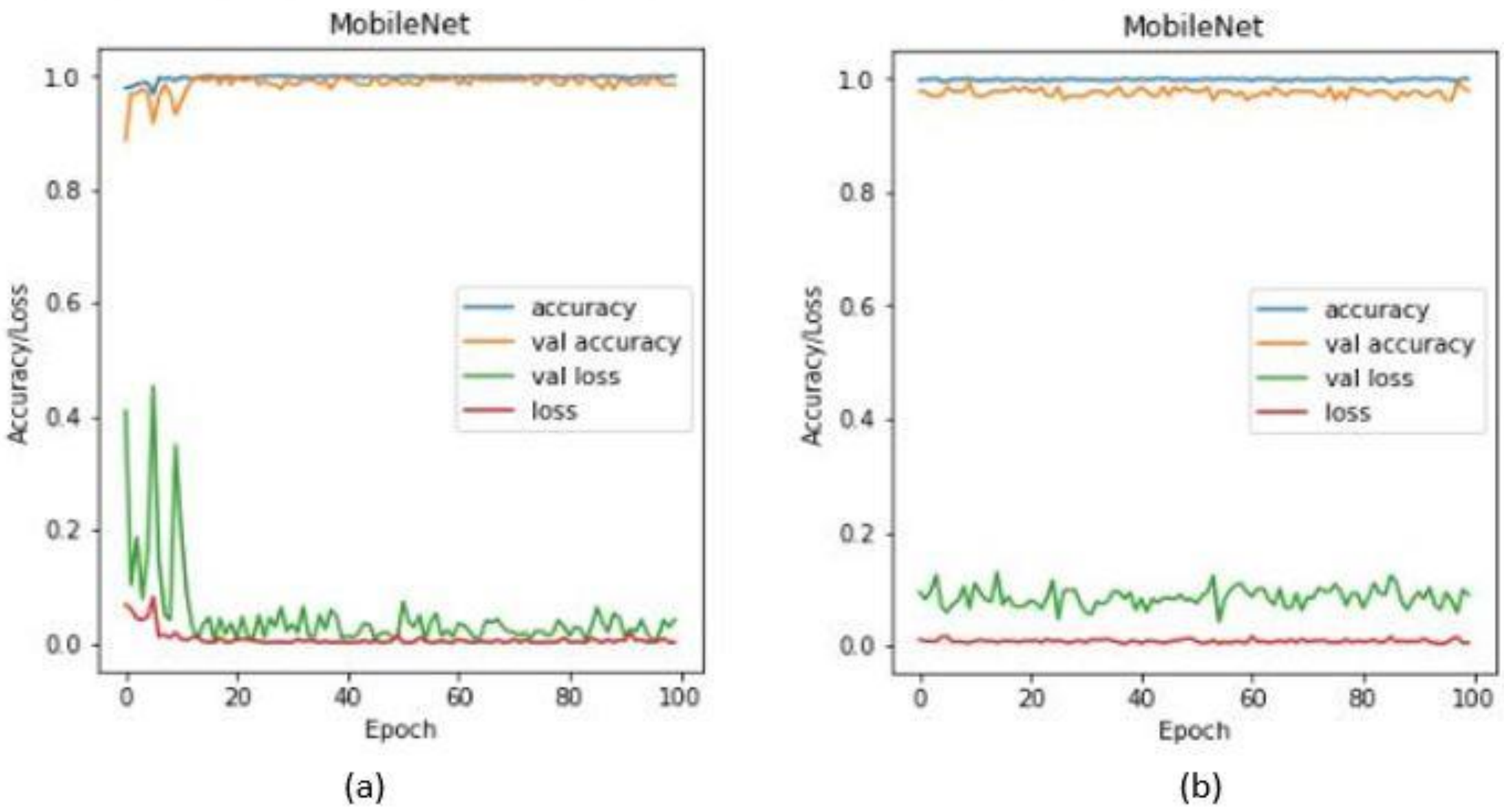

Figure 8.5 Comparing accuracy and loss between two optimizer techniques. a) Adam; and b) SGD

Table 8.2 Comparing accuracy and loss results of five optimizers during training and testing phases

| Optimizers | Accuaracy | Val-accuacy | Loss | Val-loss |
|---|---|---|---|---|
| **Adam** | 1 | 0.9849 | 0.00078 | 0.0411 |
| **SGD** | 0.9994 | 0.9847 | 0.0095 | 0.0478 |
| **Nadam** | 0.9990 | 0.9824 | 0.0254 | 0.0261 |
| **RMSprop** | 0.9924 | 0.9724 | 0.0334 | 0.2059 |
| **Adagrad** | 0.9850 | 0.9699 | 0.0703 | 0.0869 |

We can see from Table 8.2 and Figure 8.5 that the Adam optimizer outperformed the SGD optimizer for all classification outcomes, and this is further confirmed by the accuracy, loss, validation accuracy, and validation loss curves illustrated in Figure 8.5. Furthermore, the Adam optimizer achieved higher accuracy and lower loss for all classification outcomes compared to the SGD optimizer, showing a clear superiority in its performance. The observed performance may be attributed to the adaptive learning rate used by the Adam optimizer, where the learning rate for each parameter is adjusted based on its gradient, allowing it to take larger strides in different directions than SGD and thus making the optimization process more efficient.

#### 8.1.2.2 Learning rate

Manually selecting hyperparameters can be laborious and error-prone process, as the model's dynamics may change over time, rendering previously chosen hyperparameters ineffective. To address this, we employed an automated approach, as illustrated in Figure 8.6 which demonstrates how we assessed the effectiveness of different learning rates in classifying bean leaf diseases with our Mobilenet model. This comparison of learning rate performance is intended to improve disease classification in its various classes significantly. Automated hyperparameter optimization can result in better and more efficient classification results, as the model can dynamically adjust to changing conditions, resulting in improved accuracy and faster training time. To get the best results, we need to perform hyperparameter optimization systematically, taking into account all the various combinations of hyperparameters to get the most optimal solution.

One of the most crucial hyperparameters is the learning rate, which is especially crucial when configuring neural networks. The learning rate determines how much the model's current weights are updated based on the error gradient. It is used to calculate the error gradient based on the model's current state. Therefore, it is a hyperparameter that must be set with great care. Setting it too low may cause the weights never to reach an optimum value, while setting it too high might cause the model to oscillate around the desired outcome, preventing it from converging to the optimum.

Typically, the learning rate is initialized to a small value that gradually increases over time, thus allowing the model to slowly converge to the optimum while still being able to adjust quickly if it starts to diverge from the ideal—as such, correctly tuning the learning rate is of utmost importance to ensure that the model converges to the desired outcome as quickly and accurately as possible. Therefore, this section will present a detailed study to see which learning rate is more effective. To do this, we started by using different learning rates on a sample data and measuring the performance of the model, and then analyzed how the learning rate affects the model's accuracy, loss, validation accuracy, and validation loss performances. Furthermore, we evaluated the model's performance by comparing various learning rates on the same dataset, to gain further insight into how the different learning rates affect the model's performance. Then, we analyzed the data to conclude which learning rate is most effective. We used a graph to compare the performance of various learning rates on the same dataset, with our criteria of accuracy, loss, validation accuracy, and validation loss. Finally, we used the results from our detailed study to provide guidance on which learning rate is best to use in different scenarios.

As a result of the noticeable impact on performance when adjusting the learning rate, and the impressive accuracy achieved with this optimizer, we decided to use the SGD optimizer as the fundamental element for our testing in this case study. With our bean leaf disease identification model architecture in place, we conducted a study on each learning rate to demonstrate how the optimal learning rate varied for our model.

As seen in Figure 8.6 and Table 8.3, we evaluated three learning rates (0.01, 0.001, and 0.0001) on MobileNet architecture, and utilized the same training epochs (100 epochs) for each rate. Therefore, we found that by training the model with a learning rate of 0.001, we were able to achieve an impressive accuracy of 99.90%. while the learning rates of 0.0001 and 0.00001 only achieved 99.89% and 99.87% accuracies, respectively. These results show that an optimal learning rate of 0.001 significantly increases the model's accuracy and its ability to correctly identify bean leaf disease, demonstrating the importance of an optimal learning rate to achieve better results with deep learning models. Table 8.3 provides the performance results for different learning rates, while Figure 8.8 visually compares the accuracy and loss of the model for these varied learning rates.

Table 8.3 Comparison of accuracy and loss performance for different learning rates

| **Learning rate** | **Tr-accuaracy** | **Tr-loss** | **Val-accuacy** | **Val-loss** |
|---|---|---|---|---|
| **0.001** | 0.9990 | 0.0071 | 0.9774 | 0.0830 |
| **0.0001** | 0.9989 | 0.0080 | 0.9773 | 0.0714 |
| **0.00001** | 0.9987 | 0.0081 | 0.9850 | 0.0700 |

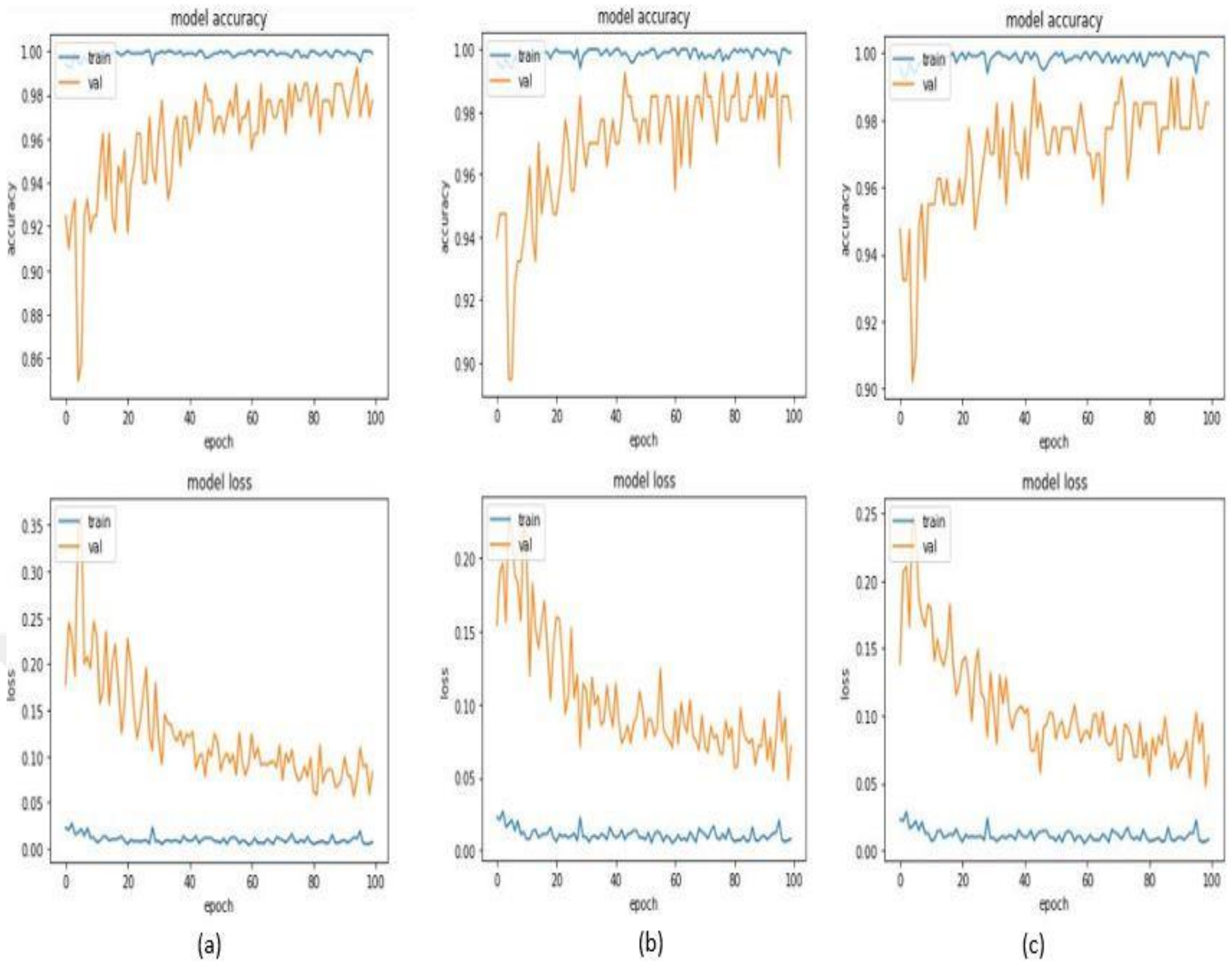

Figure 8.6 Performance evaluation of different learning rates: (a) 0.001, (b) 0.0001, and (c) 0.00001, based on accuracy and loss metrics

Furhermore, based on the observed accuracy, the performance remains consistently high across all learning rates, with accuracy varying from 99.87% to 99.90%; additionally, the loss values range from 0.0071 to 0.0081. These findings suggest that the learning rate had a discernible impact on the training accuracy, while the performance remained independent of model size. Furthermore, our findings show that the learning rate affects the accuracy of deep learning models, implying that a well-tuned learning rate may be required to achieve higher accuracy in model training. Our study revealed that an optimal learning rate of 0.001 dramatically increases the accuracy of deep learning models in correctly identifying bean leaf disease, compared to other learning rates.

#### 8.1.2.3 Batch size

The batch size plays a critical role as a hyperparameter, determining the number of images processed by the network in each training iteration. In essence, it dictates the quantity of training instances that a model works on per iteration. Modifying the batch size provides us with control over the amount of data that moves through the network in a single iteration, as well as the amount of memory the model consumes. The significance of this hyperparameter is evident in its impact on both the training time and accuracy of a model. Adjusting the batch size allows us to have more control over a model's training process, as it can reduce memory consumption and optimize the training time. Furthermore, choosing the number of batches can also have a regularization effect, which helps reduce overfitting. We can find a balance between training time, memory consumption, and model regularization by varying the batch size. Therefore, careful selection of the batch size is important for optimizing a model's training process and achieving an accurate result. Choosing an appropriate batch size can be difficult, as it requires analyzing the trade-off between regularization and training time; this decision relies on careful consideration of factors such as the dataset size, model complexity, architecture, and available computing resources.

Using the batch size strategy, we found that training error fluctuates very little during testing, but larger batches would result in overgeneralization. Additionally, we tested and compared batch sizes 32, 68, and 128 for classification accuracy, and we reported that batch size 32 provided the best results with an accuracy value of 99.90%; this indicates that a batch size of 32 is the most suitable for our model due to its consistent performance on training data and relatively high accuracy, as compared to the other batch sizes tested. Furthermore, a batch size of 32 provides the best balance of generalization and overfitting, ensuring consistent performance on training data while maintaining high classification accuracy. Therefore, the results from our experiments provide strong evidence to support that a batch size of 32 is the most suitable for our model, providing a balance of generalization and overfitting while still delivering exceptional classification accuracy.

Furthermore, Figure 8.7 and Table 8.4 shows that increasing the batch size from 32 to 128 results in a decrease in classification accuracy for this study. This highlights the importance of adjusting the batch size to ensure that the number of images processed remains consistent across each epoch, without being too high or too low.

The choice of batch size has a direct impact on the accuracy of gradient error estimations when training neural networks. Smaller batch sizes require lower learning rates to achieve optimal accuracy. Therefore, it is crucial to select a batch size that is appropriate and strikes a balance between being too small or too large. Using a batch size that is excessively small can result in greater fluctuations in model accuracy due to the reduced amount of training data being processed within each batch. Therefore, having a smaller batch size leads to better accuracy as the network is trained using fewer samples, which reduces the noise associated with the training data and hence results in a more reliable estimation of the gradient error while having an overly large batch size leads to inaccuracy due to the large amount of data being used for training, which increases the noise associated with the training data.

Consequently, the batch size should be neither too small nor too large, as this will ensure that the model is trained with a reliable estimation of the gradient error and that the networks are provided using sufficient amount of training data to enable it to learn effectively. Furthermore, this means that the best batch size is one that is appropriate for the type of data and model being used and that allows for enough training data to reduce the noise associated with the training while also not having too much noise due to a large batch size, resulting in accurate model. Table 8.4 presents a comparison of accuracy and loss across different batch sizes. Figure 8.7 visually represents the same performance information in a graph format.

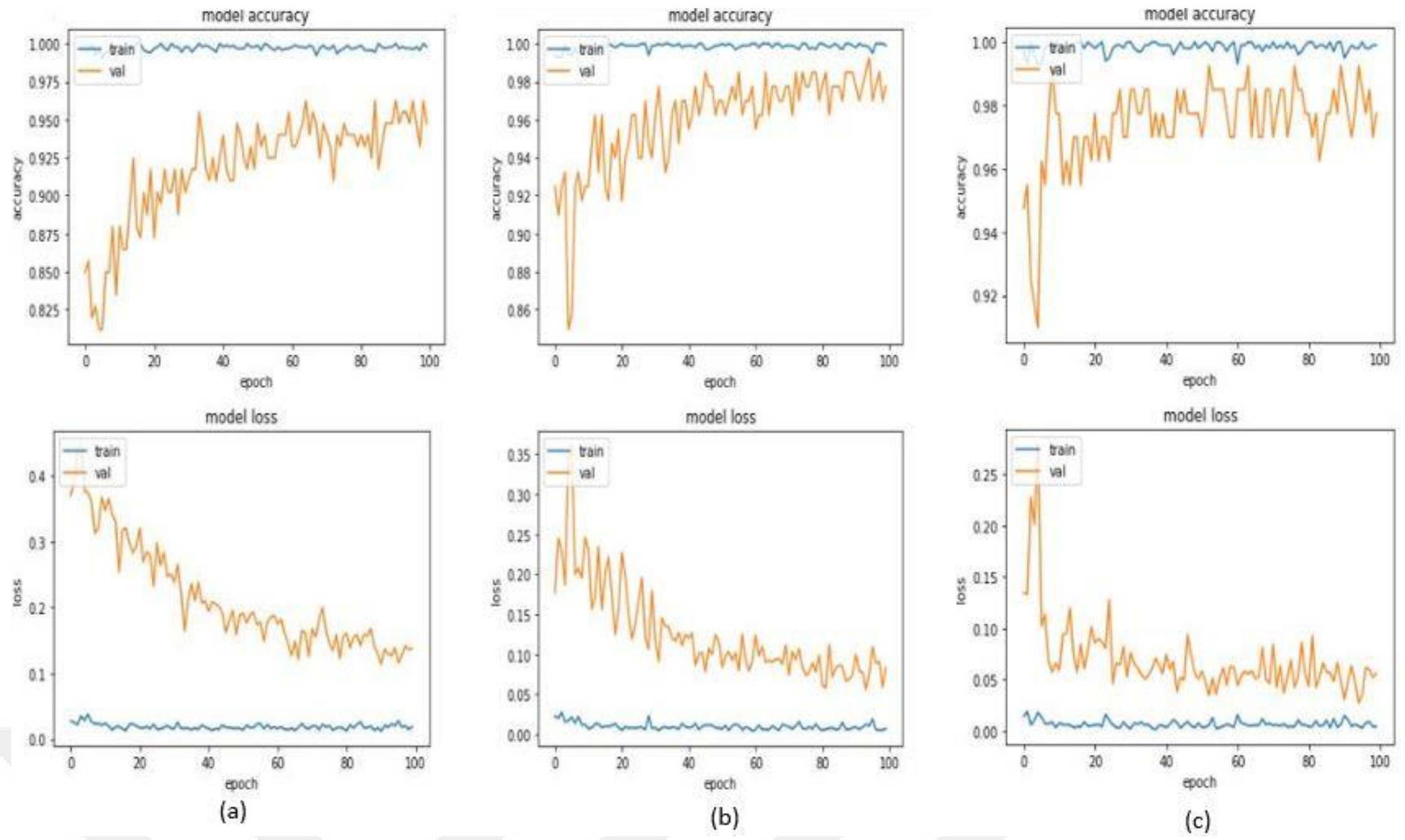

Figure 8.7 Comparing accuracy and loss using various batch sizes: (a) 128, (b) 64, and (c) 32

Table 8.4 Comparison of the accuracy and loss in various batch sizes

| Batch size | Tr-accuaracy | Tr-loss | Val-accuacy | Val-loss |
|---|---|---|---|---|
| **32** | 0.9990 | 0.0045 | 0.9774 | 0.0560 |
| **64** | 0.9989 | 0.0071 | 0.9773 | 0.0830 |
| **128** | 0.9978 | 0.00190 | 0.9474 | 0.1371 |

#### 8.1.2.4 Performance comparison of MobileNet and MobileNetV2 in identifying leaf diseases

For their prevention and control, bean leaf diseases must be accurately classified, and the objective of evaluating and comparing various infrastructure performances on a single dataset holds significant value in its own right; this is due to the fact that when model is analyzed and prepared for comparisons, detailed information that is relevant for retraining purposes is captured and recorded. Therefore, evaluating and comparing different infrastructure performances on a single dataset is essential. Moreover, this approach enables us to pinpoint the most suitable classification model that fulfills the specific data and business requirements. The best model will help us improve the accuracy of disease prediction and also help minimize the time and cost associated with data collection and model evaluation, making the entire process of bean leaf disease detection and control much more efficient. Moreover, the public dataset also allows us to understand the pattern and trend of bean leaf diseases in different parts of the world, which helps us make well-informed decisions on disease prevention and control. This is important because a single dataset can enable us to compare and analyze different infrastructure performances and help us identify the best classification model that suits our data. Therefore, it is essential to accurately classify bean leaf diseases, enabling us to make informed decisions on how to address them best.

In this section, we have conducted a test experiment for classifying images of bean leaf diseases. MobileNet and MobileNetV2, which uses depthwise separable convolution as functional building blocks, were used as the base models. The output layer and hyperparameters were modified to meet the classification condition requirements, and the dataset were trained accordingly. The experiment was carried out over 100 epochs using a learning rate of 0.001 and a batch size of 64.

As evident from the results presented in Figure 8.8, our implementation of the MobileNet model exhibited exceptional accuracy, with an average of 100% and a low loss value of 0.0112, while training for 100 epochs in just 173s. We also investigated the performance of MobileNetV2 to assess whether we could further improve the accuracy and reduce training time. Interestingly, the use of low parameter number constraints allowed MobileNetV2 to surpass MobileNet in terms of accuracy for classifying bean leaf disease. The model achieved the same level of accuracy with a lower loss value of 0.0102 and trained faster, completing 100 epochs in only 165s (as shown in Table 8.5). MobileNetV2 was then faster than MobileNet and highly effective for classification. Despite MobileNetV2 having lower parameters than MobileNet, it was still able to achieve faster training durations, some accuracy, and a reduced loss value, making it a much more desirable choice for the task of bean leaf disease classification. This is an excellent example of how, with the right architecture, it is possible to achieve higher performance even with fewer parameters without the need to increase model complexity or the risk of overfitting.

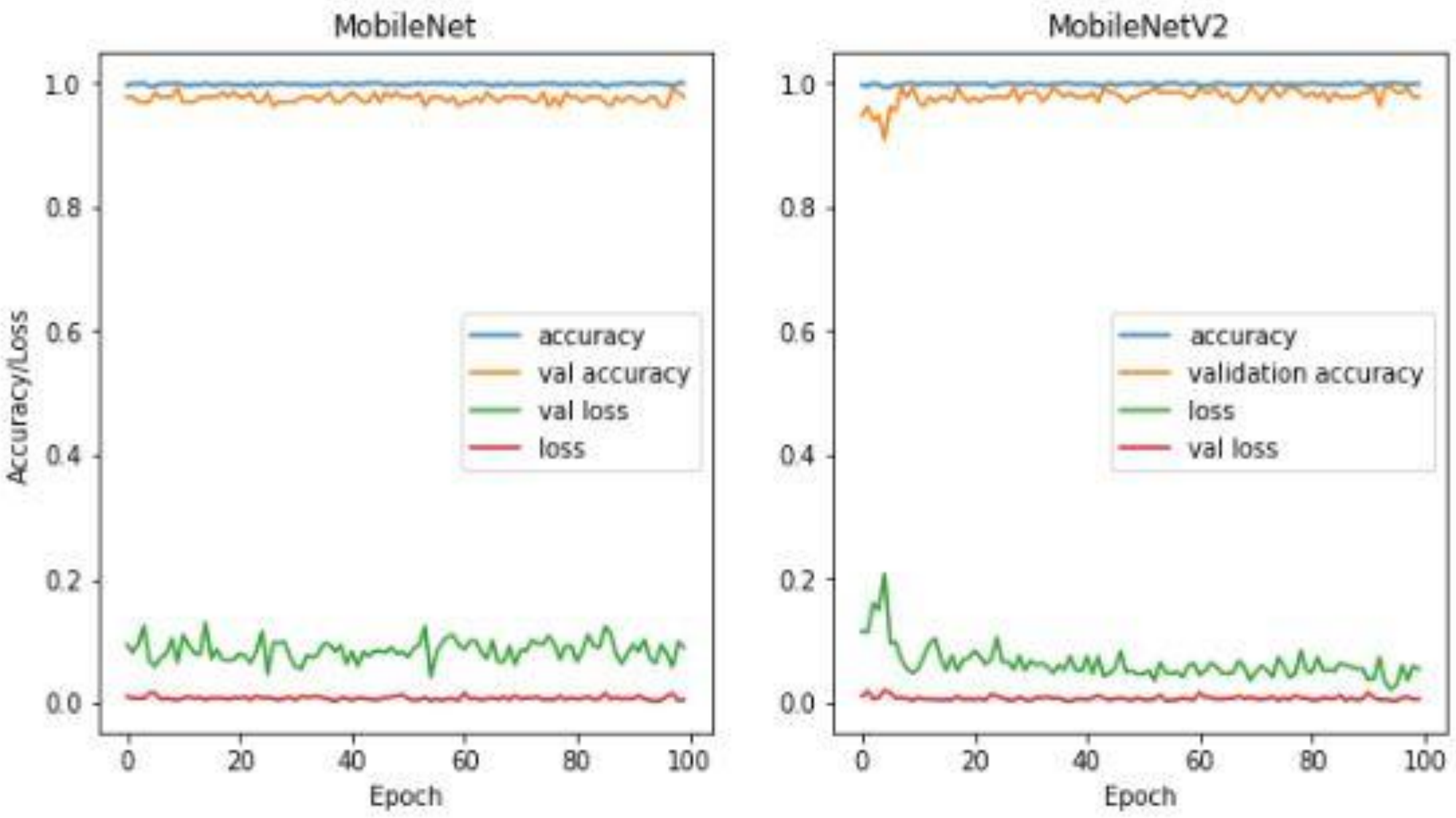

Figure 8.8 MobileNet accuracy and loss comparison to MobileV2 architecture

Table 8.5 Performance comparison between MobileNet and MobileNetV2

| Methods | Accuaracy | Val-acc | Loss | Val-loss | Epochs | Time |
|---|---|---|---|---|---|---|
| **MobileNet** | 100% | 0.9945 | 0.0093 | 0.1223 | 100 | 173s |
| **MobileNetV2** | 100% | 0.9949 | 0.0079 | 0.1119 | 100 | 165 s |

Figure 8.8 presents a unique visualization of the training progress, showcasing the performance of MobileNet and MobileNetV2 in accurately identifying bean leaf diseases. The graph includes accuracy, loss, and validation accuracy metrics for both architectures. Accuracy serves as a key metric for evaluating classification performance, specifically for three classes: healthy leaves, bean rust disease, and angular spot. Therefore, this visual representation offers valuable insights into the comparative performance of the two models in accurately classifying various types of bean leaf diseases. The graph also shows that the two architectures are relatively comparable in terms of accuracy and loss; however, MobileNetV2 outperforms MobileNet by a small margin in all three categories.

The performances in terms of accuracy and timing are especially telling, with MobileNetV2 having a consistent advantage across the three classes. Moreover, MobileNet is also efficient and effective in recognizing and identifying bean leaf diseases compared to MobileNetV2, but MobileNetV2 outperforms in terms of time and losses. Furthermore, this indicates that MobileNetV2 is better suited to accurately and quickly identify bean leaf diseases than MobileNet. It suggests that it could be utilized in real-time applications where timely recognition and classification of diseases are required. This advantage that MobileNetV2 has over MobileNet is even more pronounced when considering that it also uses less computation time and resources than MobileNet, making it a an ideal choice for real-time applications. Following the completion of 100 epochs of training, our model demonstrated convergence with an impressive accuracy range of 99% to 100%. Furthermore, the validation accuracy reached a commendable value of 97.74%. The visual representation in Figure 8.9 illustrates the model's accuracy plot at the end of each epoch, offering valuable insights into the training progress and overall performance of the model during the training process. The plot illustrates that the model's accuracy increased over time, reaching a peak at 100 epochs.

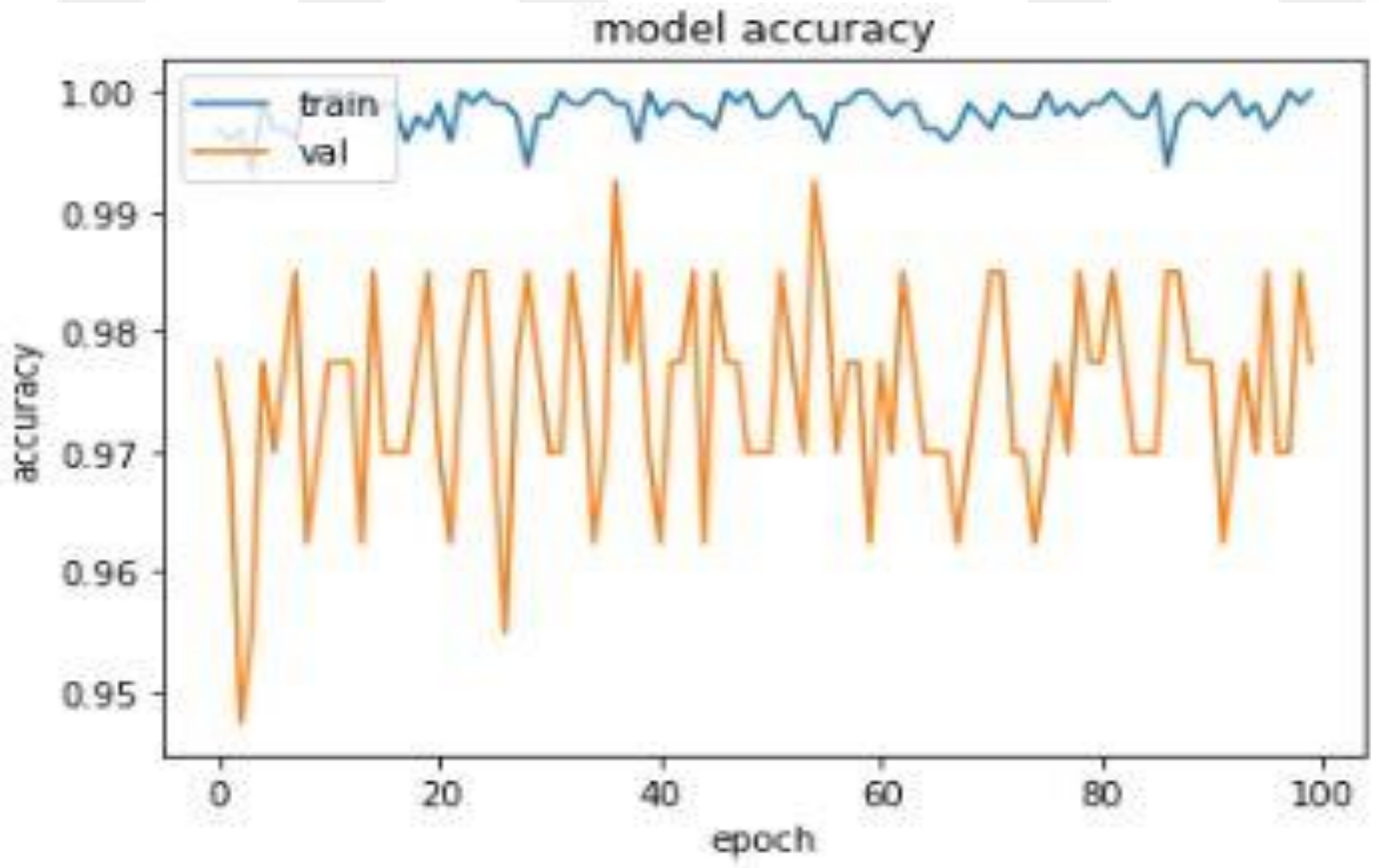

Figure 8.9 Accuracy performance of basic classifier of the model

Moreover, this suggests that the model had successfully learned from the training data and could identify the data accurately. Furthermore, from this plot, we can see that the model had achieved a steady convergence at the end of 100 epochs and had achieved its maximum accuracy at this point, which further confirms that the model had learned the data well. This high accuracy also supports our initial hypothesis that, with sufficient training, the model will be able to identify data accurately. We can also infer that, given more training data, the model would be able to learn better and eventually reach higher accuracy.

#### 8.1.2.5 Disease detection performance analysis

This section presents an extensive performance analysis of our MobileNet model specifically designed for detecting bean leaf diseases using an image leaf dataset. To thoroughly evaluate our approach's effectiveness, we employ various metrics, such as Accuracy, Recall, and F-score. In addition, we utilize the powerful tool of a confusion matrix, which provides a detailed summary of prediction outcomes for the classification task at hand. By comparing true labels and predicted labels, the confusion matrix enables us to gain valuable insights into the accuracy of our classification model and identify instances where it may encounter confusion during predictions. Our classifier is rigorously tested on three distinct classes: healthy bean leaf, bean rust disease, and angular leaf spot disease, using two labeled sets for comparison. By examining the confusion matrix, we can comprehensively evaluate the performance of our model by observing correct and incorrect predictions for each class, thereby highlighting potential improvement areas. This thorough analysis not only helps assess the accuracy of our classification model but also guides us in making informed adjustments to enhance its performance and ensure its reliability in practical applications.

During the conducted experiments, the model underwent training using the same set of hyperparameters as mentioned earlier. However, it required a few extra iterations to attain convergence and stability within the designated 100 epochs. As expected, the proposed architecture and approach, trained on the dataset images, achieved an impressive accuracy rate of 92.97% on the test set, consisting of 128 images distributed among three distinct classes. A visual representation of these results can be found in Figure 8.10. Moreover, the training accuracy of the model reached 99.87%, while the validation accuracy reached 97.44% for the purpose of classifying leaf diseases. To provide a concise summary of the classification model's performance on the three bean leaf image classes, Figure 8.10 showcases the confusion matrix.

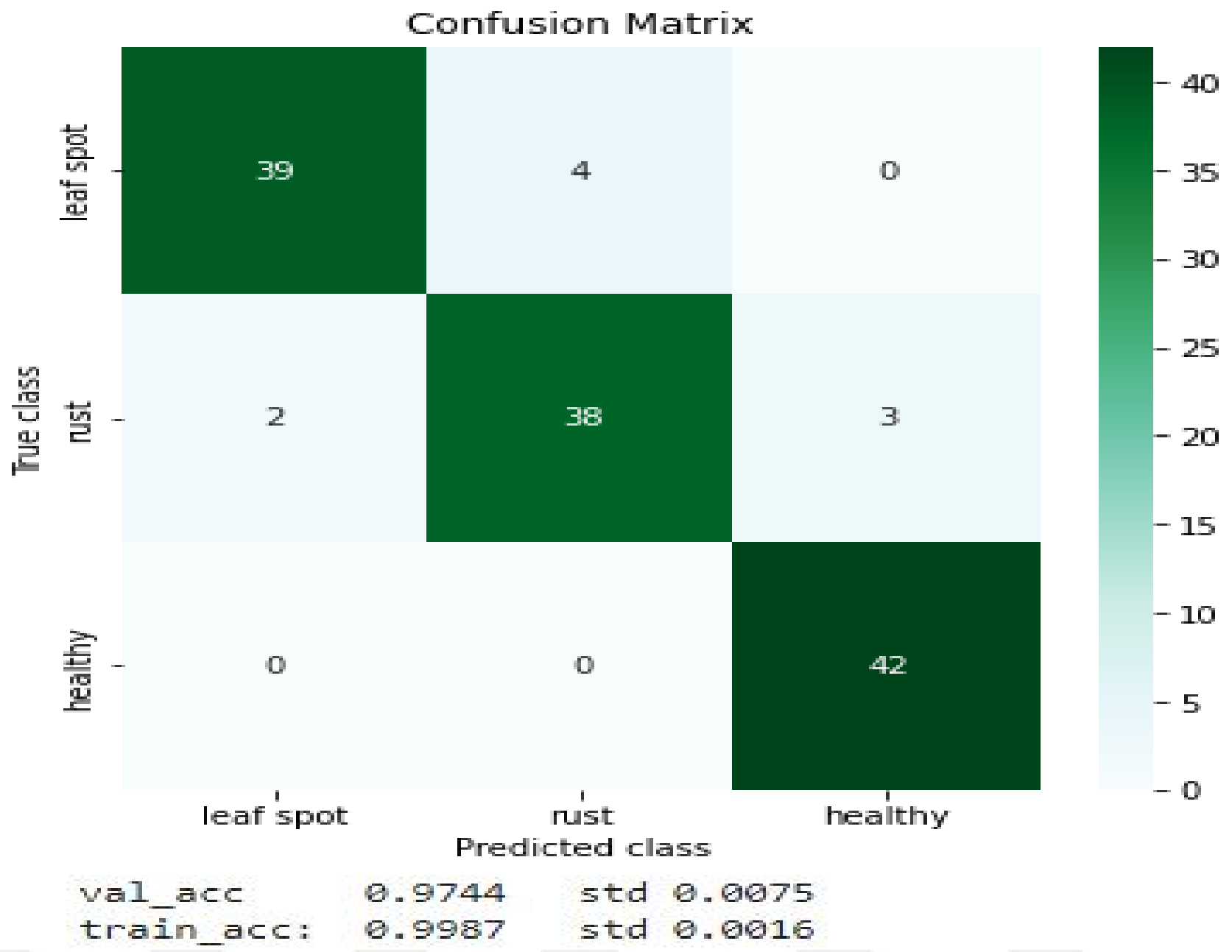

Figure 8.10 Confused matrix for a classification model

The outcomes indicated that the suggested architecture successfully distinguished between the three classes, with only 3 misclassified images out of 128 images used, demonstrating significant improvement over the existing models.

The model's performance was assessed in this study using a variety of performance metrics, to comprehensively assess the system's success, the suggested approach was evaluated utilizing different metrics, including accuracy, recall, and F1 score. Accuracy was used to evaluate the overall correctness of the model, while recall and precision were utilized to assess the number of correctly classified and rejected instances. The F1 score was then used to assess the model's overall implementation by taking into account both accuracy and recall

Since the confusion matrix provided us with all the data we needed for each class, we were able to compute the performance metrics for each class individually, as illustrated in Figure 8.11.

| | precision | recall | f1-score | support |
|---|---|---|---|---|
| 0 | 0.9512 | 0.9070 | 0.9286 | 43 |
| 1 | 0.9048 | 0.8837 | 0.8941 | 43 |
| 2 | 0.9333 | 1.0000 | 0.9655 | 42 |
| accuracy | | | 0.9297 | 128 |
| macro avg | 0.9298 | 0.9302 | 0.9294 | 128 |
| weighted avg | 0.9297 | 0.9297 | 0.9291 | 128 |

Figure 8.11 The performance metrics of three classes

The evaluation results showed that the proposed system using the MobileNet model achieved high scores for accuracy, recall, and precision, resulting in an accauarcy of 0.9297, demonstrating its overall effectiveness in image classification. The performance metrics employed in this study proved to be effective in accurately assessing the MobileNet model's performance in accurately classifying images. These results demonstrate that the MobileNet model can effectively classify images with high accuracy, leading to a successful conclusion of the study.

#### 8.1.2.6 Comparative analysis with state-of-the-art works

Comparative research is an important aspect of scientific inquiry, as it enables us to assess the effectiveness of our suggested methods compared to existing techniques. In this study, we compared suggested model with those already in use using the accuracy metric presented in Table 8.6 and Figure 8.12. As a result, the MobileNet architecture obtained remarkable accuracy of 92% for the bean crop even though it used only a small dataset size (1296 images) compared with existing works. This comparative research demonstrates that the work offered for leaf disease detection is successful. Moreover, our proposed model exhibited an impressive accuracy rate of 92%, with the training phase achieving a perfect accuracy rate of 100%. Furthermore, during the validation phase, the model achieved a commendable accuracy rate of 98.49%, which means that these results are comparable to the existing methods, especially those using the MobileNet model such as MobileNet (Abbaset al., 2021) 90%, MobileNetV2 (Ai et al., 2020) 90%, and Mobilenet (Adedoja et al., 2019) 91.7%. These results indicate that our model is a promising option for identifying leaf diseases in bean crops; this study also highlights the potential of using MobileNet architecture to accurately classify leaf diseases in bean crops. Table 8.6 compares state-of-the-art work, and figure 8.12 presents a comparative analysis of the latest advancements in the field.

Table 8.6 Performance comparison with state-of-the-art Works

| **Reference** | **Dataset** | **Dataset size** | **Method** | **Performance** |
|---|---|---|---|---|
| (Vinutha et al. 2019) [1] | PlantVillage | - | MobileNet | Accuracy = 90% |
| (Siti Zulaikha et al., 2020) [2] | PlantVillage | - | MobileNet V2 | Accuracy = 90% |
| (Ashwinkumar et al., 2022) [3] | PlantVillage | 12000 | MobileNet | Accuracy = 91.7% |
| (Mukti et al., 2019) [4] | GitHub | 87867 | ResNet50 | Accuracy = 97.83% |
| (Sambasivam et al. 2021) [5] | Kaggle | 10000 | DCNN | Accuracy = 93% |
| (Geetharam et al.,2022) [6] | PlantVillage | 54,305 | Deep CNN | Accuracy = 96.46% |
| (Agarwal et al., 2020) [7] | PlantVillage | 55,000 | CNN | Accuracy = 91.2% |
| (Yasin et al., 2021) [8] | PlantVillage | 54183 | DenseNet | Accuracy = 98% |
| (Chen et al., 2021) [9] | Institute | 8616 | B-ARNet | Accuracy = 89% |
| Our study | GitHub | 1296 | Our model | Accuracy = 92% |

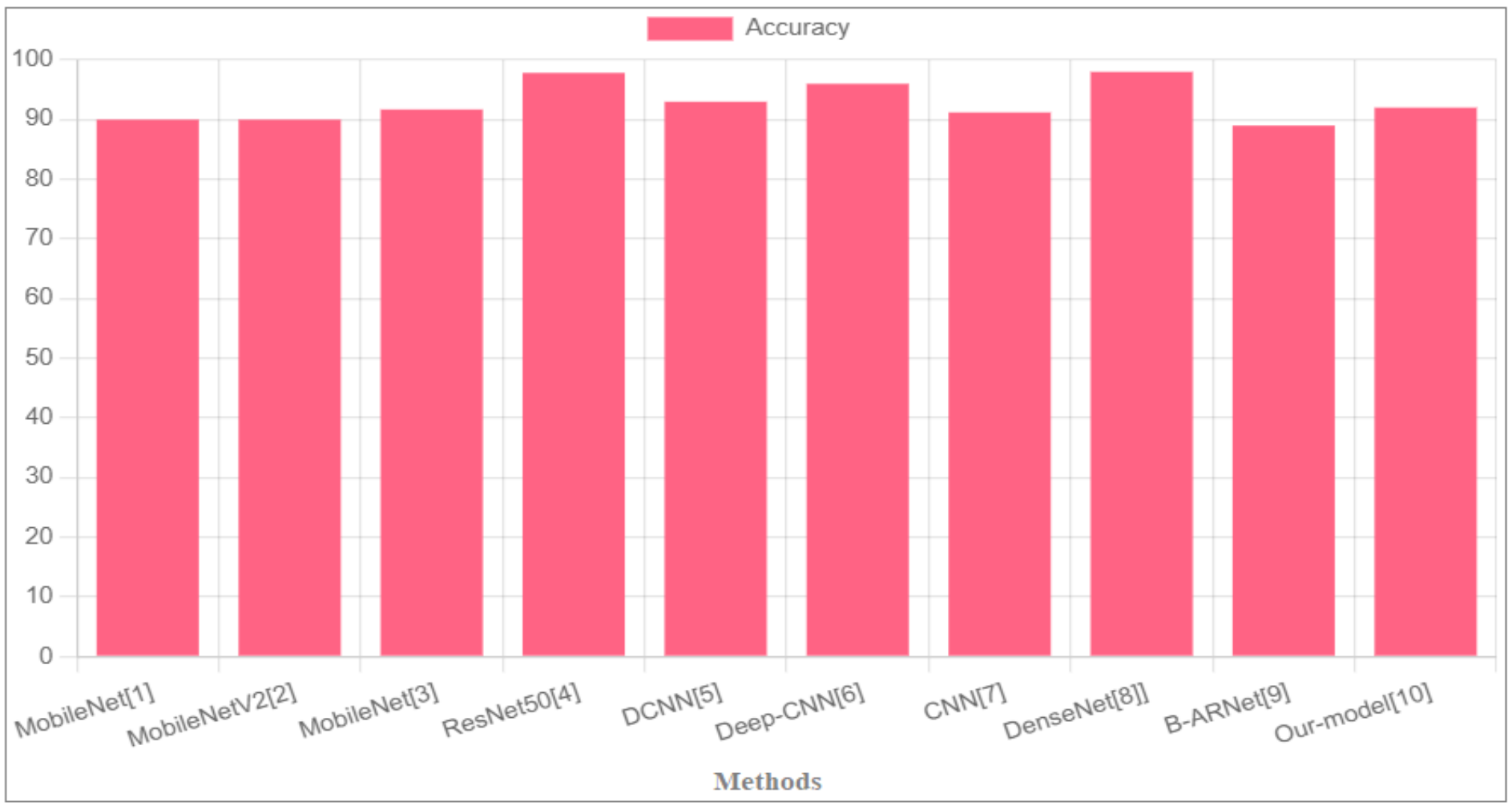

Figure 8.12 The comparative performance analysis with state-of-the-art works

Overall, the study highlights the potential of utilizing DL methods for accurate and efficient diseases detection in bean crops. Furthermore, the outcomes strongly indicate that the suggested model can be effectively utilized to detect leaf diseases in bean crops early, significantly reducing crop losses and increasing yields. Future studies could focus on expanding the dataset to include other types of crops and diseases to confirm the efficiency of the proposed model. It could also explore the application of this model for other crop types and expand its usage in precision agriculture. Overall, the proposed model has significant implications for improving crop yield and reducing the use of pesticides in agriculture. Farmers can take targeted actions to mitigate the spread of disease and protect their crops by accurately identifying and diagnosing leaf diseases. Future studies can also investigate integrating this model with other precision agriculture technologies to optimize crop management practices.

In summary, in this part of thesis we created an automatic classification model using MobileNet, bean leaf images, and an effective network architectural style to create precise models that can quickly classify the bean leaf disease into their respective classes. Our work encompasses more than just proposing a classification technique for bean leaf diseases, but also we conducted thorough performance analysis and compared multiple architectures to identify the optimal approach for disease classification; a very satisfying result of classification was achieved. This indicates the effectiveness of our proposed technique and shows that it is a suitable approach for the detetction of plant leaf diseases with reasonable accuracy, these obtained results unequivocally show that our suggested technique performs better in terms of accuracy in classifying plant leaf diseases when compared to existing methods.

Additionally, employing a batch size of 32 and a learning rate of 0.001, our model exhibited exceptional performance, attaining a training set accuracy of 100% and a validation set accuracy of 98.49%. Notably, the model demonstrated a test data accuracy of 92.97%, with the lowest training and validation accuracy rates reaching 98.50% and 94.74% correspondingly, indicating the optimal performance of our suggested approach. We also observed that the classification training accuracy is affected by changes in hyperparameters, sincluding batch size and learning rate, highlighting the importance of carefully tuning these parameters to achieve optimal model accuracy.

The results obtained from this experiment confirm the effectiveness of our proposed technique and show that it is a suitable approach for the identification of plant diseases with good accuracy, recall, precision, and values**.** In addition, the computational efficiency of our proposed technique allows it to process large datasets quickly and accurately making it an ideal solution for some real-world application.

Our suggested model study for classifying leaf diseases was effectively applied and examined, providing an excellent n terms of classification accuracy. However, although the approach is computationally successful, it has only been tested on one public bean leaf dataset. In the next section, we will provide a detailed study that will use three different datasets and the MobileNet model to analyze the impact of the datasets on the DL model's performance.

## 8.2 Impact of Datasets on The Effectiveness of Mobilenet for Leaf Disease Detection

Plant disease detection is critical in agriculture. As a result, advanced agricultural techniques should be applied to improve agriculture. For example, a system for accurately detecting plant diseases can be created to increase plant quality and quantity; this is practical since it decreases workload, particularly in large production fields. Furthermore, by using an accurate detection system, farmers can spot and address any potential issues with their crops as soon as they arise, which can potentially save them valuable resources in terms of time and money. Furthermore, the implementation of such a disease detection system can aid in preventing the space of diseases and maximizing crop yields, allowing farmers to optimize their resources more efficiently. Moreover, an accurate detection system can provide farmers with detailed data about the health of their crops, which can be used to adjust farming strategies to improve the overall productivity of their fields. However, most of these detection systems suffer from many challenges, such as the effect of dataset on the trained model's performance, the lack of scalability, The scarcity of available data, and the difficulty of setting the correct parameters. Therefore, in this work, we will present an approach to discuss these challenges, and especially, we will evaluate the impact of datasets on the model's performance of this approach by evaluating the model using three different bean leaf image datasets of varying difficulty. MobileNet is used in this study because it is fast and can achieve high performance with fewer parameters. It is essential to detect bean leaf diseases for application in real-world settings. Further, we determine how much the model's performance can change based on the datasets and present a comparative study of the three datasets.

The model's parameters are assumed to be constant for the most accurate comparisons. Finally, the models' performance is compared to the three datasets using various evaluation measures including training/validation accuracy and loss. We aim to examine how different datasets can affect the accuracy of this approach, whether MobileNet can distinguish between bean leaf diseases with various datasets, and how its performance may be improved if more datasets are available, given the challenges associated with identifying bean leaf diseases. This approach is intended to show the potential of using a CNN such as MobileNet to recognize bean leaf diseases and to develop an improved system for diagnosing and treating bean diseases.

Our research also aims to demonstrate how the GradCAM technique can improve the interpretability of a MobileNet CNN model when classifying bean leaf images. By shedding light on the model's decision-making process, our study can lead to better performance and further exploration. We also pinpoint the most crucial visual features for accurately detecting bean leaf diseases, providing practical insights for developing more effective disease detection and prevention strategies for bean crops.

This work contributes through the following tasks:

- This work aims to analyze how different datasets influence the performance of the methods, by testing the model on different bean leaf image datasets of varying difficulty. This study offers valuable insights for researchers to choose appropriate datasets for their models.
- We suggest an approach that applies a deep-learning algorithm, MobileNet, to identify bean leaf diseases for use in practical and real-world contexts. This approach is anticipated to achieve satisfactory results on three databases of annotated images of healthy and unhealthy bean leaves, significantly outperforming the baseline accuracy.
- Evaluate the method's generalizability by analyzing its performance on three distinct datasets of various difficulty.
- Analyze and assess the efficacy of the suggested method through a comparison with other advanced DL methods and enhance the precision of DL models for the detection of bean plant diseases. This can contribute to reducing the impact of the disease on bean crops and improving food security.
- Generate a novel dataset and use it to analyze the generalizability of the approach.
- Setting a benchmark for future studies: The study can set a benchmark for future studies on beans leaf disease detection based on deep learning. The findings and insights of this study can provide guidance for future researchers to improve the effectiveness of their models.
- Demonstrates how the GradCAM technique can effectively visualize the features employed by a MobileNet CNN model to classify leaf images into distinct categories.

The remaining sections of this study will be structured as follows: The system configuration and dataset utilized will be presented first, followed by a thorough explanation of the model architecture and augmentation methods. Finally, the experimental setup, performance assessment, and results obtained will all be presented and discussed at the end.

### 8.2.1 Materials and methods

This section includes a detailed discussion of the suggested method as well as the datasets that were used. The section also describes the steps taken to increase the efficacy of the suggested method's implementation.

#### 8.2.1.1 Dataset description

To assess the model's efficacy, we run it on three distinct datasets: (i) the public dataset; (ii) collected dataset; and (iii) the merged dataset, as shown in Table 2. After running the models on each of the datasets, we compared the performance to determine its efficacy, examining the precision, recall, and accuracy results.

The public dataset was obtained from GitHub. It was designed by the Makerere AI research lab and will be launched in January 2020 (Makerere AI et al., 2020). The dataset comprises bean crop leaf images obtained from a farming environment in real-world conditions using a smartphone camera. It is constituted of 1296 images separated into three categories.

In addition to the public dataset, we collected a new dataset from an agricultural area using a smartphone camera. All of the images in the gathered dataset were manually sorted into three separate classes (two unhealthy classes and one healthy class). Furthermore, we used data augmentation to increase the size of the dataset by adding more images before training the data because the custom-created dataset only has 971 images. The overall quantity of images after using the data augmentation approach is displayed in Table 8.7.

Table 8.7 Summary of the used bean datasets

| Dataset | Disease | Number of image |
|---|---|---|
| **Public Dataset** | Angular Leaf Spot | 432 |
| | Bean Rust | 436 |
| | Healthy | 428 |
| **Own Dataset** | Angular Leaf Spot | 412 |
| | Bean Rust | 416 |
| | Healthy | 403 |
| **Merged Dataset** | Angular Leaf Spot | 844 |
| | Bean Rust | 852 |
| | Healthy | 831 |

By using data augmentation, we were able to increase the variety of images within each class, allowing us to train our model better, leading to an increase in our model's accuracy. The merged dataset, which combines (hybridizes) the public and collected datasets, is the third dataset utilized in this study. The major goal is determining whether a single model will work effectively on various datasets. By merging the public and collected datasets, this study will provide a more comprehensive evaluation of the effectiveness of a single model when applied to different datasets since it will have access to both a small and large sample size of data. Moreover, this can be especially useful when attempting to develop a generalized model that can be applied across various datasets. It will facilitate a more thorough assessment of the model's performance. By merging the two datasets, this study can also better understand the relationships between different data points across different datasets, as the collected dataset can provide more detailed information that is not readily available from the public dataset.

#### 8.2.1.2 Image augmentation techniques

Data augmentation is a valuable technique that enhances the training process by increasing the variety and diversity of the data through the generation of additional examples from the original dataset. This approach plays a vital role in improving the efficiency of the model and the overall learning process. Essentially, data augmentation allows for the expansion of training data, enabling deep neural network models to handle larger and more intricate datasets, thereby enhancing their performance. Data augmentation methods attempt to provide solutions to the DL problem of limited training data by creating synthetic or "fake" data that emulates real data but looks more realistic because it has been enhanced in some way. Data augmentation has several advantages, such as enhancing the number of training examples, diversifying the set of samples that the model trains on and thus reducing overfitting, and enabling data-hungry models to be trained with smaller datasets, As a result, the generalizability of the model is enhanced, leading to improved performance while simultaneously enhancing the quality and overall effectiveness of deep learning techniques. There are many different data augmentation methods available. The most basic form of data augmentation is random rotation, which rotates each training data image around a random axis. The input data is effectively increased in size by rotating each input image. Another way to improve data is by adding noise. Noise is simply an unstructured value added to the original data set, making it more difficult for the algorithm to generalize and improve overall accuracy as it tends to reveal hidden patterns in the original data set.

Image augmentation works by applying simple modifications to image data to enhance the quality of the training dataset and the generalization of a machine-learning model. Generally, data augmentation is used in two ways: at training time (also known as supervised learning) and at inference time (also known as unsupervised learning). By utilizing data augmentation techniques, the model can learn more effectively from existing data, allowing it to make more accurate predictions on unseen data. By utilizing data augmentation, one can optimize the deep learning model's training process and overall performance.

This study used the data augmentation technique's "fill mode." In addition, we assigned it the "reflect" parameter, which fills empty values in reverse order and appears like a real and realistic image. The image dimensions were also checked to confirm that all images were the same size. All images were scaled to 224x224 pixels, then optimized and predicted using these downscaled images. Furthermore, this was necessary in order to ensure that each image had the same size, thus allowing the program to make accurate predictions while simultaneously preventing an unnecessary strain on the computational resources of the system, This optimization process was repeated for each image that went through the system, and it ensured a consistent result across all images, thus ensuring that the program could make consistent and accurate predictions. Scaling the images to the same size guaranteed a consistent output for each image. This optimization process improved the accuracy of the program's predictions while making the best use of the computational resources available to the program.

To conduct the augmentation technique on healthy and unhealthy images, we utilized the deep learning framework Keras. We rotated the images randomly between 0 and 45 degrees, shifted the width and height, and moved the image 20% sideways, up or down from its initial location. Finally, we raised the number of example images from 758 to 1231 such that each class is balanced or has an approximately similar number of images. Because balancing the data prevents the classifier from being biased toward a specific class, it is predicted that this balanced distribution would improve training accuracy and favorably influence detection outcomes. Figure 8.15 shows an example of an image created using the data augmentation method.

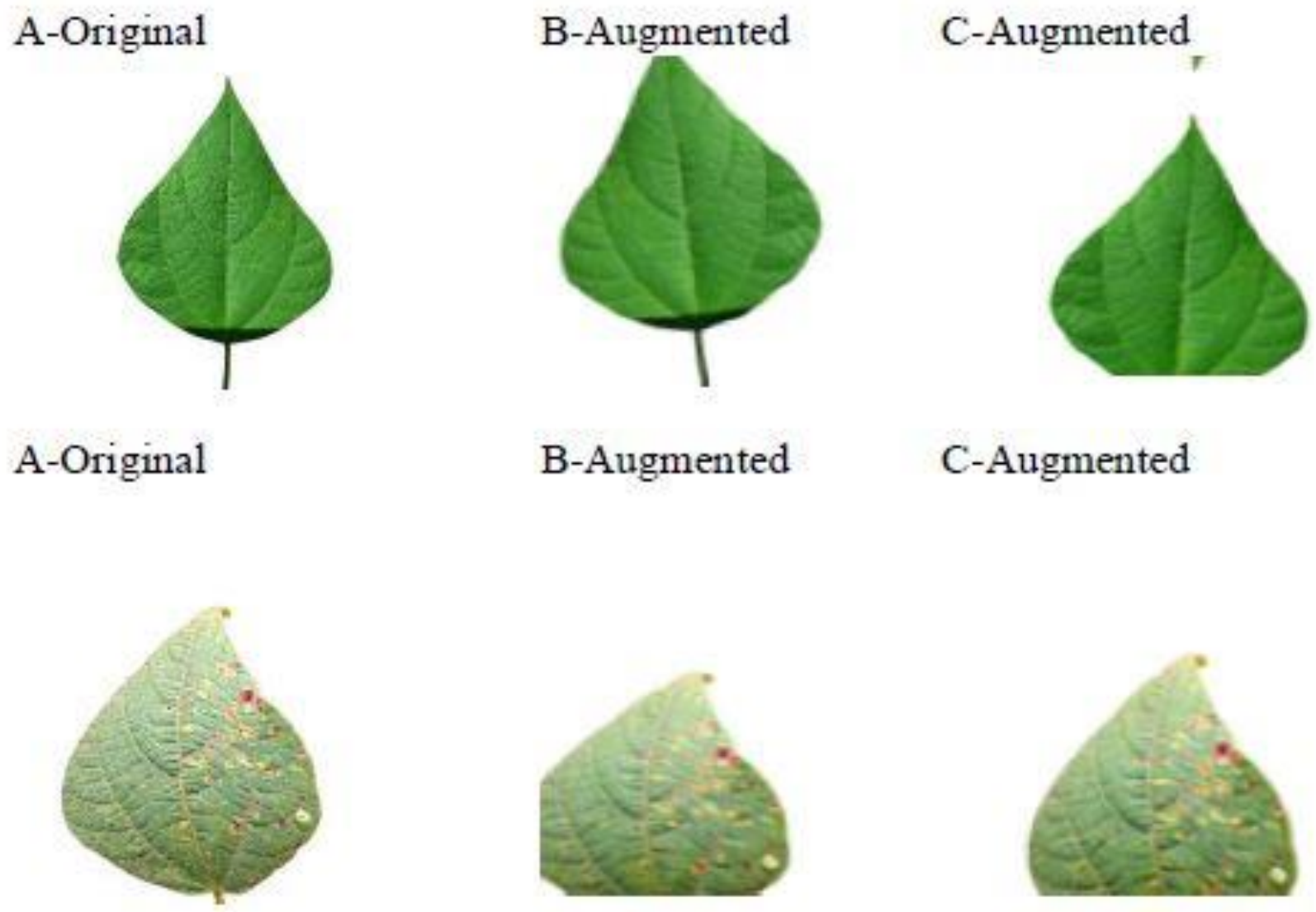

Figure 8.15 Example of data augmentation on two classes (healthy and Bean rust)

The utilization of data augmentation techniques aids in diversifying the training data and mitigating overfitting. The following table presents the data augmentation techniques implemented in the project and their associated parameters/ranges

Table 8.7 Data augmentation techniques used and their parameters

| Data Augmentation Technique | Parameter/Range |
| --- | --- |
| Horizontal flipping | Applied to the training samples |
| Rotation | Range of 0-45 degrees |
| Zooming | Range of 0.2 |
| Shifting | Range of 0.2 in both width and height |
| Fill Mode | Reflect parameter |

#### 8.2.1.3 Implementation and setup

In this study section, we employed a range of settings to assess the model's effectiveness. First, we implemented the model architecture using many thorough procedures. Next, to assess the network's performance, we will analyze the data and explore different approaches to partitioning it into three distinct sets: a test set, a validation set, and a training set. Additionally, we labeled the data and developed an input pipeline for model building (as shown in Figure 8.13).

Developing an accurate and reliable disease detection model requires the use of labeled data. Labeled data enables the model to distinguish between healthy and diseased leaves based on various features, and the accuracy of the model's output is determined by the quality and number of labeled data used in training. As depicted in Figure 8.13, labeled datasets are fundamental in developing deep learning models for disease detection. They aid in training the algorithm to identify specific patterns and features associated with the disease. It is essential to ensure that the dataset is well-labeled with accurate and consistent annotations.

Figure 8.13 provides a comprehensive example of labeled datasets for different classes of bean leaf disease. It comprises six subfigures from two different datasets, labeled with the name and index of the class it represents. These classes include healthy, rust disease, and angular leaf spot disease. The subfigures illustrate examples of each class, along with accompanying labels to recognize the type of disease in the image. For instance, the healthy class shows images of leaves that are free from any disease symptoms, while the rust disease class displays images of leaves with rust-colored spots caused by fungal infections. On the other hand, the angular leaf spot disease class contains images of leaves with angular lesions, caused by the bacterial pathogen.

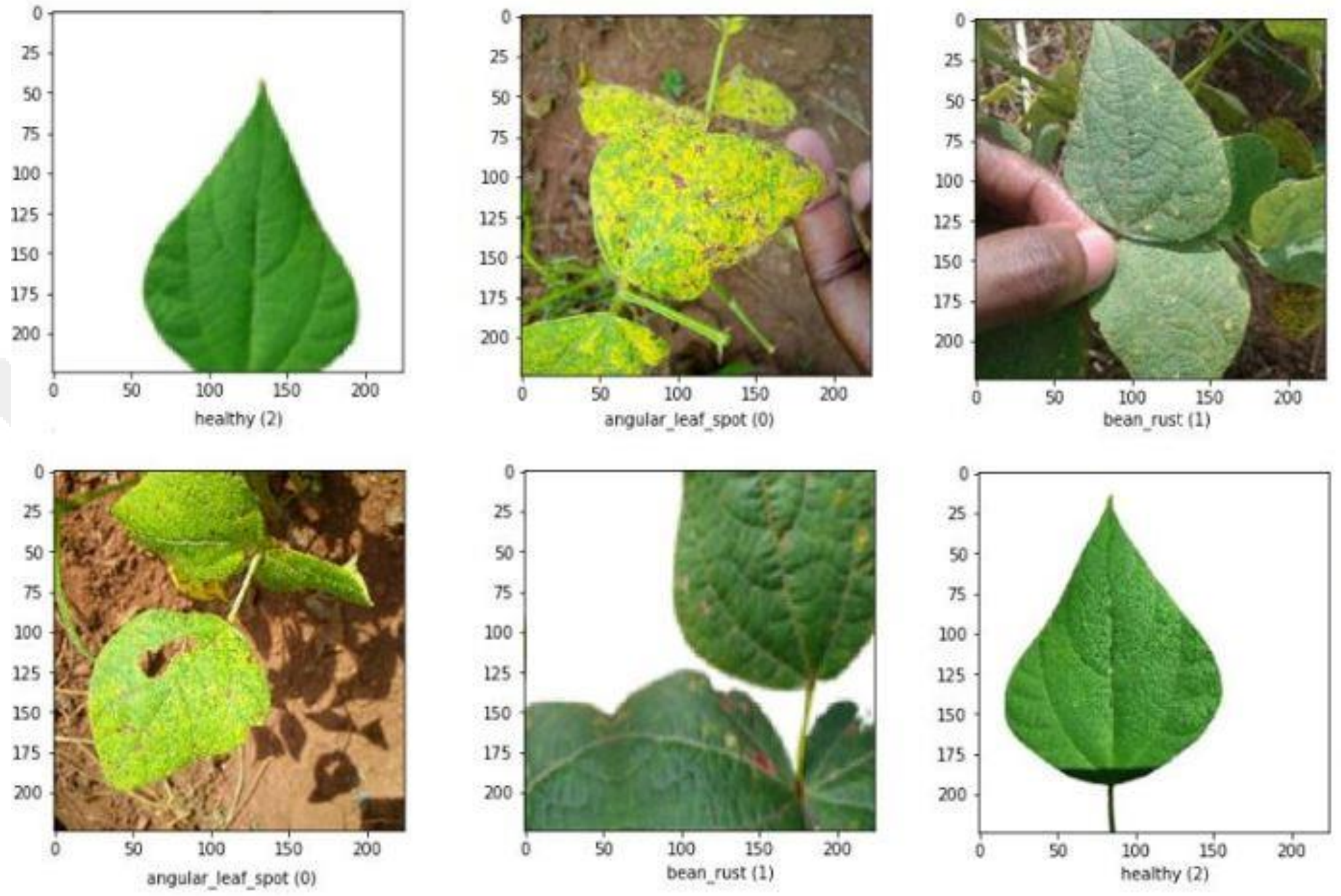

Figure 8.13 labeled datasets

Having these subfigures can aid researchers in quickly assessing the quality of the labeled dataset and verifying that each class is correctly identified. It also serves as a reference for future studies, enabling other researchers to compare their labeled datasets with the one used in this study. Ultimately, the presence of a properly annotated dataset plays a crucial role in advancing the accuracy and dependability of DL models for identifying and diagnosing bean leaf diseases.

Furthermore, the training set was utilized to fine-tune the model, while the validation set was employed to tune the hyperparameters. Finally, the test set was utilized to assess the model's performance and obtain an accurate measurement of its predictive ability. Additionally, for this research, we developed a CNN model consisting of eight layers. The model has three fully linked layers and five convolutional layers to identify bean leaf disease. After every convolution layer, we added max pooling layers and utilized two dense layers with 2 and 64 neurons as the final layers. Following those is the softmax layer, which aids in categorizing the input by providing the probability of every input relating to a given class. Following that, we ran a number of experiments to determine which parameters would be most effective for the three datasets being utilized in our specific task.

Various learning rate numbers, including 0.01, 0.001, and 0.0001, have been tested; 0.001 was determined to provide the best accuracy result. Similarly, various filters and epochs have been evaluated, and we discovered that utilizing ten filters and 100 epochs produced the best results. So, to train the present model, we employed 10 filters and the model was programmed to run for 100 epochs. However, training was designed to be stopped by checking the validation loss, i.e., if the validation loss increases for three successive epochs. In other words, when accuracy and loss stabilize, the model should come to an end. The model was configured to test each batch in training set throughout the training phase and assess performance on the validation set. The test set's performance is assessed using the final model (with final weights). The performance and outcomes of this work are displayed in the results section.

#### 8.2.1.4 Flow of the current work

In this study, three separate datasets from two different labs were used to train MobileNet models in TensorFlow using the Adam optimizer, and they were then evaluated. The schematic representation of this study is shown in Figure 8.14. The dataset was initially provided to the network as input (it might be public, collected, or combined). To prepare the data for pre-processing, the labels were one-hot encoded to represent the different classes, and image pixel normalization was performed, which enables quick calculation. The MobileNet model was then trained with the data sets, using Adam as the optimizer; after that, data augmentation was applied to reduce overfitting by enlarging the training set and augmenting the diversity and features of the training data. To compensate for the limited number of training examples in the collected dataset, we applied a data augmentation approach to enhance the dataset's diversity and increase the model's robustness.

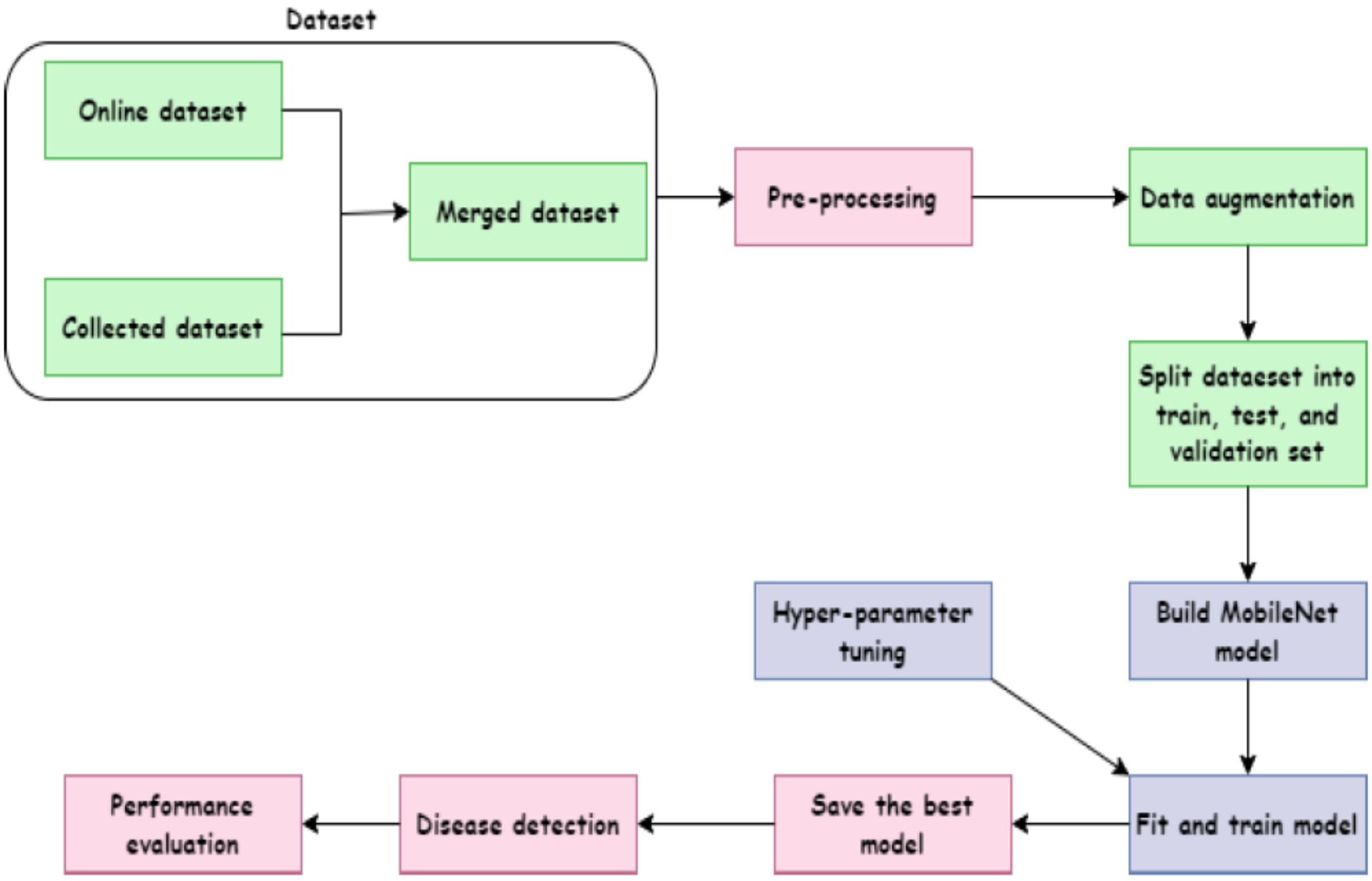

Figure 8.14 Schematic representation of the suggested approach for multi-class plant disease detetion

The input images were then partitioned into three separate sections using the 80-10-10 algorithm for validation, training, and testing. We used the trained and validation data to train the model after fine-tuning the parameters to produce a very accurate result. In order to prevent overfitting, the final layer additionally included a 30% dropout rate.

Using categorical entropy as the loss function, we built max-pooling, two dense layers, and further layers on top of the transfer learning architectures. Additionally, we developed two data generators (for both training and testing) that load the training data and transform it into training and test targets. We then preserved the top model from several runs after fitting the data and fine-tuning the hyperparameters; we used it to analyze our datasets to identify disease on bean leaves.

Finally, we assess the suggested MobileNet model's performance to determine how well the model can handle the datasets. Each dataset's findings are compared using a single model architecture to see how the data impacts the model's performance and the factors influencing each outcome. By evaluating the data, we can conclude how well the model can differentiate between different datasets and determine which factors lead to better performance. In this way, we can decide whether the MobileNet model is suitable for our particular needs. The results of our assessment help us understand the various strengths and weaknesses of using the MobileNet model for our dataset, as well as which datasets it is more suited to, in order to make the best decision possible. This assessment process helps us determine whether the MobileNet model is suitable for our particular needs and understand how our dataset affects its performance and which factors influence that performance, allowing us to make informed decisions as to how we can best use the model for our purposes.

#### 8.2.1.5 Proposed model architecture

Our customized model architecture, based on the MobileNet framework, has been tailored to identify bean leaf diseases across three diverse datasets (see Figure 8.15). When fed with a 224x224 RGB image as input, the model utilizes five convolutional layers, each consisting of ten 3x3 filters and employing the ReLU activation function. Max-pooling layers of size 2x2 follow these convolutional layers. Finally, the output from the last convolutional layer is flattened to transform the output tensor into a 1D array.

Our proposed model includes two dense layers with 2 and 64 neurons, respectively; these are followed by three fully connected layers, each consisting of 1024, 512, and 2 units. Additionally, there are three more fully connected layers consisting of 1024, 512, and 2 units, respectively. The final layer of the model employs a softmax activation function, enabling it to provide the probability predictions for the input image, classifying it as either healthy or diseased.

We train the model on three distinct datasets, which are preprocessed to ensure that the images are of uniform size and quality. The MobileNet architecture is selected due to its efficiency and ability to handle images with limited computational resources. Therefore, our proposed schema offers an efficient and effective method for detecting bean leaf disease across multiple datasets using the MobileNet architecture.

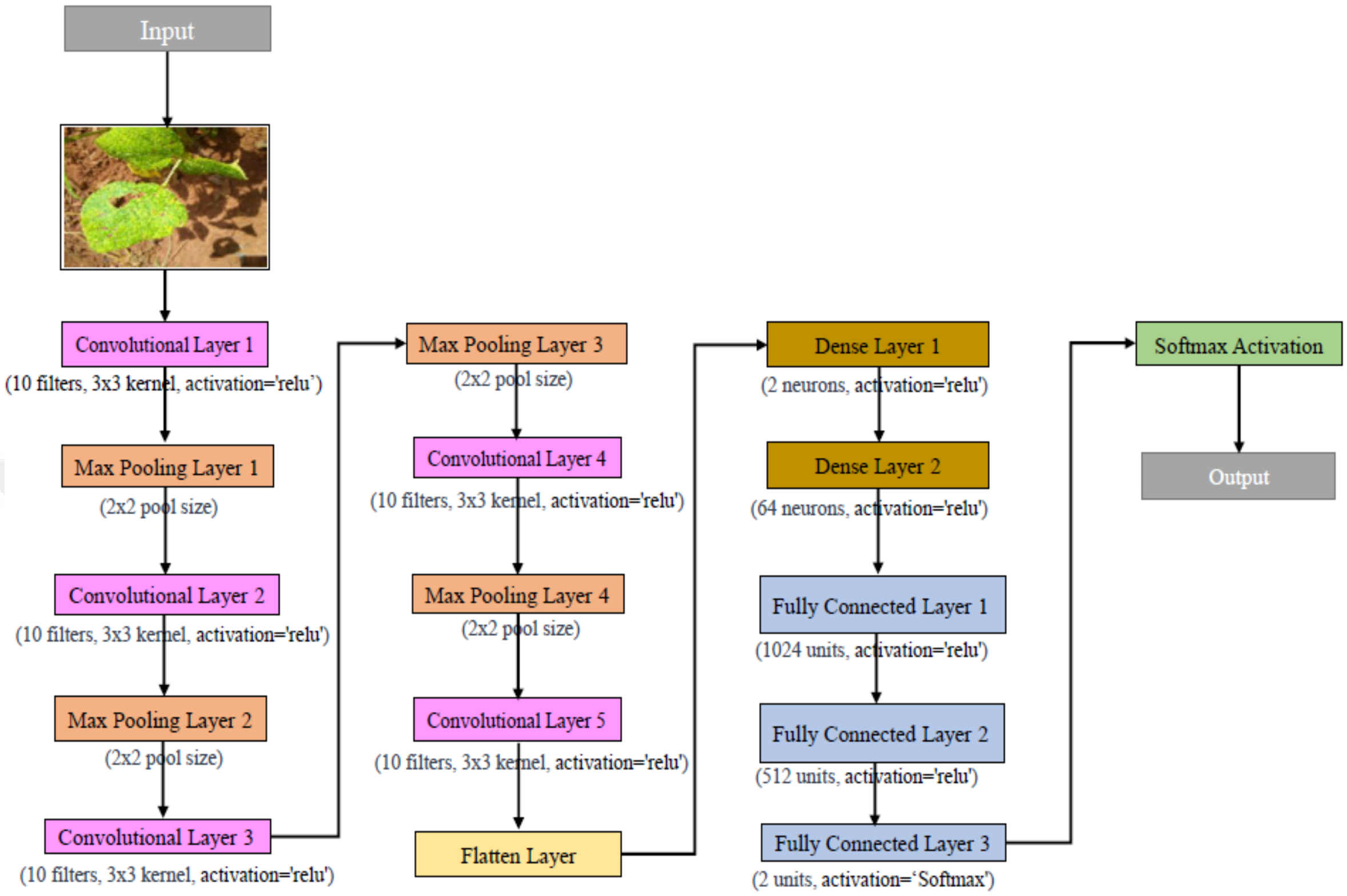

Figure 8.15 MobileNet-based architecture for beans leaf disease detection using Multiple datasets

Our proposed architecture's lightweight and efficient design makes it highly suitable for deployment on mobile devices. As such, our model architecture provides an efficient and accurate solution for detecting bean leaf disease. The use of three different datasets in training the model ensures that it is robust and generalizable to new data.

### 8.2.2 Result and discussion

In a farming environment, accurate detection of crop diseases is critical for disease prevention. As a result, running many experiments to examine the influence of datasets on a single model is essential in and of itself since it leads to numerous advantages. Furthermore, establishing an automated approach for detecting plant leaf diseases on farms is critical. But, most importantly, can a single model identify diseases in the real world? And, if so, will the model generalize across different datasets? As a result, in this part, we conducted various tests to illustrate the efficacy of the suggested technique on all three datasets (collected, public, and merged).

The dataset were divided into a training set comprising 80% of the data and a testing set comprising the remaining 20%. Initially, we ran a series of trials to determine the best settings for training. Following that, we keep those parameters constant to evaluate the datasets' effect on the single MobileNet model. The test images were randomly chosen from the dataset to eliminate selection bias. As a result, images from the test set include images of various resolutions.

All of the code was developed in Python, and experiments were conducted on Google Colaboratory (a server with a GPU library and 8 GB) running Windows 10. The MobileNet API was combined with an open-source TensorFlow library (Esmaeel et al., 2018). Subsections 8.2.2.1 and 8.2.2.1 provide a complete description of the outcomes obtained.

#### 8.2.2.1 Application on MobileNet model architecture

In this section, we used MobileNet to run a disease identification test experiment on the three different datasets. We used Softmax to modify the model output layer to output three categories and adjusted the hyperparameter values before training the model. The model was trained for 100 epochs using a fixed learning rate of 0.001. The learning rate defined as a crucial hyperparameter in the network, determining the extent to which a model's weight should be changed in response to an estimated error during the training process. An optimizer, such as Adam, was utilized to underrate the cost of the model, The batch size was set to 32, and a binary cross-entropy loss function was used to measure the accuracy of the model.  To introduce non-linearity and enable the model to learn complex data relationships, the ReLU activation function was employed. The learning rate also governs the speed at which the model converges to a global minimum.

Furthermore, we selected 32 as the model's batch size, indicating that 32 images from the training set would be utilized to assess the error gradient before changing the model weights.

The batch size refers to the number of samples processed by the network in a single iteration during the training process. In other words, a required hyperparameter specifies the number of instances to run before modifying the internal model parameters. In addition, due to the data in the bean leaf images could be noisy, we employed the Adam optimizer to manage sparse gradients in noisy situations and ensure better convergence in the training process. The Adam optimizer requires less memory in comparison to other optimization methods, so it was well-suited for this task.

To achieve the highest accuracy level, we assessed our model's performance on three datasets by analyzing its training and validation accuracy, as well as the loss. By carefully evaluating these metrics, we aimed to optimize the accuracy of our model and ensure its reliability across different datasets. This analysis helped us to evaluate the effectiveness of our model and make informed decisions on model improvements. Figure 8.16 depicts a representation of the model's accuracy and loss curves at the end of each epoch for three datasets, and Table 8.8 present the same performance results of a single model based on accuracy and loss across three datasets.

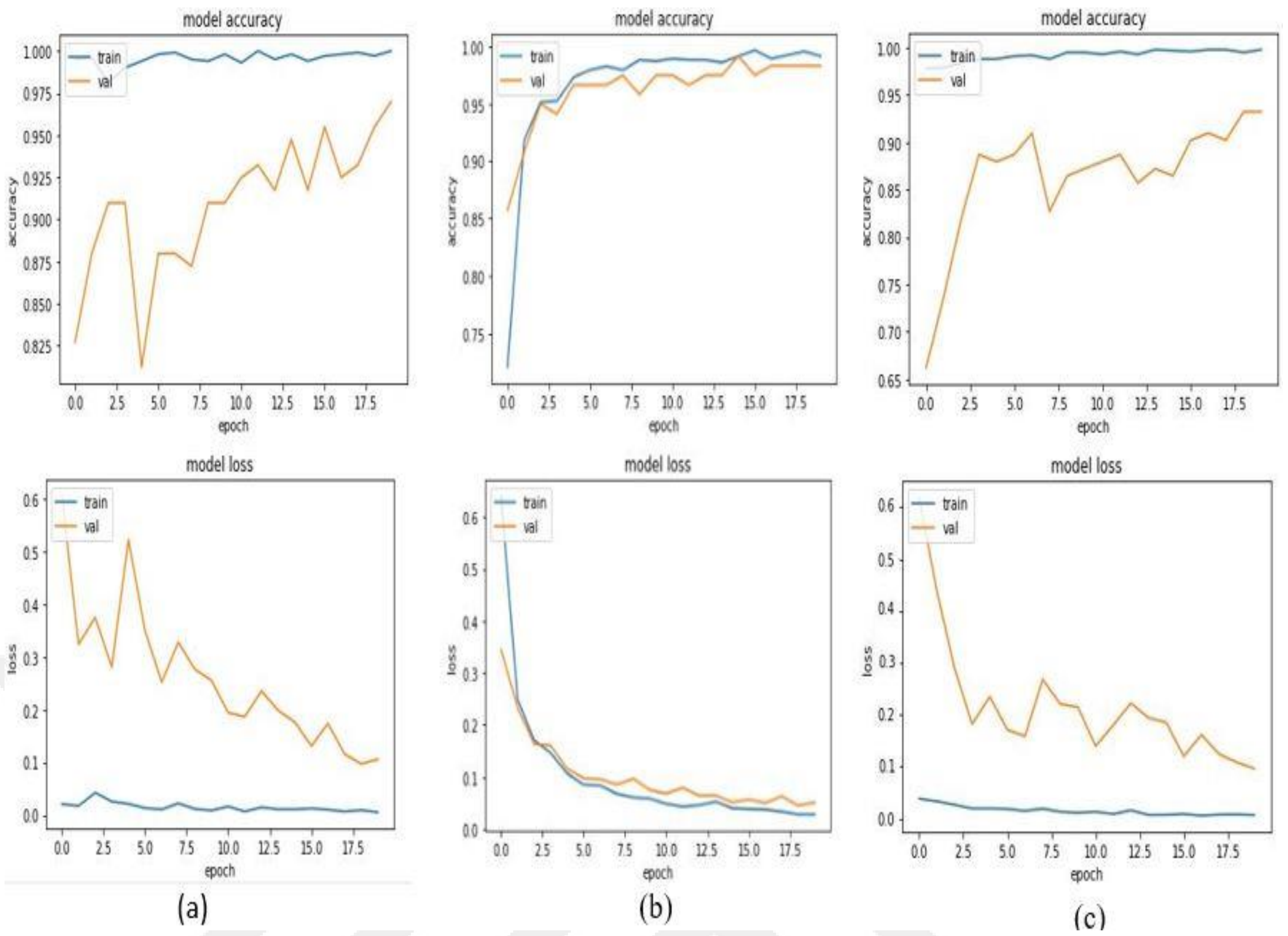

Figure 8.16 Analysis of accuracy and loss curve of a single model utilizing a (a) public, (b) collected, and (c) merged dataset)

Table 8.8 Performance results from the three datasets using a single model

| Dataset | Accuaracy | Val-acc | Loss | Val-loss |
|---|---|---|---|---|
| **Public dataset** | 1.0000 | 0.9699 | 0.0051 | 0.1055 |
| **Collected dataset** | 0.9914 | 0.9832 | 0.0286 | 0.0507 |
| **Merged dataset** | 0.9980 | 0.9323 | 0.0124 | 0.1919 |

The results reveal that the public dataset has a training accuracy of up to 100%, but the collected and merged dataset has a convergent accuracy of about 99%. These results indicate that the model can converge with reasonable accuracy (at least 99%) for all three datasets, suggesting that the model can be trained with sufficient accuracy for bean leaf disease detection, regardless of the dataset's size and diversity. Furthermore, the merged dataset also achieved an excellent training accuracy of 99.80%, indicating that combining different datasets into a single training set can achieve the best model performance.

In empirical experiments, we discovered that training on two similar datasets produces comparable outcomes. For example, in the current work, the samples from the various datasets have fairly comparable distributions and a similar appearance. Consequently, based on the model's accuracy on the public dataset, we estimate the model will perform effectively on the collected and merged datasets before analyzing the model on the data. Because a model may work effectively on two visually comparable datasets, training it on the merged dataset with the most data may increase model performance. Therefore, this suggests that combining two datasets, in which the distributions and appearance are similar, into one may result in better model performance, allowing us to make better use of the data we have and achieve higher accuracy than training on one dataset alone. With the right conditions, such as two datasets with similar distributions and appearance, merging these two datasets can yield better performance than training on one dataset alone. Additionally, when combining datasets, it is critical to consider the bias and imbalance of the two datasets; if the datasets have different levels of bias and imbalance, it is important to adjust them accordingly before merging so that the merged dataset does not reflect those differences.

The public dataset displays the maximum resolution since it can cover the whole data vector space. Furthermore, the model is well implemented because it was developed using a public dataset, which was the first dataset with access to full feature space. The model performed well on all the datasets after being trained on both sets, and the ideal model which is more robust and reliable for each dataset was produced after training on all the sets until they were all used up. Moreover, this is due to the model's ability to effectively learn from each of the datasets and use all the features to produce better results for each of the datasets.

The model is remarkable in its ability to learn from and adapt to the different datasets, despite each dataset providing its own unique set of features and characteristics, allowing the model to continually learn and improve to produce better results for each dataset. Furthermore, this is an impressive feat and serves as a testament to the model's effectiveness in using all available data for maximum performance, even when that data is from varying sources and may be more challenging to learn from than other, more homogeneous datasets.

As anticipated, the model obtained high accuracy in all datasets after training for the complete number of epochs (see Figure 8.17); specifically, the model reached 100%, 99.14%, and 99.80% accuracy for the public, collected, and merged datasets, respectively.

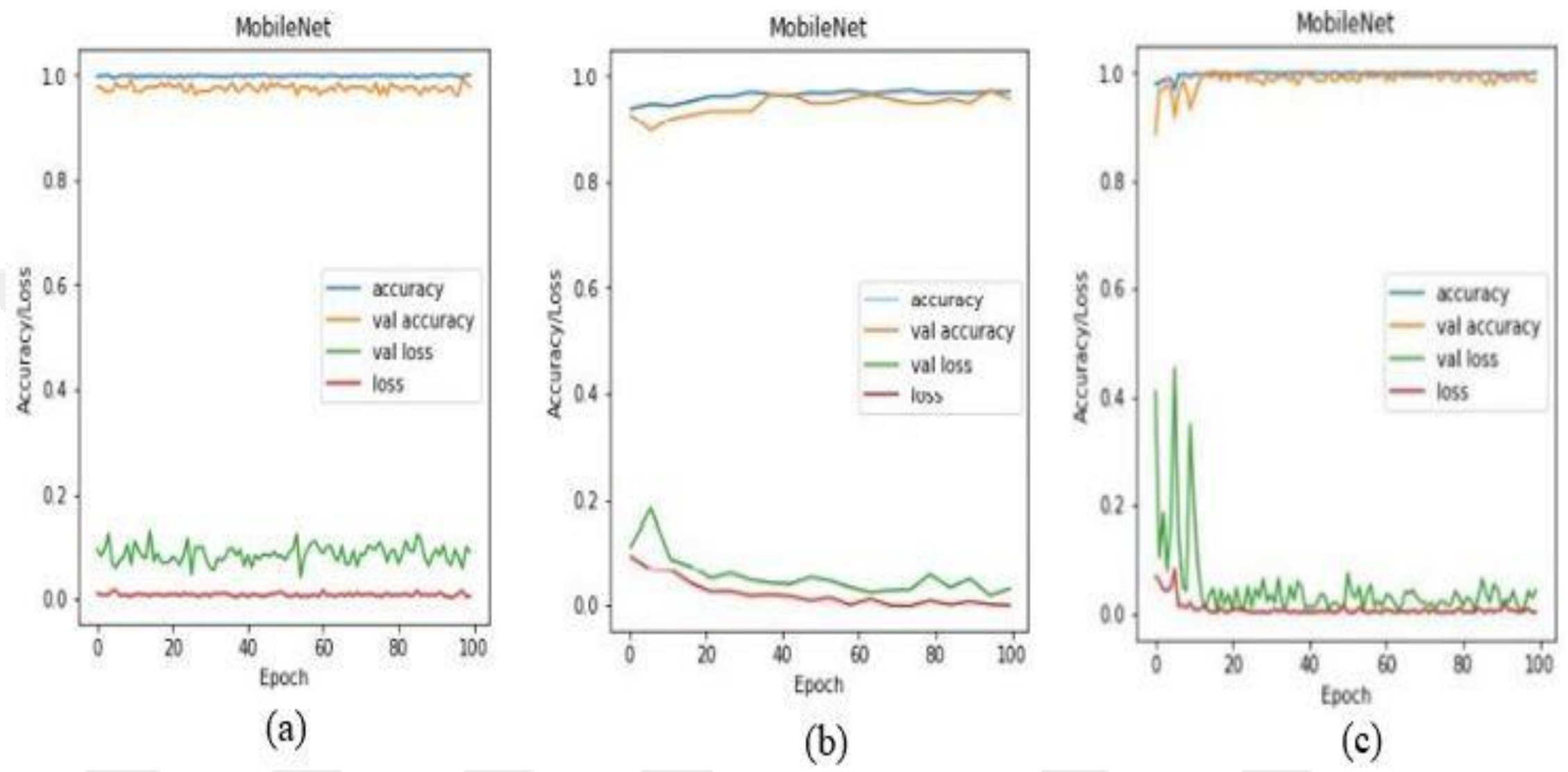

Figure 8.17 Comparison of the suggested method's accuracy and loss performance on the (a) public, (b) merged, and (c) collected datasets

The detection performance for three classes (healthy bean leaf, angular leaf spot, and bean rust) is compared and evaluated using the accuracy-based performance measure, which can be observed from the accuracy/loss curve in Figure 8.16. Moreover, the training accuracy/loss curve shows the model's development, which shows how well the model identified bean leaf disease on the three datasets. The result of the accuracy-based performance measure was promising in all datasets, which demonstrates the effectiveness of the model in detetcting the bean leaf disease

#### 8.2.2.2 Disease detection performance evaluation

Using three datasets, each of which had three classes (healthy leaf, angular leaf spot, and bean rust), we conducted comparative research to evaluate the effectiveness of bean leaf disease identification using a single model (i.e., MobileNet). In order to identify the areas where the model is prone to miscalculation when generating predictions, the performance of the model is evaluated and compared across each dataset. The objective is to comprehend how changing the dataset could affect the efficiency of the MobileNet model. We used the same hyperparameters (as previously described) to analyze the dataset and ran the algorithm across 100 epochs to verify this hypothesis.

An overview of the performance attained for the detection model is displayed using a confusion matrix, the purpose is to assess how well the model performs for each disease class. Due to the non-deterministic nature of the MobileNet model, we conducted the experiments over various runs to ensure reliable and consistent results. However, The best-performing model's findings were the only ones we provided. Figure 8.18 depicts the confusion matrix derived for the best-performing network.

Further tests with a single model are needed to demonstrate the efficiency of the suggested technique on the three bean leaf datasets. To do this, we must assess the model's performance by determining the mean values of several quantitative metrics including precision, F1-score, recall, and accuracy. These performance metrics were chosen since they are the most often utilized metrics in prior research to analyze the performance of most approaches (Szegedy et al., 2015). Furthermore, this performance assessment will help us better understand how the models perform on bean leaf datasets when compared to one another, allowing us to assess the effectiveness of our suggested technique accurately.

As a result, the performance evaluation equations presented in Eqs. 1, 2, 3, and 4 are utilized to generate performance metrics and evaluate outcomes.

1. **Accuracy**: The accuracy metric determines the proportion of correctly predicted classes among all the samples analyzed.

$$Accuracy = \frac{TP+TN}{TP+TN+FP+FN} \tag{1}$$

2. **Precision**: Precision is used to calculate the number of positive patterns that each predicted pattern in a all positive class correctly predicts.

$$Precision = \frac{TP}{TP + FP} \tag{2}$$

3. **Sensitivity (Recall):** The percentage of correctly detected positive patterns is calculated as sensitivity or recall.

$$Sensitivity(Recall) = \frac{TP}{TP + FN} \tag{3}$$

4. **F1-Score**: The F1-score is used to calculate the harmonic average of the recall and precision rates.

$$F1 - Score = \frac{Precision \times Recall}{Precision + Recall} \tag{4}$$

Where

- **TP** = True Positive
- **FN** = False Negative
- **TN** = True Negative
- **FP** = False Positive

5. **Confusion matrices:** The confusion matrix provides valuable information about the performance of a classification model by presenting the number of discrepancies between predicted and actual values. It consists of four primary categories, namely True Positives (TP), False Positives (FP), True Negatives (TN), and False Negatives (FN). This matrix, often referred to as an error matrix, displays all possible combinations of predicted and actual values. By examining the confusion matrix, we can evaluate the accuracy of the model's predictions and compare them to the ground truth values, enabling us to assess the effectiveness of the classification model.

It is a valuable tool in machine learning and can help identify areas where the model may need improvement. In addition to displaying the number of differences, confusion matrices provide insight into a model's accuracy and error rates. Their usefulness extends beyond binary classification, as they can be adapted for multi-label classification tasks. Furthermore, the values present in the confusion matrix play a crucial role in computing several essential classification metrics, such as accuracy, recall, precision, and F1-score. These metrics are fundamental in evaluating the performance of a machine learning model and determining its effectiveness in classification tasks. By analyzing these metrics, we can gain insights into how well the model is performing and make informed decisions regarding its suitability for specific applications.

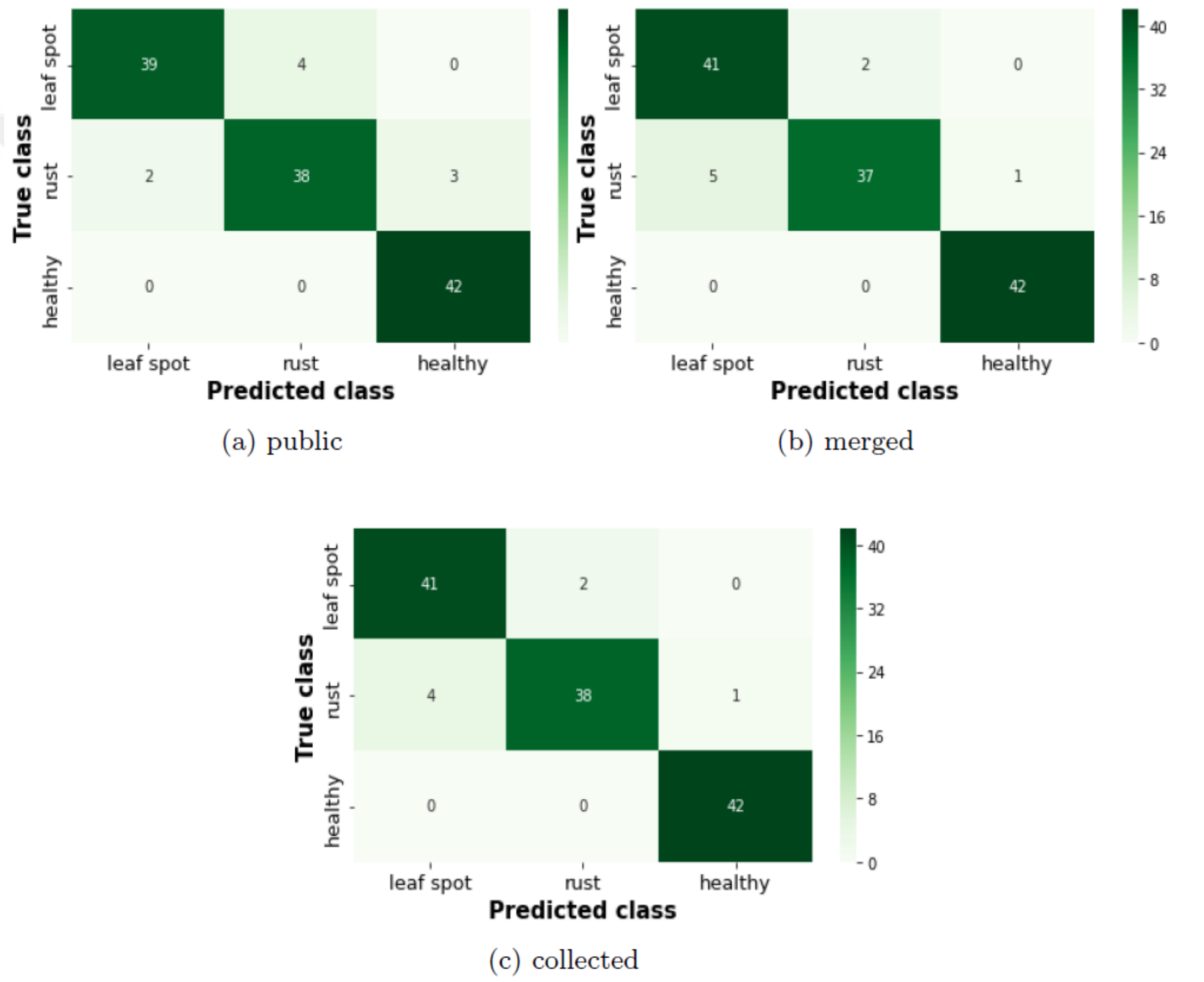

Figure 8.18 Confusion matrix of the suggested method for the a) public, b) merged, and c) collected datasets

To gauge how well our approach works, we utilized a test set comprising 128 images from the datasets for healthy leaves, angular leaf spots, and bean rust disease, resulting in a total of 384 images across the three different datasets. This method was used to assess the algorithm's performance.

First, we evaluated the approach by comparing the prediction results to the corresponding labels based on ground truth and measuring performance employing recall, precision, and F1-score metrics, considering both true positives and false negatives. We then performed an accuracy test by computing the images correctly classified by the algorithm, which was done by summing up all true positives and then dividing it by the overall number of instances in the test set. Finally, we analyzed the performance of the algorithm for each dataset separately.

The suggested technique achieved 92.97%, 93.75%, and 94.54% accuracy for the public, merged, and collected datasets, respectively. This outcomes demonstrate the proposed approach's high accuracy and precision in detecting and classifying bean leaf disease in the evaluated datasets. Figure 8.19 shows the classification report, which includes the average accuracy, F1 score, precision, and recall.

| | precision | recall | f1-score | support |
|---|---|---|---|---|
| 0 | 0.9512 | 0.9070 | 0.9286 | 43 |
| 1 | 0.9048 | 0.8837 | 0.8941 | 43 |
| 2 | 0.9333 | 1.0000 | 0.9655 | 42 |
| accuracy | | | 0.9297 | 128 |
| macro avg | 0.9298 | 0.9302 | 0.9294 | 128 |
| weighted avg | 0.9297 | 0.9297 | 0.9291 | 128 |

(a)

| | precision | recall | f1-score | support |
|---|---|---|---|---|
| 0 | 0.8913 | 0.9535 | 0.9213 | 43 |
| 1 | 0.9487 | 0.8605 | 0.9024 | 43 |
| 2 | 0.9767 | 1.0000 | 0.9882 | 42 |
| accuracy | | | 0.9375 | 128 |
| macro avg | 0.9389 | 0.9380 | 0.9373 | 128 |
| weighted avg | 0.9386 | 0.9375 | 0.9369 | 128 |

(b)

| | precision | recall | f1-score | support |
|---|---|---|---|---|
| 0 | 0.9111 | 0.9535 | 0.9318 | 43 |
| 1 | 0.9500 | 0.8837 | 0.9157 | 43 |
| 2 | 0.9767 | 1.0000 | 0.9882 | 42 |
| accuracy | | | 0.9453 | 128 |
| macro avg | 0.9460 | 0.9457 | 0.9452 | 128 |
| weighted avg | 0.9457 | 0.9453 | 0.9449 | 128 |

(c)

Figure 8.19 Performance metrics comparison of the proposed technique on (a) public, (b) merged, and (c) collected dataset

Overall, based on the model's accuracy on all three different datasets, it is reasonable to conclude that the model is performing well regardless of the dataset on which it is evaluated (i.e., the public, collected, or merged dataset), as its accuracy on each of these datasets is greater than 92%. Furthermore, the model can generalize its learning. It performs consistently well on each dataset, suggesting it can accurately make predictions and classifications across various datasets, regardless of its training environment. Therefore, our proposed approach performs excellently, with higher accuracy scores in each dataset.

#### 8.2.2.3 GradCAM technique for improved visualization of leaf disease

Convolutional Neural Networks (CNNs) face the challenge of interpreting their results. Although CNNs can precisely classify images, the specific features that the model utilizes to make predictions are often unclear. To address this issue, the Gradient CAM (Class Activation Mapping) technique is employed to visualize the features that a CNN uses for its predictions. This technique produces a heat map that identifies the significant regions within an image that contribute to the classification decision. By back-propagating the gradients of the output class with regard to the feature maps of the final convolutional layer of the CNN, the CAM heat map is created. In the generated heat map, red-shifted regions are considered important, while blue areas are deemed to have no classification significance. Hence, our study utilized the GradCAM technique to visualize the features that a MobileNet CNN model uses to classify images of bean leaves into one of three categories: Bean rust, angular leaf spot, and healthy. We selected these classes because they represent common diseases that affect bean crops, and accurately detecting them can help prevent significant yield losses. Figure 8.19 displays the CAM results of the models used in our study.

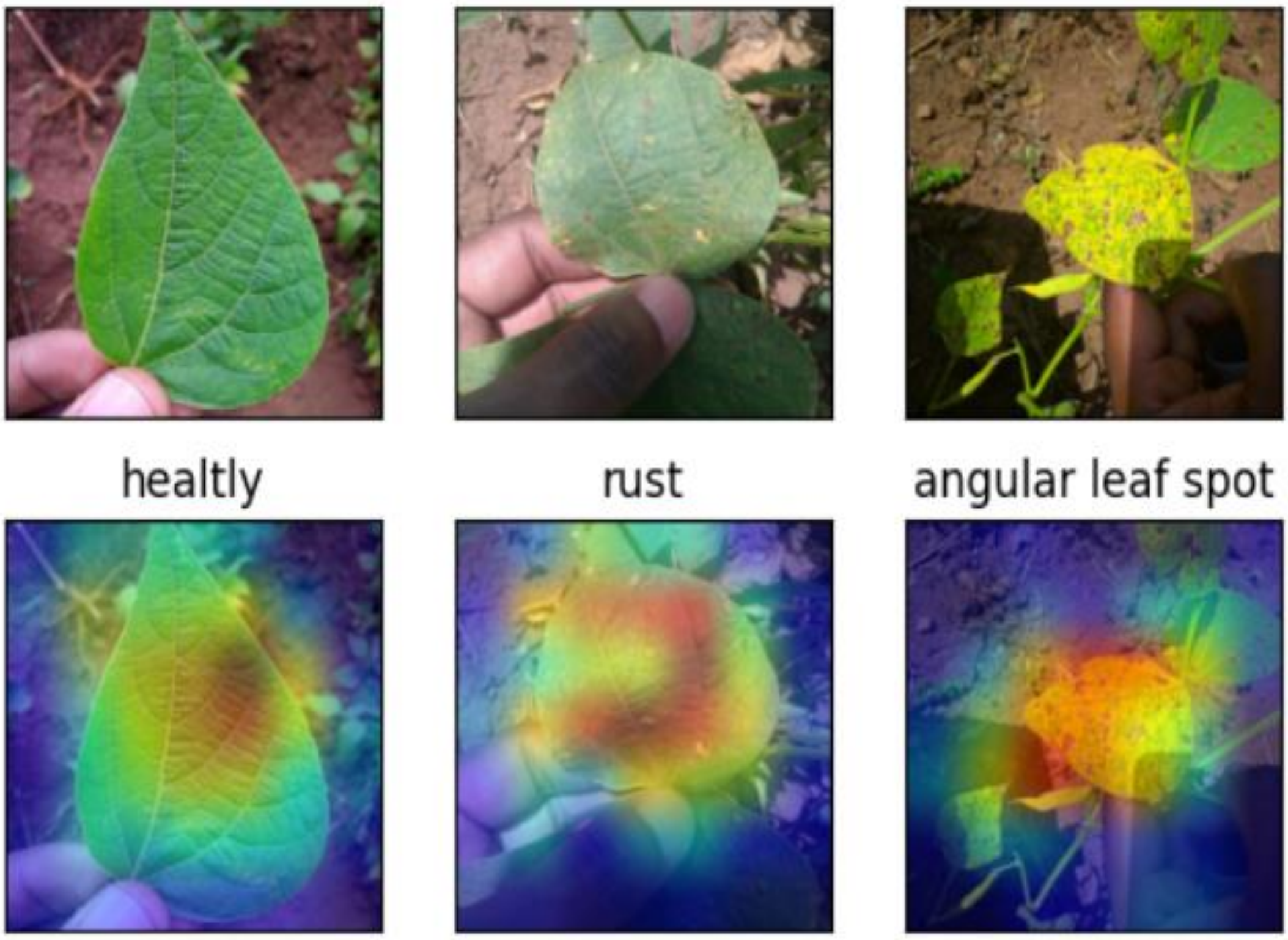

Figure 8.19 GradCAM real-time results

To enhance our comprehension of the model's decision-making process, we applied the GradCAM technique to the model's predictions. The resultant heat maps pinpointed the leaf image regions that were most crucial for the classification decision. Through an examination of these heat maps, we gained insight into the visual features that the model utilizes to classify bean leaves into their respective categories. Our findings showed that the model focused on distinct parts of the leaves for each class, indicating that the model can differentiate between the various diseases based on specific visual features.

Our study effectively showcases the GradCAM technique's effectiveness in visualizing the features employed by a MobileNet CNN model to classify bean leaf images into different categories. This enhanced comprehension of the model's decision-making mechanism can lead to improved performance and identify potential areas for further research. Additionally, our study provides valuable insights into the visual features that are essential for accurately detecting bean leaf diseases. These insights can help develop more effective detection and prevention strategies, making our findings relevant and applicable to the agricultural sector.

#### 8.2.2.4 Comparison with the State-of-the-Art methods

This section aims to assess the efficacy of the proposed model for detecting bean leaf diseases and compare it with existing cutting-edge approaches. To accomplish this, we conducted training and evaluation using three distinct datasets: a public dataset, a merged dataset, and a collected dataset. The classification report results revealed that our suggested model exhibited high accuracy, recall, precision, and F1-score across all three datasets, affirming its efficacy in detecting bean leaf diseases.

To further verify the effectiveness of our model, we compared it with existing cutting-edge approaches for the same task, based on accuracy results. Additionally, we performed a two-sample (unpaired) t-test and found that there is an important difference between our model's performance and the cutting-edge approaches (all $p < .001$) on all three datasets, with the public dataset achieving the highest accuracy and the merged dataset achieving the lowest accuracy.

Nonetheless, our model demonstrated excellent performance on each of the datasets, achieving over 92% accuracy. In Table 8.9, we use "+" and "-" signs to indicate statistically significant and insignificant differences, respectively, between our proposed model and the cutting-edge methods.

Table 8.9 A comparison between the suggested MobileNet model and the state-of-the-art techniques. Take note that column ST indicates a statistically significant test

| Ours | State-of-the-art models | ST |
|---|---|---|
| 92.97% | GoogleNet 65.08% (Szegedy et al., 2015) | + |
| | InceptionV3 68.80% (He et al., 2016) | + |
| | ResNet-18 71.32% (Zhang et al., 2018) | + |
| | ShuffleNet 63.37% (Abed et al., 2018) | + |
| | Pretrained-GoogleNet 91.36% (Szegedy et al., 2015) | + |
| | Pretrained-InceptionV3 93.37% (He et al., 2016) | - |
| | Pretrained-ResNet-18 94.69% (Zhang et al., 2018) | - |
| | Pretrained-ShuffleNet 91.82% (Abed et al., 2018) | + |

Moreover, comparing the results with state-of-the-art methods (refer to Table 8.9) demonstrated that the proposed approach exhibited significant superiority over the majority of cutting-edge approaches ($p < .001$) at $\alpha = 0.05$. This further emphasizes the reliability and accuracy of our suggested approach compared to cutting-edge methods, underscoring its effectiveness and potential applicability in various tasks.

In summary, in this study, we introduced a new approach for automating the detection of leaf diseases on bean crops by utilizing the MobileNet architecture and three diverse datasets. Our study produces several key contributions to the field of plant disease detection. Firstly, our proposed method achieved high accuracy (>92%) on all three datasets, demonstrating its robustness and effectiveness in detecting bean leaf diseases. Secondly, our method showed strong generalization capabilities across datasets, with comparable accuracy results. Thirdly, our approach exhibited substantial superiority over multiple cutting-edge DL methods for bean leaf disease detection on all three datasets. Lastly, we demonstrated the effectiveness of the GradCAM technique in visualizing the most salient visual features for accurate disease detection. These insights can guide the development of more effective detection and prevention strategies, promoting sustainable agricultural practices. Therefore, our study provides a promising direction for automating plant disease detection and encourages further research in this field.

In the next section, we will introduce another detailed study describing our new CNN model for plant disease detection, which is accurate and efficient, providing better accuracy and scalability than existing models.

## 8.3 A Novel Convolution Neural Network-Based Approach for Seeds Image Classification

Image analysis of seeds has become an essential tool for biodiversity protection. With the growing concern for conserving plant genetic resources, seed banks are established worldwide to collect and preserve seeds from different plant species. Image analysis of seeds is used to identify and classify seeds accurately, which is critical for preserving and managing seed collections.

Seed images can be analyzed for various properties such as size, shape, texture, and color, which can provide valuable information about the genetic diversity of plant species. In addition, the analysis of seed images can help identify and characterize different seed varieties, assess the quality of seeds, and identify seed traits that are important for crop improvement. Moreover, seed image analysis can be used to track the movement of seeds and prevent the illegal trade of endangered plant species. The use of image analysis for seed identification and classification can also aid in the development of new technologies for seed processing and seed quality control. As a result, it is now challenging to identify and categorize all plant species worldwide. For example, classifying and recognizing seed images is challenging for several reasons, including:

1. **Variation in seed shape and size:** Seeds come in various shapes and sizes, even within the same species, making it difficult to classify them accurately. This variability can make identifying and differentiating seeds from different species hard.
2. **Seed coat complexity**: Seed coats can be complex, with various surface textures, colors, and patterns, making it challenging to capture all the relevant information in a single image. Additionally, the seed coat may have cracks, which can further complicate the analysis.
3. **Seed orientation**: Seed orientation can vary, making it challenging to capture a standardized image. Even slight variations in the angle at which the seed is captured can result in significant changes in the features of interest.
4. **Limited image data**: Obtaining large datasets of high-quality seed images can be challenging, especially for rare or endangered plant species. More data is needed to train deep learning algorithms accurately.
5. **Intra- and inter-species variability:** Seeds within a species may vary significantly, and some species may have similar-looking seeds, making it difficult to distinguish between them accurately.
6. **Computational challenges**: Processing large numbers of images and extracting relevant features can be computationally expensive and time-consuming, requiring specialized hardware and software.

Overall, these challenges highlight the need for a comprehensive approach to seed image analysis, combining expertise in plant biology, computer vision, and deep learning. Therefore, developing new models and algorithms will assist us in better comprehending fundamental seed traits and the interrelationships that regulate these traits. It will also increase our capacity to generate more precise and effective classifications in other areas of image classification and reduce the difficulties those areas face.

This part of thesis focuses on creating and proposing a novel CNN architecture for seed image classification, with Brassica seeds as the primary use case. By adjusting the number and characteristics of image layers, we could also assess the effectiveness of our CNN model and extract features. The objective was to determine the model's architecture and the ideal training conditions for these issues, which we subsequently used to solve the classification challenges for Brassica seed image problems. Finally, using various metrics, we assessed the impacts of the suggested designs and training settings on raising performance.

The proposed approach has significant implications for seed quality control and agricultural practices. It provides an automated and efficient method for seed classification, which can save time and reduce errors associated with the manual classification. Moreover, the proposed approach exhibits the potential to be extended to other plant species, thereby making valuable contributions towards enhancing the agricultural industry as a whole.

As a result, this study details the whole research process of developing and applying a novel CNN model for image classification in higher dimensional spaces, specifically to categorize Brassica seed images into 10 groups. Also, a new Brassica dataset that did not already exist is created as part of this work to test our CNN architectures' effectiveness on it. Digital microscopy captured images of the seeds at 1600 x 1200 pixels at 96 dpi in natural light. Many images of each kind were chosen randomly to form a dataset. Furthermore, to ensure that the data was consistent, images of the seeds were taken under similar lighting and capture conditions, and all images were visually inspected and cleaned of noise. In addition, the seeds were classified into their respective species based on their physical characteristics and the presence of distinguishing features such as the shape of the seed or stem structure, color and texture of the seed coat, and any unique marks or indentations, allowing for precise identification of each seed sample.

This work contributes through the following tasks:

1. Using advanced deep-learning techniques, we introduced a novel CNN model for identifying and classifying brassica seeds. The suggested technique is developed to support farmers in resolving their issues.
2. Present the whole process research of developing and implementing a novel CNN model for image classification in higher dimensional spaces.
3. We fine-tune our model by changing parameters such as the learning rate, to verify that our model performs comparably well on the dataset and does not miss the optimal solution.
4. Evaluate the suggested method's performance using several measures and compare it to pre-trained existing deep learning approaches. This is a critical insight for soft applications, for instance, when the model is applied in an agricultural environment.
5. Create a new seeds dataset, and use it to evaluate the method's generalizability.

The created model was tested using a collected Brassica seed dataset and several assessment metrics. This method assisted us in identifying and implementing the best-performing architectures and training parameters for predicting class labels. Furthermore, it enables us to adjust the training model and identify the optimal combination of architectural parameters, network topology, and weight values for predicting class labels. Therefore, this approach also has the potential to be extended to other problems with similar characteristics, as it can be implemented to other complex deep-learning problems, including image classification and segmentation. Moreover, the proposed approach can handle complex data by providing a comprehensive solution to identify the optimal model architecture, allowing us to achieve improved performance.

#### 8.3.1 Research materials and methods

This section included a detailed explanation of the suggested model and the datasets utilized. The section also discusses the steps taken to increase the effectiveness of the suggested method implementation, and it ends with a description of the suggested CNN model architecture.

##### 8.3.1.1 Dataset and training process

This section discussed how the dataset for this study was generated and the training method. Furthermore, 10 different types of Brassica seeds were selected to construct the dataset for this study. Brassica Nigra, Brassica Napus Var Annua, Brassica Napus Var Oleifera, Brassica Oleracea Gongylodes, Brassica Oleracea Rapa Brassica, Brassica Oleracea Rapa Var Gongylodes, and Brassica Rapa subsp. rapa were the seeds chosen. The seeds were then put through a training procedure that involved using an image classification algorithm, aiming to classify each seed correctly. The classification accuracy of the training procedure was then evaluated and compared with the actual frequency distribution of the dataset.

It's significant to note that the number of images collected for each seed type was around 600. A total of 6065 images were classified into 10 classes, with 50% of them trained, 30% tested, and 20% validated. We converted each image in this dataset to 128x128 pixels based on the input requirements of the suggested model. After preprocessing, we used the same model for each category to identify the correct class label (i.e., whether the image belongs to the seeds or the background category). We then calculated the model's accuracy on each class and the total model accuracy to assess its performance on the given dataset and used the obtained accuracy results to compare the efficacy of different classification models.

Therefore. In this research, we will also create a new Brassica dataset that is not already available and then we evaluate how well our CNN architectures perform on it. Images of the seeds were collected in daylight using digital microscopy at 1600 x 1200 pixels and 96 dpi. The dataset was created by selecting a significant number of images from each class at random. It was designed to put our new CNN architectures to the test and study the results, which will help us understand how CNN models are employed in agricultural data classification tasks. Table 8.10 depicts the dataset's frequency distribution.

Table 8.10 Brassica seed image dataset description

| ID | Brassica class | Number of Images |
|---|---|---|
| 1 | Brassica Napus Var Annua | 610 |
| 2 | Brassica Napus Var Oleifera | 475 |
| 3 | Brassica Nigra | 653 |
| 4 | Brassica Oleracea Gongylodes | 667 |
| 5 | BrassicaO leraceaLCAV rubra | 650 |
| 6 | Brassica Oleracea Rapa Brassica | 612 |
| 7 | Brassica Oleracea Var Gongylodes | 562 |
| 8 | Brassica Rapa | 562 |
| 9 | Brassica Rapa Oleifera | 494 |
| 10 | Brassica rapa subsp. rapa | 640 |

The analysis of this research includes data preparation and collection, which is a work that requires close attention throughout the analysis process. So, collecting the information is a necessary stage in creating and building a CNN model. Therefore, the initial stage of this work is to prepare the data for a CNN. The second phase is data analysis, which entails feature engineering, extracting the target function, determining the functional form of the primary components that may effectively characterize the existing system, as well as determining the functional form of the target function. Our implementation technique in this study was based on many processes. In this approach, model creation and revision come first, then the collection of a seed dataset. Finally, the classification process is done, and the results are evaluated. After these steps, we conduct the data analysis process, in which we review and discuss the results, provide feedback, and modify the model according to our observations. Finally, the model is tested and retested with these modifications to ensure accuracy and consistency.

The data analysis process is an essential part of model creation and revision as it allows us to refine the model to better fit the data and ensure that the model provides reliable and valid results, allowing us to make better decisions and progress forward with the project. The methodology followed in this study is shown in Figure 8.19.

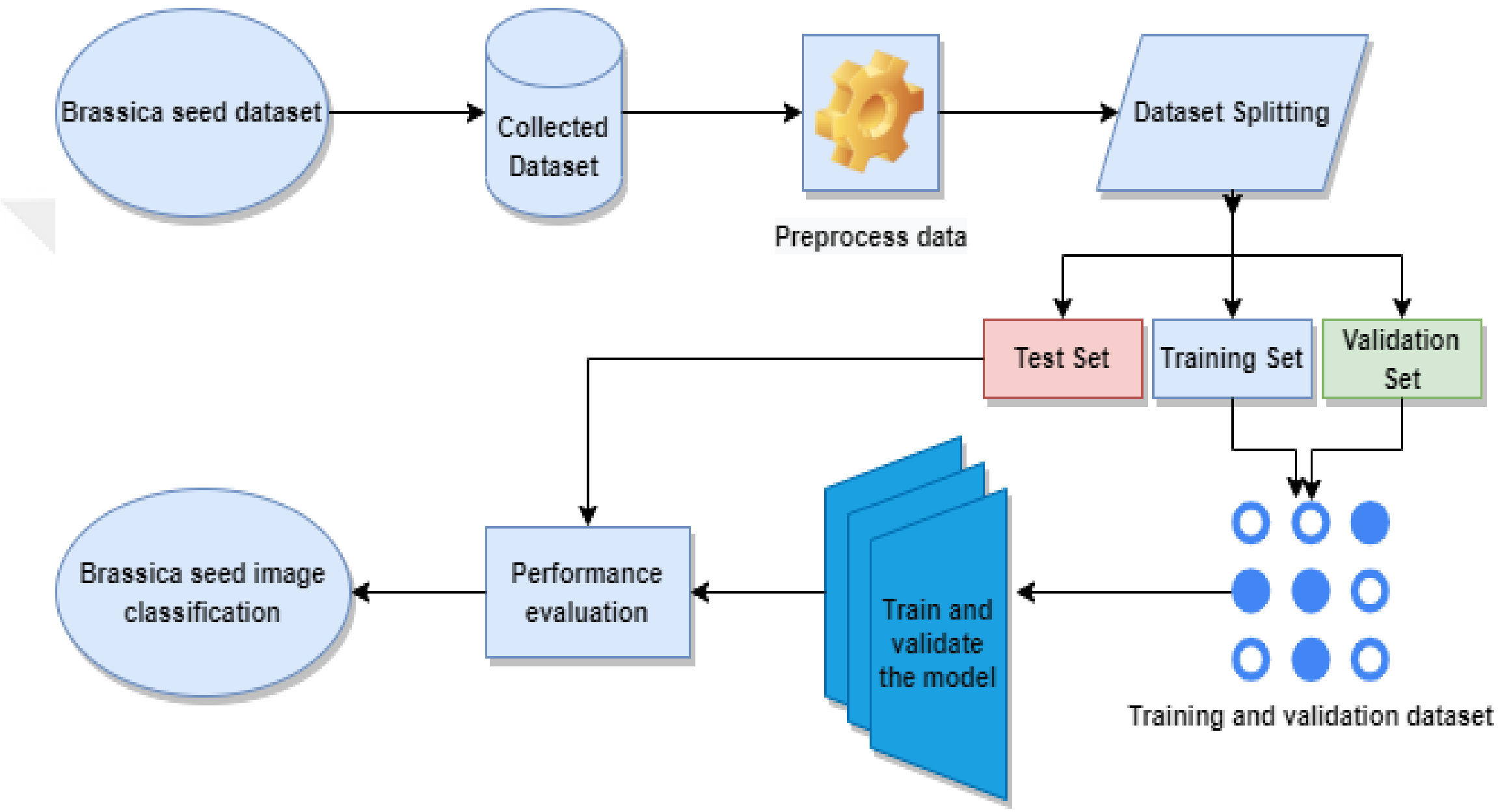

Figure 8.20 Basic steps for the proposed CNN model design flow

The suggested system process is depicted in Figure 8.20. Brassica seed datasets are gathered from various sources in the first stage, and all of these data pieces are then preprocessed to give the classifier algorithm useful input. We employed our model for seed class identification after data splitting (training, validation, and testing), which may be performed using a learning method. The generated model was then analyzed for correctness, and several performance measures were used to evaluate and analyze it (see the result and discussion sections). Once the model had been evaluated and analyzed, it could be used to make predictions and decisions about a new seed dataset, allowing us to identify new seeds accurately.

Once the process flow was successfully implemented, a more in-depth analysis of the data was necessary to ensure that all features were properly represented for prediction and to improve the accuracy of seed class classification. As we were attempting to classify the Brassica seed type into 10 categories, for example, this included evaluating the range of data values, checking for missing values, inspecting features to determine whether they should be normalized or standardized, and identifying correlations between seed properties, such as texture and moisture, as well as other physical attributes. The following subsection discusses how this study was applied to improve seed type recognition accuracy.

#### 8.3.1.2 Implementation

This section focuses on setting up lab tests for a Brassica seed classification system utilizing a novel model architecture and the TensorFlow framework. Various steps are needed to apply the suggested model architectures, starting with dataset collection and finishing with performance evaluation and classification. The model setup consists of several steps, including data analysis and creating an input pipeline to enable a classifier to predict classes. The input pipeline consists of pre-processing techniques, such as image cropping and resizing, to create the desired data set for classification, followed by model selection, training, and optimization. Once the input pipeline has been completed, we trained and tested a model on the Brassica dataset to assess its performance using various metrics, including recall, precision, and accuracy.

After training and testing the model, further improvements can be made to increase its accuracy and precision, either by further fine-tuning the parameters and hyperparameters of the model or by using different algorithms and techniques such as ensembling, or by applying regularization and other methods to reduce overfitting and bias.

Since the learning strategy of our new model fits into administered learning, we additionally labeled the data (as seen in Figure 8.21). We have observed that by using our learning approach, model training may be sped up while also allowing for better model performance. The capacity to perform more quickly and inexpensively is the main advantage of the model's speed and scalability.

As a result of the labeled data, our model offers faster and more cost-effective training, allowing us to attain better outcomes in a fraction of the time, which has implications for both industry and academic research. Furthermore, The use of data labeled improves the performance of the model we use and provides valuable insights into how to address diverse tasks across various domains, improving the efficiency of model training and overall performance. Figure 8.20 presents an example of a Brassica-labeled dataset.

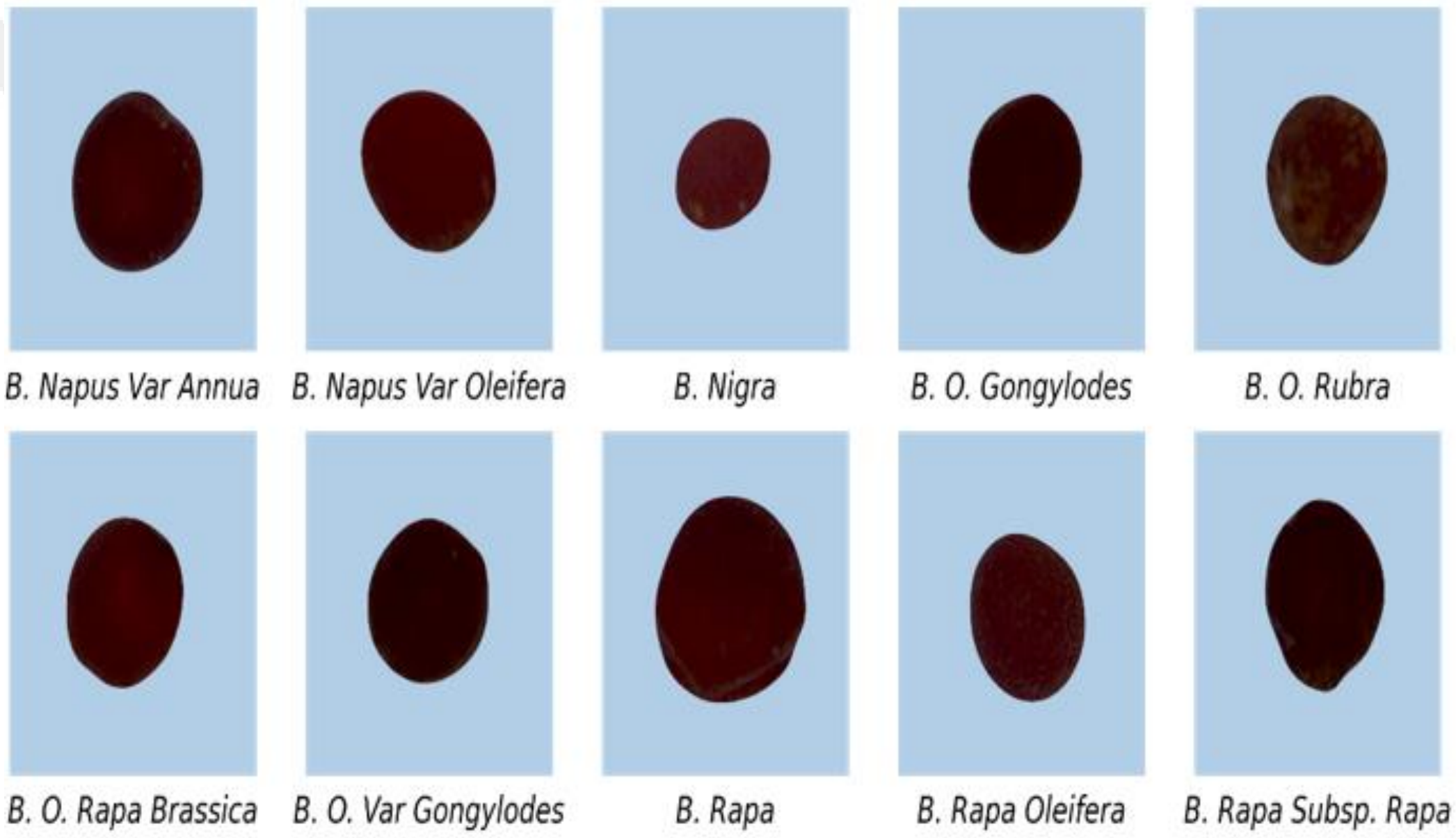

Figure 8.21 A glimpse of the Brassica labeled dataset of ten classes

In this work, our model contains 24 layers for image classification, including an input and output layer, and each image was used many times throughout the training stage. During the training of the classification model, each batch is encountered precisely once per epoch by the classification model, and at the end of each epoch, the model's performance on the validation set is evaluated. To configure the model for the dataset, we tuned hyperparameters such as the optimizer, learning rate, batch size, and number of epochs. Based on the dataset's sample size, the batch size was adjusted to 64. Adam was chosen as the optimizer for the model architecture. Additionally, the training process involved setting the learning rate to 0.001 and training the model for a total of 200 epochs. The learning rate epochs number were determined based on the dataset's sample size, and the learning rate was fine-tuned to balance learning time and accuracy, aiming for maximum accuracy while maintaining reasonable learning time. With these parameters set, the model was assessed and tested on an independent test set, using recall, precision, and accuracy to determine the overall model performance.

Even though, in practice, it is advisable to terminate the model when the loss and accuracy both stabilize, it is expected to observe a decline in both the training and validation loss at each epoch. Moreover, training and validation accuracy would be less accurate since more data would be introduced to our model if we could decrease the learning rate by increasing the time step length. Although a slight adjustment in the learning rate may not appear to be much, it could significantly impact how well the model system works. As a result, the weight normalization for the most recent period should be large. Therefore, it should be changed until the last epoch, when the loss function remains stable.

Numerous thorough tests were performed to accurately analyze the performance and demonstrate the efficacy of our suggested approach for performing seed classification over a wide variety of seeds. The experiments were carried out using a Dell N-series laptop equipped by a 2.5 GHz Intel i6 CPU, and 8 GB of RAM, in addition to the new CNN model.

The suggested method is developed in Python and uses the open-source and free TensorFlow package. All implementations were completed on a personal computer with a GPU using Google Collab. The suggested model's effectiveness was assessed on real-world datasets, accompanied by a comprehensive evaluation that compared it to existing methods. The results clearly indicate that the presented approach outperforms existing methods in terms of accuracy for image classification tasks.

### 8.3.1.3 The proposed CNN model architecture

This part of this thesis suggests a novel CNN-based model for effectively classifying Brassica seed images. The new model consists of a total of 24 layers, with the first layer being the input layer, which feeds data into the network. This layer will handle images with a resolution of 128x128x3. Furthermore, the complete process of this model architecture is divided into two stages: feature extraction and classification. The first step has successive convolutional, activation, and pooling layers. Ultimately, in the second stage, there are dense, dropout, fully connected, SoftM.ax, and output layers. The CNN network is described in detail in Figure 8.22 and Figure 8.23.

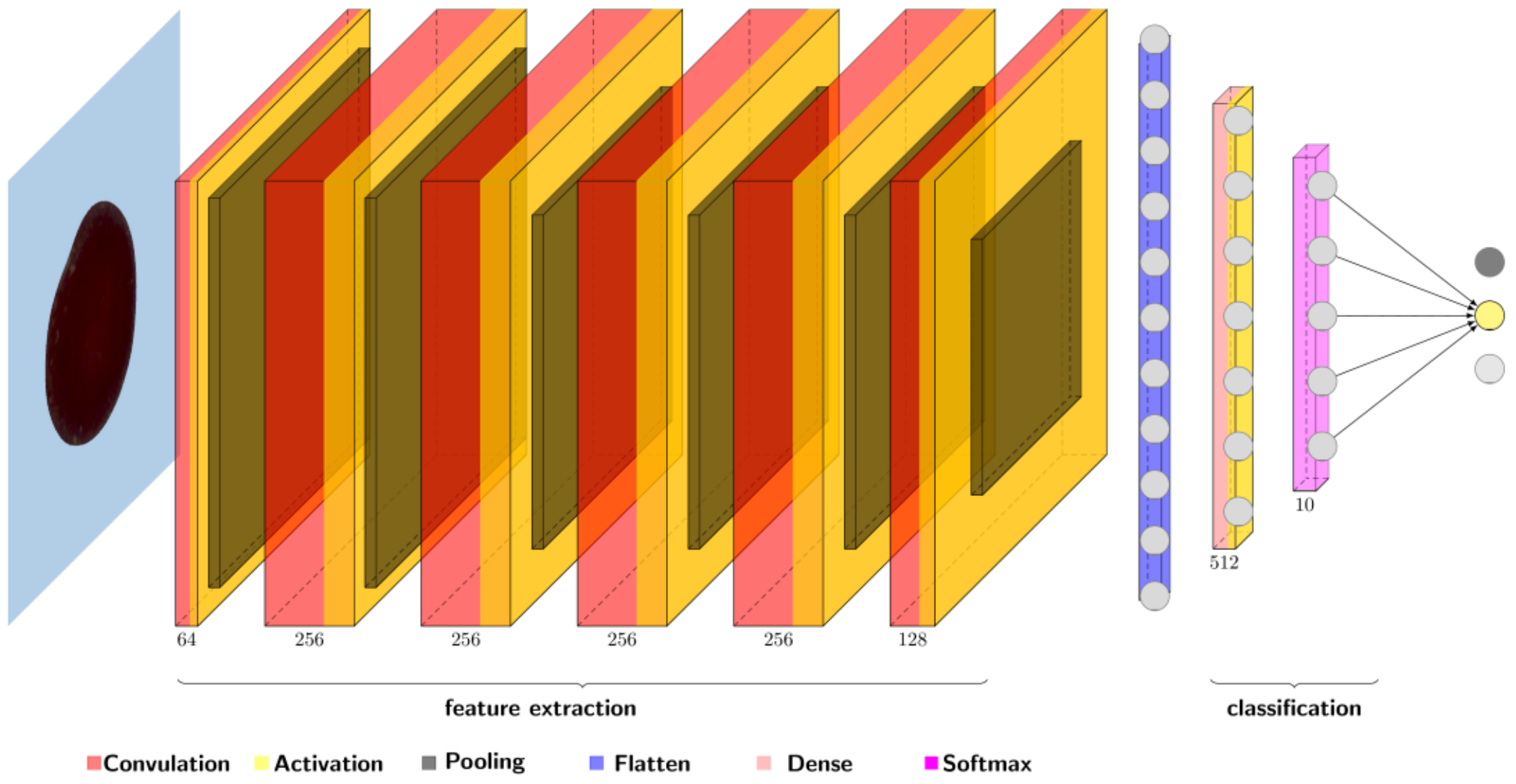

Figure 8.22 Visual representation of the proposed CNN model for classifying brassica seeds

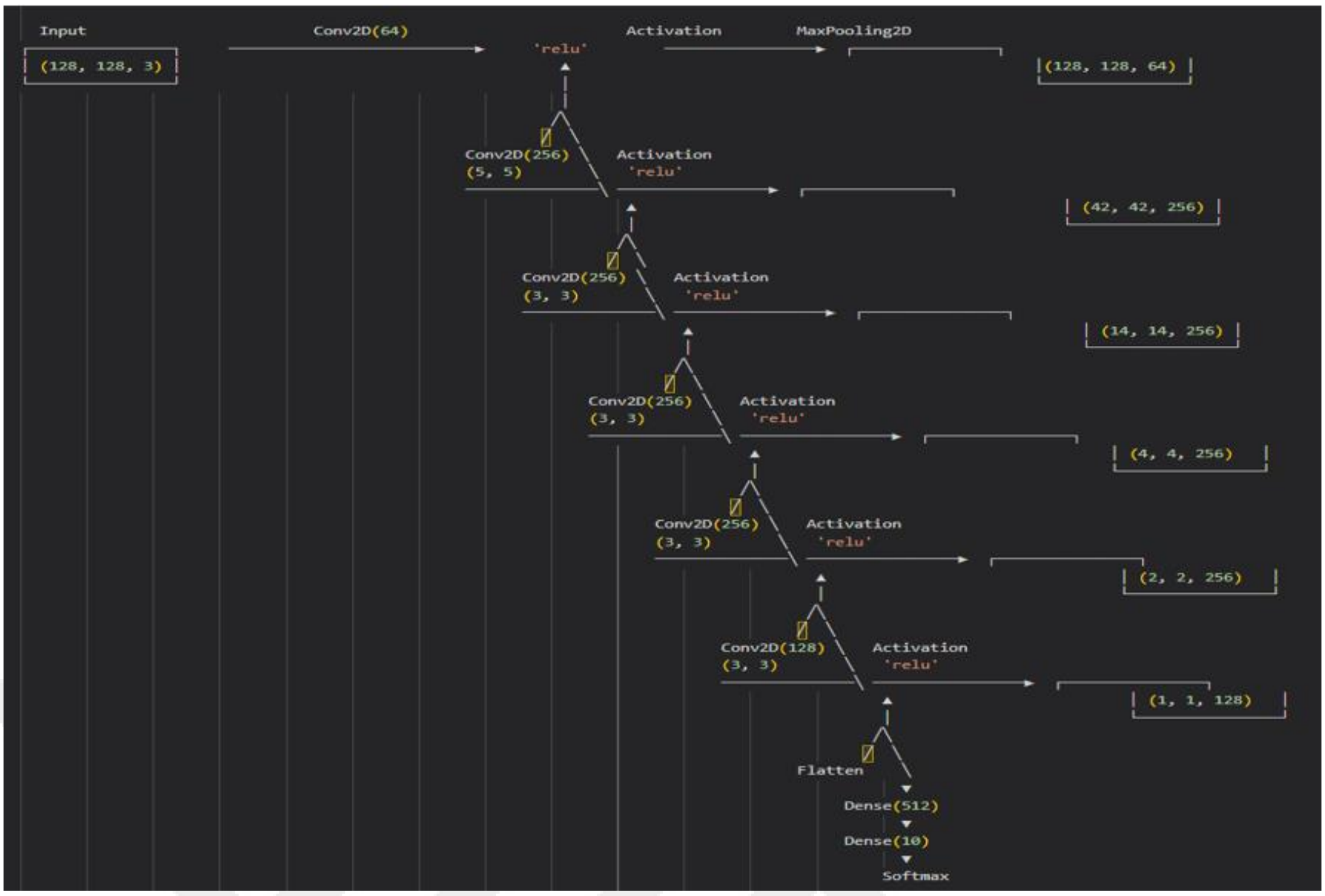

Figure 8.22 Proposed CNN architecture for classifying brassica seeds

As shown in both Figures, the new model comprises a total of 23 layers, designed to optimize the classification of Brassica types. The initial layer serves as the input layer, receiving data in the form of 128x128x3 resolution images. Subsequently, a convolutional layer with 64 filters is applied, aiming to detect various visual patterns and features in the input image. The 'relu' activation function is then employed, introducing non-linearity to the network.

To reduce spatial dimensions while preserving vital information, a max pooling layer is utilized. This layer aids in decreasing computational complexity and offers a degree of translation invariance. Following this, a series of convolutional layers with increasing numbers of filters (256) and corresponding activation functions are employed. The filter sizes for these convolutional layers are specified as 5x5, 3x3, 3x3, and 3x3, respectively. Each convolutional layer, in this model architecture, is accompanied by a subsequent max pooling layer, except for the last one. At the final convolutional layer, a 1x1 convolutional layer with 128 filters and a 'relu' activation function is implemented to capture higher-level features from the preceding layers. Notably, the kernel size for this layer is 3x3, deviating from the 1x1 size mentioned in the architecture representation.

After the 1x1 convolutional layer, the output is flattened to transform the multidimensional feature maps into a one-dimensional vector. This vector is then passed through two fully connected layers, each containing 512 neurons. To obtain the final class probabilities for the 10 possible Brassica types, the softmax activation function is applied. By utilizing the softmax function, the fully connected layers generate a predictive distribution of Brassica seeds based on the prepared dataset. To address overfitting concerns, pooling layers are commonly inserted between consecutive convolutional layers. This helps reduce the parameter count and regulate data processing within the network. This pooling layer aids in preventing the model from overly adapting to the training data, enhancing generalization capabilities.

The novelty of this architecture lies in the utilization of multiple Conv2D layers with different filter sizes (5x5, 3x3) and the resulting spatial dimensions of the outputs (42x42, 14x14, 4x4, 2x2). Unlike the standard CNN architecture that typically employs one or two Conv2D layers with the same filter size and spatial dimensions, this model captures more diverse and complex features from the input image, thereby improving accuracy. Furthermore, unlike the traditional approach of downsampling through pooling layers, this architecture employs identity mappings. This strategy preserves the spatial dimensions throughout the network, enabling the model to capture fine-grained spatial information at each stage. This system can increase the recognition and classification of Brassica types by allowing the model to capture intricate patterns and specific details unique to Brassica classification tasks.

Overall, the proposed model showcases a distinctive design with multiple Conv2D layers of varying filter sizes and spatial dimensions, offering a novel perspective on Brassica classification. The decision to maintain spatial resolution throughout the network, instead of downsampling, facilitates the capture of intricate details and patterns, potentially leading to improved accuracy and performance in Brassica classification tasks.

Based on the provided dataset, Softmax outputs fully connected layers to produce a prediction distribution of 10 Brassica seeds. Also, we frequently insert a pooling layer between convolutional layers that precede each other to minimize the number of parameters and data processing, as well as to control overfitting. Our model thus included a pooling layer to take overfitting into account. To enhance the accuracy of feature selection, we also employed layer pooling to account for the consistent layer addition. We used dense layers because convolutional layers aim to extract distinguishable characteristics and features, even though the neurons are already flattened before they are fed into the dense layer. However, fully connected layers try to categorize the features. Therefore, adding more layers to the dense section can enhance the classification performance of our network using the collected data. Fully connected layers can be compelling in this regard, as they can identify patterns within the data that are not apparent from a mere visual inspection, giving the network a much greater understanding of what it is working with and allowing it to make more complex decisions when classifying new data patterns within the data that are not apparent, giving the network a much greater understanding of what it is working with and allowing it to take more complex decisions when classifying new data.

Further, adding too many layers to the dense section can result in overfitting and cause our model to become too reliant on the collected data, which can lead to poorer classification performance, especially when working with new data, so It is imperative to strike the right balance between the number of layers in our model and its ability to generalize. Therefore, to make the best use of fully connected layers and enhance the classification performance of our network, we must carefully consider both the number of layers and their complexity. With this in mind, we should aim to optimize the number of layers and their complexity by adjusting our hyperparameters accordingly, paying close attention to the impact of our adjustments on the ability of model to generalize to new data, in order to ensure that it can make effective decisions when processing data that it has not seen before.

The convolutional layer of the suggested network is a 6 x 3 convolution. Furthermore, in this dataset, the receptive field is big enough to extract image features. However, more subtle traits must be retrieved to categorize Brassica seeds more precisely. Therefore, to maximize the performance of the developed model, it was crucial to meticulously select the optimal network depth and accordingly decrease its size. Additionally, network architecture may significantly affect how well a model classifies data. As a result, the network's convolutional layers were made better by making them more pertinent to the model classification goal. Using a pertinent network depth reduced the developed model's size without sacrificing its accuracy. Its accuracy was improved by optimizing the convolutional layers, allowing for a more efficient system that could classify data more effectively.

On the other hand, we classified images of brassica seeds using three popular cutting-edge CNN models (InceptionV3, DenseNet121, and ResNet152) and compared their performance to our model. We employed a global average pooling with a fatten layer and a fully connected for each model, then a SoftMax layer with 10 outputs for every epoch. Finally, we investigated whether this network was susceptible to overfitting by carefully adjusting the number of hidden omponents in three networks: Inceptionv3, Densnet121, and Resnet152.

Comparing the performance of multiple CNN models is vital to identify the architectural decisions that yield the highest performance. As the more conventional methods of feature engineering, dimensionality reduction, and super-resolution become less effective, a potential direction for future study is the comparison of CNNs with deep convolutional layers and dense pooling layers, as well as the work on CNN models in general. The network diagrams for these three models are shown in Figure 8.23.

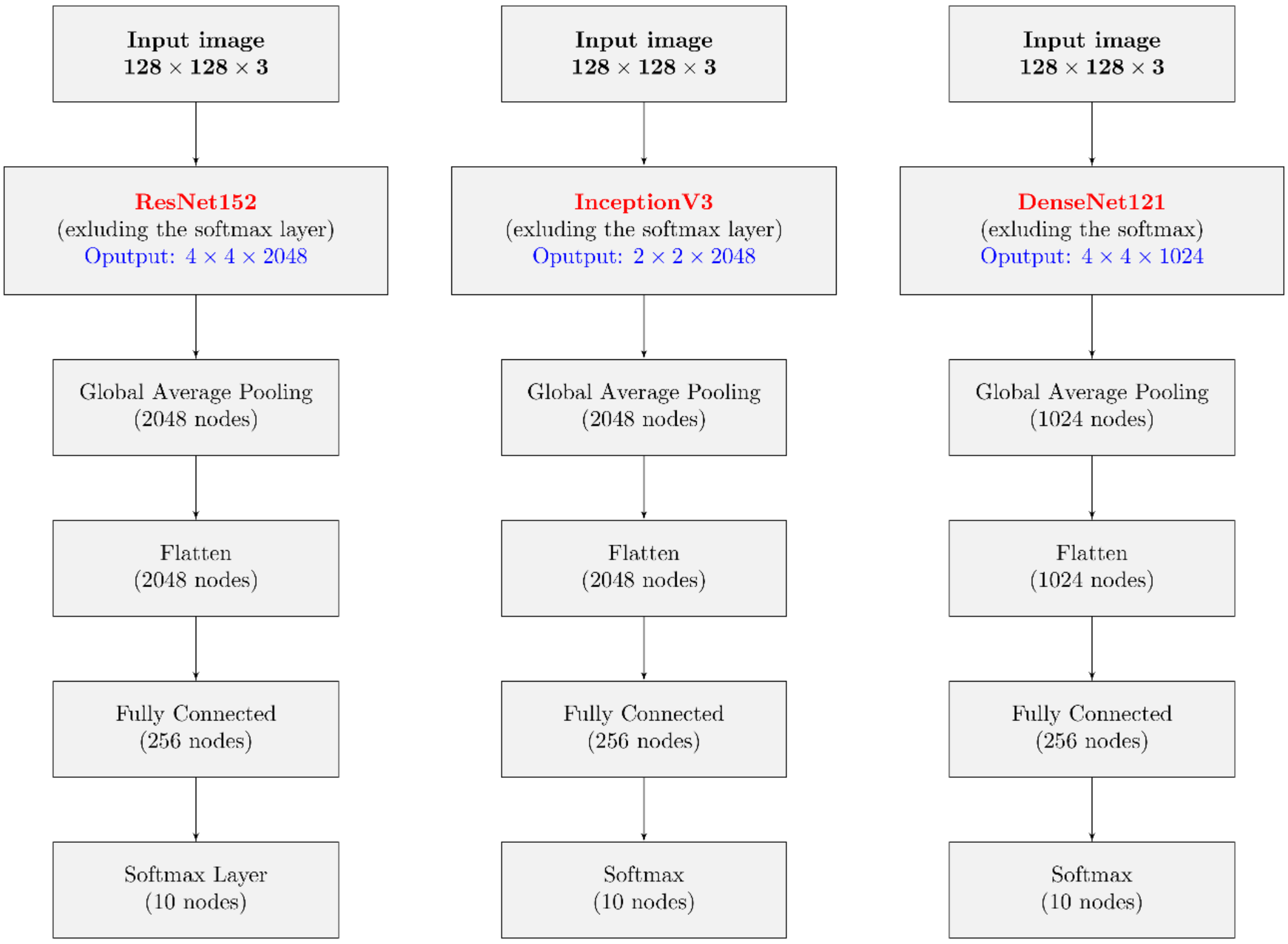

Figure 8.23 Architectural schematics for three models

We evaluated and compared these models' performance using a dataset of Brassica seed images. Using our Brassica seed dataset, we trained all model layers using the baseline training approach. We also used pre-trained weights that were randomly initialized for these three models. The training and validation datasets were trained for 200 epochs using a batch size 64. We utilized a learning rate of 0.001, an Adam optimizer, and fine-tuned additional hyperparameters to optimize the model's performance.

We conducted experiments with different batch sizes and epochs to determine the optimal values for model training. As shown in Figure 8.24, we present the results of our proposed model trained using batch sizes of 8, 16, 32, and 64. It was possible to see in Figure 8.24’a, b that the training time per epoch decreased while the testing accuracy increased and batch sizes increased. Throughout the model training, a batch size of 64 generated the most effective results. Testing accuracies at various model training epochs were measured in Figure 8.25. Testing accuracy increased gradually with epochs until it reached 200, so the epochs were chosen for 200 iterations, providing a balance between training time and accuracy. This suggests that using a batch size of 64 and training the model for 200 epochs may be optimal for achieving good performance. It is crucial to acknowledge that these optimal parameters may vary depending on the specific dataset and model architecture being used. The results from this study provide valuable insights into selecting optimal parameters for training neural networks. Nevertheless, it is important to remember that generalizing these findings to other models and datasets may not be appropriate. Therefore, it is recommended to perform similar experiments on other architectures with various datasets to identify the best parameters for each case.

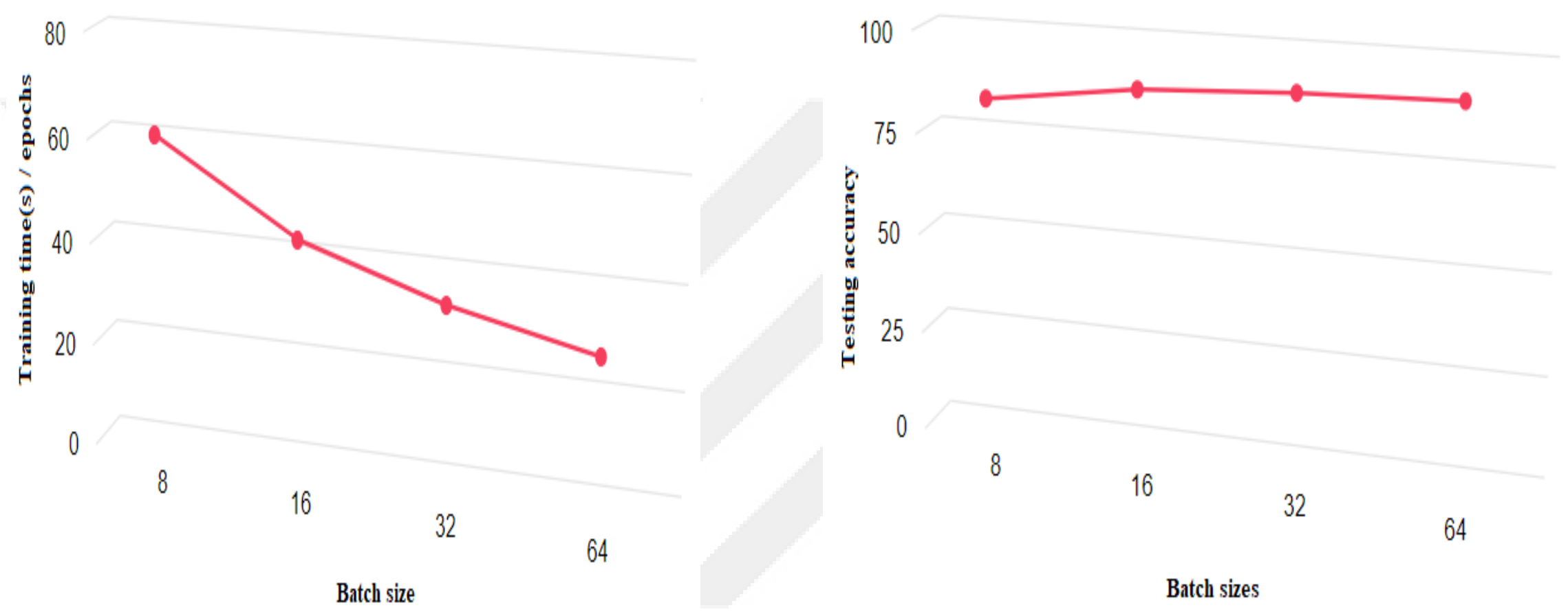

(a) (b)

Figure 8.24 The effect of batch sizes on the model's performance is shown in (a) batch size vs. training time per epoch and (b) batch size vs. the model's testing accuracy

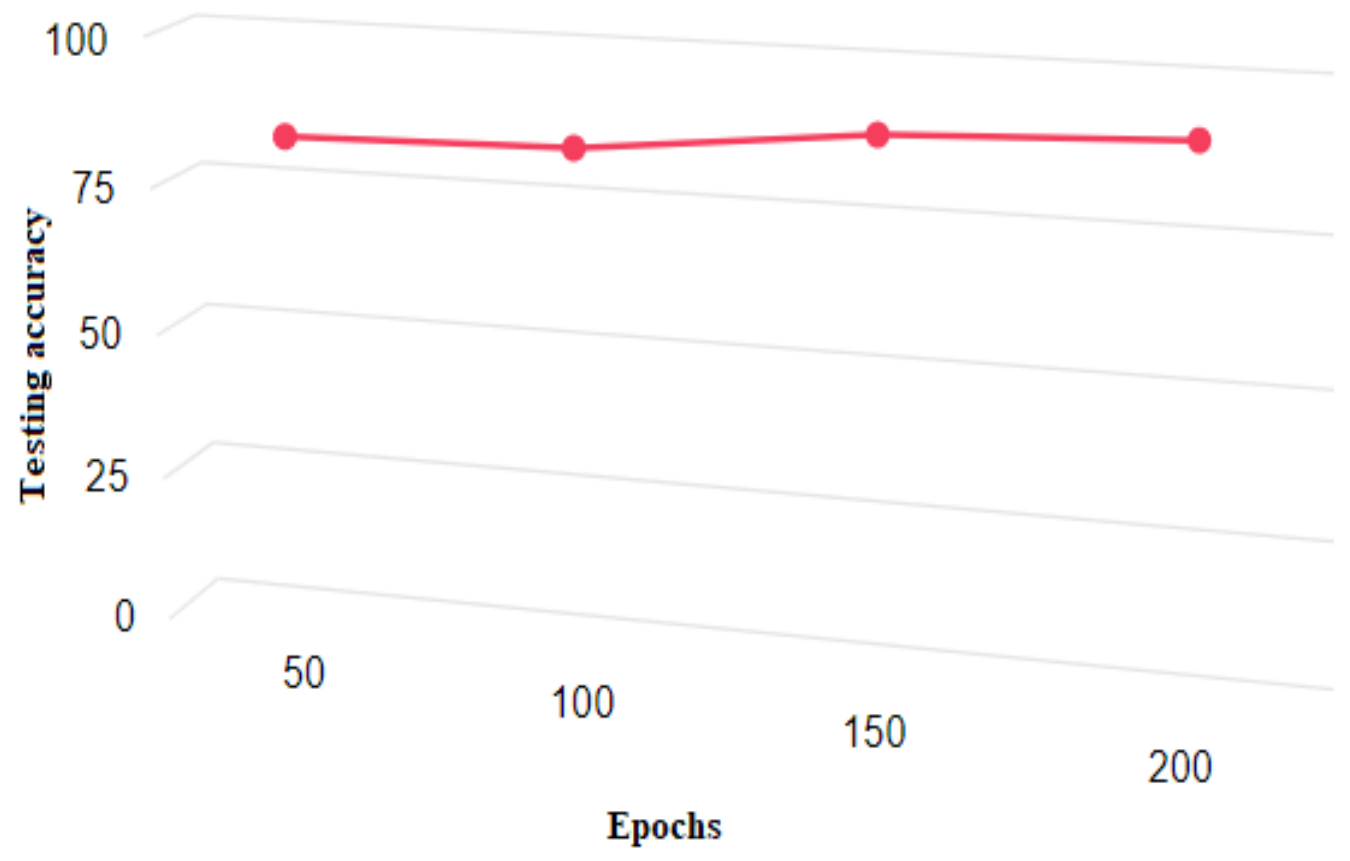

Figure 8.25 Effect of epochs on testing accuracy

In the next section, a detailed examination is carried out to scrutinize the findings from the training and validation stages.

### 8.3.2 Result and discussion

This section details the experiments conducted to assess the accuracy and efficiency of our suggested approach using the Brassica seed dataset.

#### 8.3.2.1 Application of the proposed model architecture

In this part, we conducted a Brassica seed classification test experiment on the collected dataset using the suggested model. Before training the model, we updated the hyperparameter values and modified the model output layer using Softmax to output ten classifications. The model was set up to run for 200 epochs with a 0.001 learning rate. Understanding how to analyze the impacts of learning rate on model performance and choosing an acceptable learning rate for a model are essential for preventing overfitting.

To make sure the model was not overfitting, we monitored the accuracy and loss after each epoch, ensuring that the accuracy was increasing and the loss was decreasing, as well as using early stopping, where the model stops running if there is no improvement in accuracy or loss for a set number of epochs. We also monitored the validation accuracy of the model to ensure that it was not overfitting by assessing how well it performed on data that it had not seen before while at the same time keeping an eye on the training accuracy of the model. Moreover, This allowed us to adjust the learning rate and ensure that the model was running optimally while also preventing overfitting, which is a significant issue that can occur when the model begins to "memorize" the training data rather than learning general patterns from it, which could lead to poor performance on test data or new data.

We adjusted additional hyperparameters like the optimizer and batch size to create a model for our dataset. Therefore, the batch size was changed to 64, and Adam was selected as the model's optimizer based on the dataset's sample size. In order to guarantee that the model generalizes successfully, the hyperparameter adjustment approach is crucial. Ensuring the model fits the complete dataset for the prediction task is another essential factor to consider while fine-tuning the model. Then, we assessed the performance on the data utlizing the accuracy curve to evaluate its performance and determine that it could correctly predict and classify Brassica seed kinds in a shorter training time. Figure 8.26 depicts the suggested architecture's accuracy and loss of graphics.

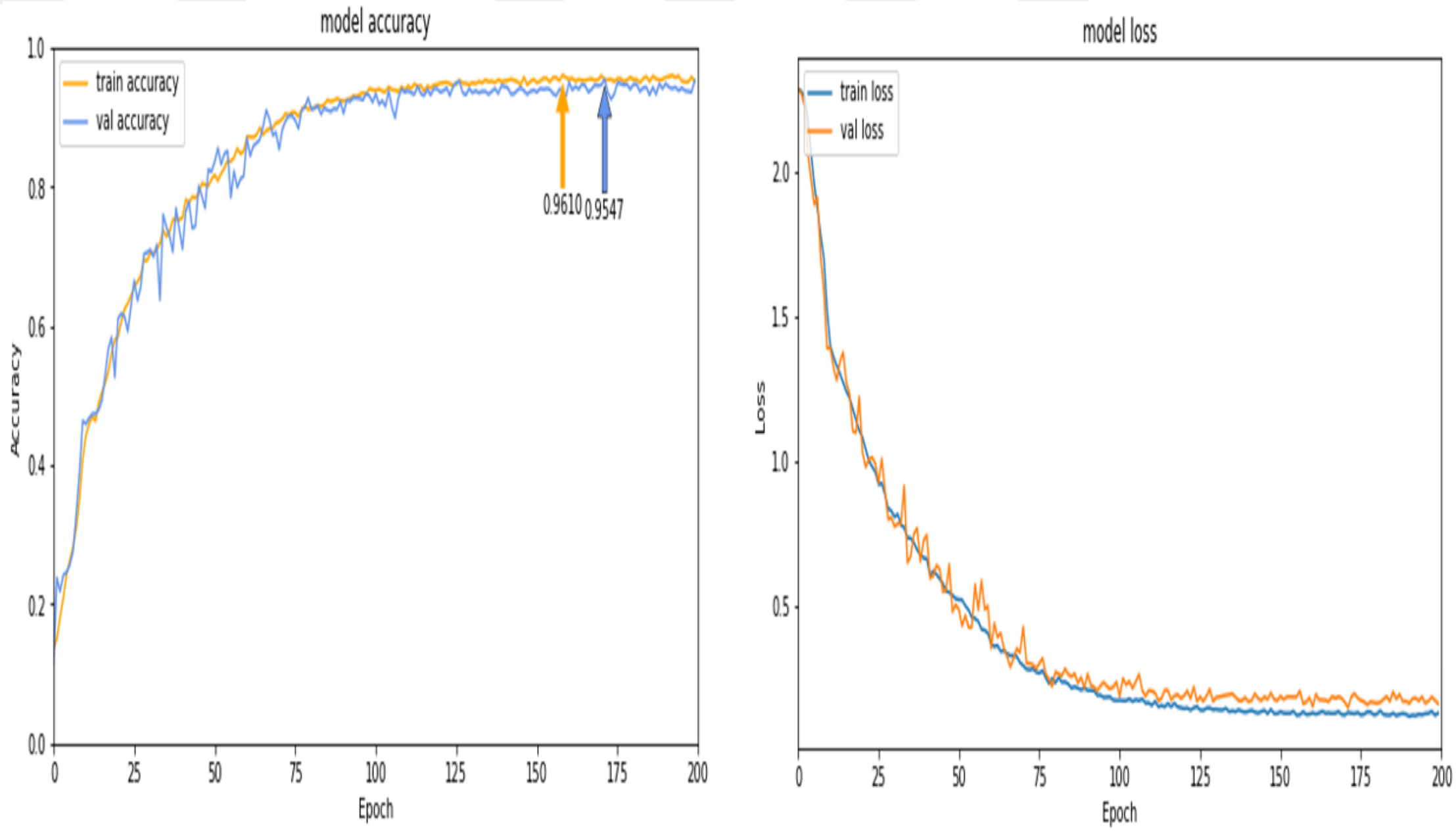

Figure 8.26 Accuracy of loss, validation, and training of the suggested model

The suggested models were clearly accurate due to their high training and validation accuracy. As a result, we discovered that the models were accurate when we evaluated the average training and validation accuracies, which are 96.10% and 95.47%, respectively, as well as the training and validation losses, which are 0.3478 and 0.4390, respectively. These high training and validation accuracies demonstrate that the proposed models can accurately predict different dataset items, which is a desirable trait of any neural network model, making them suitable for various tasks. Moreover, the training and validation losses were relatively low, indicating that the model could find a good optimum in training, meaning that the dataset items can be predicted accurately with minimal chances of overfitting or underfitting. Although we expected some variance between the training and validation accuracies due to the increasing complexity of the problem, they were still close enough to demonstrate that our models could learn effectively and thus make accurate predictions on the task. Overall, the results of our evaluation were promising, demonstrating that our models have the potential to be used for this task.

The performance of the model remained constant throughout the training and validation stages because of the pre-processing techniques included in the model. Data was first collected, and attempts were made to distribute the data among all classes. Additionally, by ensuring that the model did not considerably stray from its training performance, the dropout strategy significantly improved its validation performance, reducing overfitting and allowing the model to generalize better.

#### 8.3.2.2 Performance evaluation of the proposed model for image classification of Brassica seeds classes

The confusion matrix was employed in this work to offer a clear view of the accuracy and methods by which our classification model gets confused while making predictions. The confusion matrix in this study comprises four metrics, each of which assesses classification accuracy and seeks to anticipate how each combination of predictor and target attributes will behave for one given class value. The effectiveness of the CNN model was assessed through the utilization of the confusion matrix. This matrix provides information on the true class labels of the samples as well as the predicted class labels generated by the CNN classifier. The confusion matrix data is incredibly useful in this context as it demonstrates where misclassifications occur and helps to identify weaknesses in the model, as well as providing a means of identifying which class labels were the most difficult to classify correctly. Consequently, we presented the outcomes of testing the classifier that distinguishes among 10 classes of Brassica seeds utilizing the two labeled datasets. The four metrics most frequently used in this study are the ones used in the previous study. In this study, the letters TP and TN stand for correctly identifying

Brassica seeds, whereas the letters FP and FN stand for incorrectly identifying them. The models' confusion matrices are shown in Figure 8.27.

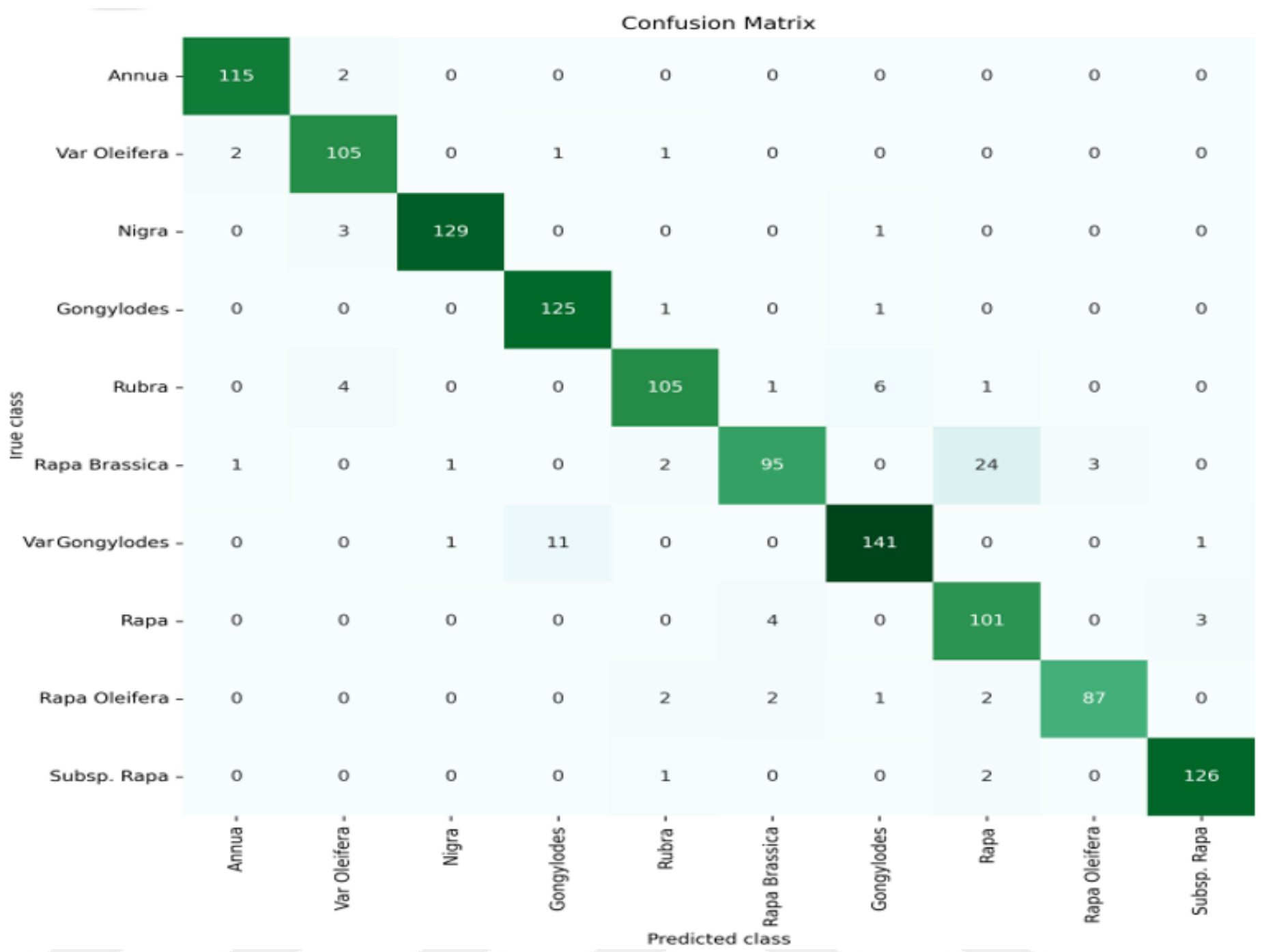

Figure 8.27 Confusion matrices for the suggested models using the dataset for Brassica seeds

The suggested CNN model effectively predicted the images of 10 classes, according to the proposed approach and architecture trained with the Brassica dataset image. Hence, the tasks of analysis, assessment, and validation of our method were completed. The presented CNN model, which was utilized to classify the input image into 10 classes, produced satisfactory results, as shown in Figure 27. Using 1214 combined images from 10 distinct classes After 200 epochs, The model attained a classification accuracy of 95.56% for the Brassica seeds during training and 94.21% during validation, as shown in Figure 8.26. Therefore, the suggested CNN model was able to successfully classify Brassica seeds into 10 distinct classes, providing reliable results with an acceptable accuracy rate, both during the training and validation processes.

The results of the suggested model were assessed by specific statistical metrics of the confusion matrix, including accuracy, recall, precision, and F1-score. These performance metrics were chosen because they were the most regularly used measures to assess the outcomes and performance of most approaches in prior research (Hochreiter et al., 1997) Consequently, the previous study's performance assessment equations were employed to construct performance measures and evaluate results.

As the confusion matrix provided us with all the essential parameters for each class, we computed the performance criteria for the classes, which are displayed in Figures 8.23 and 8.24. In addition, looking at the accuracy obtained by the model on this collected dataset, it is safe to say that the model is doing well regardless of the dataset it is evaluated on (i.e., the collected dataset), as the achieved accuracy is 93% on this dataset for a total test number image of 1214. Furthermore, throughout the training process of the suggested method, the model attained an accuracy of 95.56%, while the validation phase produced an accuracy of 94.21%.

In addition, we computed recall, precision, and f1 scores for each class using the confusion matrix, as presented in Figure 8.23. The high values obtained for each class's precision, recall, and f1 scores indicate that the suggested model can accurately classify and distinguish between different classes, making it well-suited for real-world applications. Figure 8.28 represents the proposed model's overall performance measure, while Figure 8.29 exhibits a curve indicating how well the metric performance works for both the training and validation sets.

| | precision | recall | f1-score | support |
|---|---|---|---|---|
| Annua | 0.9746 | 0.9829 | 0.9787 | 117 |
| Var Oleifera | 0.9211 | 0.9633 | 0.9417 | 109 |
| Nigra | 0.9847 | 0.9699 | 0.9773 | 133 |
| Gongylodes | 0.9124 | 0.9843 | 0.9470 | 127 |
| Rubra | 0.9375 | 0.8974 | 0.9170 | 117 |
| Rapa Brassica | 0.9314 | 0.7540 | 0.8333 | 126 |
| VarGongylodes | 0.9400 | 0.9156 | 0.9276 | 154 |
| Rapa | 0.7769 | 0.9352 | 0.8487 | 108 |
| Rapa Oleifera | 0.9667 | 0.9255 | 0.9457 | 94 |
| Subsp Rapa | 0.9692 | 0.9767 | 0.9730 | 129 |
| accuracy | | | 0.9300 | 1214 |
| macro avg | 0.9314 | 0.9305 | 0.9290 | 1214 |
| weighted avg | 0.9332 | 0.9300 | 0.9297 | 1214 |

| | | |
|---|---|---|
| val_acc | 0.9421 | std 0.0052 |
| train_acc: | 0.9556 | std 0.0039 |

Figure 8.28 The overall performance of the suggested model

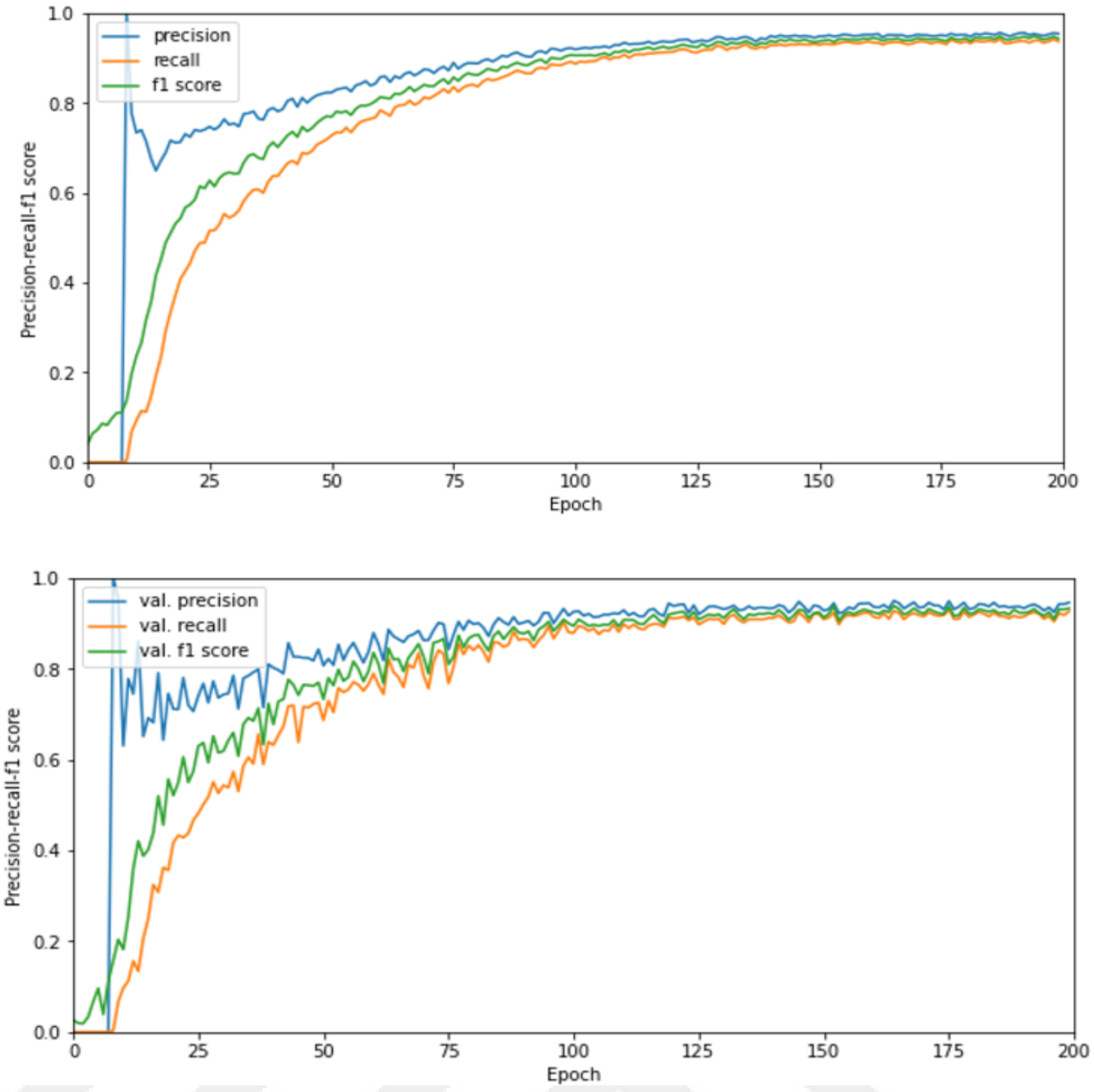

Figure 8.29 A schematic representation of the suggested model's training and validation outcomes

Figure 8.29 depicts the recall, accuracy, F1 score, and support for each seed class described in the suggested model. The support value represents the number of samples of each class during the training, while accuracy is the proportion of correctly predicted samples by the model over the entire dataset. It is important to note that, as illustrated in Figure 8.29, the model achieved the highest possible values for each seed class and in all metrics (recall, precision, and F1-score) in the training and validation sets, except for class 5 recall and class 7 precision, which were mixed up due to texture similarities, and also that this might be due to the camera's light settings, indicating that the model could be improved if it were trained in different light settings and with a larger number of instances for each class, which would provide a better understanding of the object classes and lead to a more accurate model. Despite the mixed-up results for class 5 recall and class 7 precision, overall, the model achieved very good results in all metrics, indicating that the model is indeed very successful in terms of predicting the different classes with a high degree of accuracy. But it is expected that the performance of the model would be more successful in the overall test set, and to make a more accurate evaluation, future experiments should be performed in other environmental conditions, such as different camera light settings, to assess the robustness of the model.

#### 8.3.2.3 Comparaison of the proposed CNN model to pre-trained state-of-the-art deep learning methods

In this paper, we study and assess the performance of our CNN model by comparing its classification performance to that of pre-trained models. As a consequence, as demonstrated in Table 8.10, we reported close results between the pre-trained models and our suggested model. The transfer learning settings and architectures shown in Figure 8.23 were chosen. Brassica was trained with Inception-v3, Densnet 121, and Resnet 152. Moreover, the ideal parameters shown in Figure 8.23 were employed to minimize over-fitting during training and to save time. The networks were trained over 200 epochs. Table 8.11 and Figure 8.30 displays the classification results for all kinds of Brassica seeds using the various models.

Table 8.11 Overall performance of CNN architectures

| Method | Accuaracy | Precision | Recall | F1 score |
|---|---|---|---|---|
| **Our model** | 0.930 | 0.9078 | 0.930 | 0.9026 |
| **Resnet152** | 0.7334 | 0.8613 | 0.7334 | 0.7279 |
| **Inceptionv3** | 0.8471 | 0.8745 | 0.8471 | 0.8212 |
| **DenseNet121** | 0.9003 | 0.9245 | 0.9003 | 0.9011 |

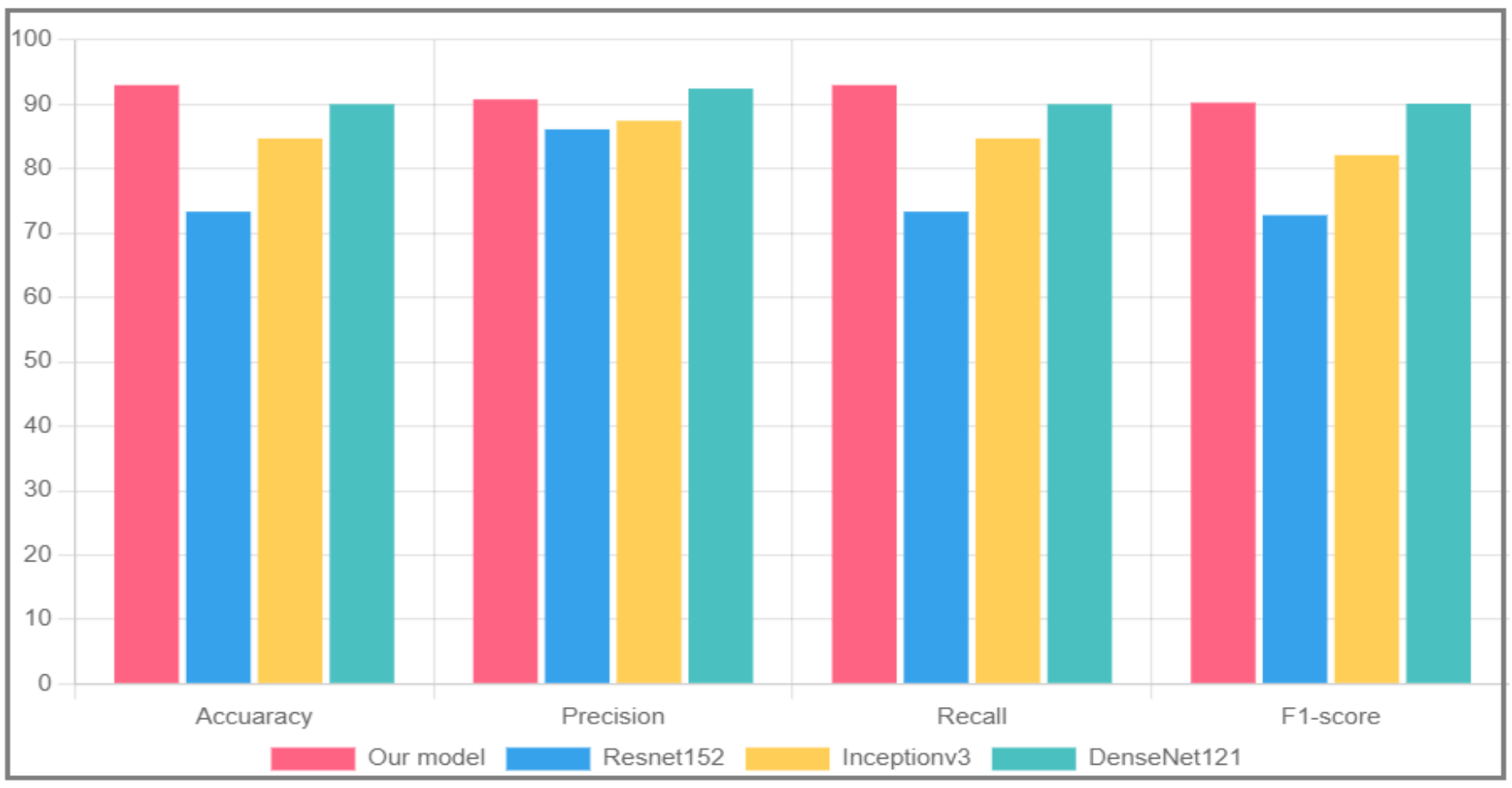

Figure 8.30 Overall performance of CNN architectures

Table 8.11 and Figure 8.30 shows that Densent121 obtained the best accuracy of 90.03% among pre-trained models, Inceptionv3 reached 84.71%, and Resnet152 obtained the lowest (73.34%).

The results indicate that the Densent121 model achieves a classification accuracy of 93%, which is lower than the suggested model. However, our experimental findings unequivocally demonstrate that our proposed model surpasses pre-trained models in accuracy, average precision, recall, and f1-score. For instance, Densent121 reported an average accuracy, recall, and f1-score of 90.03%, 90.03, and 90.11%, respectively, while our model achieved 93.30%, 93.30%, and 90.26% for similar parameters. These results suggest that our model has better image classification capabilities than pre-trained models, indicating the potential of our approach to accurately identify and classify images.

According to these results, it is conceivable to train a suggested model to execute classification tasks satisfactorily and more effectively than previously trained models and to do so more accurately. Additionally, it implies that the recommended learning techniques may be used to successfully extract low and high-level features from the image dataset under study. The capacity of the suggested model to analyze large amounts of data more quickly than other deep learning techniques is one of its benefits. Moreover, the offered techniques can improve performance by combining feature selection with transfer learning. Implementing the proposed model and its associated techniques provided evidence for performance improvement compared with other methods, demonstrating the potential of the suggested model to outperform existing deep learning approaches, allowing for better performance and accuracy in image classification tasks.

After that, in addition to model accuracy, we employed class accuracy, which appears to be more descriptive. Finally, we conducted a comparison of the classification performance between the suggested model and pre-trained models by analyzing the accuracy, F1-score, and recall for each seed class as specified in the suggested model. The results are shown in Figure 8.31, which demonstrates that the values of these performance measures are close among all models, and that all models achieved high values for each seed class, except for the seed classes indicated by the red line.

The presence of the red line clearly indicates a significant difference in the performance of the model, classes with a red line reveal that all models' performances become confused and unstable, except for our model, which maintained strong performance throughout all classes. Moreover, this again demonstrates how steady and efficient our model is compared to others. Furthermore, this can be be attributed to the fact that our model is trained with an optimal set of hyperparameters and a unique architecture, which offers it a significant advantage over all the other models and provides it with better generalization capabilities. These results were very impressive as the performance of all other models decreased. However, our model still managed to maintain its high performance, highlighting that it is well-suited for tasks requiring strong generalization capabilities. Using multiple models, Figure 8.25 depicts the classification results for all Brassica seed types.

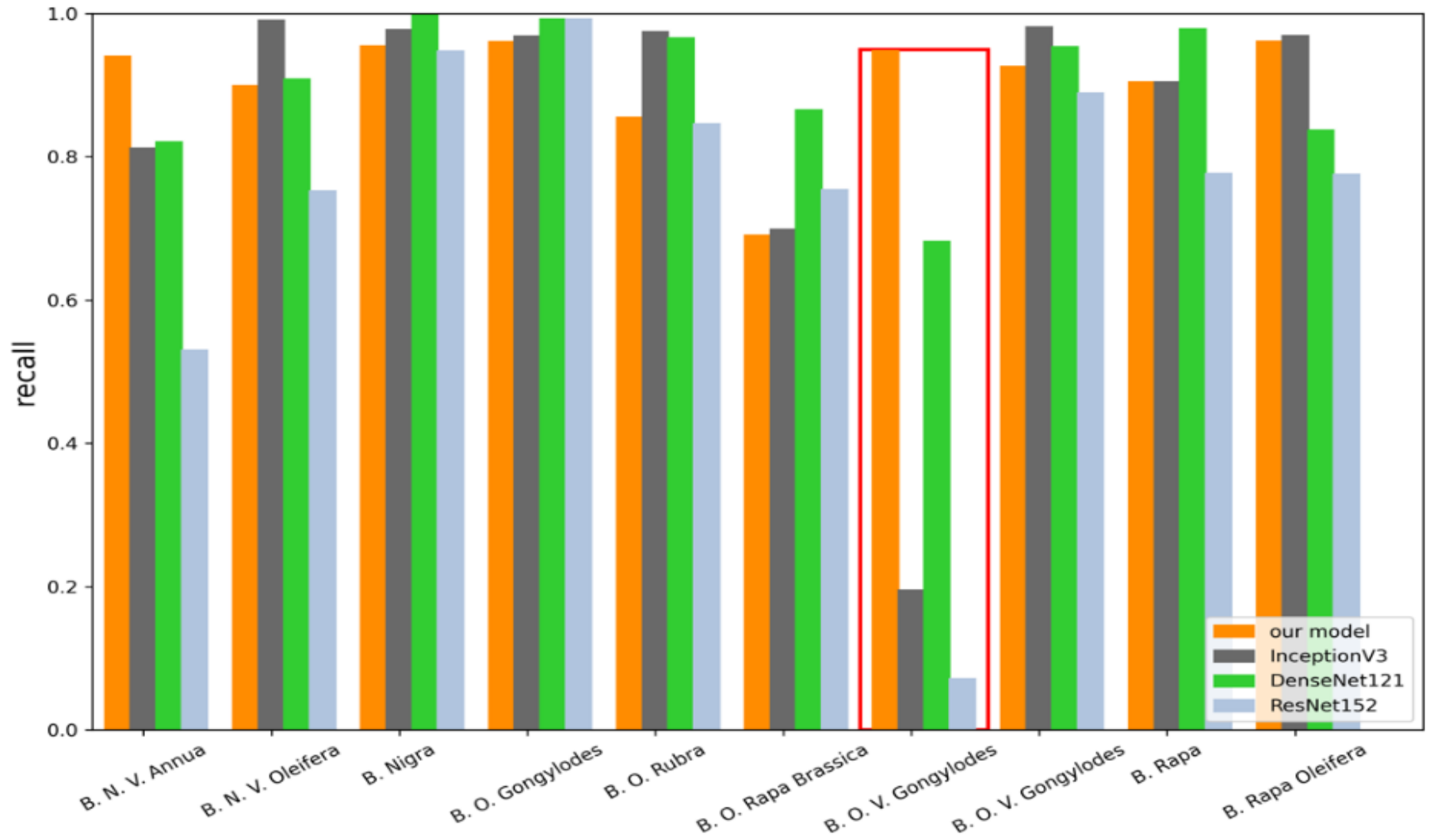

**(a) Recall**

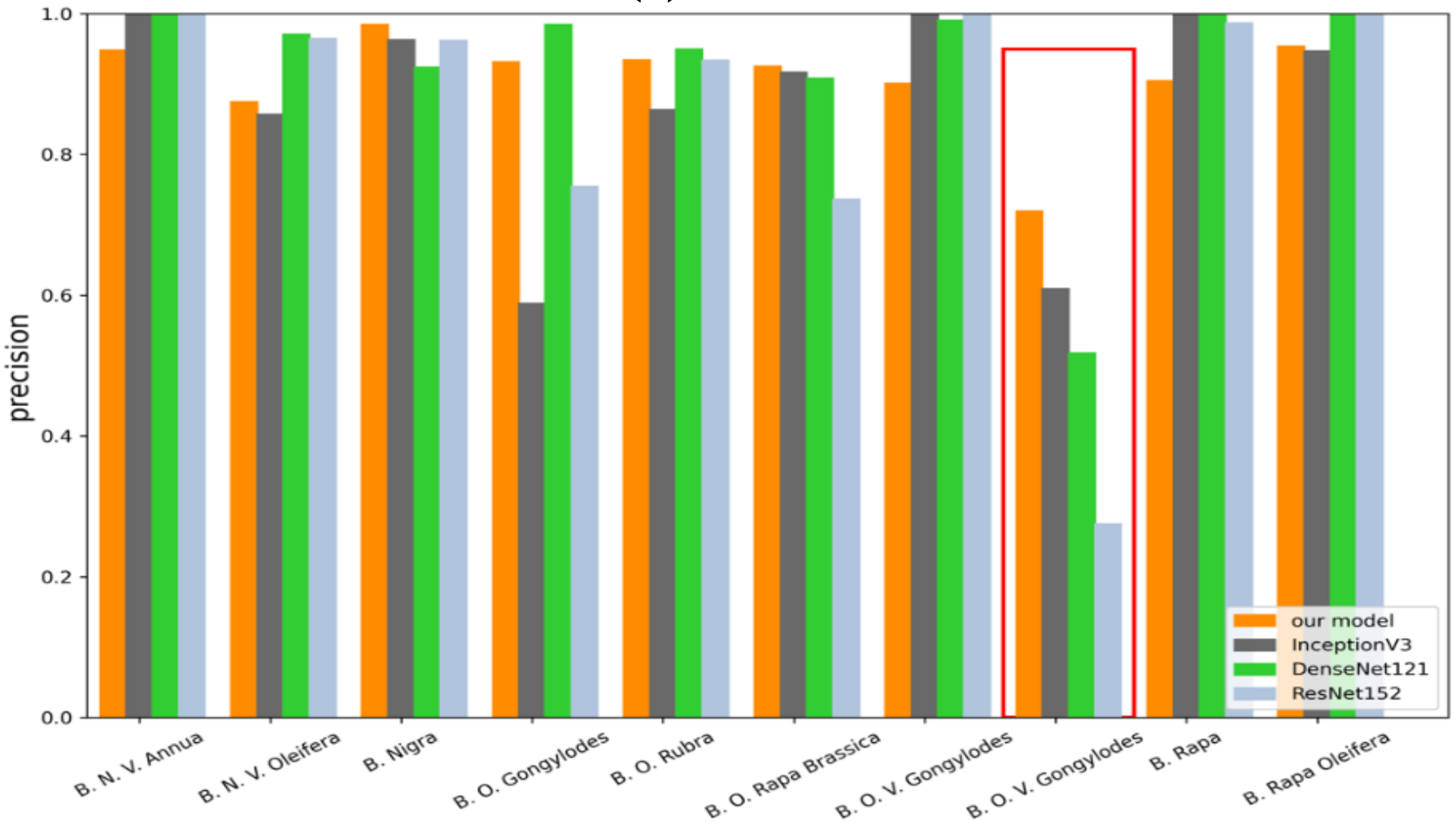

**(b) Precision**

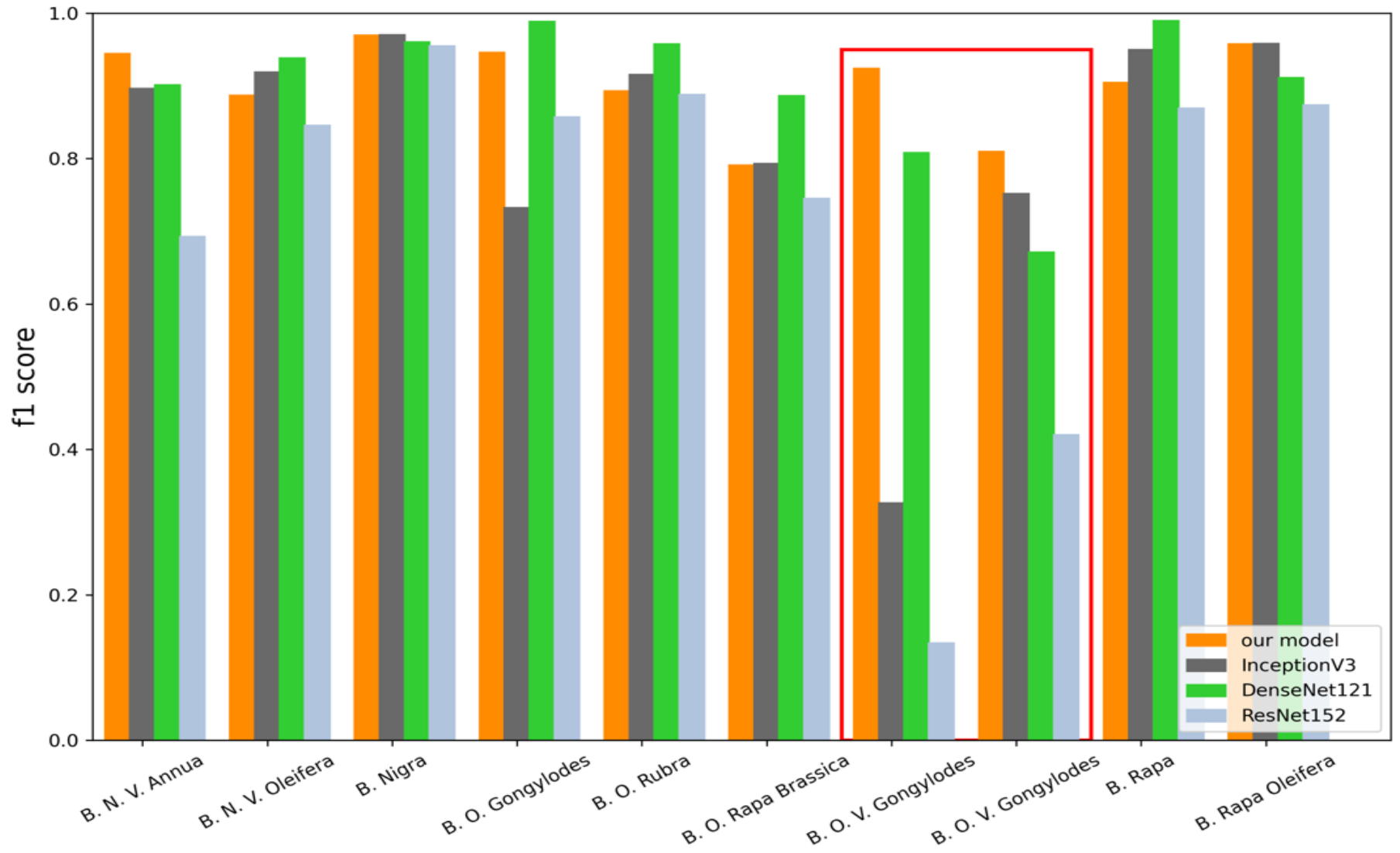

**(c) F1 score**.

Figure 8.31 A comparative analysis of classification performance using the following performance measures: a) recall, b) precision, and c) f1 score

According to Figure 8.31, until the classes were denoted by a red line, which led to poor differentiation, the classification accuracy of four different models for 10 Brassica seeds was controlled and convergent in all classes. Densnet121 reported the second-best performance, and our model's accuracy was over 90%. These results showed that the Densnet 121, Resnet 152, and Inception v3 models needed to be better adapted to these variations, possibly due to textural similarity. Until now, the results of our model classification were still highly positive and revealed the highest values attainable for each seed class. Thus, the experimental study showed that our model had strong generalization capability in the Brassica seed dataset compared to the pre-trained models' architectures with updated weights and fine-tuning.

To evaluate the robustness of our suggested approach, we utilized the Brassica dataset as a benchmark for training our models using high-level features. The outcomes showed that, on this dataset, our model has remarkable generalization performance. This outcome gave us confidence in our approach and suggested that the model could generalize well for other datasets, prompting us further to assess the performance in different scenarios.

The suggested model performs better in accuracy and performance when compared with various networks in the literature. A maximum variation in classification accuracy of more than 2% was also found. However, it demonstrates that creating a new model and establishing the network is feasible and practicable owing to its improved classification outcomes. The proposed network structure successfully combines depth and width in a natural way to create an optimal network model for image classification, which is advantageous for locating workable solutions for the most challenging classification tasks and will provide stable and reliable solutions for future applications in a variety of fields. Table 8.12 provides a comparison of our findings to those of previous research.

Table 8.12 A comparison of the proposed model with various state-of-the-art research

| Reference | Crop | Dataset | Method | Accuaracy |
|---|---|---|---|---|
| (Gulzar et al., 2021) | 14 kinds of seeds | 2733 images | VGG16 | 99% |
| (Keya et al., 2022) | 5 kinds of seeds | 1250 images | CNN | 87%-89% |
| (Salimi et al., 2020) | Sugar beet (5 kinds) | 2000 images | MSI | 82% |
| (Minah et al., 2021) | Brassica rapa | 1056 images | New AI models | 87.72% |
| (Dubey et al., 2021) | 3 types of wheat | Wheat dataset | CNN | 84% - 94% |
| Proposed approach | 10 Brassica classes | Our dataset 6065 images | Our model<br>Resnet152<br>Inceptionv3<br>DenseNet121 | 93%<br>73.34%<br>84.71%<br>90.03% |

To conclude the analysis of this work, this work was concerned with creating and recommending a novel CNN model for tasks involving the classification of Brassica seed images. Moreover, we assessed and compared the effectiveness of the suggested method to different pre-trained models, including Densent121, Inceptionv3, and Resnet152, revealing that our model could significantly increase the accuracy of CNNs in predicting expression values. The impact of the suggested architectures and training settings on performance improvement was also assessed using a variety of metrics. The outcomes of the suggested method demonstrated up to 93% accuracy for our model. While Inceptionv3 scored 84.71%, Resnet152 scored the lowest (73.34%), and Densnet121 reported 90.03%.

Moreover, we assessed the impact of our suggested architecture and training settings on performance improvement utilizing a variety of metrice including precision, F1 score, recall, and accuracy. Our suggested architecture and optimized training settings resulted in an important enhancement in performance over other exesting models, demonstrating increased precision, F1 score, and accuracy; this is a testament to the effectiveness of our approach and its ability to yield accurate results. Furthermore, our approach is also noteworthy for its scalability, as the performance improvements are achieved with minimal changes to the underlying architecture and settings; this allows our model to be easily integrated into existing systems without requiring extensive modifications.

In addition, a new Brassica dataset that did not exist will be created as part of this project to assess how well our CNN architectures perform on it. The data gathered will be used to analyze and compare the performance of different CNN architectures, such as Inceptionv3, Resnet152, and Densnet12. Finally, the results of this research can be used as a model for other visual object identification studies, which means that the practical study presented in this work can be readily applied to classify other seed images. The scalability of our approach will not only lead to better results. However, it will also pave the way for further developments in this field, providing researchers with a new benchmark dataset to test their model performance. Furthermore, the results of this research could be used to identify and develop different techniques to optimize object detection, leading to higher accuracy and more efficient networks.

## 8.4 Deep Multi-Scale Convolutional Neural Networks for Automated Classification of Multi-class Leaf Diseases

Deep learning has revolutionized the agricultural industry, offering tremendous potential for improving various aspects of crop management and production, offering potential benefits such as increased productivity, cost reduction, and improved sustainability through early disease detection, optimized crop yields, and precision agriculture. In the pursuit of advancing the field of plant health assessment through deep learning, this thesis not only focuses on the critical task of plant leaf disease detection but also incorporates seed classification as an integral component of our research. Recognizing that plant health encompasses various facets of a plant's lifecycle, we extend our investigation beyond leaves to seeds, as they play a pivotal role in plant reproduction and overall vitality. Moreover, this work on seed classification lays the groundwork for future model development, paving the way for a more comprehensive and holistic approach to plant health assessment. By integrating both leaf disease detection and seed classification, this research aims to provide a more nuanced and encompassing understanding of plant health, ultimately contributing to more effective solutions for the agricultural industry. In this research, we suggest a novel approach that utilizes a Deep Multi-Scale Convolutional Neural Network (DMCNN) to automatically classify multi-class leaf diseases in tomatoes. The DMCNN architecture incorporates parallel streams of CNNs at different scales, merged at the end to generate a single output. To improve model performance, data augmentation techniques are applied during preprocessing of tomato leaf images before feeding them into the DMCNN model for disease classification. Our proposed approach is thoroughly assessed on a dataset of tomato plant images with 10 distinct disease classes and compared to cutting-edge models. The findings demonstrate that our proposed DMCNN model excels in accuracy, precision, recall, and F1 score compared to other models. Remarkably, our model reaches an exceptional accuracy rate of 99.1%, surpassing the accuracy of all other models evaluated on the identical dataset. This investigation highlights the immense potential of deep learning techniques for automating disease classification in agriculture, providing invaluable insights for early disease detection and prevention of crop loss.

This work provides a deep learning-based method for automated classification of multi-class leaf diseases in tomatoes using DMCNN. The significance of this proposed study stems from the fact that tomato plants are affected by several diseases that can significantly reduce crop yield, quality, and economic value. Traditional methods for detecting and classifying these diseases are often time-consuming and require expert knowledge, which can be a limiting factor in large-scale crop production. Recent developments in deep learning methods, particularly CNN, have demonstrated impressive achievements in various computer vision tasks, such as object recognition and classification. However, most existing studies in plant disease detection revolves around single-channel and same-resolution images, which may not capture the complete information required for accurate disease detection and classification.

The suggested DMCNN framework is designed to overcome these limitations of existing tomato leaf disease detection and classification methods, this framework aims to improve the accuracy and efficiency of disease classification by leveraging multiple channels of information. Specifically, the study focuses on developing a DMCNN architecture capable of accurately classifying 10 different categories of tomato diseases. To assess the performance of the proposed framework, a publicly available dataset comprising 11,000 tomato plant images, which offer diverse channels and scales of information, is utilized. The study also aims to optimize the model's hyperparameters to further enhance its performance. Additionally, the effectiveness of the proposed approach is compared with other cutting-edge methods to evaluate its implementation in the field of plant disease classification.

The study presented in this paper makes five key contributions:

1. It proposes a new deep multi-scale convolutional neural network (DMCNN) architecture that utilizes multiple information channels to classify tomato plant diseases accurately.
2. The DMCNN architecture consists of parallel streams of CNNs at different scales, which are merged at the end to generate a single output.
3. The performance of the suggested framework is assessed using a diverse and extensive dataset of tomato leaf images, and its performance is compared against other advanced techniques.
4. The study provides insights into the feature importance of the suggest model, which can aid understand the underlying mechanisms of disease classification.
5. Extensive analysis is performed to exhibit the proposed model's robustness towards various factors, ultimately enhancing its reliability and generalizability.

The results of this research show the potential impact of deep learning-based strategies on revolutionizing the field of plant pathology and crop management. The suggested approach holds potential for various applications in precision agriculture and sustainable crop management, which could lead to improved crop productivity and food security. By leveraging the benefits of deep learning algorithms and multiscale imaging, the proposed approach offers improved accuracy and efficiency in disease classification, which can help mitigate crop losses and reduce the economic impact of plant diseases. The study also provides insights into the feature importance of the proposed model, which can aid in understanding the underlying mechanisms of disease classification. Overall, this study highlights the significant contributions of DL techniques in addressing the challenges of leaf disease detection and classification, with potential applications in various fields of agriculture and crop management.

### 8.4.1 Research materials and methods

This section of this work offers a detailed explanation of the suggested model and the datasets employed in this work. It outlines the various steps taken to enhance the efficacy of the suggested approach and presents the architecture of the DMCNN model.

#### 8.4.1.1 Dataset and Pre-Processing

##### 8.4.1.1.1 Dataset description

In our study, we used a diverse and extensive dataset of 11,000 images from 10 different categories, encompassing healthy tomatoes and various diseases such as early blight, bacterial spots, leaf molds, late blight, and mosaic viruses. The images were sourced from the popular Kaggle dataset, which is widely used in deep-learning research. To ensure an even distribution of the 10 classes, we carefully curated the dataset with 1100 images per class, providing a reliable representation of the different tomato leaf diseases.

To develop and analyze our suggested DMCNN for the automatic detection of multi-class leaf diseases in tomatoes, we divided the dataset into 10:90 testing and training sets. To prevent overfitting and optimize the model, we further partitioned 10% of the training set as validation; this allowed us to closely monitor the training process and fine-tune the model's hyperparameters, ultimately achieving optimal performance.

The dataset was meticulously curated to include image data of tomato plant leaves with multiple diseases and healthy leaves captured under various lighting conditions and diverse orientations. This dataset diversity made it an excellent choice for training and evaluating Multi-Scale Convolutional Neural Networks (MSCNNs), allowing us to to catch features at multiple scales and handle input image variations. To present a visual representation of the data available for training and testing our DL model, we included the Figure 8.32 in our study. This collection displays image data of tomato diseases from the dataset, each belonging to a specific disease category. Analyzing these images helped us gain insights into the characteristics and features of different diseases, which, in turn, aided us in developing a more precise and efficient model.

The dataset used in this study offers a dependable portrayal of the various types of tomato leaf diseases necessary for precise training and evaluation of DL models. The inclusion of diverse samples in the dataset, along with the careful selection of training, validation, and testing subsets, facilitated the development and precise assessment of the performance of the proposed DMCNN architecture.

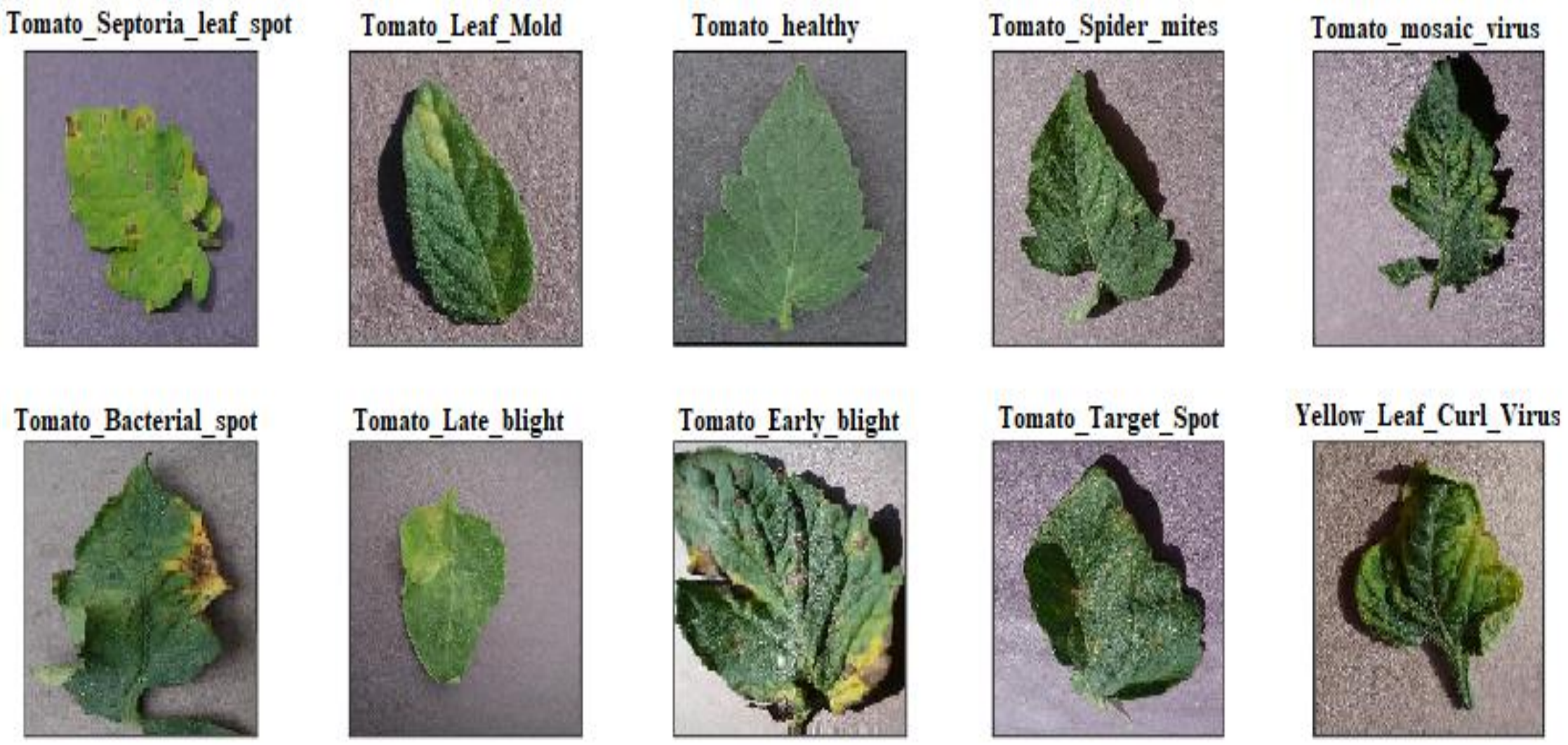

Figure 8.32 Class-wise image subsets

#### 8.4.1.1.2 Dataset preprocessing

The success of DLmodels in accurately categorizing plant diseases heavily relies on the quality of the dataset used to generate the training data. As a result, data preprocessing is a critical component of the deep learning pipeline. It includes cleaning, transforming, and formatting the data to ensure optimal learning by the model. Techniques such as filtering and image resizing used in data preprocessing can substantially improve the model's performance in the field of plant disease classification.

In this study, several preprocessing steps were applied to the dataset to assure that the images were in a suitable format for analysis. First, we resized the images to a fixed resolution. Additionally, we utilized data augmentation methods like rotation, zooming, and flipping to artificially increase the overall amount of data and reduce overfitting. These techniques helped us create a more robust dataset, which is essential to deep learning model success.

To train a DMCNN model on this dataset, we utilized images at various scales. Specifically, we applied a multi-scale approach where each image was resized to multiple scales, and the network processed them separately. This allowed the network to detect objects at different scales and capture more fine-grained details, thus improving the model's accuracy in disease classification. Table 13 provides details of the specific techniques and scales employed in the multi-scale approach for the tomato leaf dataset.

Table 8.13 Multi-scale approach for image resizing

| Scale | Technique |
|---|---|
| 224x224 | Resizing to a fixed scale |
| 256x256 | Resizing to a fixed scale |
| 128x128 | Resizing to a fixed scale |

The data preprocessing steps conducted in this study were crucial in ensuring that the dataset was suitable for training an accurate and efficient deep learning algorithm for identifying plant leaf diseases. Moreover, utilizing a multi-scale approach with various preprocessing techniques could improve the model's performance and result in excellent disease classification outcomes.

The training samples were subjected to several transformations, including horizontal flipping, applied to the training samples, while rotation was carried out within 20 degrees. Furthermore, zooming was done within a range of 0.2, and shifting was done within a 0.2 range in both width and height. These data augmentation methods aid in broadening the diversity of the training set and preventing overfitting. Table 8,14 presents the data augmentation techniques utilized in the project and their corresponding parameters/ranges.

The implementation of DL models in plant disease identification is heavily influenced by data size and data composition, as well as the effectiveness of the preprocessing techniques applied. In this study, after preprocessing techniques, the dataset contained 12500 images, with 1250 images per class, representing 10 different classes of plant diseases. The balanced distribution of samples across classes ensures that the model does not exhibit bias towards any particular class, which can affect its accuracy.

Table 8.13 Data augmentation techniques and their parameters

| Data Augmentation Technique | Parameter/Range |
|---|---|
| Horizontal flipping | Yes |
| Rotation | 20 degrees |
| Zooming | 0.2 range |
| Shifting | 0.2 range in width and height |

A large dataset provides more examples for the model to learn from, and appropriate preprocessing techniques, such as image resizing and data augmentation, improve the model's ability to capture fine-grained details and avoid overfitting. These techniques helped to create a more robust dataset, which was critical for training an accurate and efficient deep learning method to treat and identify plant diseases.

Overall, data size and composition, along with the preprocessing methods used, are essential factors in the efficiency of the DL model. By accurately identifying and classifying plant leaf diseases, farmers can take early steps to prevent crop loss and increase yield, making this research an important contribution to the agricultural industry.

### 8.4.2 Implementation (Experiment Setup)

This section provides information on how our proposed model was implemented for automated classification of multi-class diseases in tomatoes, using Deep Multi-scale Convolutional Neural Networks (DMCNN).

Our model architecture includes multi-scale convolutional filters, batch normalization, and dropout layers. We used the PyTorch deep learning framework on a high-performance machine with an NVIDIA GeForce RTX 3090 GPU and 64GB RAM. In addition, python libraries like NumPy, Pandas, and Matplotlib were utilized for visualization and manipulation of data. During the training process, our model utilized a batch size of 64 and a learning rate of 0. 001. Moreover, To optimize the learning process, we employed a cosine annealing learning rate scheduler; the Adam optimizer was used with a weight decay rate of 0.0001. The model underwent training for a total of 100 epochs. Early stopping based on validation accuracy was implemented to prevent overfitting. To enhance the reliability of our findings, we conducted all experiments using a five-fold cross-validation approach.

To assess the perormance of our model, we employed multiple evaluation metrics including F1-score, recall, precision, and accuracy. Additionally, we visualized the confusion matrix to gain insights into the classification performance of our model. Detailed information regarding the model setup is provided in Table 8.14.

Table 8.14 Model Details

| Option | Value | Details |
|---|---|---|
| Implementation Details | Deep Learning Framework | PyTorch |
| | RAM | 64GB |
| | GPU | NVIDIA GeForce RTX 3090 |
| | Libraries | NumPy, Pandas, Matplotlib |
| Model Training Details | Batch Size | 64 |
| | Learning Rate | 0.001 |
| | Learning Rate Scheduler | Cosine Annealing |
| | Optimizer | Adam |
| | Weight Decay | 0.0001 |
| | Epochs | 100 |
| | Early Stopping | Patience of 100 epochs based on validation accuracy |
| | Cross-validation | 5-fold |
| Evaluation Metrics | Accuracy | |
| | Precision | |
| | Recall | |
| | F1-score | |
| | Confusion Matrix | Visualization |

The ability of our proposed approach to handle multi-class classification of leaf diseases, coupled with its scalability and adaptability, makes it a valuable tool in the field of precision agriculture. By providing accurate and automated methods for disease classification, our approach can contribute to sustainable crop management and improve crop yield. Furthermore, the potential applications of our approach are wider than the tomato crop alone. The deep multi-scale convolutional neural network architecture used in our model can be applied to other crops and agricultural settings, leading to more precise and efficient methods of disease classification. Therefore, our study highlights the importance of leveraging the power of deep learning in agriculture and demonstrates how it can contribute to sustainable crop management. As the field of precision agriculture continues to grow and evolve, we believe that our approach can significantly improve crop yield and reduce crop disease's economic and environmental impact.

### 8.4.3 Proposed model architecture

The proposed deep multi-scale CNN model is a powerful tool for automated disease classification in tomato crops, leveraging deep learning power and multi-scale features to accurately detect different diseases.

The suggested deep multi-scale CNN model architecture utilizes a multi-scale feature extraction module, followed by a global feature fusion module, for disease classification in tomato leaf images. The multi-scale feature extraction module comprises convolutional layers with varying filter sizes that enable the extraction and separation of features across a range of scales. These convolutional layers' outputs are combined and fed through a pooling layer to decrease the features' spatial dimensions. The output feature map is subsequently fed into the global feature fusion module. The global feature fusion module integrates the multi-scale features into a unified feature representation using fully connected layers. The model's disease classification task is achieved by passing the output of the global feature fusion module through a softmax layer.

To train our model, a large and diverse dataset of tomato leaf images captured by a camera was used. The dataset was preprocessed and augmented to prevent overfitting and ensure a balanced class distribution. A learning rate of 0.001 was utilized with the Adam optimizer to train the model for 100 epochs. Finally, the PyTorch framework was used to implement the model. Figure 8.33 provides a visualization of the proposed deep multi-scale CNN model architecture.

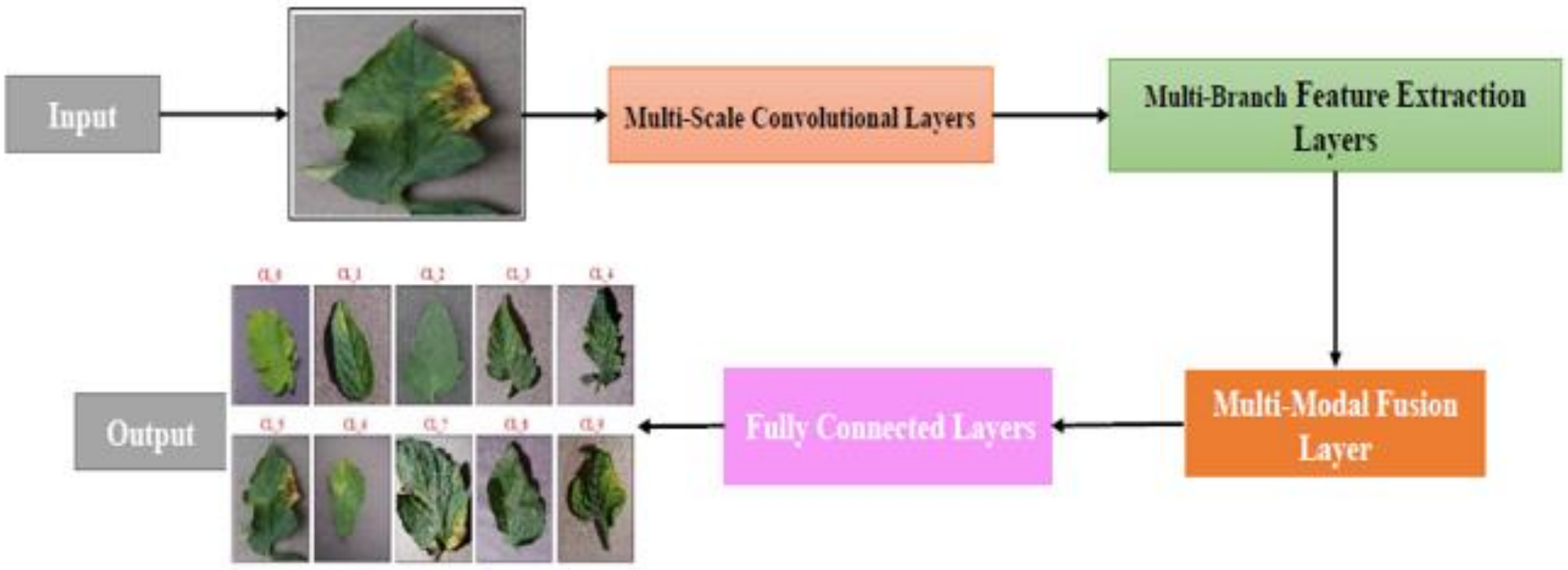

Figure 8.33 Proposed deep multi-scale CNN model architecture

The proposed deep multi-scale CNN model architecture is designed to classify tomato plant leaf diseases by incorporating various elements of the image into a single model. By employing multi-scale convolutional layers, the model can extract features at different scales, enabling the identification of both local and global information within the tomato image.

The output of the multi-scale convolutional layers is then passed through multi-branch feature extraction layers that learn representations specific to different spectral channels, including texture, shape, and color. Next, these features are fused together by a multi-modal fusion layer, which combines the information from each branch to form a unified feature representation. Finally, Fully connected layers are utilized to classify the fused features. The model is trained by reducing the cross-entropy loss between the predicted probabilities and the actual labels, which enables it to produce a probability distribution for each class of tomato plant leaf disease. The suggested deep multi-scale CNN model architecture is designed to leverage the benefits of large dataset images and capture different aspects of the tomato plant leaf image for precise classification. Figure 8.34 presents a detailed architecture of the suggested deep multi-scale CNN model.

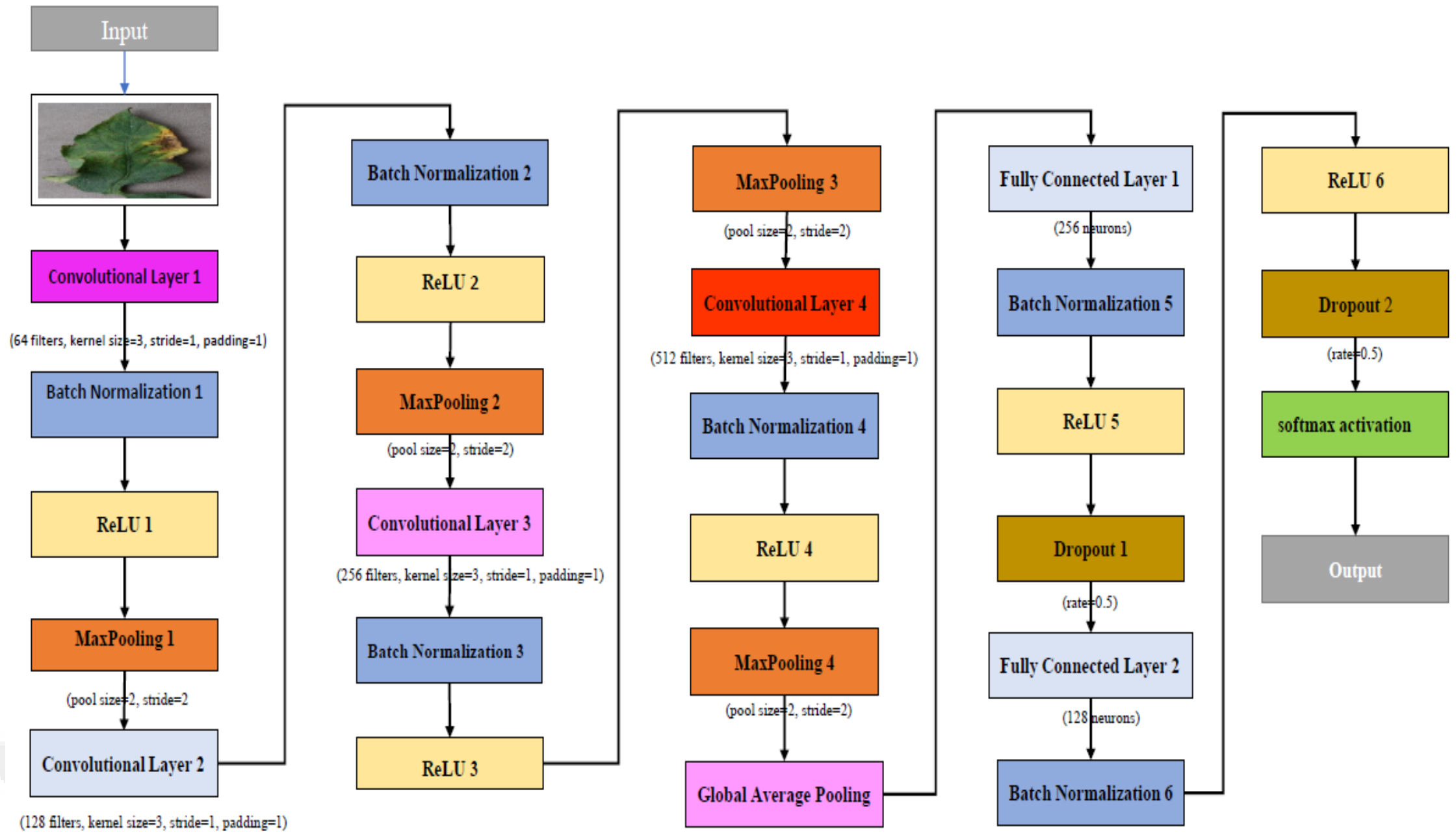

Figure 8.34 Flowchart of the Proposed deep Multi-Scale CNN Architecture for multi-class Tomato Leaf Disease classification

The proposed model architecture for multi-scale image classification tasks with a tomato picture as input and the predicted class label as output is represented schematically. In this schema, the terms "Batch Normalization," "ReLU," and "MaxPooling" refer to the batch normalization layers, rectified linear activation functions, and maximum pooling layers, respectively. Then, "Fully Connected Layer" refers to a dense layer with a specified number of neurons, and "Dropout" refers to dropout layers with a specified dropout rate.

The architecture comprises four convolutional layers with increasingly smaller filter sizes, following each convolutional layer, batch normalization and ReLU activation functions are applied, followed by max pooling. To address overfitting and reduce the complexity of the model, a global average pooling layer is employed after the fourth convolutional layer, which helps to reduce the dimensionality of the feature maps. Following the global average pooling layer, the architecture incorporates two fully connected layers with 256 and 128 neurons correspondingly. Batch normalization and ReLU activation are employed after each fully connected layer to enhance the model's ability to generalize and alleviate overfitting. To further mitigate overfitting, two dropout layers with a dropout rate of 0.5 are introduced before the output layer. The final output is obtained by applying a softmax activation function, which generates a probability distribution for the various classes of tomato plant leaf disease.

The suggested architecture is designed for multi-scale image classification tasks with a tomato picture as input and the predicted class label as output. Figure 8.35 provides an overview illustration of the proposed model architecture Multi-Branch Network and Figure 8.36 provides a detailed description of its implementation using Multi-Branch Network for Tomato leaf disease classification.

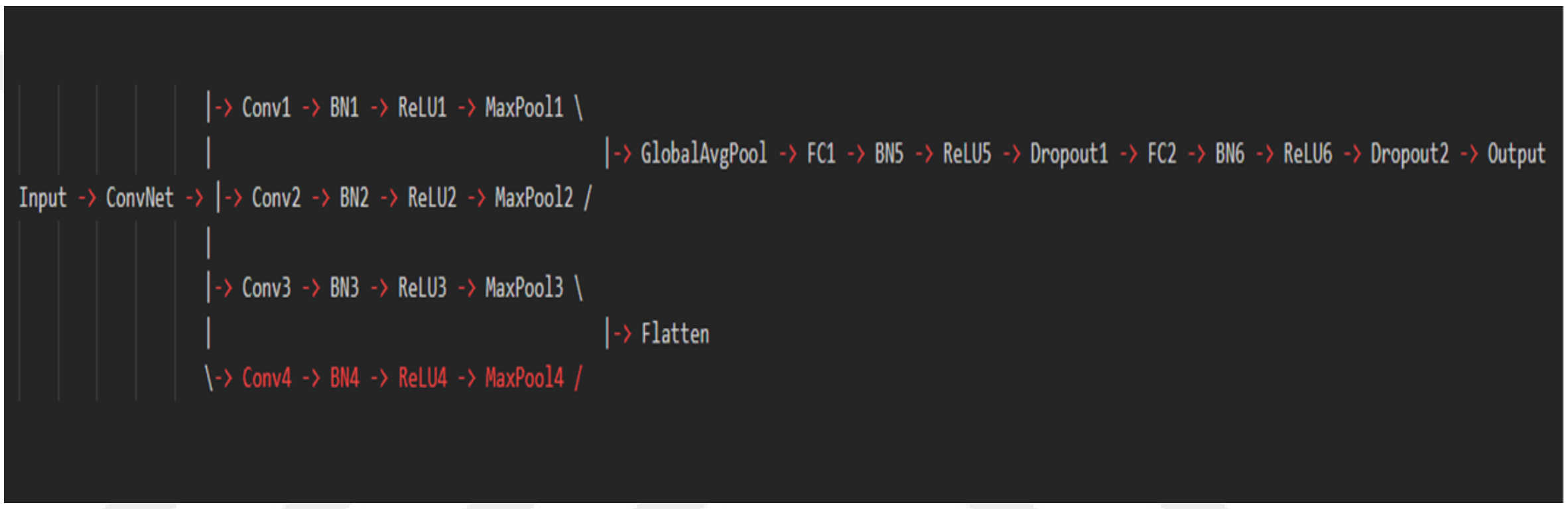

Figure 8.35 An overview representation of the proposed architecture employing a Multi-Branch Convolutional Neural Network

The proposed deep Multi-Scale CNN Architecture, utilizing a Multi-Branch CNN, is depicted in Figure 8.35 and Figure 8.36. It has been specifically designed for tomato disease classification. The network encompasses four convolutional layers, typically accompanied by batch normalization, ReLU activation, and max pooling. Each convolutional layer's output is routed through a distinct branch, with the first branch terminating in global average pooling, then a fully connected layer and dropout. The remaining three branches' outputs are pooled using max pooling, concatenated, and then passed through fully connected layers, batch normalization, Rectified linear activation, dropout, and a softmax output layer.

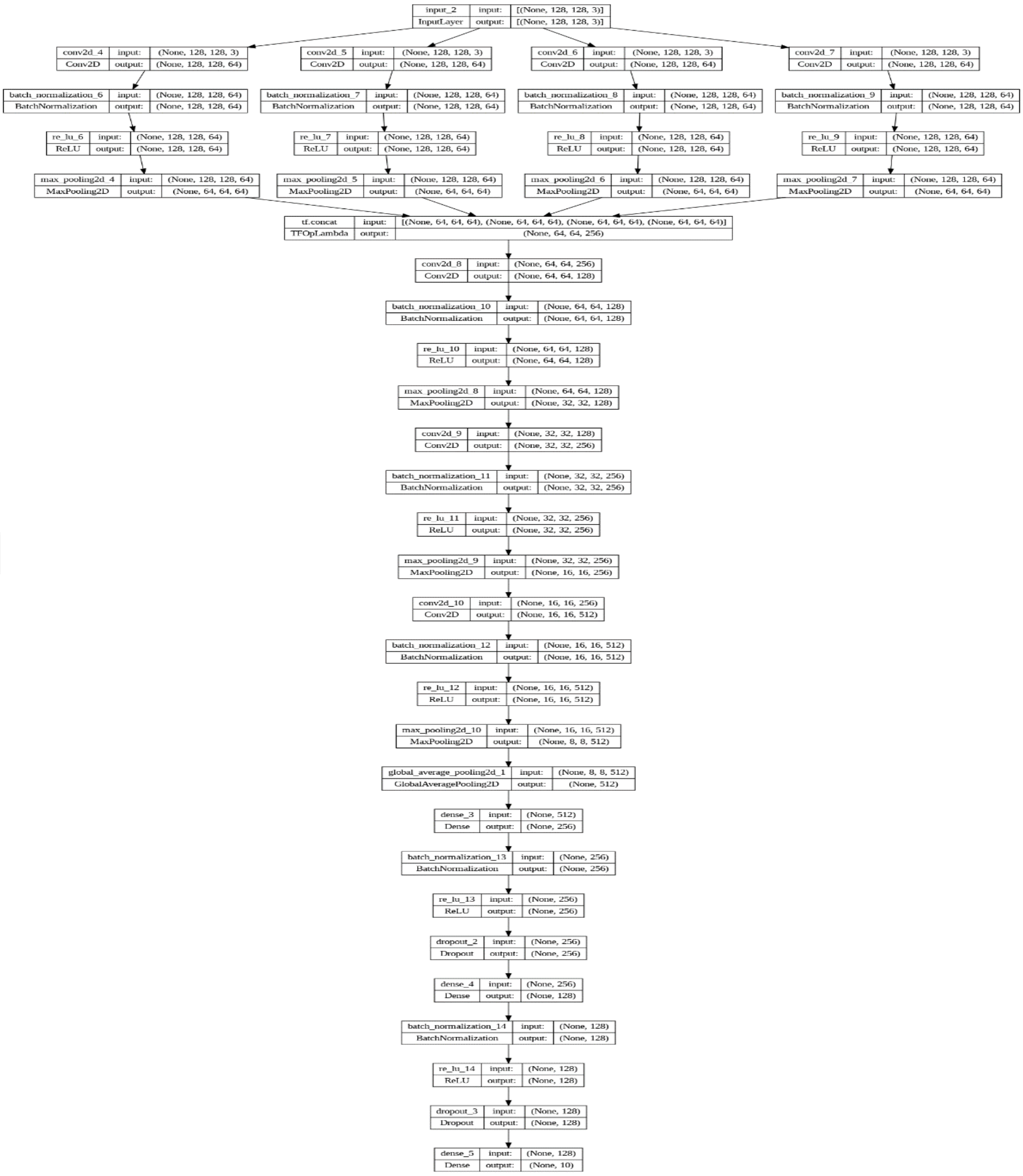

Figure 8.36 A detailed visual depiction of the proposed deep Multi-Scale CNN Architecture using Multi-Branch CNN

This architecture is an innovative approach to image classification, and its effectiveness has been demonstrated. By using a multi-branch CNN, more detailed information can be extracted from each layer of the input image, resulting in more accurate and reliable classification of tomato diseases. Batch normalization and ReLU activation are critical components of this architecture. Batch normalization normalizes the input to each layer, increasing the model's accuracy rate. At the same time, ReLU activation introduces non-linearity, allowing the network better to capture the complex relationships between various input image features. Dropout layers are also important for preventing overfitting, a common problem in machine learning. By randomly dropping out units during training, dropout layers force the model to learn more robust and generalizable visualizations of the input data, Enhancing its performance on unseen data.

The proposed deep multi-scale architecture is an advanced approach to classifying tomato diseases. Its use of a multi-branch CNN, batch normalization, ReLU activation, and dropout layers makes it a highly effective tool for this task. This architecture contributes to the development of more accurate and automated methods for disease classification in agriculture.

### 8.4.4 Pre-trained models' architectures

In this work, we conducted a comprehensive comparative valuation of our proposed deep Multi-Scale CNN model's performance with several commonly used pre-trained models for image classification tasks, including AlexNet, VGG16, InceptionV3, and ResNet50. This analysis was crucial for evaluating theeffectiveness of our proposed model and understanding the strengths and weaknesses of each model. The tuning details of different pre-trained models on the tomato leaf dataset are presented in Table 8.15.

Table 8.15 Tuning details of different pre-trained models on Tomato Leaf dataset

| Model Name | VGG16 | AlexNet | InceptionV3 | ResNet50 |
|---|---|---|---|---|
| Total Layers | 16 | 8 | 22 | 50 |
| Max Pool Layers | 5 | 3 | 3 | 1 |
| Filter size | 3 | - | 1x1, 3x3, 5x5 | 3x3 |
| Stride | 2x2 | - | 2x2 | 2x2 |
| Dense Layers | 3 | 2 | - | 3 |
| Dropout Layers | 2 | 2 | - | 2 |
| Flatten Layers | 1 | 1 | - | 1 |

As shown, each pre-trained model has a unique architecture with varying numbers of layers and different configurations of pooling, convolution, and dense layers. These differences can affect the model's performance and efficiency for a given task, which emphasizes the importance of comparing the efficacy of our suggested DMCNN model with these pre-trained models on the Tomato dataset.

Through our comparative analysis, we can determine whether our proposed model outperforms or falls short of cutting-edge models. The results of this comparison will provide valuable insights that will help us refine and optimize our model further. A detailed discussion and analysis of the performance comparison will be presented in the next sections of this study.

### 8.4.5 Experimental results

This section presents the experimental methodology employed to evaluate the efficacy of the proposed approach on the dataset consisting of tomato plant leaf images. Detailed results of these experiments are provided in section 4.1, where a comprehensive analysis of the findings is presented. Furthermore, subsections 4.1.1 and 4.1.2 delve into specific aspects of the results, providing a detailed and insightful interpretation of the outcomes obtained.

#### 8.4.5.1 Implementing the proposed model architecture

In this section, we have explored the application of DMCNNs for automating the classification of leaf diseases in tomatoes. To assess the efficacy of our suggested approach, we utilized a dataset comprising 125,000 images across 10 distinct categories, including healthy leaves and nine types of diseases. The images were evenly distributed among the 10 classes, each containing 1,250 images.

The results of our evaluation provided valuable insights into the model's performance, identified areas for improvement and refined the model further. Our proposed DMCNNs model demonstrated exceptional accuracy, as evident from its training and validation accuracy. The average training and validation accuracies were found to be 99.24% and 99.15%, respectively, which indicates the models' robust performance. Moreover, the training and validation losses were observed to be 0.2918 and 0.3699, respectively, further emphasizing the accuracy of the models. To illustrate these results, Figure 8.37 and Table 8.16 display the accuracy and loss of the proposed architecture. Therefore, the performance evaluation demonstrated that our proposed DMCNNs model is highly effective in achieving its intended goals and can be considered a robust solution.

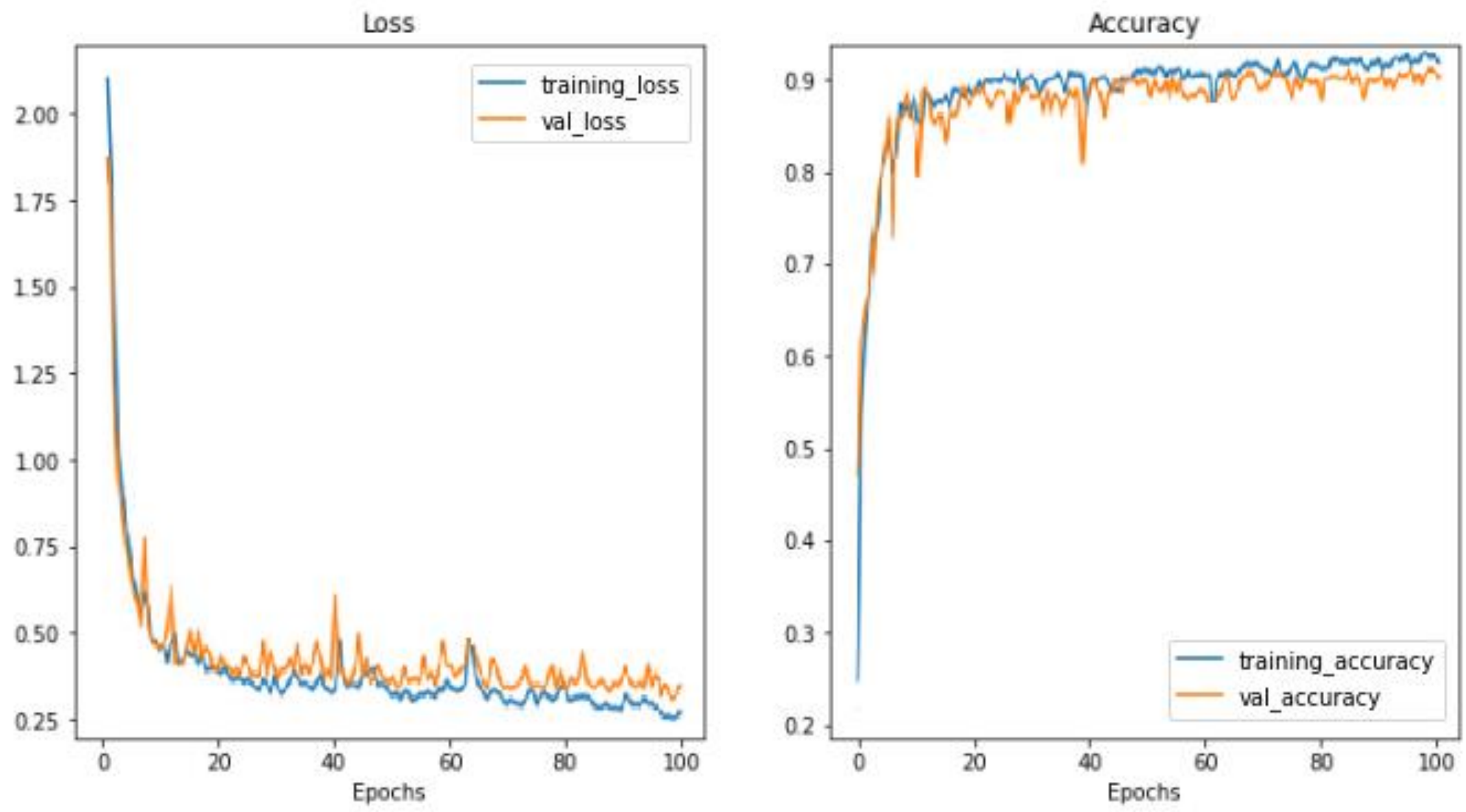

Figure 8.37 Analysis of epoch vs. Accuracy/Loss plots of the proposed model on train and validation datasets

Table 8.16 Observed training and validation accuracies and losses

| **Metrics** | **Training** | **Validation** |
|---|---|---|
| Accuracy | 99.24% | 99.15% |
| Loss | 0.2918 | 0.3699 |

The observed training and validation accuracies and losses for the proposed models provide strong evidence of their accuracy and effectiveness. The average training accuracy of 99.24% indicates that the model performed exceptionally well on a training set, correctly classifying most images. This high training accuracy suggests that the model could learn and capture the relevant patterns and features in the dataset, leading to its robust performance.

The average validation accuracy of 99.15% confirms the model's accuracy on data and its ability to generalize well. This is a crucial aspect of any reliable model since it ensures that it can accurately classify new data that it has not seen before. Therefore, the high validation accuracy demonstrates the model's robustness and reliability, making it an effective solution for classifying tomato leaf diseases. Moreover, the training and validation losses of 0.2918 and 0.3699, respectively, highlight the model's ability to effectively learn the features and patterns within the dataset, resulting in minimized loss values. This indicates that the model has successfully captured important information from the data and can accurately classify the images. The close proximity of the training and validation loss values also suggests that the model demonstrates a strong generalization capability, showcasing its ability to effectively classify diverse instances beyond the training data. The observed training and validation accuracies and losses collectively provide compelling evidence of the effectiveness of the proposed model in classifying tomato leaf diseases with high precision and generalization ability.

The accuracy and loss graphics illustrated in Figure 8.37 visually represent the proposed architecture's performance, emphasizing the model's stability and consistency throughout the training process. The accuracy and loss plots for both the training and validation datasets demonstrate consistent and stable performance of the proposed model throughout the training process. Furthermore, the absence of significant spikes or dips in the curves indicates that the model successfully learned the features and patterns of the dataset without encountering overfitting or underfitting issues. Therefore, the accuracy and loss curves demonstrate that the model performed well throughout the training process. Consequently, the high accuracy rates and low losses observed in the proposed model and the consistency shown in the accuracy and loss graphics provide strong evidence of the performance and accuracy of the suggested approach in classifying tomato plant leaf diseases.

To optimize the model's performance, careful selection of the batch size and number of epochs is critical. In an experimental study described in Figure 8.38, we tested four different batch sizes (8, 16, 32, and 64) to measure the training time for each epoch and testing accuracy. The results showed that as the batch size increased, the training time for each epoch decreased while the testing accuracy continued to increase. Figures 8.38a and b clearly demonstrate this relationship, highlighting the benefits of using larger batch sizes during model training. This finding suggests that increasing the batch size can lead to faster convergence and better testing accuracy, but it is important to balance this against the risk of overfitting. Ultimately, selecting the appropriate batch size and a number of epochs requires careful experimentation and consideration of the specific model architecture and dataset.

By training the model, we found that a batch size of 64 was the most effective, generating the highest testing accuracy while minimizing training time. However, further analysis of testing accuracy at different model training epochs, as shown in Figure 8.39, revealed that accuracy gradually increased up to 100 epochs. As such, a batch size of 64 and 100 training epochs could be optimal for this particular model. Therefore, this study provides valuable insights into selecting optimal parameters for training neural networks.

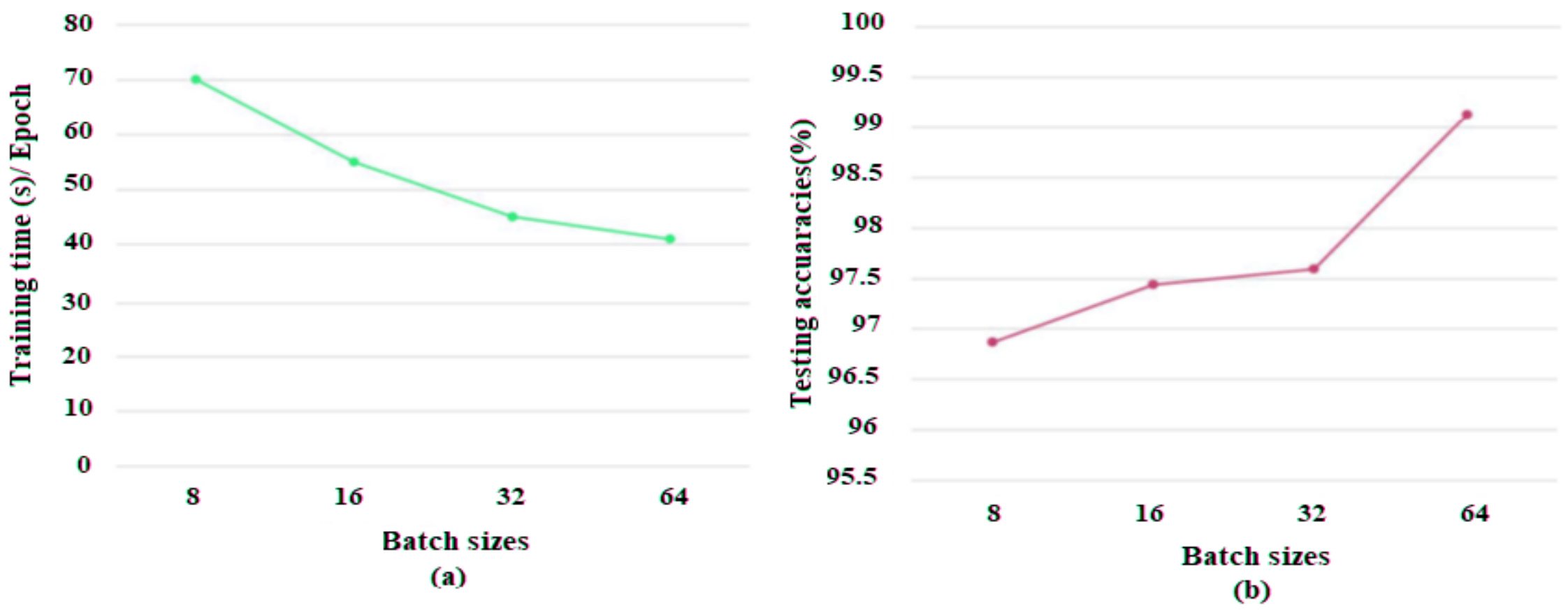

Figure 8.38 Exploring batch sizes impact on system performance: (a) Analyzing Training Time and Epochs vs Batch Size, and (b) Assessing Testing accuracy for Varying Batch Sizes

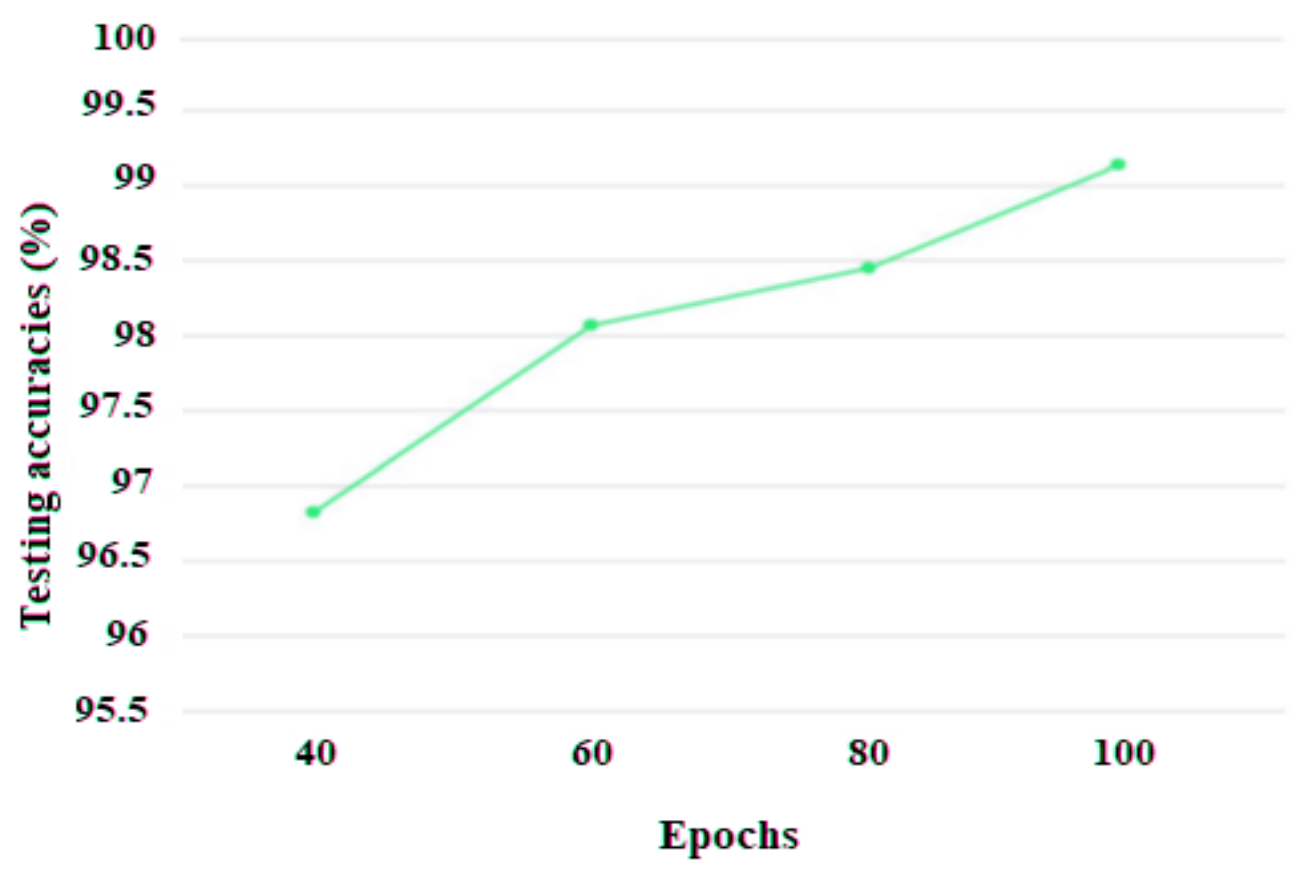

Figure 8.39 The Influence of epochs on testing accuracies: An analysis of the relationship between model training iterations and accuracy

By experimenting with different batch sizes and epochs, we can identify the best values for a particular model and dataset, significantly improving the model's performance. Moreover, it is essential to recall that the optimal parameters may vary depending on the specific circumstances, and generalizing the findings of one study to other models and datasets may not be appropriate. Therefore, it is recommended to perform similar experiments on other architectures with various datasets to identify the best parameters for each case.

#### 8.4.5.2 Performance evaluation of the developed model

In this research project, we utilized the confusion matrix to thoroughly understand the accuracy and potential sources of confusion for our classification model while making predictions. The confusion matrix included four metrics that helped us measure the accuracy of classifications and forecast the behavior of each predictor and target attribute pair for a given class value. Using the confusion matrix, we could effectively evaluate the efficacy of our DMCNN model, identifying its strengths and weaknesses in detecting tomato leaf diseases.

Our study analyzed the performance of the classifier that distinguished ten classes of tomato leaf diseases using four metrics, namely true negatives (TN), false positives (FP), true positives (TP), and false negatives (FN). TP and TN indicated the correct identification of tomato leaf diseases, while FP and FN indicated incorrect identification. We illustrated the confusion matrices for the models in Figure 8.40, clearly visualizing the true class values in the sample data and the class values predicted by the CNN classifier. The experimental results demonstrate that our proposed DMCNN achieved an impressive accuracy of 99.1%, showcasing the effectiveness and high performance of the model. Moreover, using the confusion matrix was crucial in accurately classifying tomato leaf diseases using our DMCNN model. It enabled us to identify areas for improvement and optimize the model for better performance in real-world applications.

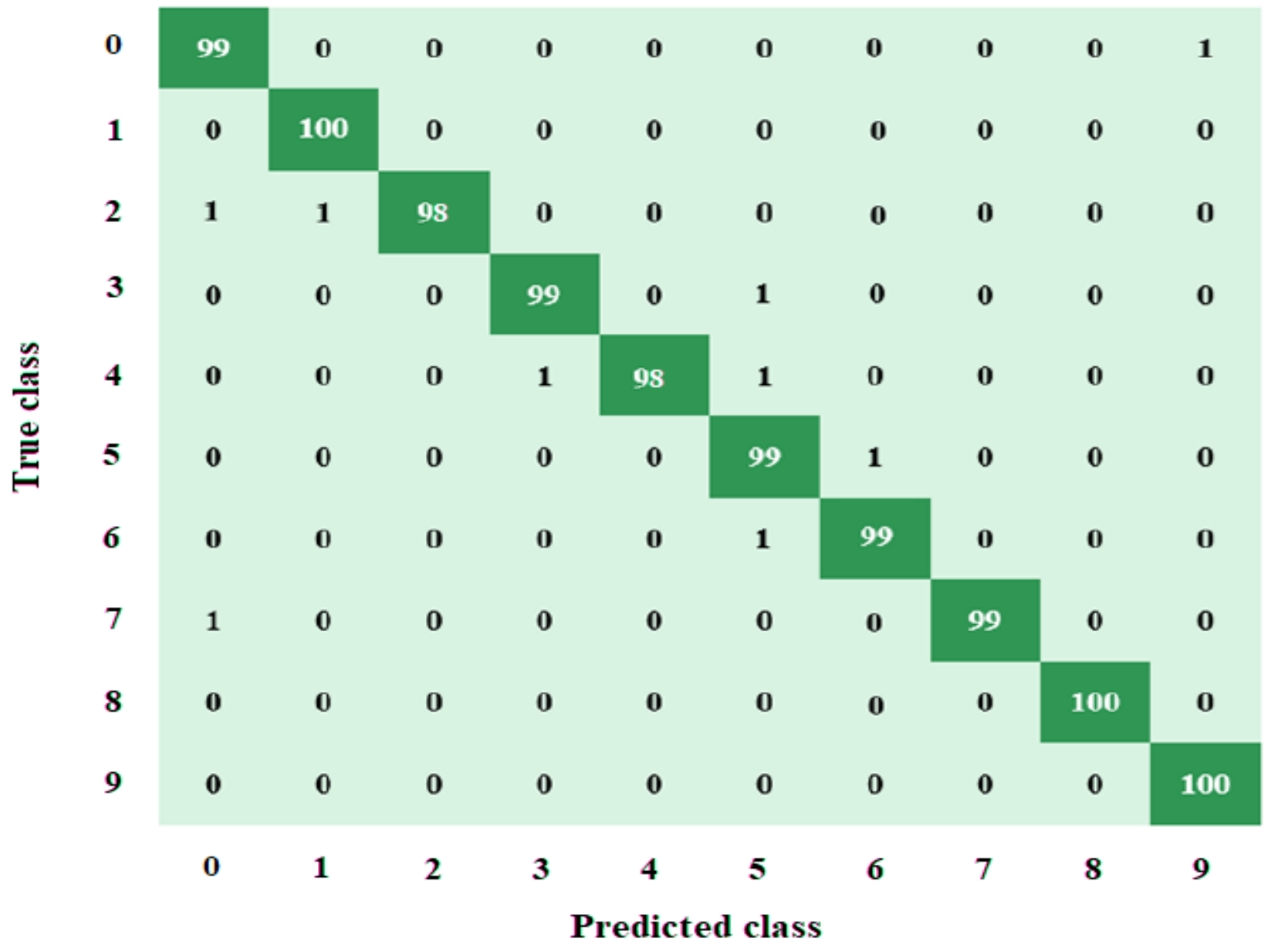

Figure 8.40 The Influence Confusion matrix for detection of tomato leaf diseases. ‘0’, ‘1’, ‘2’, ‘3’, ‘4’, ‘5’, ‘6’, ‘5’, ‘8’, and ‘9’ represent bacterial spot, leaf mold, late blight, early blight, spider mite, Septoria leaf spot, mosaic virus, target spot, yellowmcurl virus, and healthy leaves, respectively

A confusion matrix is an effective and practical method for evaluating how well a classification model is performing. In our research, we utilized the confusion matrix as a valuable tool to assess the performance of our DMCNN in classifying 10 distinct types of tomato leaf diseases. Our model demonstrated an impressive accuracy rate of 99.1% with very few misclassifications. This high accuracy indicates that the DMCNN is a robust and dependable tool for identifying tomato leaf diseases. Furthermore, the confusion matrix provided detailed information on true negatives, false positives, true positives, and false negatives. This helped us gain insights into the strengths and weaknesses of our model, allowing us to refine it further and enhance its performance. Using the confusion matrix was critical in evaluating the accuracy of our model and understanding how we can optimize it to achieve even better results. Overall, the confusion matrix is a practical and effective tool for evaluating how well a classification model performs. Our research provided us with valuable insights into the accuracy and potential areas of improvement of our proposed DMCNN model for classifying tomato leaf diseases.

In addition to overall accuracy, we conducted a thorough evaluation of our model by assessing accuracy, F1 score, recall, and precision for each individual class. These metrics are crucial as they offer a more comprehensive understanding of the model's performance across different classes. Precision measures the proportion of true positive predictions among all positive predictions, while recall calculates the proportion of true positive predictions among all actual positive cases. The F1 score, being the harmonic average of recall and precision, provides a balanced measure of the model's performance.

Therefore, we used the performance evaluation equations in Eq1, Eq2, Eq3, and Eq4.  to calculate performance metrics and evaluate the results. These equations helped us obtain a more detailed assessment of the DMCNN model's performance for each class of tomato leaf disease, enabling us to pinpoint areas where the model could be improved further.

$$Accuracy = \frac{TP+TN}{TP + TN + FP + FN} \quad (1)$$

$$Precision = \frac{TP}{TP + FP} \quad (2)$$

$$Sensitivity(Recall) = \frac{TP}{TP + FN} \quad (3)$$

$$F1 - Score = \frac{Precision \times Recall}{Precision + Recall} \quad (4)$$

Where

- **TP** = True Positive
- **FN** = False Negative
- **TN** = True Negative
- **FP** = False Positive

Assessing these metrics for each class allows us to identify specific weaknesses or strengths of the model for particular classes, which can guide further improvements or adjustments to the model. In our research, we evaluated these metrics for each class of tomato leaf disease on the testing data, including accuracy, recall, precision, and F1-score. These metrics helped us gain information into the performance of the DMCNN model for each class and allowed us to identify areas where the model could be improved further. The outcomes of these evaluations are presented in Table 8.17, providing a detailed analysis of the model's performance for each class of tomato leaf disease.

Table 8.17 The test dataset's class accuracy, recall, precision, and F1-score values for each class of tomato leaf disease

| Class | Precision | Recall | F1 score |
|---|---|---|---|
| Bacterial spot | 0.99 | 0.98 | 0.99 |
| Late blight | 1.0 | 0.99 | 1.0 |
| Early blight | 0.98 | 1.0 | 0.99 |
| Leaf mold | 0.99 | 0.99 | 0.99 |
| Spider mite | 0.98 | 1.0 | 0.99 |
| Septoria leaf spot | 0.99 | 0.97 | 0.98 |
| Target spot | 0.99 | 0.99 | 0.99 |
| Mosaic virus | 0.99 | 1.0 | 0.99 |
| Yellow curl virus | 1.0 | 1.0 | 1.0 |
| Healthy | 1.0 | 0.99 | 1.0 |
| Accuracy | | | 0.991 |
| Macro Avg | 0.991 | 0.991 | 0.992 |
| Weighted Avg | 0.9918 | 0.9911 | 0.9916 |

Comparing our DMCNN model with pre-trained models was a critical element of our study, as it allowed us to evaluate the efficiency and performance of our approach. We compared our model with commonly used models for image classification tasks, including AlexNet, VGG16, InceptionV3, and ResNet50, and their architectures are explained in Table 8.15.

Our analysis revealed that our proposed DMCNN outperformed all other models regarding accuracy, F1-score, recall, and precision. Our model achieved an accuracy of 0.991, indicating that 99.1% of the test dataset's samples were correctly classified. The precision of our model was 0.991, indicating that the percentage of true positive predictions was 99.1% of all positive predictions. The recall of our model was 0.991, indicating that the percentage of true positive predictions between all actual positive instances was 99.1%. Finally, the F1-score of our model was 0.992, indicating that it performed admirably in terms of precision and recall.

Among the compared models, InceptionV3 and ResNet50 had the lowest accuracy and F1-score. InceptionV3 reported an accuracy of 0.89, while ResNet50 reached an accuracy of 0.88. AlexNet and VGG16 performed better than InceptionV3 and ResNet50 but fell short of our proposed DMCNN performance. AlexNet achieved an accuracy of 0.95, while VGG16 attained an accuracy of 0.91. The results indicate that the presented DMCNN model exhibits superior classification performance compared to the other models. The comparison results, including class accuracy, recall, F1-score, and precision, have been presented in Tables 8.18 and Figure 8.41.

Table 8.18 The Comparative Analysis: Proposed Model vs. Pre-Trained Models

| Method | Accuracy | Precision | Recall | F1 score |
|---|---|---|---|---|
| AlexNet | 0.95 | 0.96 | 0.96 | 0.95 |
| VGG16 | 0.91 | 0.93 | 0.91 | 0.92 |
| InceptionV3 | 0.89 | 0.90 | 0.89 | 0.89 |
| ResNet50 | 0.88 | 0.89 | 0.88 | 0.88 |
| Proposed DMCNN | 0.991 | 0.991 | 0.991 | 0.992 |

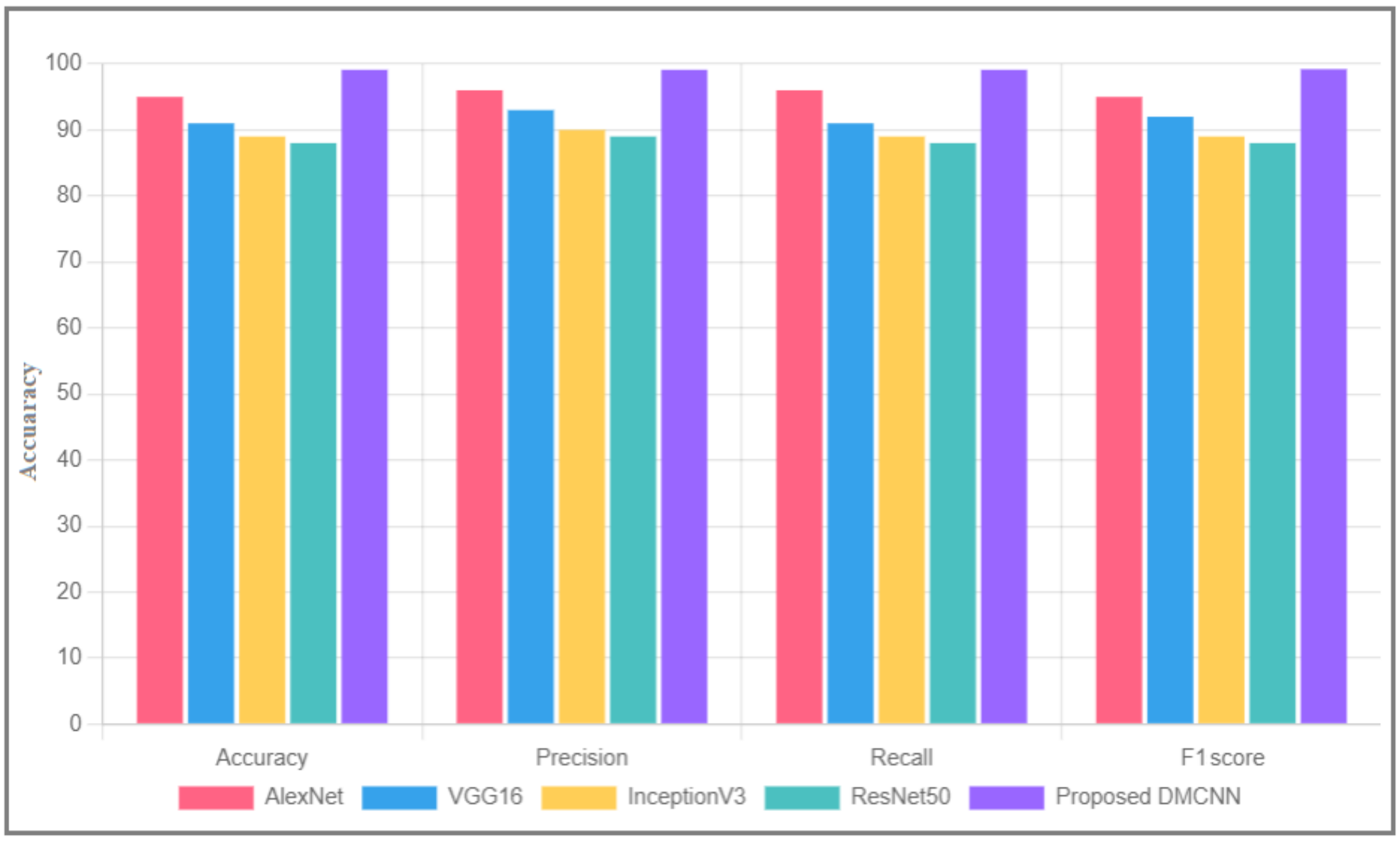

Figure 8.41 The Overall performance of different architectures

The excellent classification performance exhibited by our DMCNN model in comparison to pre-trained models in our study highlights its potential to efficiently and effectively classify tomato plant leaf diseases with crucial implications for the agriculture industry. Accurate disease classification enables farmers to take timely and appropriate measures to control and prevent the spread of diseases in their crops.

Furthermore, the practicality and efficacy of our DMCNN model can extend to other areas of agriculture, such as disease classification in other crops. Accurate and efficient deep-learning models for agriculture can improve sustainable farming practices and have far-reaching benefits. Overall, our study's findings demonstrate the potential of our proposed DMCNN model as a reliable tool for accurately and efficiently classifying tomato leaf diseases and its potential for wider applications in the agriculture industry. Developing more advanced deep learning models for agricultural purposes can lead to increased efficiency and productivity in the sector.

#### 8.4.5.3 Comparison of various state-of-the-art approaches to our proposed model: an evaluation of performance

A comprehensive evaluation was performed, comparing the proposed model with 10 existing DL-based methods utilized for the classification of tomato leaf diseases. The obtained results and the performance assessment of our model are presented in Table 8.19, offering significant insights into its efficacy and distinguishing characteristics, and Figure 8.42 alongside the results obtained from comparing it with other existing cutting-edge approaches.

Our analysis reveals that the suggested model outperforms all other contemporary techniques in classification performance. The model achieved 0.991 as an accuray, which is higher than the accuracy of all other models considered. This result indicates that our proposed model can accurately classify various types of tomato leaf diseases, and can provide an effective tool for plant pathologists and farmers to assess the health condition of tomato crops.

The evaluation encompassed a diverse set of models, comprising both transfer learning-based approaches and models developed from scratch. The outcomes indicate that our model outperforms all other models, irrespective of their development approach. This highlights the notable superiority of our model when compared to other modern approaches in the classification of tomato leaf diseases. In summary, our study demonstrates the potential of DL methods in accurately and efficiently classifying tomato plant leaf diseases. The proposed model's superior performance suggests that it can be an effective tool for plant pathologists and help farmers in assessing the health condition of tomato crops.

Table 8.19 Comparison of different existing work with our proposed model

| **Reference** | **Crop** | **Dataset** | **Method** | **Accuaracy** |
|---|---|---|---|---|
| (Agarwal et al., 2020) [19] | 10 tomato plant classes | 17,500 images | CNN based model | 91.2% |
| (Gadekallu et al., 2021) [20] | 10 tomato plant classes | 18,160 images | Hybrid deep model | 86%-94% |
| (Intan et al., 2023) [21] | 10 tomato plant classes | Kaggle | Dense CNN | 95.7% |

Table 8.19 Comparison of different existing work with our proposed model (continue)

| Reference | Crop | Dataset | Method | Accuaracy |
|---|---|---|---|---|
| (Agarwaal et al., 2020) [22] | 10 tomato plant classes | Plant Village | Vgg-16 | 91.2% |
| (Wang et al., 2019) [23] | 10 tomato plant classes | Plant Village | AlexNet | 95.62% |
| (Kaur et al., 2019) [24] | 7 tomato plant classes | Plant Village | ResNet-101 | 98.8% |
| (Kaushik et al., 2020) [25] | 6 tomato plant classes | Plant Village | ResNet-50 | 97.01% |
| (Trivedi et al., 2021) [26] | 9 tomato plant classes | 3000 images | Deep-CNN | 98.49% |
| (Vijay et al., 2021) [27] | Tomato plant | Plant Village | XAI-CNN | 98.5% |
| (Ozbılge et al., 2021) [28] | Tomato plant | ImageNet | Compact-CNN | 99.70% |
| Proposed DMCNN | 10 classes of tomato plant | 12500 images | DMCNN | 99.1% |

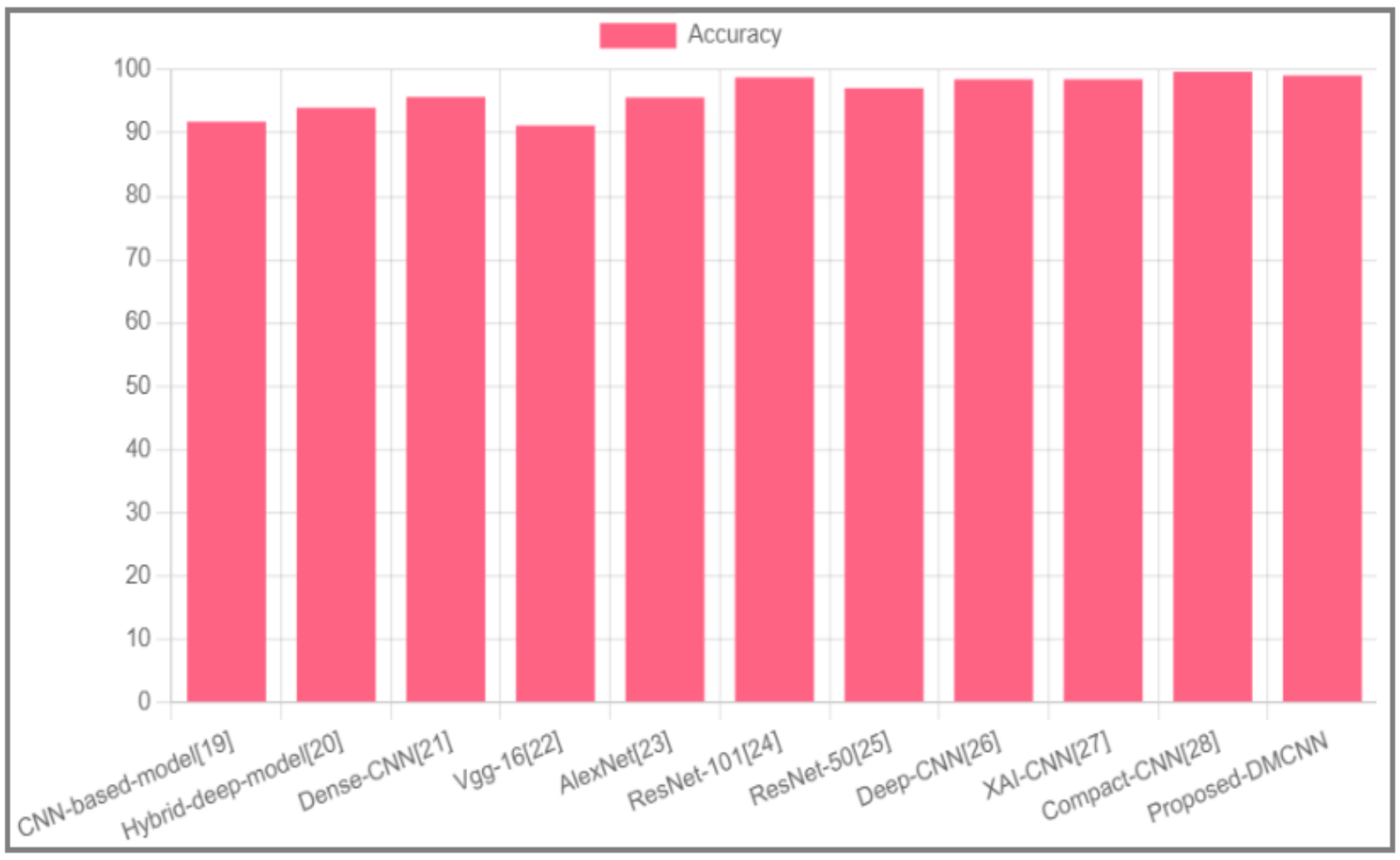

Figure 8.42 A comparative analysis with state-of-the-art-works

The proposed model has the potential to make substantial contributions to the field of plant pathology by accurately classifying tomato diseases. It represents a promising step towards achieving accurate and efficient identification and categorization of tomato diseases, which is crucial for disease management in agriculture. Although the proposed model has demonstrated excellent performance in the experiments, it is important to note that there is still potential for further enhancements. For example, future research could focus on incorporating additional data sources, such as environmental factors, to improve the model's accuracy and generalization capability. Furthermore, optimizing hyperparameters and exploring novel deep learning architectures could lead to even better performance.

The comparative analysis presented in Table 8.20 highlights the importance of evaluating the performance of new models against existing approaches. The remarkable performance of the proposed model in classifying tomato diseases, surpassing alternative methods, underscores its potential for practical implementation in the field of plant pathology. This model holds great promise as a valuable resource for plant pathologists and farmers, enabling them to accurately assess the health status of tomato crops and implement effective strategies to prevent and manage disease outbreaks. Table 8.20 provides a comprehensive performance comparison between the proposed DMCNN model and other cutting-edge approaches in the specified domain. The findings reveal the superior performance of the DMCNN model over most of the advanced methods listed in the table. Additionally, the ST column reports the results of a statistical significance test, bolstering the robustness of the outcomes.

Table 8.20 Comparison of the proposed DMCNN model with existing approaches. The ST column reports the results of a statistical significance test, where + indicates that our method outperforms the baseline with $p < 0.05$, - indicates that our method does not outperform the baseline with $p \geq 0.05$, and N/A indicates that no statistical test was performed in the original paper

| **Ours** | **State-of-the-art methods** | **ST** |
|---|---|---|
| 99.1% | 91.2% CNN based mode (Agarwal et al., 2020) | + |
| | 86%-94%% Hybrid deep model (Gadekallu et al., 2021) | + |
| | 95.65% AlexNet (Durmus et al., 2017) | + |
| | 91.2% Vgg-16 (Agarwaal et al., 2020) | + |
| | 95.62%AlexNet (Wang et al., 2019) | + |
| | 98.8% ResNet-101 (Kaur et al., 2019) | + |
| | 97.01% ResNet-50 (Kaushik et al., 2020) | + |
| | 98.49% Deep-CNN (Trivedi et al., 2021) | + |
| | 98.5%% XAI-CNN (Vijay et al., 2021) | + |
| | 99.70%% Compact-CNN (Ozbılge et al., 2021) | - |

The "+" symbol in the ST column signifies that the performance improvement of the suggested DMCNN model compared to the baseline is statistically significant, with a significance level of $p < 0.05$. This indicates that the superiority of the DMCNN model is supported by statistically significant results.

In Table 8.20, the ST column presents the outcome of a statistical significance test, a critical element in assessing deep learning models' performance. Such tests are crucial in determining if performance differences between models result from chance or a statistically significant improvement. The test results confirm that the suggested DMCNN model surpasses existing methods significantly in terms of performanc, highlighting the model's superiority. These findings have significant implications for deep learning, demonstrating the DMCNN model's potential to solve sequential data problems.

Our DMCNN model has demonstrated superior performance in the automated classification of multi-class plant diseases. One potential reason for this is its ability to extract multi-scale features from input images. This ability to capture both the fine-grained details and the overall context of the input image has proven particularly effective for accurately classifying multi-class plant diseases. Additionally, the model was trained on a different dataset, which may have contributed to its robustness and generalization performance. Our study showcases the effectiveness of DMCNNs for automated disease classification in agriculture, particularly for multi-class leaf diseases in tomatoes. This approach may greatly improve the accuracy, performance, and reliability of disease identification, which can significantly impact crop yields and quality. Further research can explore the application of our model to other crops and the development of more advanced models for disease detection in agriculture.

Despite the promising results, it is important to acknowledge the study's limitations. First, the suggested model was evaluated on a single dataset, which may not represent all tomato diseases across different regions and growing conditions. Therefore, future studies should investigate the generalizability of the proposed model across different datasets and growing conditions to further validate its effectiveness.

To summarize, our study presents a highly effective method for accurately classifying diseases in tomato plants using a novel deep multi-scale CNN architecture. The suggested model attains an accuracy of 99.1% and outperforms other existing methods regarding various evaluation metrics. Our study also highlights the importance of multi-scale imaging and deep DL for improving the classification of leaf diseases.

The proposed appraoch has significant potential for practical applications in the agricultural industry, including precision agriculture, where it can detect and prevent disease spread in crops more efficiently and effectively. In the future, we plan to evaluate the practical applicability of the suggested model and its potential for identifying and categorizing diseases in various crops. Additionally, we will conduct research on employing transfer learning approaches to enhance the performance of our model.

# 9. THESIS RESULTS AND DISCUSSION

The utilization of DL methods for detecting plant leaf diseases has garnered considerable attention in recent years, as it holds the promise of enhancing agricultural productivity and mitigating the adverse effects of diseases on crops. In this thesis, different facets of plant leaf disease detection using DL have been explored and analyzed. Firstly, we have emphasized the importance of plant leaf disease analysis in identifying the disease type and developing an effective control strategy. Deep learning methods can automate this process and provide accurate and reliable results, which is essential for the timely and efficient management of plant diseases.

Furthermore, we provided a detailed review of deep learning applications in agriculture. Deep learning methods can revolutionize how we manage crops, improve food security, and reduce the negative impact of diseases on agriculture. This thesis has detailedly studied deep learning and its applications in plant disease detection, including image-based classification, real-time disease detection, and disease severity estimation. Additionally, a case study showcasing the application of DL in leaf disease detection was presented, highlighting the technology's potential in the field of agriculture. Nevertheless, it is crucial to acknowledge that the effective implementation of deep learning in agriculture necessitates a meticulous assessment of various factors such as data accessibility, computing resources, and the requirement for specialized expertise. Therefore, evaluating these factors before adopting deep learning methods in agriculture is crucial to ensure successful and efficient implementation. Despite these challenges, deep learning holds tremendous promise in revolutionizing agriculture and addressing various challenges faced by the industry.

We also discussed the future directions of deep learning in agriculture, highlighting the need for further research to develop more accurate and efficient DL models for leaf disease detection. In addition, it is crucial to consider the significance of data quality and quantity in training these models. Addressing these concerns will contribute to developing more accurate and efficient deep-learning models that can significantly impact agriculture and food security.

In addition to a detailed study of deep learning applications in agriculture, we presented a new approach called "Plant Leaf Diseases Detection Using MobileNet Model", In this approach, MobileNet, a high-performing convolutional neural network known for its efficiency, is employed to identify various types of diseases based on leaf images, utilizing an effective network architecture. The method is specifically designed tackle the inherent challenges associated with plant disease detection, such as a large number of classes, the complexity of the plant structure, and the variability of environmental conditions. The model takes leaf images as input and extracts meaningful features to classify the disease type accurately. In addition, the network architecture of the model is carefully optimized to achieve a high level of accuracy, while also ensuring that the model size remains compact enough for deployment on resource-constrained mobile devices. Furthermore, we explored the different hyperparameters of the model to find the best architecture configuration, including learning rate, batch size, and number of epochs. By optimizing these hyperparameters, we achieved an accuracy of 92.97%, which is a significantly higher level when compared to the baseline model, demonstrating the proposed approach's effectiveness. The model holds significant potential to make a valuable contribution towards reducing the adverse impact of diseases on crops, thereby improving food security by enabling earlier detection and intervention, before the disease spreads to other crops. Overall, this approach shows promise for the practical implementation of deep learning in agriculture, and future research could explore its potential for other agricultural applications.

In addition to our previous approach, we introduced another study called “the impact of datasets on the effectiveness of MobileNet for leaf disease detection”; we aimed in this study to assess the generalizability of the suggested method by analyzing its performance on three different datasets of varying difficulty levels. We also conducted a comparative analysis of our suggested approach with other leading cutting-edge deep learning methods to evaluate its efficacy. This study provides valuable insights for future researchers to improve the effectiveness of their models by choosing appropriate datasets for their models and applying the best-performing approach. Through this study, we analyzed the impact of datasets on the model's performance by testing the model on different bean leaf image datasets of varying difficulty levels. This information can be helpful for researchers and practitioners to choose appropriate datasets for their models. Additionally, we suggested an approach that applies the MobileNet deep-learning algorithm to identify leaf diseases for use in practical and real-world contexts. This approach is expected to achieve satisfactory accuracy results on three databases of annotated images of healthy and unhealthy leaves (Collected leaf dataset = 94.53%, Public leaf dataset = 92.97%, Merged dataset = 93.75%), outperforming the baseline accuracy significantly. Furthermore, the evaluation of our presented method's generalizability across three different datasets with varying levels of difficulty, coupled with a comparative analysis against other advanced deep learning approaches, we were able to significantly enhance the accuracy of DL models for leaf disease detection, thereby making a substantial contribution to the field.

Furthermore, we presented a new method and research called "A Novel Convolution Neural Network-Based Approach for Seeds Image Classification." The approach was designed to identify and classify seeds, assisting farmers in addressing their problems. The study provided a detailed account of the entire research process, from developing and implementing a novel CNN model for image classification in higher dimensional spaces to fine-tuning the model by changing parameters such as the learning rate. We evaluated its performance using several measures to ensure that our model performs well on the dataset and does not miss the optimal solution. Furthermore, we compared it to pre-trained advanced deep learning approaches, the overall accuracy achieved for the proposed approach is 93% compared with DenseNet121 = 90.03%, InceptionV3 = 84.71%, and ResNet152 = 73.34%. This demonstrates the superior performance of our novel approach compared to existing pre-trained advanced deep learning models. This insight is valuable for practical applications, particularly in the agricultural industry. Furthermore, the study contributed to generating novel datasets, which we used to assess the method's generalizability.

Finally, CNN, a type of Deep Learning technique, have demonstrated exceptional performance across a diverse set of tasks. However, studies in leaf disease detection and classification mainly concentrate on single-channel and same-resolution images, which may not capture the complete information needed for accurate disease detection. Furthermore, To address this limitation, this thesis proposes a DMCNN based framework for automatically classifying leaf diseases. The proposed approach leverages multiple channels of information to enhance the accuracy of disease classification. This study introduces a new area of research with five key contributions: proposing a deep multi-scale CNN architecture that leverages multiple channels of information for accurate disease classification, developing a DMCNN architecture consisting of parallel streams of CNNs at different scales, evaluating the proposed framework using a large and diverse dataset of leaf images, providing insights into the feature importance of the proposed model, and conducting significant analysis to exhibit the resilience of the suggested model to various factors. This study's contributions highlight the promise of utilizing deep learning and multi-scale imaging for identifying and categorizing plant diseases. By utilizing multiple information channels, this approach achieves enhanced accuracy and efficiency in disease detection and classification, with an overall accuracy reported at 99.1%, significantly outperforming the existing pre-trained methods.

However, DL-based methods for plant disease detection also faces challenges and limitations, such as the need for large and diverse datasets. Additionally, ethical considerations regarding the use of automated systems in agriculture must be addressed in future research and development. While automation can increase productivity and reduce costs, it may also raise questions about the role of human expertise in agriculture and the potential impact on employment in the sector. These challenges and limitations must be carefully considered and addressed in future studies to ensure that plant disease identification using DL is a viable and sustainable solution.

In summary, the application of DL methods in plant disease detection holds immense potential in transforming the agriculture industry and safeguarding global food security. By enhancing the efficiency of disease detection, effective plant disease management, reduced crop losses, and improved food security can be achieved. However, it is imperative to tackle the challenges and limitations inherent in deep learning approaches, such as the requirement for diverse and representative datasets and ethical considerations in its utilization. Continued research and development in this field can lead to the establishment of more sustainable and resilient agricultural practices, ultimately benefiting future generations. The utilization of DL in plant leaf disease detection represents a dynamic and rapidly advancing field of research. It holds great promise for improving agricultural productivity and mitigating the adverse impact of diseases on crops.

The research presented in this thesis offers valuable insights into the development and assessment of DL models for plant disease detection and classification. A crucial contribution of this thesis is the creation of more precise and efficient DL models for disease detection and classification, with the potential to revolutionize the agriculture sector and address food security challenges. Additionally, creating new datasets in the plant leaf and seed domains is another significant contribution, facilitating the training and evaluation of more robust deep learning models. Through the utilization of deep learning, this thesis offers insights into the future of disease management and crop production, highlighting the potential for reduced crop losses, improved yields, and more sustainable agriculture practices. Continued research and development in this area are essential to unlock the full potential of DL in agriculture and ensure food security for future generations.

# 10 CONCLUSION

The utilization of deep learning (DL) methods for detecting plant leaf diseases represents a promising avenue for transforming the agricultural landscape. This thesis has delved into various facets of plant leaf disease detection using DL, emphasizing its potential to enhance agricultural productivity and mitigate the adverse effects of diseases on crops, we embarked on a comprehensive exploration of plant leaf disease detection and identification using advanced deep learning methodologies. Recognizing the critical importance of early disease detection for crop health and productivity, we aimed to address the shortcomings of existing detection methods.

Throughout this study, we undertook several significant initiatives. First and foremost, we designed and implemented novel deep learning approaches tailored specifically for plant leaf disease detection and identification. These approaches were founded upon meticulous data curation and model architecture optimization, striving to achieve superior accuracy and efficiency in disease detection. In parallel, we engaged in a thorough examination of the challenges plaguing current disease detection methodologies. We sought to enhance the robustness and precision of deep learning models to ensure their effectiveness across a wide spectrum of real-world scenarios. This encompassed the development of techniques to accommodate images of varying quality, environmental conditions, and disease progression stages. Our research also delved into the critical aspect of distinguishing between similar diseases and benign conditions, aiming for a level of precision that could significantly reduce false positives and ensure accurate disease identification.

In addition to refining pre-existing deep learning models, we introduced a novel CNN-based models explicitly designed for plant leaf disease detection. Rigorous evaluations, comparative analyses, and parameter-tuning algorithms were all part of our efforts to optimize and fine-tune these models for peak performance. Moreover, we ventured into the realm of dataset creation, both by utilizing existing datasets from the literature and by generating our own, targeting new datasets and conditions that had not previously been explored in the literature. This strategic dataset expansion was instrumental in assessing the adaptability and generalization capabilities of our models.

As a result of our diligent research and development efforts, we successfully implemented and evaluated our proposed leaf disease detection approaches, yielding highly satisfactory performance outcomes. These outcomes lay the foundation for practical applications of deep learning in agriculture, with the potential to revolutionize disease management, reduce crop losses, and enhance food security.

Looking ahead, our work points to several promising avenues for future research. The successful implementation of these technologies will necessitate collaboration between researchers and industry experts to address cost, scalability, and user-friendliness issues. Additionally, the ongoing collection of diverse and comprehensive datasets will be pivotal in advancing the adaptability of the models. Finally, ethical considerations surrounding the automation of disease detection in agriculture should not be overlooked. Future research should delve into the implications for human expertise and employment within the agricultural sector as automation continues to gain traction.

In summary, our endeavors in this study mark a significant step forward in the quest to harness the potential of deep learning for plant leaf disease detection. We have laid the groundwork for a future where technology plays a pivotal role in safeguarding crop health, ensuring food security, and promoting sustainable farming practices.

## REFERENCES

A. Review of Deep Learning Research https://scite.ai/reports/10.3837/tiis.2019.04.001 https://www.tutorialspoint.com/artificial_neural_network/artificial_neural_network_hopfield.htm

Abadi, M.; Barham, P.; Chen, J.; Chen, Z.; Davis, A.; Dean, J.; Devin, M.; Ghemawat, S.; Irving, G.; Isard, M.; et al. TensorFlowg: a system for fLarge-Scaleg machine learning. In Proceedings of the 12th USENIX Symposium on Operating Systems Design and Implementation (OSDI 16), Savannah, GA, USA, 2–4 November 2016; pp. 265–283.

Abbas, A.; Jain, S.; Gour, M.; Vankudothu, S. Tomato plant disease detection using transfer learning with C-GAN synthetic images. Comput. Electron. Agric. 2021, 187, 106279.

Abdel-Hamid, O.; Mohamed, A.R.; Jiang, H.; Deng, L.; Penn, G.; Yu, D. Convolutional neural networks for speech recognition. IEEE/ACM Trans. Audio Speech Lang. Process. 2014, 22, 1533–1545.

Abed.S.; Esmaeel, A.A.; A novel approach to classify and detect bean diseases based on image processing, 2018 IEEE Symposium on Computer Applications & Industrial Electronics (ISCAIE) (2018) 297–302. URL: DOI:10.1109/ISCAIE.2018.8405488.

Adedoja, A.; Owolawi, P.A.; Mapayi, T. Deep learning based on NASNet for plant disease recognition using leave images. In Proceedings of the 2019 International Conference on Advances in Big Data, Computing and Data Communication Systems (icABCD), Winterton, South Africa, 5–6 August 2019; pp. 1–5.

Agarwal, M., Singh, A., Arjaria, S., Sinha, A., and Gupta, S., "ToLeD: Tomato leaf disease detection using convolution neural network," Proc. Comp. Sci, vol.167, pp.293-301, 2020, doi:10.1016/j.procs.2020.03.225

Agarwal, M.; Arjaria, S.; Sinha, A.; Singh, A.; Gupta, S.J.P.C.S. ToLeD- Tomato leaf diseases detection using convolution neural network. 2020, 167, 293-301. DOI: 10.1016/j.procs.2020.03.225

Agarwal, M.; Gupta, S.K.; Biswas, K. Development of Efficient CNN model for Tomato crop diseases identification. Sustain. Comput. Inform. Syst. 2020, 28, 100407. https://doi.org/10.1016/j. Suscom.2020.100407

Agarwal, Mohit, Suneet Kr Gupta, and K. K. Biswas. "Development of Efficient CNN model for Tomato crop disease identification." Sustainable Computing: Informatics and Systems 28 (2020): 100407.

Agrios, G. N. (2005). Plant pathology (5th ed.). Elsevier Academic Press.

Ahmad, I.; Hamid, M.; Yousaf, S.; Shah, S.T.; Ahmad, M.O. Optimizing Pretrained Convolutional Neural Networks for Tomato Leaf Disease Detection. Complexity 2020, 2020, 8812019:1–8812019:6.

Ahmad, J.; Jan, B.; Farman, H.; Ahmad,W.; Ullah, A. Disease Detection in Plum Using Convolutional Neural Network under True Field Conditions. Sensors 2020, 20, 5569.

Ahmad, W., Shah, S., Irtaza, A. (2020) Plants disease phenotyping using quinary patterns as texture descriptor. KSII Trans Internet Inf Syst 14(8):3312–3327

Ai, Y.; Sun, C.; Tie, J.; Cai, X. Research on recognition model of crop diseases and insect pests based on deep learning in harsh environments. IEEE Access 2020, 8, 686–693.

Alexandre, B., Corentin,B., Élie,F., Thierry, G., Fadi,J., Béchara,W., Pascal.R.,2022.Effects of dataset size and interactions on the prediction performance of logistic regression and deep learning models. v.21, URL.https://doi.org/10.1016/j.cmpb.2021.106504.

Alexey Averkin, Sergey Yarushev.; Evolution of Artificial Neural Networks, https://libeldoc.bsuir.by/bitstream/123456789/30338/1/Averkin_Evolution.PDF. 2018.

Alom, M.Z., Taha, T.M., Yakopcic, C., Westberg, S., Sidike, P., Nasrin, Hasan, M., Van Essen, B.C., Awwal, A.A.S., Asari, V.K., 2019. A state-of-the-art survey on deep learning theory and architectures. Electronics 8.https://www.mdpi.com/2079-9292/8/3/292. doi:10.3390/electronics8030292.

Alom, M.Z., Yakopcic, C., Hasan, M., Taha, T.M., Asari, V.K., 2019. Recurrent residual unet for medical image segmentation. J. Med. Imag. (Bellingham, Wash.) 6, 014006.

Alom, M.Z.; Taha, T.M.; Yakopcic, C.; Westberg, S.; Sidike, P.; Nasrin, M.S.; Hasan, M.; Van Essen, B.C.; Awwal, A.A.; Asari, V.K.A state-of-the-art survey on deep learning theory and architectures. Electronics 2019, 8, 292.

Amara, J.; Bouaziz, B.; Algergawy, A. A deep learning-based approach for banana leaf diseases classification. In Proceedings of the Datenbanksysteme für Business, Technologie und Web (BTW 2017) - Workshopband, Stuttgart, Germany, 6–10 March 2017.

Amara, J.; Bouaziz, B.; Algergawy, A. A deep learning-based approach for banana leaf diseases classification. In Proceedings of the Datenbanksysteme für Business, Technologie und Web (BTW 2017)—Workshopband, Stuttgart, Germany, 6–10 March 2017.

Aravind, K.R., Raja, P., Anirudh, R., Mukesh, K.V., Ashiwin, R., and Vikas Vikas, G., “Grape crop disease classification using transfer learning approach,” in Proceedings of the International Conference on ISMAC in Computational Vision and Bio-Engineering 2018 (ISMAC- CVB), vol.30. Springer, Cham. Jan. 2019, pp. 1623-1633, doi:/10.1007/978-3-030-00665-5150

Arivazhagan, S.; Ligi, S.V. Mango leaf diseases identification using convolutional neural network. Int. J. Pure Appl. Math. 2018, 120, 11067–11079.

Arya, S.; Singh, R. A Comparative Study of CNN and AlexNet for Detection of Disease in Potato and Mango leaf. In Proceedings of the IEEE International Conference on Issues and Challenges in Intelligent Computing Techniques (ICICT), Ghaziabad, India, 27–28 September 2019; Volume 1, pp. 1–6.

Ashwinkumar, S., S. Rajagopal, V. Manimaran, and B. Jegajothi. "Automated plant leaf disease detection and classification using optimal MobileNet based convolutional neural networks." Materials Today: Proceedings 51 (2022): 480-487.

Aston Zhang, Zachary C. Lipton, Mu Li, and Alexander J. Smola. 2020. Dive into Deep Learning. Retrieved August 15, 2020 from https://d2l.ai/d2l-en.pdf.

Atila, Ü.; Uçar, M.; Akyol, K.; Uçar, E. Plant leaf disease classification using EfficientNet deep learning model. Ecol. Inform. 2021, 61, 101182.

Badrinarayanan, V., Kendall, A., Cipolla, R., 2017. Segnet: A deep convolutional encoder-decoder architecture for image segmentation. IEEE Trans. Pattern Anal. Mach. Intell. 39, 2481–2495.

Bahrampour S, Ramakrishnan N, Schott L and Shah M (2015) Comparative study of deep learning software frameworks. arXiv preprint arXiv:1511.06435.

Barbedo, J. G. A. (2017). Plant disease diagnosis: Challenges and opportunities. Applied Plant Pathology, 1(1), 1-10.

Barbedo, J.G. Factors influencing the use of deep learning for plant disease recognition. Biosyst. Eng. 2018, 172, 84–91.

Barbedo, J.G.A. A review on the main challenges in automatic plant disease identification based on visible range images. Biosyst. Eng. 2016, 144, 52–60.

Barbedo, J.G.A. Plant disease identification from individual lesions and spots using deep learning. Biosyst. Eng. 2019, 180, 96–107.

Barbedo, J.G.A., "Impact of dataset size and variety on the effectiveness of deep learning and transfer learning for plant disease classification," Comput. Electron. Agric. vol.153, pp.46–53, October 2018, doi: 10.1016/j.compag.2018.08.013.

BENGIO, Y., 2009. "Learning deep architectures for AI. Foundations and trends, in Machine Learning", Dept. IRO, Universite de Montreal, 2(1): 1-127.

Borhani, Y.; Khoramdel, J.; Najafi, E. A deep learning based approach for automated plant disease classification using vision transformer. Sci. Rep. 2022, 12, 11554.

Brahimi, M., Boukhalfa, K., and Moussaoui, A., ``Deep learning for tomato diseases: Classi_cation and symptoms visualization," *Appl.Artif. Intell.*, vol. 31, no. 4, pp. 299_315, Apr. 2017, doi: 10.1080/08839514.2017.1315516.

Brahimi, M.; Boukhalfa, K.; Moussaoui, A. Deep learning for tomato diseases: classification and symptoms visualization. Appl. Artif. Intell. 2017, 31, 299–315.

Canziani A, Paszke A and Culurciello E (2016) An analysis of deep neural network models for practical applications. arXiv preprint arXiv:1605. 07678 [cs.CV].

Canziani, A.; Paszke , A.; and E. Culurciello, "An Analysis of Deep Neural Network Models for Practical Applications," arXiv preprint, vol. arXiv:1605.07678, 2016.

Canziani, A.; Paszke, A.; Culurciello, E. An analysis of deep neural network models for practical applications. arXiv 2016, arXiv:1605.07678.

Channamallikarjuna Mattihalli, Edemialem Gedefaye, Fasil Endalamaw, Adugna Necho., Plant leaf diseases detection and auto-medicine., Internet of Things 1–2 (2018) 67–73.

Chen, J., Liu, Q., and Gao, L., "Visual Tea Leaf Disease Recognition Using a Convolutional Neural Network Model," Symmetry, vol.11, no.3, pp.343, Mar. 2019, doi: 10.3390/sym11030343.

Chen, J.; Liu, Q.; Gao, L. Visual tea leaf disease recognition using a convolutional neural network model. Symmetry 2019, 11, 343.

Chen, J.; Liu, Q.; Gao, L. Visual tea leaf disease recognition using a convolutional neural network model. Symmetry **2019**, 11, 343.

Chen, J.; Zhang, D.; Zeb, A.; Nanehkaran, Y.A**.** Identification of rice plant diseases using lightweight attention networks. Expert Syst. Appl. 2021, 169, 11451

Chen, L., Papandreou, G., Kokkinos, I., Murphy, K., Yuille, A.L., 2018. Deeplab: Semantic image segmentation with deep convolutional nets, atrous convolution, and fully connected crfs. IEEE Trans. Pattern Anal. Mach. Intell. 40, 834–848.

Chen, Xiao, Guoxiong Zhou, Aibin Chen, Jizheng Yi, Wenzhuo Zhang, and Yahui Hu. "Identification of tomato leaf diseases based on combination of ABCK-BWTR and B-ARNet." Computers and Electronics in Agriculture 178 (2020): 105730.

Chen, Y.; Xu, K.; Zhou, P.; Ban, X.; He, D. Improved cross entropy loss for noisy labels in vision leaf disease classification. IET Image Process. 2022, 16, 1511–1519.

Chollet, F. Xception: Deep learning with depthwise separable convolutions. In Proceedings of the IEEE Conference on Computer Vision and Pattern Recognition, Honolulu, HI, USA, 21–26 July 2017; pp. 1251–1258.

Chouhan, S. S., A. Kaul, U. P. Singh, and S. Jain, "Bacterial foraging optimization-based radial basis function neural network (BRBFNN) for identification and classification of plant leaf diseases: An automatic approach towards plant pathology," IEEE Access, vol.6, pp.8852-8863, 2018, doi:10.1109/ACCESS.2018.2800685.

Chouhan, S.S., Pratap Singh, U., and Jain, S., "Automated Plant Leaf Disease Detection and Classification Using Fuzzy Based Function Network," Springer Science, vol.1757–1779, pp.121, 19 July 2021, doi:10.1007/s11277-021-08734-3.

Chouhan, S.S., Pratap Singh, U., Sharma, U., and S.Jain, "Leaf disease segmentation and classification of Jatropha Curcas L. and Pongamia Pinnata L. biofuel plants using computer vision based approaches," elsevier journal Measurement, vol.171, pp.108796, February 2021, doi:10.1016/j.measurement.2020.108796.

Cruz, A.C.; Luvisi, A.; De Bellis, L.; Ampatzidis, Y. X-FIDO: An effective application for detecting olive quick decline syndrom with deep learning and data fusion. Front. Plant Sci. 2017, 8, 1741.

Cubuk, E.D.; Zoph, B.; Shlens, J.; Le, Q.V. Randaugment: Practical automated data augmentation with a reduced search space. In Proceedings of the IEEE/CVF Conference on Computer Vision and Pattern Recognition Workshops, Seattle, WA, USA, 14–19 June 2020; pp. 702–703.

Dang, L.M.; Syed, I.H.; Suhyeon, I. Drone agriculture imagery system for radish wilt. J. Appl. Remote Sens. 2017, 11, 16006.

Darwish, A.; Ezzat, D.; Hassanien, A.E. An optimized model based on convolutional neural networks and orthogonal learning particle swarm optimization algorithm for plant diseases diagnosis. Swarm Evol. Comput. 2020, 52, 100616.

Del Ponte, M., Fernandes, J. M. C., & Bergamin Filho, A. (2009). Relationship between crop type and pathogen groups on the severity of soybean diseases in Brazil. Crop Protection, 28(4), 313-320.

Deng J, Dong W, Socher R, Li LJ, Li K and Fei-Fei L (2009) Imagenet: a large-scale hierarchical image database. 2009 IEEE Conference on Computer Vision and Pattern Recognition (CVPR).Miami,FL,USAIEEE, pp. 248-255. (http://host.robots.ox.ac.uk/pascal/VOC/).

Diaz, G., Fokoue-Nkoutche, A., Nannicini, G., Samulowitz, H., 2017. An effective algorithm for hyperparameter optimization of neural networks. IBM J. Res. Dev. 61 https://doi.org/10.1147/JRD.2017.2709578.

Durmu¸ s, H.; Gune¸ s, E.O.; Kırcı, M. Diseases detection on the leaves of the tomato plant by using deep learning. In Proceedings of the 2017 6th International Conference on Agro-Geoinformatics Fairfax, VA, USA, 7–10 August 2017; pp. 1–5. DOI:10.1109/Agro -Geoinformatics.2017.8047016

Durmus, H.; Günes, E.O.; Kirci, M. Disease detection on the leaves of the tomato plants by using deep learning. In Proceedings of the 6th International Conference on Agro-Geoinformatics, Fairfax, VA, USA, 7–10 August 2017; pp. 1–5.

Dyrmann, M.; Karstoft, H.; Midtiby, H.S. Plant species classification using deep convolutional neural network. Biosyst. Eng. 2016 151, 72–80.

E. Ülker, Derin Öğrenme ve Görüntü Analizinde Kullanılan Derin Öğrenme Modelleri, Gaziosmanpaşa Bilimsel Araştırma Dergisi, 6:3 (2017) 85-104.

Esmaeel, A. A., “A novel approach to classify and detect bean diseases based on image processing,” in 2018 IEEE Symposium on Computer Applications & Industrial Electronics (ISCAIE). IEEE, 2018, pp. 297–302.

Fang, T.; Chen, P.; Zhang, J.; Wang, B. Crop leaf disease grade identification based on an improved convolutional neural network. J. Electron. Imaging 2020, 29, 013004.

Ferentinos, K. Deep learning models for plant disease detection and diagnosis. Comput. Electron. Agric. 2018, 145, 311–318.

Ferentinos, K.P. Deep learning models for plant disease detection and diagnosis. Comput. Electron. Agric. 2018, 145, 311–318.

Ferentinos, K.P. Deep learning models for plant disease detection and diagnosis. Comput. Electron. Agric. 2018, 145, 311–318.

Frank Rosenblatt. The perceptron: A probabilistic model for information storage and organization in the brain. Psychological review, 65(6):p.386, 1958.

Fuentes A, Yoon S, Kim SC, Park DS. A robust deep-learning-based detector for real-time tomato plant diseases and pests’ recognition. Sensors. 2022; 2017:17.

Gadekallu, T.R.; Reddy, M.P.K.; Lakshmanna, K.; Rajput, D.S.; Bhattacharya, S.; Jolfaei, A.; Singh, S.; Alazab, M.J.J.o.R.-T.I.P. A novel 729 PCA whale optimizationbased deep neural networks model for classification of tomato plant diseases using GPU. 2021, 18, 1383-1396. DOI:10.1007/s11554-020-00987-8.

Gaikwad, S.S.; Rumma, S.S.; Hangarge, M. Fungi affected fruit leaf disease classification using deep CNN architecture. Int. J. Inf. Technol. 2022.

Gajjar, R.; Gajjar, N.P.; Thakor, V.J.; Patel, N.P.; Ruparelia, S. Real-time detection and identification of plant leaf diseases using convolutional neural networks on an embedded platform. Vis. Comput. 2022, 38, 2923–2938.

Geetharamani, G., and Arun Pandian."Identification of plant leaf diseases using a nine-layer deep convolutional neural network." Computers Electrical Engineering 76 (2019): 323-33

Geetharamani, G.; Pandian, A. Identification of plant leaf diseases using a nine-layer deep convolutional neural network. Comput. Electr. Eng. 2019, 76, 323–338.

Girshick, R., Donahue, J., Darrell, T., Malik, J., 2014. Rich feature hierarchies for accurate object detection and semantic segmentation. In: 2014 IEEE Conference on Computer Vision and Pattern Recognition, pp. 580–587.

Goodfellow, I., Bengio, Y., Courville, A., 2016. Deep Learning. MIT Press. http://www. deeplearningbook.org.

Ha, J.G.; Moon, H.; Kwak, J.T.; Hassan, S.I.; Dang, M.; Lee, O.N.; Park, H.Y. Deep convolutional neural network for classifying Fusarium wilt of radish from unmanned aerial vehicles. J. Appl. Remote Sens. 2017, 11, 042621.

Hanh, B.T.; Manh, H.V.; Nguyen, N.V. Enhancing the performance of transferred efficientnet models in leaf image-based plant disease classification. J. Plant Dis. Prot. 2022, 129, 623–634.

Hassan, S.M.; Maji, A.K.; Jasinski, M.F.; Leonowicz, Z.; Jasi ´ nska, E. Identification of Plant-Leaf Diseases Using CNN and Transfer-Learning Approach. Electronics 2021, 10, 1388.

He, K.; Gkioxari, G.; Dollár, P.; Girshick, R. Mask R-CNN. In Proceedings of the IEEE International Conference on Computer Vision, ser. ICCV ’17, Venice, Italy, 22–29 October 2017; pp. 2980–2988.

Hebb, D. O. (1949). The organization of behavior; a neuropsychological theory. A Wiley Book in Clinical Psychology., 62-78.

Hendrycks, D.; Mu, N.; Cubuk, E.D.; Zoph, B.; Gilmer, J.; Lakshminarayanan, B. Augmix: A simple data processing method to improve robustness and uncertainty. arXiv 2019, arXiv:1912.02781.

Hinton GE, Osindero S, Teh Y-W. A fast learning algorithm for deep belief nets. Neural Comput. 2006;18(7):1527–54.

Ho, D.; Liang, E.; Chen, X.; Stoica, I.; Abbeel, P. Population based augmentation: Efficient learning of augmentation policy schedules. In Proceedings of the International Conference on Machine Learning, PMLR, Long Beach, CA, USA, 9–15 June 2019; pp. 2731–2741.

Howard, A. G; Zhu , M.; Chen , B.; Kalenichenko D.; Wang ,W. T. Weyand, M. Andreetto, and H. Adam, ``MobileNets: Efficient convolutional neural networks for mobile vision applications," 2017, arXiv:1704.04861

Howlader, M.R.; Habiba, U.; Faisal, R.H.; Rahman, M.M. Automatic Recognition of Guava Leaf Diseases using Deep Convolution Neural Network. In Proceedings of the 2019 International Conference on Electrical, Computer and Communication Engineering (ECCE), Cox's Bazar, Bangladesh, 7–9 February 2019; pp. 1–5.

Hubel DH, Wiesel TN (1959) Receptive fields of single neurones in the cat's striate cortex. J Physiol. Doi: 10.1113/jphysiol. 1959.sp006308.

Hubel DH, Wiesel TN (1962) Receptive fields, binocular interaction and functional architecture in the cat's visual cortex. J Physiol 160:106–154. doi: 10.1113/jphysiol. 1962.sp006837.

Hubel, D.H.; Wiesel, T.N. Receptive fields, binocular interaction and functional architecture in the cat's visual cortex. J. Physiol. 1962, 160, 106–154.

Ian Goodfellow, Yoshua Bengio, and Aaron Courville. 2017. Deep Learning. MIT Press, Cambridge, MA.

Indu, V.T.; Priyadharsini, S.S. Crossover-based wind-driven optimized convolutional neural network model for tomato leaf disease classification. J. Plant Dis. Prot. 2021, 129, 559–578.

Intan, N.Y.;Naufal A.A.; Akik,H. Mobile Application for Tomato Plant Leaf Disease Detection Using a Dense Convolutional Network Architecture,Computation 2023, 11(2), 20; https://doi.org/10.3390/computation11020020

Ioffe, S.; Szegedy, C. Batch normalization: Accelerating deep network training by reducing internal covariate shift. In Proceedings of the International Conference on Machine Learning, PMLR, Lille, France, 7–9 July 2015; pp. 448–456.

Javidan, S.M.; Banakar, A.; Vakilian, K.A.; Ampatzidis, Y. Diagnosis of Grape Leaf Diseases Using Automatic K-Means Clustering and Machine Learning. SSRN Electron. J. 2022, 3, 100081.

Jia, Y.; Shelhamer, E.; Donahue, J.; Karayev, S.; Long, J.; Girshick, R.; Guadarrama, S.; Darrell, T. Caffe: Convolutional architecture for fast feature embedding. In Proceedings of the 22nd ACM International Conference on Multimedia, Orlando, FL, USA, 3–7 November 2014; pp. 675–678.

Jiang, J.; Liu, H.; Zhao, C.; He, C.; Ma, J.; Cheng, T.; Zhu, Y.; Cao, W.; Yao, X. Evaluation of Diverse Convolutional Neural Networks and Training Strategies for Wheat Leaf Disease Identification with Field-Acquired Photographs. Remote Sens. 2022, 14, 3446.

Jiang, P.; Chen, Y.; Liu, B.; He, D.; Liang, C. Real-time detection of apple leaf diseases using deep learning approach based on improved convolutional neural networks. IEEE Access 2019, 7, 69–80.

Jin, H.; Li, Y.; Qi, J.; Feng, J.; Tian, D.; Mu,W. GrapeGAN: Unsupervised image enhancement for improved grape leaf disease recognition. Comput. Electron. Agric. 2022, 198, 107055.

Kamilaris, A.; Prenafeta-Boldú, F.X. Disaster monitoring using unmanned aerial vehicles and deep learning. arXiv 2018, arXiv:1807.11805.

Karami, O., & Alishiri, A. (2014). Application of serological techniques in plant disease diagnosis. International Journal of Advanced Biological and Biomedical Research, 2(8), 2383-2393.

Karhunen J, Raiko T, Cho KH. Unsupervised deep learning: a short review. In: Advances in independent component analysis and learning machines. 2015; p. 125–42.

Karlekar, A.; Seal, A. SoyNet: Soybean leaf diseases classification. Comput. Electron. Agric. 2020, 172, 105342.

Karthik R, Hariharan M, Anand Sundar, Mathikshara Priyanka, Johnson Annie, Menaka R. Attention embedded residual CNN for disease detection in tomato leaves. Applied Soft Comput. 2020. https ://doi.org/10.1016/j.asoc.2019.10593 3.

Kaur, M.; Bhatia, R. Development of an improved tomato leaf dis eases detection and classification method. In Proceedings of the IEEE Conference on Information, and Communication Technology Baghdad, Iraq, 15–16 April 2019; pp. 1–5. DOI:10.1109/CICT48419.2019.9066230.

Kaushik, M.; Ajay, R.; Prakash, P.; Veni, S. Tomato Leaf-Disease Detection using Convolutional Neural Networks with Data Augmentation. Proceedings of the 2020 5th International Conference on Communication and Electronics Systems (ICCES), Coimbatore, India, 10–12 June 2020; pp. 1125–1132. DOI:10.1109/ICCES4 8766.2020.9138030

Kawasaki, Y.; Uga, H.; Kagiwada, S.; Iyatomi, H. Basic study of automated diagnosis of viral plant diseases using convolutional neural networks. In Proceedings of the International Symposium on Visual Computing; Springer: Berlin/Heidelberg, Germany, 2015; pp. 638–645

Kaya, Yasin, and Ercan Gursoy. "A novel multi-head CNN design to identify plant diseases using the fusion of RGB images." Ecological Informatics (2023): 101998.

Khan, A., Sohail, A., Zahoora, U., Qureshi, A.S., 2019. A survey of the recent architectures of deep convolutional neural networks, CoRR abs/1901.06032. http://arxiv.org/abs/1901.06032. ArXiv:1901.06032.

Kianat, J.; Khan, M.A.; Sharif, M.; Akram, T.; Rehman, A.; Saba, T. A joint framework of feature reduction and robust feature selection for cucumber leaf diseases recognition. Optik 2021, 240, 166566

Kim, P. Matlab Deep Learning: With Machine Learning, Neural Networks and Artificial Intelligence; Apress: New York, NY, USA, 2017; p. 130.

Kitchenham, B., Pretorius, R., Budgen, D., Brereton, O.P., Turner, M., Niazi, M., Linkman, S., 2010. Systematic literature reviews in software engineering–a tertiary study. Inf. Softw. Technol. 52 (8), 792–805.

Krizhevsky A, Sutskever I and Hinton GE (2012) Imagenet classification with deep convolutional neural networks. In Pereira F, Burges CJC, Bottou L and Weinberger KQ (eds), Advances in Neural Information Processing Systems 25. Harrahs and Harveys, Lake Tahoe, US: Neural Information Processing Systems Foundation, Inc., pp. 1097–1105.

Krohling, R.; Esgario, J.; Ventura, J.A. BRACOL—A Brazilian Arabica Coffee Leaf images dataset to identification and quantification of coffee diseases and pests. Mendeley Data 2019, V1.

Kurmi, Y.; Saxena, P.; Kirar, B.S.; Gangwar, S.; Chaurasia, V.; Goel, A. Deep CNN model for crops' diseases detection using leaf images. Multidimens. Syst. Signal Process. 2022, 33, 981–1000.

Lecun, Y., Bottou, L., Bengio, Y., Haffner, P., 1998. Gradient-based learning applied to document recognition. Proc. IEEE 86, 2278–2324. https://doi.org/10.1109/5.726791.

LeCun, Y.; Boser, B.; Denker, J.S.; Henderson, D.; Howard, R.E.; Hubbard, W.; Jackel, L.D. Backpropagation applied to handwritten zip code recognition. Neural Comput. 1989, 1, 541–551.

LeCun, Y.; Bottou, L.; Bengio, Y.; Haffner, P. Gradient-based learning applied to document recognition. Proc. IEEE 1998,86, 2278–2324.

Lee, S. H., Chan, C. S., Wilkin, P., and Remagnino, P. (2015). « Deepplant: Plant identification with convolutional neural networks ». In: 2015 IEEE International Conference on Image Processing (ICIP). IEEE, pp. 452–456 (cit. on p. 52).

Li, X.; Li, S. Transformer Help CNN See Better: A Lightweight Hybrid Apple Disease Identification Model Based on Transformers. Agriculture 2022, 12, 884.

Li, Y.; Nie, J.; Chao, X. Do we really need deep CNN for plant diseases identification? Comput. Electron. Agric. 2020, 178, 105803.

Liang, W., Zhang, H., Zhang, G. F., and Cao, H.X., "Rice blast disease recognition using a deep convolutional neural network, "nature.Sci. Rep, vol.9, pp.2869, 27 February 2019, doi:/10.1038/s41598-019-38966-0.

Liang, W.j.; Zhang, H.; Zhang, G.F.; Cao, H.x. Rice blast disease recognition using a deep convolutional neural network. Sci. Rep. 2019, 9, 1–10.

LifeCLEF (2019). Multimidea retrival in CLEF. https://www.imageclef.org//. (Visited on 12/10/2019) (cit. on p. 52).

Lim, S.; Kim, I.; Kim, T.; Kim, C.; Kim, S. Fast autoaugment. Adv. Neural Inf. Process. Syst. 2019, 32, 6665–6675.

Lin, T., Goyal, P., Girshick, R., He, K., Doll´ar, P., 2017. Focal loss for dense object detection. In: 2017 IEEE International Conference on Computer Vision (ICCV), pp. 2999–3007.

Liu, B., Zhang, Y., D. He, and Y. Li, "Identification of Apple Leaf Diseases Based on Deep Convolutional Neural Networks," Symmetry, vol.10, no.1, pp.11, Dec. 2017, doi: 10.3390/sym10010011.

Liu, B., Zhang, Y., He, D., Li, Y., 2017. Identification of apple leaf diseases based on deep convolutional neural networks. Symmetry 10. https://doi.org/10.3390/ sym10010011. https://www.mdpi.com/2073-8994/10/1/11.

Liu, B.; Ding, Z.; Tian, L.; He, D.; Li, S.; Wang, H. Grape Leaf Disease Identification Using Improved Deep Convolutional Neural Networks. Front. Plant Sci. 2020, 11, 1082.

Liu, B.; Zhang, Y.; He, D.; Li, Y. Identification of apple leaf diseases based on deep convolutional neural networks. Symmetry 2017, 10, 11.

Liu, B.; Zhang, Y.; He, D.; Li, Y. Identification of apple leaf diseases based on deep convolutional neural networks. Symmetry, 2017, 10, 11.

Liu, J., Wang,X., 2020.Early recognition of tomato gray leaf spot disease based on Mo-bileNetv2-YOLOv3 mode Plant Methods. v.16, no.83.URL.https://plantmethods. biomedcentral.com/articles /10.1186/s13007-020-00624-2.

Long, J., Shelhamer, E., Darrell, T., 2015. Fully convolutional networks for semantic segmentation. In: 2015 IEEE Conference on Computer Vision and Pattern Recognition (CVPR), pp. 3431–3440.

Lu, J.; Hu, J.; Zhao, G.; Mei, F.; Zhang, C. An In-field Automatic Wheat Disease Diagnosis System. Comput. Electron. Agric. 2017,142, 369–379.

Lu, Y.; Yi, S.; Zeng, N.; Liu, Y.; Zhang, Y. Identification of rice diseases using deep convolutional neural networks. Neurocomputing 2017, 267, 378–384.

M.A. Kizrak and B. Bolat Derin Öğrenme ile Kalabalık Analizi Üzerine Detaylı Bir Araştırma, Bilişim Teknolojileri Dergisi, 11:3 (2018) 263-286.

M.S. Vinutha, R.Kharbanda, B.Rashmi, S.N.Rajani and K.P.Rajesh, "crop monitoring: Using mobilenet models," Intern.Res.Jour.Engand Tech(IRJET), vol.6, pp.290-296, may 2019.

M.Z. Siti, M.A.Zulkifley, M. M.Stofa, A.M.Kamari and N.A.Mohamed, "classification of tomato leaf diseases using Mobilenetv2," Appl. Artif. Intell, vol.9, pp.290–296, June 2020, doi:110.11591/ijai. v9.i2.pp290- 296

Makerere AI Lab and AIR Lab National Crops Resources Research Institute (NaCRRI). (2020). iBean. [Online]. Available: https://github.com/AILab-Makerere/ibean/blob/ master/ README.md.

McCulloch, W.S., Pitts, W., 1943. A logical calculus of the ideas imminent in nervous activity. Bull. Math. Biophys. 115–133.

Memon, M.S.; Kumar, P.; Iqbal, R. Meta Deep Learn Leaf Disease Identification Model for Cotton Crop. Computers 2022, 11, 102.

Milioto, A.; Lottes, P.; Stachniss, C. Real-Time Blob-Wise Sugar Beets VS Weeds Classification for Monitoring Fields Using Convolutional Neural Networks. ISPRS Ann. Photogramm. Remote Sens. Spat. Inf. Sci. 2017, 4, 41–48.

Mingxing Tan, Quoc V. "Le EfficientNet: Rethinking Model Scaling for Convolutional Neural Networks», https://arxiv.org/abs/1905.11946, 2019

Mishra, S.; Sachan, R.; Rajpal, D. Deep convolutional neural network based detection system for real-time corn plant disease recognition. Procedia Comput. Sci. 2020, 167, 2003–2010.

Mkonyi, L.; Rubanga, D.; Richard, M.; Zekeya, N.; Sawahiko, S.; Maiseli, B.; Machuve, D. Early identification of Tuta absoluta in tomato plants using deep learning. Sci. Afr. 2020, 10, e00590.

Mohanty, S.P.; Hughes, D.P.; Salathé, M. Using deep learning for image-based plant disease detection. Front. Plant Sci. 2016, 7, 1419.

Møller, M., Moller, M.F., 1993. A scaled conjugate gradient algorithm for fast supervised learning. Neural networks 6, 525-533. Neural Network. 6, 525–533. https://doi.org/10.1016/S0893-6080(05)80056-5.

Mosavi A., Varkonyi-Koczy A. R.: Integration of Machine Learning and Optimization for Robot Learning. Advances in Intelligent Systems and Computing 519, 349-355 (2017).

Muhammad,E., Chowdhury,H., Tawsifur,R., Amith,K., Mohamed.A.,Aftab,U.K., Muhammad,S.K., Nasser,A., Mamun,B.I., Mohammad,T.I., Sawal,H.A.,2021.i. Automatic and Reliable Leaf Disease Detection Using Deep Learning Techniques, AgriEngineering, v.3, pp.294–312.URL.https://doi.org/ 10.3390/ agriengineering 3020020.

Mukti, I.Z.; Biswas, D. Transfer learning based plant diseases detection using ResNet50. In Proceedings of the 4th IEEE International Conference on Electrical Information and Communication Technology (EICT), Khulna, Bangladesh, 20–22 December 2019; pp. 1–6.

Mukti, I.Z.; Biswas, D. Transfer learning-based plant diseases detection using ResNet50. In Proceedings of the 4th IEEE International Conference on Electrical Information and Communication Technology (EICT), Khulna, Bangladesh, 20–22 December 2019; pp. 1–6.

Mukti, Ishrat Zahan, and Dipayan Biswas. "Transfer learning based plant diseases detection using ResNet50." In 2019 4th International conference on electrical information and communication technology (EICT), pp. 1-6. IEEE, 2019

Nachtigall, L.G.; Araujo, R.M.; Nachtigall, G.R. Classification of apple tree disorders using convolutional neural networks. In Proceedings of the 28th IEEE International Conference on Tools with Artificial Intelligence (ICTAI), San Jose, CA, USA, 6–8 November 2016; pp. 472–476.

Nagi, R.; Tripathy, S.S. Deep convolutional neural network based disease identification in grapevine leaf images. Multimed. Tools Appl. 2022, 81, 24995–25006.

Naik, B.N.; Ramanathan, M.; Palanisamy, P. Detection and classification of chilli leaf disease using a squeeze-and-excitation-based CNN model. Ecol. Inform. 2022, 69, 101663.

Nandhini, M.; Kala, K.U.; Thangadarshini, M.; Verma, S.M. Deep Learning model of sequential image classifier for crop disease detection in plantain tree cultivation. Comput. Electron. Agric. 2022, 197, 106915.

Nkemelu, D.K.; Omeiza, D.; Lubalo, N. Deep convolutional neural network for plant seedlings classification. arXiv 2018, arXiv:1811.08404.

Oppenheim, D.; Shani, G. Potato disease classification using convolution neural networks. Adv. Anim. Biosci. 2017, 8, 244–249.

Osdaghi, E., Safaie, N., & Alizadeh, A. (2017). Prevalence and pathogenicity of fungal pathogens associated with almond trees decline in Iran. Crop Protection, 101, 48-56.

Oyewola, D.O.; Dada, E.G.; Misra, S.; Damaševiˇcius, R. Detecting cassava mosaic disease using a deep residual convolutional neural network with distinct block processing. PeerJ Comput. Sci. 2021, 7, e352.

Ozbılge, E.; Ulukok, M.K.; Toygar, O.; Ozbılge, E. Tomato Disease Recognition Using a Compact Convolutional Neural Network. IEEE Access 2022, 10, 77213–77224. DOI:10.1109/ACCESS.2022.3192428.

Pan SJ and Yang Q (2010) A survey on transfer learning. IEEE Transactions on Knowledge and Data Engineering 22, 1345–1359.

Pandey, A.; Jain, K. A robust deep attention dense convolutional neural network for plant leaf disease identification and classification from smart phone captured real world images. Ecol. Inform. 2022, 70, 101725.

Pandian, J.A.; Kanchanadevi, K.; Kumar, V.D.; Jasi ´ nska, E.; Goˇ no, R.; Leonowicz, Z.; Jasinski, M.L. A Five Convolutional Layer Deep Convolutional Neural Network for Plant Leaf Disease Detection. Electronics 2022, 11, 1266.

Pandian, J.A.; Kumar, V.D.; Geman, O.; Hnatiuc, M.; Arif, M.; Kanchanadevi, K. Plant Disease Detection Using Deep Convolutional Neural Network. Appl. Sci. 2022, 12, 6982.

Pantazi, X.E., Moshou, D., and Tamouridou, A.A., "Automated leaf disease detection in different crop species through image features analysis and One Class Classifiers," Comput. Electron. Agric. vol.156, pp.96–104, January 2019, doi:/10.1016/j.compag.2018.11.005.

Partel, V.; Kim, J.; Costa, L.; Pardalos, P.M.; Ampatzidis, Y. Smart Sprayer for Precision Weed Control Using Artificial Intelligence: Comparison of Deep Learning Frameworks. In Proceedings of the International Symposium on Artificial Intelligence and Mathematics, ISAIM 2020, Fort Lauderdale, FL, USA, 6–8 January 2020.

Pearlstein, L.; Kim, M.; Seto, W. Convolutional neural network application to plant detection, based on synthetic imagery. In Proceedings of the 2016 IEEE Applied Imagery Pattern RecognitionWorkshop (AIPR), Washington, DC, USA, 18–20 October 2016; pp. 1–4.

Picon, A., Alvarez-Gilaa, A., Seitzcd, M., Ortiz-Barredob, A., Echazarra, J., and Johannesd, A., "Deep convolutional neural networks for mobile capture device-based crop disease classification in the wild," Comput.Electron. Agric, vol.161, pp.280-290, June 2019, doi: 10.1016/j.compag.2018.04.002

Picón, A.; Alvarez-Gila, A.; Seitz, M.; Ortiz-Barredo, A.; Echazarra, J.; Johannes, A. Deep convolutional neural networks for mobile capture device-based crop disease classification in the wild. Comput. Electron. Agric. 2019, 161, 280–290.

Prabu, M.; Chelliah, B.J. Mango leaf disease identification and classification using a CNN architecture optimized by crossoverbased levy flight distribution algorithm. Neural Comput. Appl. 2022, 34, 7311–7324.

Ramcharan, A.; Baranowski, K.; McCloskey, P.; Ahmed, B.; Legg, J.; Hughes, D.P. Deep learning for image-based cassava disease detection. Front. Plant Sci. 2017, 8, 1852.

Ramcharan, A.; McCloskey, P.; Baranowski, K.; Mbilinyi, N.; Mrisho, L.; Ndalahwa, M.; Legg, J.; Hughes, D.P. A mobile-based deep learning model for cassava disease diagnosis. Front. Plant Sci. 2019, 10, 272.

Rangarajan, A.K.; Purushothaman, R. Disease classification in eggplant using pre-trained vgg16 and msvm. Sci. Rep. 2020, 10, 1–11.

Rangarajan, A.K.; Purushothaman, R. Disease Classification in Eggplant Using Pre-trained VGG16 and MSVM. Sci. Rep. 2020, 10, 2322.

Rangarajan, A.K.; Purushothaman, R.; Ramesh, A. Tomato crop disease classification using pre-trained deep learning algorithm. Procedia Comput. Sci. 2018, 133, 1040–1047. [

Ravi, V.; Acharya, V.; Pham, T.D. Attention deep learning-based large-scale learning classifier for Cassava leaf disease classification. Expert Syst. 2022, 39, e12862.

Redmon, J. Darknet: Open Source Neural Networks in C. 2013–2016. Available online: http://pjreddie.com/darknet/ (accessed on 13 June 2022).

Redmon, J., Divvala, S., Girshick, R., Farhadi, A., 2016. You only look once: Unified, real-time object detection. In: 2016 IEEE Conference on Computer Vision and Pattern Recognition (CVPR), pp. 779–788. https://doi.org/10.1109/CVPR.2016.91.

Redmon, J.; Farhadi, A. Yolov3: An incremental improvement. arXiv 2018, arXiv:1804.02767

Ren, S.; He, K.; Girshick, R.; Sun, J. Faster r-cnn: Towards real-time object detection with region proposal networks. In Proceedings of the Advances in Neural Information Processing Systems 28 (NIPS 2015), Montreal, QC, Canada, 7–12 December 2016.

Richey, B., Majumder, S., Shirvaikar, M. and N. Kehtarnavaz, " Real-time detection of maize crop disease via a deep learning-based smartphone app," In Real-Time Image Processing and Deep Learning, International Society for Optics and Photonics, Proceedings of the SPIE, vol.11401, pp.2020, 22 April 2020, doi:10.1117/12.2557317.

Ronneberger, O., Fischer, P., Brox, T., 2015. U-net: Convolutional networks for biomedical image segmentation. In: Navab, N., Hornegger, J., Wells, W.M.

Rosenfeld, A., & Richardson, A. (2019). Explainability in human–agent systems. *Autonomous Agents and Multi-Agent Systems*, *33*, 673-705.

Rumelhart, D.E., Hinton, G.E., Williams, R.J., 1986. In: McClelland, J.L. (Ed.), Learning Internal Representations by Error Propagation. MIT Press, Cambridge. Sadegh.

Russel, N.S.; Selvaraj, A. Leaf species and disease classification using multiscale parallel deep CNN architecture. Neural Comput. Appl. 2022.

Ruth, J.A.; Uma, R.; Meenakshi, A.; Ramkumar, P. Meta-Heuristic Based Deep Learning Model for Leaf Diseases Detection. Neural Process. Lett. 2022..

S Medina, J.R., Garrido, J.M., G´omez-Martín, M.E., Vidal, C., 2004. Armor damage nalysis using neural networks. Coastal structures 236–248. Menezes

Sahu, P., Chug, A., Singh, A.P., Singh,D.,Singh,R.P.,2021. Deep Learning Models for Beans Crop Diseases: Classification and Visualization Techniques, International Journal of Modern Agriculture, v.10, No.1, ISSN: 2305-7246.URL.http://modernjournals.com/index.php/ijma/article/view/670

Sahu, P.; Chug, A.; Singh, A.P.; Singh, D.; Singh, R.P. Deep Learning Models for Beans Crop Diseases: Classification and Visualization Techniques. Int. J. Mod. Agric. 2021, 10, 796–812.

Sambasivam, G. A. O. G. D., and Geoffrey Duncan Opiyo. "A predictive machine learning application in agriculture: Cassava disease detection and classification with imbalanced dataset using convolutional neural networks." Egyptian informatics journal 22, no. 1 (2021): 27-34.

Schmidhuber, J. Deep learning in neural networks: An overview. Neural Netw. 2015, 61, 85–117.

Sembiring, A., Away, Y., Arnia, F., and Muharar, R.,, “Development of concise convolutional neural network for tomato plant disease classification based on leaf images,” In: Journal of Physics: Conference Series. ICIASGA, Jawa Timur, Indonesia, vol.1845, pp.4-5, November 2020, doi:10.1088/1742-6596/1845/1/012009.

Sepp Hochreiter; Jürgen Schmidhuber (1997). "Long short-term memory". Neural Computation. 9 (8): 1735–1780. doi:10.1162/neco.1997.9.8.1735.

Shao, L., Zhu, F., Li, X., 2015. Transfer learning for visual categorization: A survey. IEEE Trans. Neural Netw. Learn. Syst. 26, 1019–1034. https://doi.org/10.1109/TNNLS.2014.2330900.

Shehu, H. A.; A. Rabie, M. H. Sharif et al., “Artificial intelligence tools and their capabilities,” PLOMS AI, vol. 1, no. 1, 2021.

Shijie,J., Peiyi , J., Siping , H., and Haibo, L., “Automatic detection of tomato diseases and pests based on leaf images,” Chinese Automation Congress,Jinan, China, 2017, doi: 10.1109/CAC.2017.8243388.

Shorten, C., Khoshgoftaar, T.M., 2019. A survey on image data augmentation for deep learning. J. Big Data 6, 60. https://doi.org/10.1186/s40537-019-0197-0.

Shrivastava, V.K.; Pradhan, M.K.; Minz, S.; Thakur, M.P. Rice plant disease classification using transfer learning of deep convolution neural network. Int. Arch. Photogramm. Remote Sens. Spat. Inf. Sci. 2019, XLII-3/W6, 631–635.

Sibiya, M.; Sumbwanyambe, M. A computational procedure for the recognition and classification of maize leaf diseases out of healthy leaves using convolutional neural networks. AgriEngineering 2019, 1, 119–131.

Simonyan K and Zisserman A (2014) Very deep convolutional networks for large-scale image recognition. arXiv preprint arXiv, 1409.1556 [cs.CV]. Szegedy C, Ioffe S, Vanhoucke V and Alemi AA (2017) Inception-v4, Inception-ResNet and the impact of residual connections on learning. In Proceedings of the Thirty-First AAAI Conference on Artificial Intelligence (AAAI-17). Palo Alto, CA, USA: AAAI, pp. 4278–4284.

Singh, A.K.; Ganapathysubramanian, B.; Sarkar, S.; Singh, A. Deep learning for plant Stress phenotyping: Trends and future perspectives. Trends Plant Sci. 2018, 23, 883–898.

Singh, R.K.; Tiwari, A.; Gupta, R.K. Deep transfer modeling for classification of Maize Plant Leaf Disease. Multimed. Tools Appl.2022, 81, 6051–6067.

Singh, U.P.; Chouhan, S.S.; Jain, S.; Jain, S. Multilayer Convolution Neural Network for the Classification of Mango Leaves Infected by Anthracnose Disease. IEEE Access 2019, 7, 43721–43729.

Sladojevic, S.; Arsenovic, M.; Anderla, A.; Culibrk, D.; Stefanovic, D. Deep neural networks-based recognition of plant diseases by leaf image classification. Comput. Intell. Neurosci. 2016, 2016, 1–10.

Sladojevic, S.; Arsenovic, M.; Anderla, A.; Culibrk, D.; Stefanovic, D. Deep neural networks based recognition of plant diseases by leaf image classification. Comput. Intell. Neurosci. 2016, 2016, 3289801

Sledevic, T., 2019. Adaptation of convolution and batch normalization layer for cnn implementation on fpga. In: 2019 Open Conference of Electrical, Electronic and Information Sciences (eStream), pp. 1–4.

Srinivasan, R., Ramasamy, R., Prabha, A. L., & Prasad, P. V. (2013). Relationship between crop type and pathogen group in rice, tomato and cucumber crops of India. Indian Journal of Plant Protection, 41(1), 47-53.

Stanford University Machine Learning Course. Available online. https://www.coursera.org /learn/machine-learning/home.

Subetha, T.; Khilar, R.; Christo, M.S. A comparative analysis on plant pathology classification using deep learning architecture Resnet and VGG19. Mater. Today Proc. 2021, Epub ahead of printing.

Subramanian, M.; Shanmugavadivel, K.; Nandhini, P.S. On fine-tuning deep learning models using transfer learning and hyper-parameters optimization for disease identification in maize leaves. Neural Comput. Appl. 2022, 34, 13951–13968.

Sun, H.; Zhai, L.; Teng, F.; Li, Z.; Zhang, Z. qRgls1. 06, a major QTL conferring resistance to gray leaf spot disease in maize. Crop. J. 2021, 9, 342–350.

Sun, X.; Li, G.; Qu, P.; Xie, X.; Pan, X.; Zhang, W. Research on plant disease identification based on CNN. Cogn. Robot. 2022, 2, 155–163.

Szegedy C, Liu W, Jia Y, Sermanet P, Reed S, Anguelov D, Erhan D, Vanhoucke V and Rabinovich A (2015) Going deeper with convolutions. In IEEE Conference on Computer Vision and Pattern Recognition. Piscataway, NJ, USA: IEEE, pp. 1–9.

Szegedy, C.; Vanhoucke, V.; Ioffe, S.; Shlens, J.; and Wojna, Z., "Rethinkin the inception architecture for computer vision," in Proceedings of the IEEE conference on computer vision and pattern recognition, 2016, pp. 2818–2826.

Szegedy, C.; W. Liu, Y. Jia, P. Sermanet, S. Reed, D. Anguelov, D. Erhan, V. Vanhoucke, and A. Rabinovich, "Going deeper with convolutions," in Proceedings of the IEEE conference on computer vision and pattern recognition,2015, pp. 1–9.

Tesauro, G., 1992. "Practical issues in temporal difference learning. Machine Learning", IBM Thomas J. Watson Research Center, pp. 8, 257-277.

Thapa, R.; Zhang, K.; Snavely, N.; Belongie, S.; Khan, A. The Plant Pathology Challenge 2020 data set to classify foliar disease of apples. Appl. Plant Sci. 2020, 8, e11390.

Toda, Y.; Okura, F. How Convolutional neural networks diagnose plant disease. Plant Phenomics 2019, 2019, 9237136.

Traore, B.B.; Kamsu-Foguem, B.; Tangara, F. Deep convolution neural network for image recognition. Ecol. Inform. 2018, 48, 257–268.

Trivedi, N.K.; Anand, A.; Aljahdali, H.M.; Gautam, V.; Villar, S.G.;Goyal, N.; Anand, D.; Kadry, S. Early Detection and Classification of Tomato Leaf Diseases Using High Performance Deep Neural Network. Sensors 2021, 21, 7987. https://doi.org/10.3390/s21237987

Tugrul, B.; Elfatimi, E.; Eryigit, R. Convolutional Neural Networks in Detection of Plant Leaf Diseases: A Review. *Agriculture* 2022, *12*, 1192. https://doi.org/10.3390/ agriculture12081192.

University, P. S. (2019). PlantVillage. https://plantvillage.psu.edu//.(Visited on 12/10/2019) (cit. on p. 52).

URL-1, http://buyukveri.firat.edu.tr/2018/04/17/derin-sinir-aglari-icin aktivasyonfonksiyonlari/, Derin Sinir Ağları İçin Aktivasyon Fonksiyonları-Büyük Veri ve Yapay Zekâ Laboratuvarı. 10 Mayıs 2018.

Vallabhajosyula, S.; Sistla, V.; Kolli, V.K.K. Transfer learning-based deep ensemble neural network for plant leaf disease detection. J. Plant Dis. Prot. 2021, 129, 545–558.

Vijay, N. Detection of Plant Diseases in Tomato Leaves: With Focus on Providing Explainability and Evaluating User Trust. Master's Thesis, University of Skovde, Skovde, Sweden, September2021.https://www.divaportal.org/smash/record.jsf?pid=diva2%3A1593851&dswid=4788.

Wang, F.; Rao, Y.L.; Luo, Q.; Jin, X.; Jiang, Z.H.; Zhang, W.; Li, S. Practical cucumber leaf disease recognition using improved Swin Transformer and small sample size. Comput. Electron. Agric. 2022, 199, 107163.

Wang, G.; Sun, Y.;Wang, J. Automatic image-based plant disease severity estimation using deep learning. Comput. Intell. Neurosci. 2017, 2017, 2917536.

Wang, X., Shrivastava, A., Gupta, A., 2017. A-fast-rcnn: Hard positive generation via adversary for object detection, 2017. ArXiv:1704.03414.

Wang, Y.; Zhang, H.; Liu, Q.; Zhang, Y. Image classification of tomato leaf diseases based on transfer learning. J. China Agric. Univ.2019, 24, 124–130.

Wei, K.; Chen, B.; Zhang, J.; Fan, S.;Wu, K.; Liu, G.; Chen, D. Explainable Deep Learning Study for Leaf Disease Classification.Agronomy 2022, 12, 1035.

Wu, S. G., Bao, F. S., Xu, E. Y., Wang, Y.-X., Chang, Y.-F., and Xiang,Q.-L. (2007). « A leaf recognition algorithm for plant classification using probabilistic neural network ». In: 2007 IEEE international symposium on signal processing and information technology. IEEE, pp. 11–16 (cit. on p. 52).

Wulff, E. G., Sultana, T., & Castañeda-Álvarez, N. P. (2014). Plant disease diagnosis: Successes, challenges, and opportunities. Annual Review of Phytopathology, 52, 489-512.

Xiang, Q.; X. Wang, R. Li, G. Zhang, J. Lai, and Q. Hu, ``Fruit image classification based on MobileNetV2 with transfer learning technique," in Proc. 3rd Int. Conf. Comput. Sci. Appl. Eng., vol. 6, Oct. 2019, pp. 17, doi: 10.1145/3331453.3361658.

Xu, Y.; Kong, S.; Gao, Z.; Chen, Q.; Jiao, Y.B.; Li, C. HLNet Model and Application in Crop Leaf Diseases Identification. Sustainability 2022, 14, 8915.

Yadav, S.; Sengar, N.; Singh, A.; Singh, A.; Dutta, M.K. Identification of disease using deep learning and evaluation of bacteriosis in peach leaf. Ecol. Inform. 2021, 61, 101247.

Yakkundimath, R.; Saunshi, G.; Anami, B.S.; Palaiah, S. Classification of Rice Diseases using Convolutional Neural Network Models. J. Inst. Eng. Ser. B 2022, 103, 1047–1059.

Yin, H.; Gu, Y.H.; Park, C.J.; Park, J.H.; Yoo, S.J. Transfer Learning-Based Search Model for Hot Pepper Diseases and Pests. Agriculture 2020, 10, 439.

Yu, H.; Cheng, X.; Chen, C.; Heidari, A.A.; Liu, J.; Cai, Z.; Chen, H. Apple leaf disease recognition method with improved residual network. Multimed. Tools Appl. 2022, 81, 7759–7782.

Zeng, Q.; Ma, X.; Cheng, B.; Zhou, E.; Pang, W. Gans-based data augmentation for citrus disease severity detection using deep learning. IEEE Access 2020, 8, 882–891.

Zeng, W.; Li, H.; Hu, G.; Liang, D. Lightweight dense-scale network (LDSNet) for corn leaf disease identification. Comput. Electron. Agric. 2022, 197, 106943.

Zhang, H.; Tang, Z.; Xie, Y.; Gao, X.; Chen, Q. A watershed segmentation algorithm based on an optimal marker for bubble size measurement. Measurement 2019, 138, 182–193.

Zhang, K. He.; Ren, X.S.; and Sun, J., "Deep residual learning for image recognition," in Proceedings of the IEEE conference on computer vision and pattern recognition, 2016, pp. 770–778.

Zhang, K.; Wu, Q.; Chen, Y. Detecting soybean leaf disease from synthetic image using multi-feature fusion faster R-CNN. Comput. Electron. Agric. 2021, 183, 106064.

Zhang, S.; Zhang, S.; Zhang, C.; Wang, X.; Shi, Y. Cucumber leaf disease identification with global pooling dilated convolutional neural network. Comput. Electron. Agric. 2019, 162, 422–430.

Zhang, W.; Hansen, M.F.; Volonakis, T.N.; Smith, M.; Smith, L.; Wilson, J.; Ralston, G.; Broadbent, L.; Wright, G. Broad-leaf weed detection in pasture. In Proceedings of the 3rd IEEE International Conference on Image, Vision and Computing (ICIVC), Chongqing, China, 27–29 June 2018; pp. 101–105.

Zhang, X.; Zhou, X.; Lin, M.; and Sun, J., "Shufflenet: An extremely efficient convolutional neural network for mobile devices," in Proceedings of the IEEE conference on computer vision and pattern recognition, 2018, pp. 6848–6856.